\documentclass[acmtog, nonacm]{acmart}

\usepackage{cleveref}
\usepackage{multirow}
\usepackage{subcaption}
\usepackage{tabularx}
\usepackage{svg}
\usepackage[export]{adjustbox}
\usepackage{lipsum}  
\usepackage[percent]{overpic}

\AtBeginDocument{%
  }

\begin{document}

\title{Learning Fine-to-Coarse Cuboid Shape Abstraction}

\author{Gregor Kobsik}
\email{kobsik@cs.rwth-aachen.de}
\author{Morten Henkel} 
\author{Yanjiang He}
\author{Victor Czech}
\author{Tim Elsner}
\author{Isaak Lim}
\author{Leif Kobbelt}
\email{sekretariati8@cs.rwth-aachen.de}
\affiliation{
  \institution{Visual Computing Institute | RWTH Aachen University}
  \city{Aachen}
  \country{Germany}
}

\renewcommand{\shortauthors}{Kobsik et al.}

\begin{abstract}
    The abstraction of 3D objects with simple geometric primitives like cuboids allows to infer structural information from complex geometry.
    It is important for 3D shape understanding, structural analysis and geometric modeling. 
    
    We introduce a novel fine-to-coarse unsupervised learning approach to abstract collections of 3D shapes.
    Our architectural design allows us to reduce the number of primitives from hundreds (fine reconstruction) to only a few (coarse abstraction) during training.
    This allows our network to optimize the reconstruction error and adhere to a user-specified number of primitives per shape while simultaneously learning a consistent structure across the whole collection of data.
    We achieve this through our abstraction loss formulation which increasingly penalizes redundant primitives.
    Furthermore, we introduce a reconstruction loss formulation to account not only for surface approximation but also volume preservation. 
    Combining both contributions allows us to represent 3D shapes more precisely with fewer cuboid primitives than previous work.
    
    We evaluate our method on collections of man-made and humanoid shapes comparing with previous state-of-the-art learning methods on commonly used benchmarks. 
    Our results confirm an improvement over previous cuboid-based shape abstraction techniques.
    Furthermore, we demonstrate our cuboid abstraction in downstream tasks like clustering, retrieval, and partial symmetry detection.
\end{abstract}



\begin{teaserfigure}
    \rotatebox[origin=c]{90}{\LARGE \textbf{Training}}
    \hspace{0.1in}
    \begin{adjustbox}{width=3.1in}
        



\begin{tabular}{c}
    \begin{overpic}[trim=0 0 0 -4.5in]{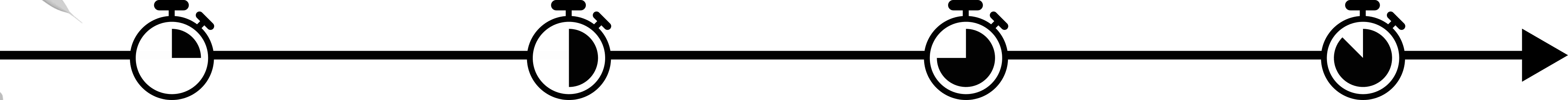}
        
        \put(-3,2){\includegraphics[scale=2.0]{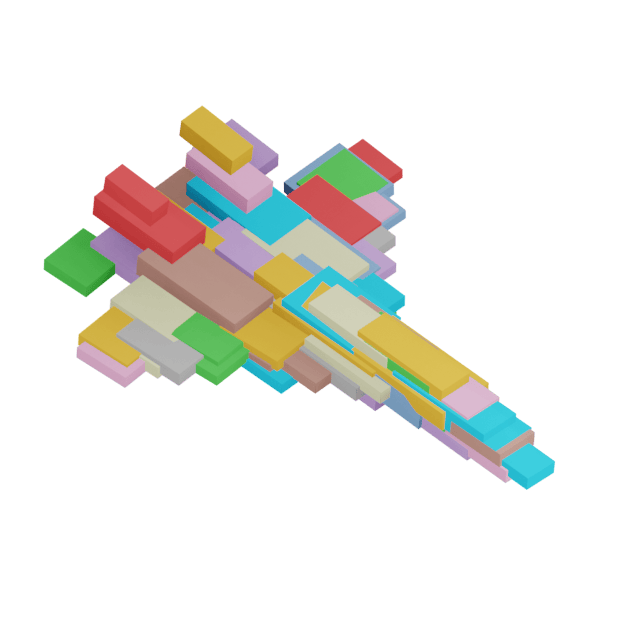}}
        \put(22,2){\includegraphics[scale=2.0]{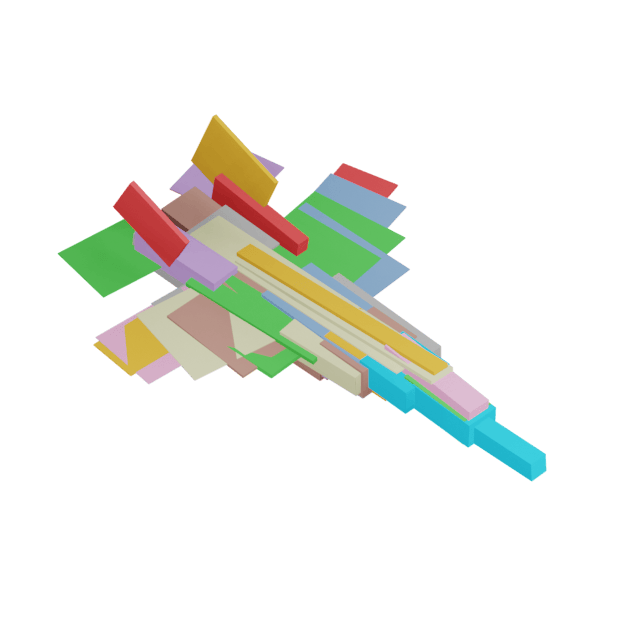}}
        \put(47,2){\includegraphics[scale=2.0]{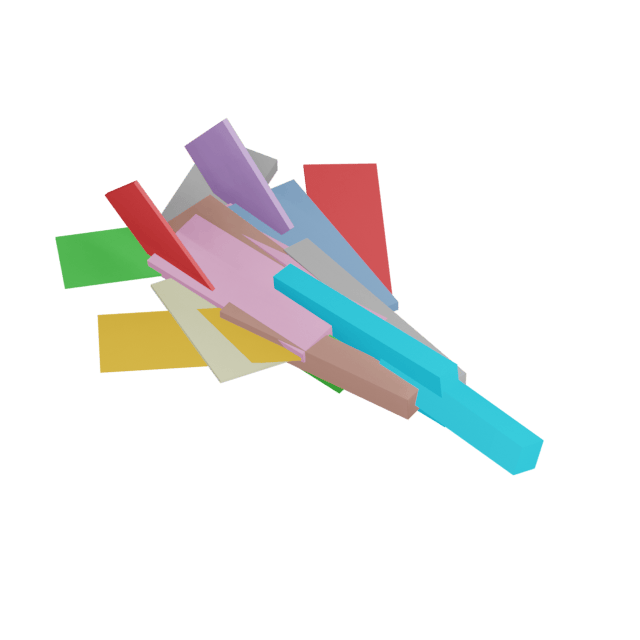}}
        \put(72,2){\includegraphics[scale=2.0]{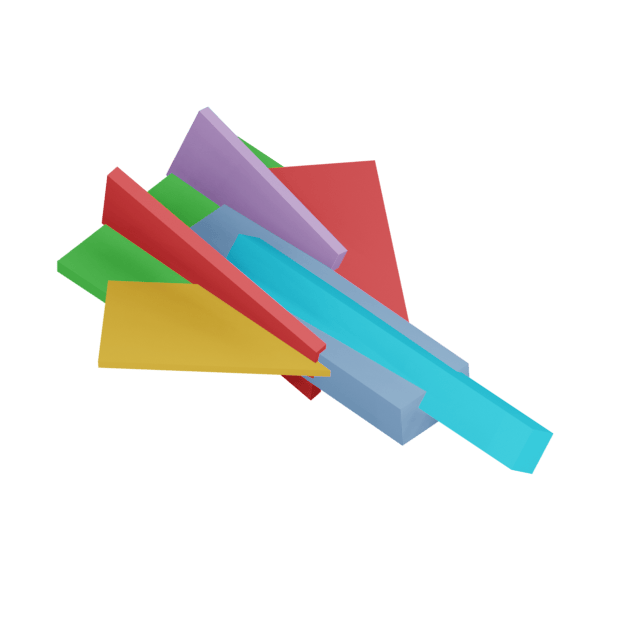}}


        
    \end{overpic} \\
\end{tabular}
    \end{adjustbox}
    \hfill
    \rotatebox[origin=c]{90}{\LARGE \textbf{Results}}
    \hspace{0.1in}
    \begin{adjustbox}{width=3.1in}
        \begin{tabular}{cccc}
    
    \begin{overpic}[]{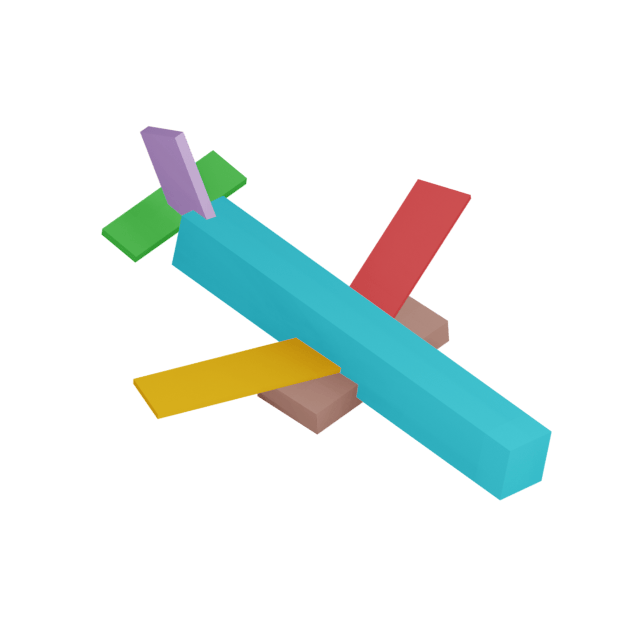}
        \put(-20,-20){\includegraphics[scale=0.8]{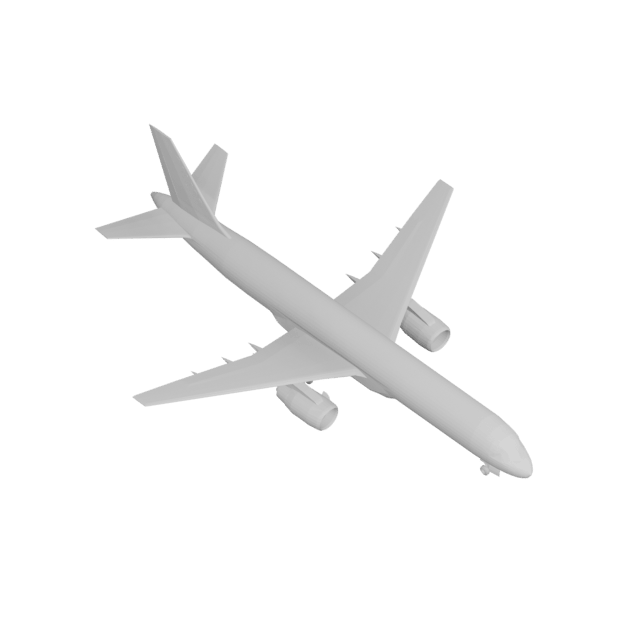}}
    \end{overpic} &
    \begin{overpic}[]{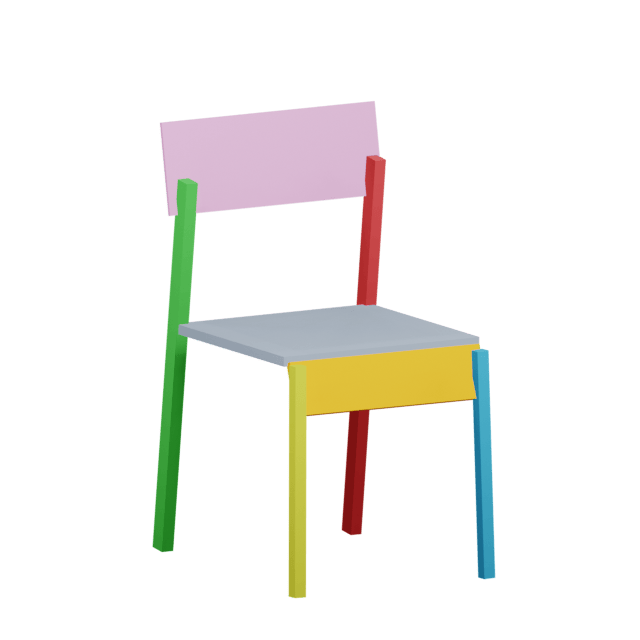}
        \put(-20,-20){\includegraphics[scale=0.7]{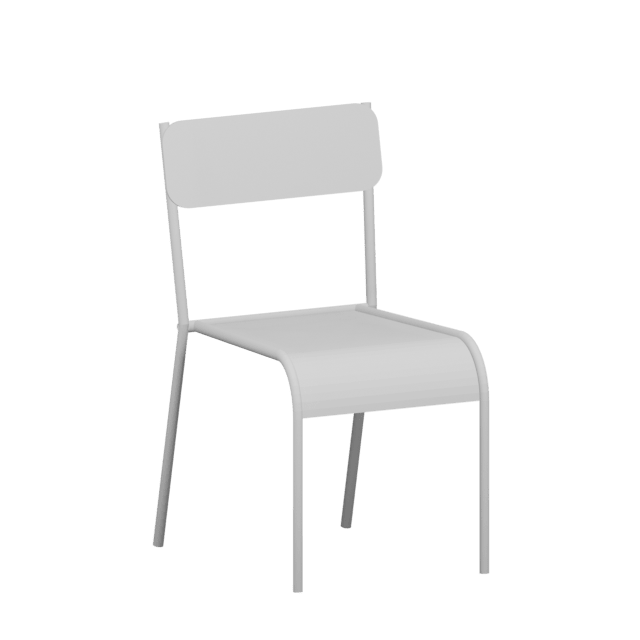}}
    \end{overpic} &  
    \begin{overpic}[]{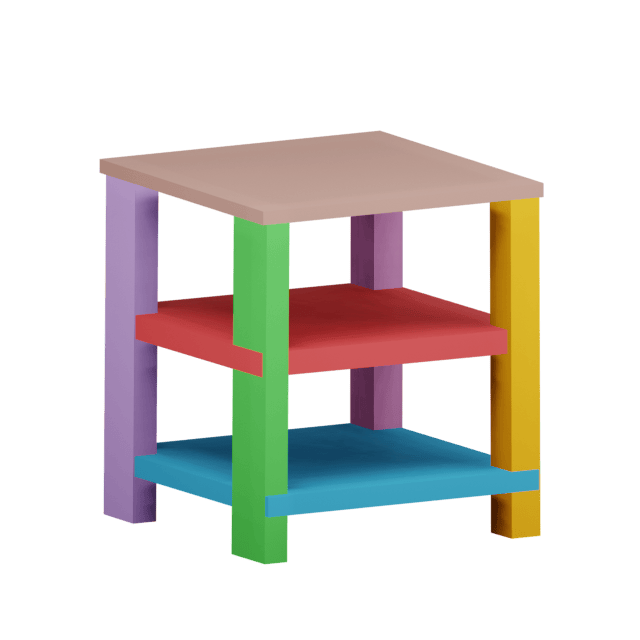}
        \put(-20,-20){\includegraphics[scale=0.7]{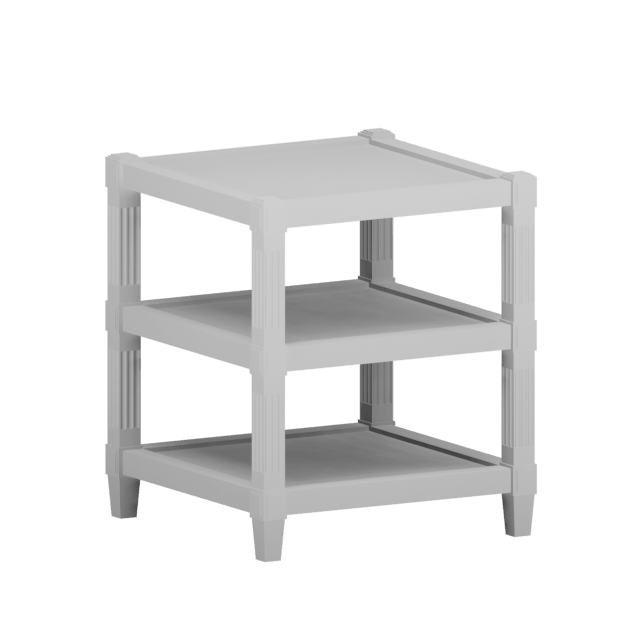}}
    \end{overpic} & 
    \begin{overpic}[]{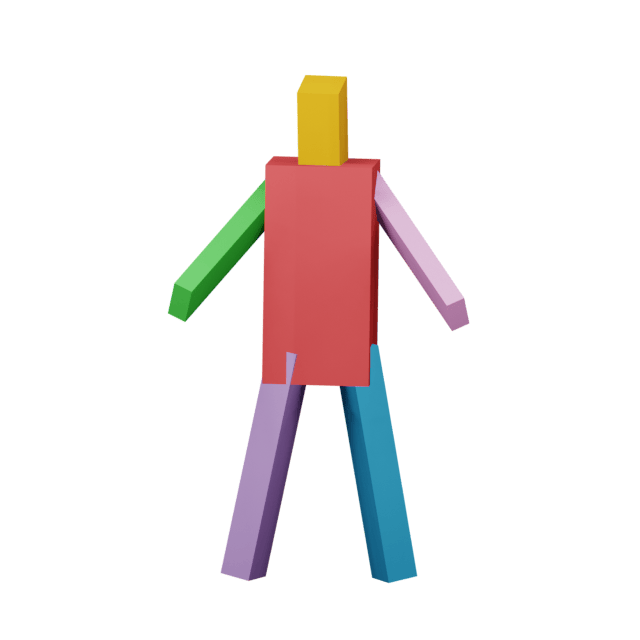}
        \put(-20,-20){\includegraphics[scale=0.7]{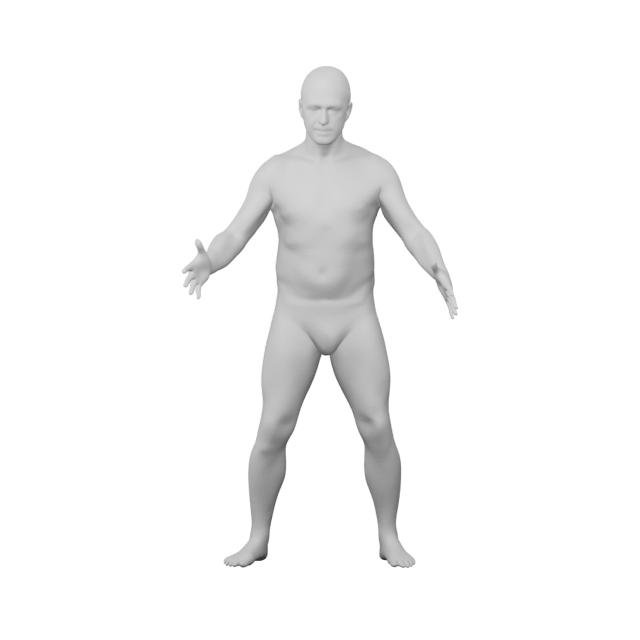}}
    \end{overpic} \\

    \begin{overpic}[]{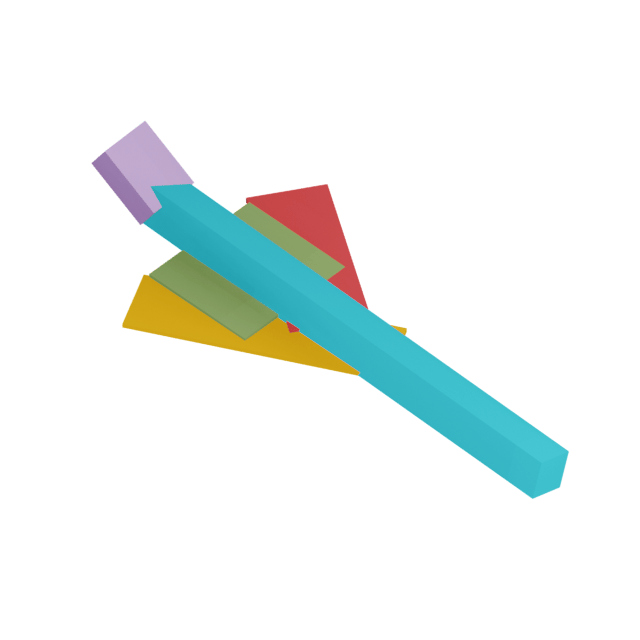}
        \put(-20,-20){\includegraphics[scale=0.8]{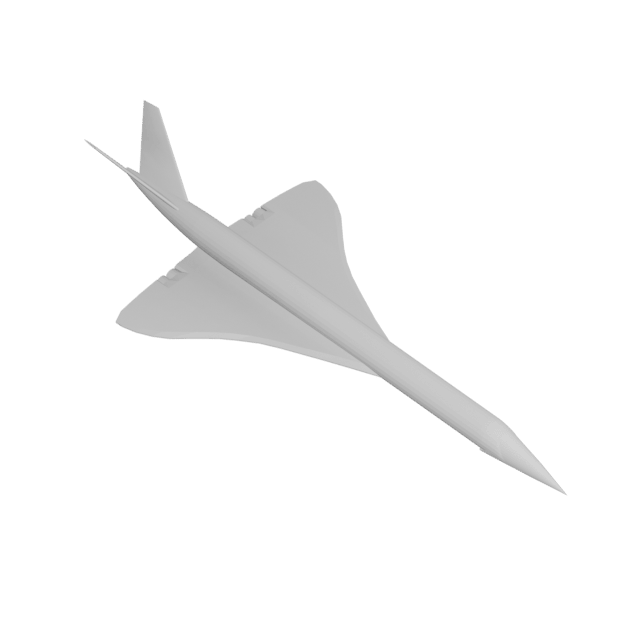}}
    \end{overpic} & 
    \begin{overpic}[]{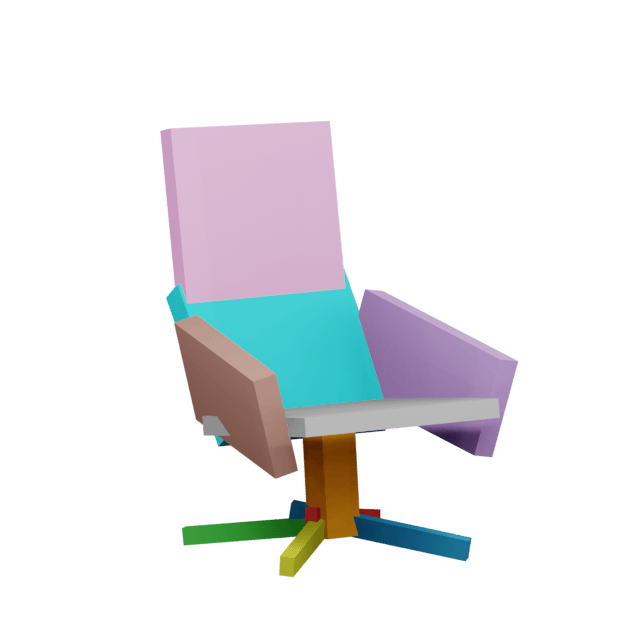}
        \put(-20,-20){\includegraphics[scale=0.7]{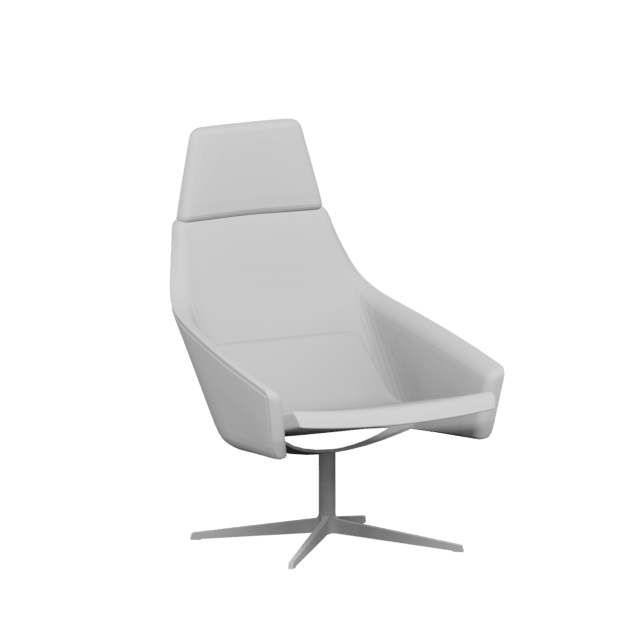}}
    \end{overpic} & 
    \begin{overpic}[]{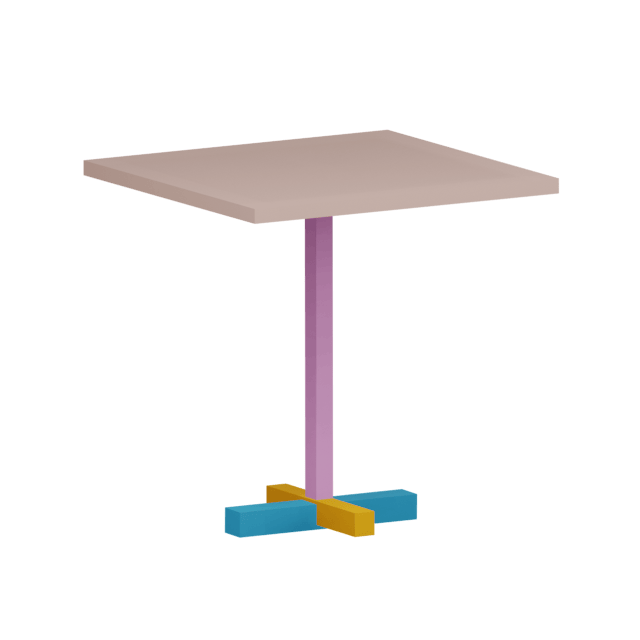}
        \put(-20,-20){\includegraphics[scale=0.7]{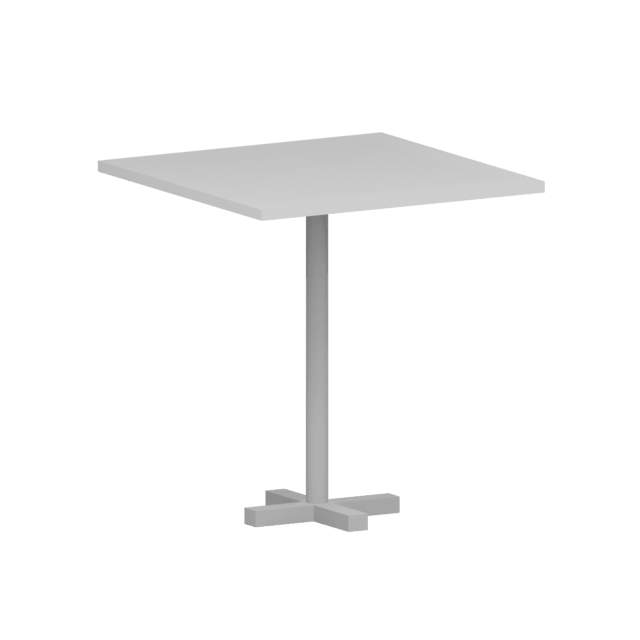}}
    \end{overpic} & 
    
    \begin{overpic}[]{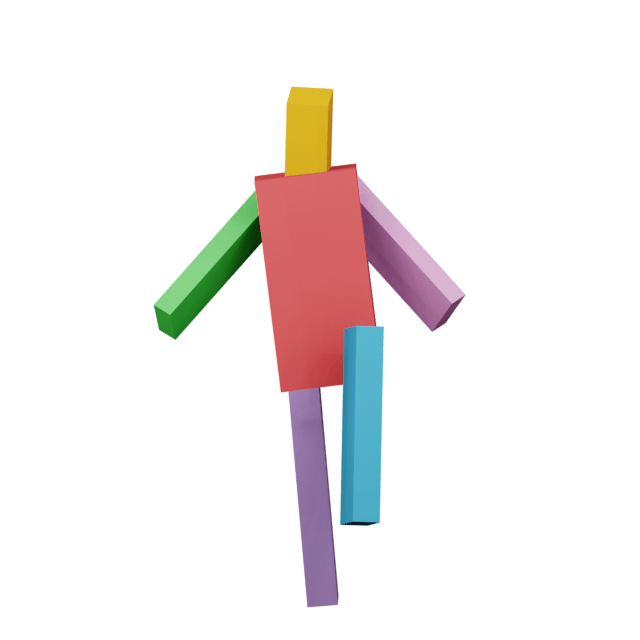}
        \put(-20,-20){\includegraphics[scale=0.7]{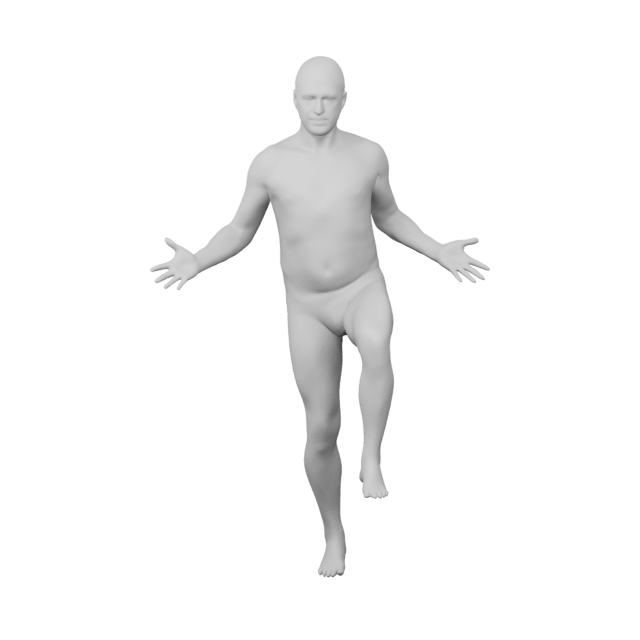}}
    \end{overpic} \\
       
\end{tabular}
        
    \end{adjustbox}
    \caption{Left: During training the number of primitives decreases from hundreds of cuboids to only a few ones. Colors are based on a global cuboid index. \\
    Right: After discarding and merging redundant cuboids we obtain a structurally consistent abstraction of a shape. Colors indicate distinct parts per category.}
    \label{fig:teaser}
\end{teaserfigure}


\maketitle

\section{INTRODUCTION}
\label{sec:introduction}

The decomposition of 3D objects into their underlying structural elements is an longstanding problem in the field of computer graphics and computer vision.
Real world objects can often be decomposed into geometrically distinct parts.
Exemplary, a chair can be decomposed into a seat, a backrest and four legs, while a human consists of a head, a torso, as well as two arms and legs (cmp. \Cref{fig:teaser}).
Obtaining the underlying structure of objects, allows us to argue about 3D shapes by analyzing the relationships between their components and automatically classify collections of them into different sub-categories, e.g. chairs without a backrest or tables with a single leg support.
The desired abstraction should adhere to two target objectives.
It should be \textit{expressive} to represent the object as truthfully as possible, but also \textit{compact} to abstract the general structure of the object \cite{HCA_Sun}.
Both objectives are non-trivial to satisfy simultaneously. 
An optimal solution to a compact representation is most often contrary to an expressive one and vice versa.

While this is challenging to automate, humans tend to decompose objects into its components naturally \cite{RbC_Biederman}.
Much research has been invested into learning the decomposition of 3D objects given manual annotations, either with structural information \cite{Structurenet_Mo, Annotation_Yi} or semantic labels \cite{shapenet_chang, PointNet_Qi, PointNeXt_Qian}.
However, annotating data, especially in 3D, is a time consuming process. 

Unsupervised techniques focus on the geometric properties of the object without human annotations \cite{VP_Tulsiani, HCA_Sun, CAS_Yang}.
In this paper we propose an unsupervised cuboid-based shape abstraction method.
In contrast to the previous methods, we do not optimize our model with a fixed number of primitives, but initially use an extensive amount of cuboid primitives to capture all details with a very fine-granular reconstruction, satisfying the expressiveness requirement first.
Then, we gradually trade reconstruction quality of our model for compactness by reducing the amount of available primitives during training.
This forces our model to learn a compact abstraction of the target shape using a prescribed number of primitives by discarding or merging redundant ones while approximating the shape as faithfully as possible.
During training individual primitives cover increasingly larger areas of the shape continuously improving the fitting to their geometric parts.
Ultimately, resulting in each cuboid representing a single distinct part of the object.

We choose cuboids as our representative primitive as it is one of the most basic 3D shapes. 
Even though superquadrics \cite{SQR_Paschalidou} or part templates \cite{BAE-Net_Chen} allow for a more precise approximation of the object our focus lies on the abstraction of the underlying shape structure and not shape reconstruction.
Cuboids offer the least number of degrees of freedom, are easily understandable by the user, and thus widely used in other structure-based algorithmic and learning approaches \cite{Struct_Mitra, Structurenet_Mo, PASTA_Li}.

Our key contributions are:
\begin{itemize}
    \item Novel training scheme for unsupervised shape abstraction with a fine-to-coarse loss functions, which allows discarding or merging redundant primitives.
    \item Simple user control over the abstraction quality by defining a target number of primitives.
    \item State-of-the-art performance on the structural shape reconstruction benchmarks for cuboid-based methods.
    \item Applying our results to downstream task like co-segmentation, clustering, retrieval, or partial symmetry detection.
\end{itemize}


\section{RELATED WORK}
\label{sec:related_work}


\subsection{Algorithmic Primitive Fitting}
Primitive fitting and shape abstraction techniques have a long tradition in 3D shape analysis \cite{RPM_Solina}.
They allow extracting structural information of only a single shape to obtain e.g. a label-agnostic segmentation of the shape, but correspondences between abstracted primitives within collections of shapes are not given \cite{SQS_Chevalier, SegMat_Lin}.
They miss on the extraction of global knowledge from a dataset and need to be re-run on unseen samples.


\subsection{Supervised Structural Learning}
Some research has been performed on supervised learning for structured representation of collections of 3D shapes \cite{GRASS_Li, SDM-Net_Gao, Structurenet_Mo, PASTA_Li}.
These works use the underlying structure to either interpolate between different shapes using the learned latent space or to generate novel 3D shapes following the learned structure of the data.
Conditioning a generative model on an abstract representation leads to more precise and structure-preserving results.
Furthermore, it allows to prescribe the final shape in an intuitively controllable manner.
Unfortunately, this data cannot be easily obtained and all approaches rely on annotated datasets.
These methods can benefit greatly from an automatic cuboid abstraction of 3D shapes.


\subsection{Unsupervised Shape Abstraction}
To extract the structure of geometric objects one cannot always depend on annotated data.
Unsupervised techniques focus on geometric features of a shape, learn a latent representation, and finally extract an approximate reconstruction of the input using a limited number of primitives.
One defining part of each method is the selection of the representative primitive.

\subsubsection{Cuboids} 
The most simple choice of a volumetric primitive is a cuboid \cite{VP_Tulsiani, HCA_Sun, CAS_Yang}. 
They allow intuitive interpretability of their parameters and straight-forward processing, but sacrifice reconstruction quality per primitive due to the limited expressiveness of the shape.
\cite{VP_Tulsiani} showed that they are still a valid choice for shape abstraction.
\cite{HCA_Sun} continued the line of work and extended it to hierarchical compositions of cuboids. 
They fit cuboids at multiple levels and use a cuboid selection module to obtain the optimal abstraction.
It is a coarse-to-fine approach that considers first cuboids at the coarsest level and subdivides them if necessary.
Contrary, we propose a fine-to-coarse approach where we start with a large number of primitives and discard or merge redundant ones. 
\cite{CAS_Yang} propose to couple shape abstraction together with segmentation.
They introduce two decoders, where the first one learns the abstraction and the second one learns to associate cuboids with the nearest points and couple them by their loss formulation.
Furthermore, they observe that shape abstraction methods suffer often from degenerate primitives reconstructing only the surface of the shape and not its volume.
To resolve this problem \cite{CAS_Yang} introduce a loss formulation incorporating surface normals of the input shape as well as the normals of the cuboid surface.
We propose to use a simpler approach to resolve this issue and do not consider only the surface of the shape but also enforce volume preservation of the shape with our loss directly.

\subsubsection{Superquadrics} 
An extension to cuboids provides the class of superquadrics. 
By only incorporating two more parameters they are able to extend the primitive types by ellipsoids, spheres, cylinders, octahedra and interpolations in between. 
They offer the advantage of a continuous, differentiable parametrized representation, but are rather challenging to edit intuitively or sample uniformly \cite{EDS_Maurizio, SSQN_Ferreira}.
Nevertheless multiple methods \cite{SQR_Paschalidou, HSQ_Paschalidou, DSQ_Li} used them to abstract 3D shapes improving the reconstruction quality while using a similar or lower number of primitives than previous work.

\subsubsection{Deformable part templates} 
The most general class of primitives can be characterized by deformable part templates.
They learn to deform a simple sphere mesh \cite{NeuralParts_Paschalidou}, cuboids and cylinders \cite{DPF-Net_Shuai} or a learned implicit function \cite{DAE-Net_Chen}.
The methods mostly rely on a branched autoencoder architecture \cite{BAE-Net_Chen} where the decoder consists of multiple sub-networks which learn a representation of one commonly occurring part each.
Thus the number of learned parts is determined a-priori by the selected number of decoders.
Even though they have the most representative power per primitive and thus the best reconstruction quality, the final reconstructed parts cannot be easily processed or modified by the user as each vertex position of the template mesh is deformed separately by the neural network.


\section{CUBOID SHAPE ABSTRACTION}
\label{sec:learning_cuboid_abstraction}

In the following section we introduce our learning-based fine-to-coarse shape abstraction approach. 
Given a surface point cloud $X_S$ and a volume point cloud $X_V$ our goal is to train a neural network that predicts a set of cuboid primitives $\{C_i\}_{i=1,...,M}$ which approximates the corresponding 3D shape as closely as possible.
Each cuboid is identified by its unique index $m$ and parametrized by a rotation represented as an unit quaternion $r \in \mathbb{R}^4$, a translation $t \in \mathbb{R}^3$, scaling $s \in \mathbb{R}^3$ as well its existence probability $\gamma \in [0, 1]$.
During inference time we only require  a surface point cloud $X_S$ as input to our network.


\subsection{Architecture Design}
\label{sec:architecture_design}

\begin{figure}
  \centering
  \includegraphics[width=0.48\textwidth]{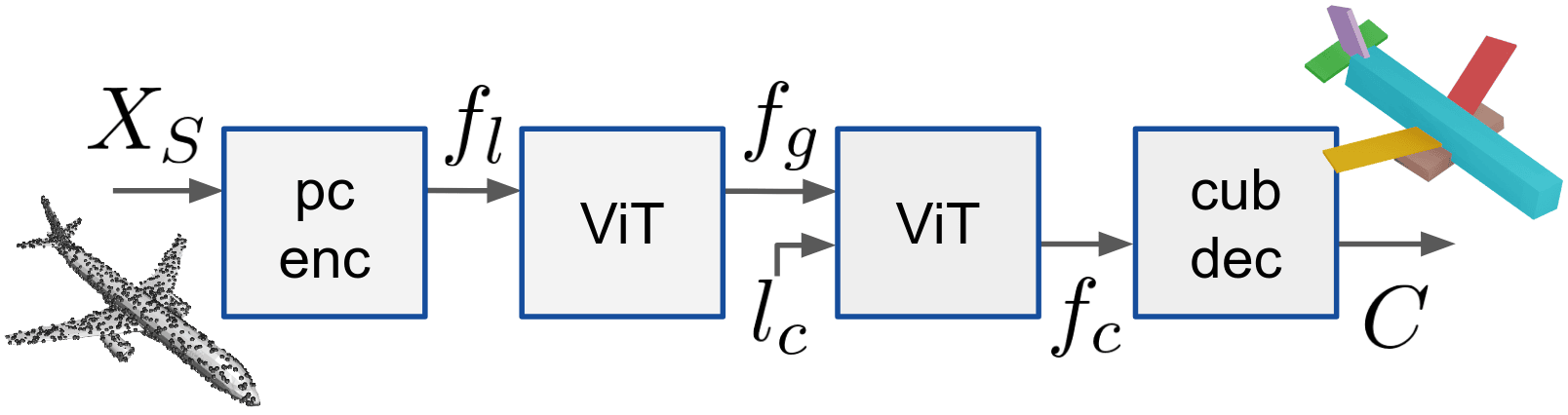}
  \caption{Architecture overview: a surface point cloud $X_S$ is processed by a point cloud encoder to obtain local features $f_l$. These features exchange information between each other throught a ViT to obtain global features $f_g$. Next, they are concatenated with learnable cuboid latents $l_c$ and passed throught a second ViT to obtain cuboid features $f_c$. These features are decoded into a set of cuboid primitives $C$ which approximate $X_S$.}
  \label{fig:architecture}
\end{figure}

Our method uses a simple and flexible architecture. It consists of a point cloud encoder, two transformer blocks and finally a cuboid prediction head (cmp. \Cref{fig:architecture}).
Due to the nature of transformers operating on sets of tokens it allows us to use a dynamic number of latent features, which can either represent the input shape or the cuboid primitives.
With this architecture we do not need to specify a fixed number of primitives a-priori like \cite{VP_Tulsiani, HCA_Sun, CAS_Yang, SQR_Paschalidou, DSQ_Li, BAE-Net_Chen} and can even discard redundant primitives to speedup the training and inference times.
The encoder part of the architecture is inspired by \cite{3DILG_Zhang} and the cuboid decoder by \cite{CAS_Yang}.

\subsubsection{Local shape features}

The input to our network is a point cloud $X_S \in \mathbb{R}^{2048 \times 3}$ sampled from a surface mesh.
First, we select a subset of $N$ points using furthest point sampling. 
Next, for each selected sample we compute the $K$ nearest neighbors to extract a point cloud patch, which is finally embedded into the latent space using a small PointNet to obtain a local shape feature $f_{l} \in \mathbb{R}^{N \times 128}$.
We use a relatively low value of $N = 128$ and $K = 32$ in all our experiments.
Please refer to \cite{3DILG_Zhang} for more details on the point cloud encoder.

\subsubsection{Global shape features}

To obtain global shape features $f_{g} \in \mathbb{R}^{N \times 128}$ we pass $f_{l}$ through a small Vision Transformer (ViT) \cite{ViT_Dosovitskiy}.
The self-attention modules exchange information between local tokens to extract global context about the shape.

\subsubsection{Abstract cuboid features}

We compute the abstract cuboid features $f_{c} \in \mathbb{R}^{N \times 128}$ in an auto-decoder fashion conditioned on $f_{g}$.
At the beginning of the training we define learnable cuboid latents $l_c \in \mathbb{R}^{M \times 128}$ as random vectors and optimize them with the weights of our network jointly.
Next, we concatenate $[f_g, l_c]$ and pass it through the second ViT block to allow each cuboid token to attend to the global tokens.
To additionally allow exchange of information among the cuboid tokens we use self-attention layers.
From the output token sequence we discard all tokens at the positions of $f_g$ and the remaining tokens form the abstract cuboid features $f_c$.

We empirically obtained better results using two small ViTs computing $f_{g}$ first, rather than concatenating $[f_l, l_c]$ directly and using one large ViT.
Details are analyzed in \Cref{sec:ablation_study_and_analysis}.

\subsubsection{Predicted cuboid primitives}
Each $f_c^m$ corresponds directly to a cuboid primitive $C_m$. 
The parameters $p_m = [r_m, t_m, s_m, \gamma_m]$ are computed by fully connected layers each of which share their parameters across all cuboids.
As we use the same cuboid prediction head, we refer to \cite{CAS_Yang} for more details.


\subsection{Loss Formulation}
\label{sec:loss_formulation}

We formulate the training of our network as a gradual transition between a shape reconstruction task and a shape abstraction task by continually reducing the number of used primitives.
In the beginning we fit an excessive number of cuboids to the input shape which allows to tightly approximate the target shape with a minimal reconstruction error.
At the end of the training process the number of primitives has been drastically reduced, thus the remaining cuboids form an abstraction of the shape.
The reconstruction loss $\mathcal{L}_{rec}$ tries to reconstruct the shape with the available primitives as precisely as possible, while the abstraction loss $\mathcal{L}_{abs}$ modulates the descriptive power of our representation by continually decreasing the number of available primitives.
Both losses are coupled by the existence probability $\gamma$ of each cuboid.
This fine-to-coarse approach allows us to overcome local minima introduced through suboptimal initialization of the neural network in contrast to directly optimizing a low, but fixed number of primitives.
Our loss formulation is geometrically motivated and thus allows to extract structural information based purely on geometric features. 

\subsubsection{Reconstruction loss}

We use a Chamfer distance based loss formulation as our reconstruction loss. 
For the sake of completeness, we recall \Cref{eq:paschalidau_1} - \eqref{eq:paschalidau_4} from \cite{SQR_Paschalidou}.
To compute the loss from the primitives to the target $P = \{p_j\}_{j=1,...J} \rightarrow X = \{x_i\}_{i=1,...,I}$, we compute the closest neighbor $x$ of $p_j^m$ in $P$ and its distance $\Delta_j^m$.
Next, we average the distance across all points of all cuboids as
\begin{equation}
    \mathcal{L}_{P \rightarrow X}(P, X) = \sum^M_{m=1} \frac{S_m}{J} \sum^J_{j=1}  \Delta^m_j \gamma_m 
\label{eq:paschalidau_1}
\end{equation}
with
\begin{equation}
    \Delta_j^m = \min_{x \in X} \lVert x - p_j^m \rVert_2
\label{eq:paschalidau_2}
\end{equation}
and weight its influence by its size $S_m$.
Similarly, the loss from the target point cloud to the primitives $X \rightarrow P$ is computed as
\begin{equation}
    \mathcal{L}_{X \rightarrow P}(X, P) = \sum_{i = 1}^I \sum_{m=1}^M \Delta_i^m \gamma_m \prod^{m-1}_{\bar{m}=1} (1-\gamma_{\bar{m}}),
\label{eq:paschalidau_3}
\end{equation}
where $\gamma_{\bar{m}}$ is a shorthand notation to denote that there exists a primitive closer than primitive $m$ and
\begin{equation}
    \Delta_i^m = \min_{p \in P^m} \lVert x_i - p \rVert_2.
\label{eq:paschalidau_4}
\end{equation}
Note, that the distances $\Delta^m$ are weighted with the existence probability $\gamma_m$ as only primitives that exist should contribute to the reconstruction error.
 \\

In contrast to \cite{SQR_Paschalidou, CAS_Yang, HCA_Sun, DSQ_Li} we do not consider only the surface of the shape but also aim to preserve its spatial volume explicitly.
Given that the above loss formulation depends only on a target point cloud $X = \{x_i\}$ and a sampled point cloud $P^m = \{p_j^m\}$ of each predicted primitive, we extend it to model volume preservation.
Furthermore, we do not use a fixed number of samples per primitive as \cite{SQR_Paschalidou}, but increase it anti-proportional to the number of remaining primitives.
This allows us to keep a constant sample density throughout the training while dynamically adjusting the primitive count.

To compute the surface loss $\mathcal{L}_{surf}$ we sample the surface of each primitive uniformly to obtain $P_S$ and compare it with the target surface point cloud $X_S$:
\begin{equation}
    \mathcal{L}_{surf} = \alpha \mathcal{L}_{P \rightarrow X}(P_S, X_S) + \beta \mathcal{L}_{X \rightarrow P}(X_S, P_S).
\end{equation}

To compute the volume loss $\mathcal{L}_{vol}$ we sample the volume of each primitive uniformly to obtain $P_V$ and compare it with the target volume point cloud $X_V$.
This loss is defined as
\begin{equation}
    \mathcal{L}_{vol} = \alpha \mathcal{L}_{P \rightarrow X}(P_V, X_V) + \beta \mathcal{L}_{X \rightarrow P}(X_V, P_V).
\end{equation}
We weight both losses with default values $\alpha = 1.2$ and $\beta = 0.8$ used in \cite{SQR_Paschalidou}, respectively.
Furthermore, we compute the box size $S_m$ to weight the loss either as the surface area or its volume.
We use the same number of samples for the surface and volume and weight both losses with $\lambda_{vol}$ and $\lambda_{surf}$.
The resulting reconstruction loss is given by
\begin{equation}
    \mathcal{L}_{rec} = \lambda_{vol} \mathcal{L}_{vol} + \lambda_{surf} \mathcal{L}_{surf}.
\end{equation}

\subsubsection{Abstraction loss}

During training we optimize the abstraction loss $\mathcal{L}_{abs}$ starting with a high number of primitives and progressively reduce their count.
Pruning a single primitive affects some portions of the shape and will force other neighboring primitives to approximate the affected region. 

To control the number of primitives we define a target function $\Gamma(\phi) \in \mathbb{R}$.
In the simplest case it is formulated as a linear interpolation
\begin{equation}
    \Gamma(\phi) = \kappa_{max} - (\kappa_{max} - \kappa_{min}) \cdot \phi
\end{equation}
between the maximum $\kappa_{max}$ and minimum $\kappa_{min}$ number of primitives depending on the progress of the training $\phi = \frac{epoch_{cur}}{epoch_{max}}$.
We analyze other target functions (cmp. \Cref{eq:target_fuction}) in \Cref{sec:ablation_study_and_analysis}.
To follow $\Gamma(\phi)$ closely we use the binary cross entropy as the primitive abstraction loss
\begin{equation}
    \mathcal{L}_{abs} = \sum_i - t_i \log \gamma_i - (1 - t_i) \log (1 - \gamma_i),
\end{equation}
where $t_i = 1$ for $\Gamma(\phi)$ cuboids with the largest existence probability $\gamma_i$, otherwise $t_i = 0$. \\

\subsubsection{Final loss.}
The final loss used to optimize the cuboid prediction network is a sum of the previously defined losses
\begin{equation}
    \mathcal{L} = \mathcal{L}_{rec} + \lambda_{abs} \mathcal{L}_{abs}.
\end{equation}

Additionally, we restrict the number of processed cuboids by masking out $l_c^m$ with $\gamma_c^m < 0.01$ in every prediction at the end of each epoch. 
This speeds up the training and inference times.

\subsection{Merging primitives.}
During training the reconstruction loss $\mathcal{L}_{rec}$ approximates the target shape, while the abstraction loss $\mathcal{L}_{abs}$ reduces the number of used primitives.
These are contradicting objectives. 
As more primitives allow for a better reconstruction quality, $\mathcal{L}_{rec}$ is prone to maximizing the number of used primitives and pursues a different optimum than $\mathcal{L}_{abs}$.
Especially, when two cuboids try to approximate the shape of a cylinder they tend to overlap and rotate by 45° along the rotation symmetry axis, e.g. the corpus of an airplane or a round tabletop of a table. 
The initial approximation can often be simplified by merging overlapping cuboids to obtain structurally equivalent representation using less primitives without a significant degradation in reconstruction quality.
We merge cuboids at inference time as a post-processing step.

Cuboids that overlap completely or in part can be detected by computing the ratio $V^{ratio}$ of the sum of their volumes and the volume of their bounding box.
The ratio $V^{ratio} > 1$ if they are overlapping along an axis, 
or  $V^{ratio} \approx 1$ if they are touching each other along one face.
In both cases they can be merged into a single cuboid.

In the following we consider only cuboids with existence $\gamma > 0.5$. 
To check which cuboids should be merged we first compute the oriented bounding box $BB_{ij}$ for every existing cuboid pair $C_i$, $C_j$.
Next, we compute the volume ratio 
\begin{equation}
    V^{ratio}_{ij} = \frac{V(C_i) + V(C_j)}{V(BB_{ij})}
\end{equation}
and select the pair with the largest ratio.
If the ratio is above a threshold $V^{ratio}_{ij} > \theta_{merge}$, we substitute the parameters of $C_i$ with the parameters of $BB_{ij}$ and set the existence probability $\hat{\gamma}_i = 1, \hat{\gamma}_j = 0$.
We repeat this process, until all valid cuboids are merged, also considering the updated $C_i$ as a candidate.
To counteract inaccurate alignment of thin cuboids, we estimate the volume ratio $V^{ratio}$ using cuboids with an increased size by a small fraction $\epsilon_{size}$.
In this fashion we transform the initial set of cuboids $C$ into a merged set of cuboids $\hat{C}$ representing our cuboid shape abstraction.
We keep track of the ancestors of $\hat{C}$ by using a binary encoding that defines which cuboids of the initial predicted set $C$ were merged together.


\section{EXPERIMENTAL EVALUATION}
\label{sec:experimental_evaluation}

\textit{Datasets.} 
We use common datasets to benchmark cuboid shape abstraction methods.
For all experiments we use three categories from the ShapeNet dataset \cite{shapenet_chang}: plane (3640), chair (5929), table (7555).
Furthermore, we use a subset of the DFAUST dataset \cite{dfaust_bogo} including humans (4179) for the reconstruction benchmark.
We use only every tenth frame from DFAUST to exclude repetitive poses and match the size of the ShapeNet classes.
Similar to previous data, we split this dataset randomly with a ratio of 4:1 between the train and the test set.

\textit{Reference methods.}
We compare our approach with two state-of-the-art cuboid-based shape abstraction methods: \cite{HCA_Sun} (HCA) and \cite{CAS_Yang} (CAS).
Furthermore, we compare with \cite{DPF-Net_Shuai} ($\text{DPF}_{PPM}$) using their primitive prediction module to compute cuboid primitives.
Whenever applicable we use the published model weights, otherwise we train the method five times for every category using the standard hyper-parameters and report the best performing run in terms of Chamfer distance.
For $\text{DPF}_{PPM}$ we reduce the batch size to $8$ to train on our GPUs.

\textit{Hyper-Parameters.}
We train all models on a single RTX 2080Ti with 11GB of VRAM.
If not otherwise noted, our models are trained for $1000$ epochs with a batch size of $16$ using AdamW with a learning rate of $1e\text{-}3$ and a cosine annealing scheduler with warm-up for the first $5$ epochs.
We set $\lambda_{vol} = 1e1$, $\lambda_{surf}=1e0$ and $\lambda_{abs} = 1e\text{-}3$.
Our model uses two ViTs with $6$ layers and an embedding size of $128$.
We decided to use a half-cosine scheme for the cuboid target function $\Gamma(\phi)$ with $\kappa_{max} = 128$ and $\kappa_{min} = 7$ to closely constrain the final number of cuboids to previous work.
For chairs we use $\kappa_{min} = 10$.
The merge threshold $\theta_{merge}$ is set to $1.2, 1.4, 1.0$ and $1.0$ for the plane, chair, table and human class, respectively.


\subsection{Structured Shape Reconstruction}
\label{sec:structured_shape_reconstruction}

For the reconstruction benchmark we report three metrics to evaluate the quality of the abstraction. 
Chamfer distance (CD) is used to measure the surface reconstruction quality. 
We uniformly sample $2048$ points on the ground-truth shape surface and the predicted cuboid surface.
To measure how well the abstraction does approximate the spatial volume of the original shape we compute Intersection over Union (IoU) on a uniformly sampled grid with the size of $128^3$.
Furthermore, we report the mean number of used cuboid primitives (Num).
Using an extensive number of primitives allows to easily adhere to the shape surface and volume as every primitive needs to approximate only a small portion of the shape. 
The reconstruction becomes much more challenging with a limited number of primitives.
We regress our network to limit the mean number of cuboids within the typical range of previous work.
All metrics are reported in \Cref{tab:reconstruction}.

We see that our method achieves the best scores for CD and IoU on almost all classes while using least or second least amount of cuboids on average.
\cite{HCA_Sun} are able to abstract the data with a lower number of primitives at the cost of a sub-optimal reconstruction performance. 
In cases where we match or undercut their number of primitives, we are able to obtain a better reconstruction quality measured in terms of CD and IoU.
Comparing with \cite{CAS_Yang} we obtain a similar CD using substantially less primitives. 
In all cases we are able to significantly improve upon the volume preservation.
This can be attributed to the chosen loss function, which in \cite{CAS_Yang} focuses on the reconstruction of the surface and their normal direction, but does not account for the shape volume. 
Starting the reconstruction with an extensive amount of primitives allows us to capture fine details of the shape first and thus more easily overcome a suboptimal initialization leading to local minima.
Preserving those features during the optimization of the network is easier than discovering them with a limited number of primitives.

We provide qualitative samples in \Cref{fig:qualitative_airplane_chair} and \Cref{fig:qualitative_table_human}.
Our reconstruction adheres faithfully to the geometry of the ground truth shape.
Especially, for more complex shapes like fighter jets, office chairs, and one legged tables we are able to reconstruct the shape more accurately than the state-of-the-art.
This is directly reflected by our CD and IoU scores.
\cite{CAS_Yang} try to overcome this issue by over-segmenting these challenging shapes.
This leads to degenerated primitives, e.g. chair and table legs or barely recognizable human feet (cmp. \Cref{fig:qualitative_table_human}).
Inspecting the table and human class more closely, we observe that other methods \cite{HCA_Sun, CAS_Yang} struggle with joints between distinct primitives.
The cuboids are often disconnected.
By directly modeling volume preservation in our volume loss function $\mathcal{L}_{vol}$ we are able to overcome this problem.
We observe that \cite{DPF-Net_Shuai} struggle to extract a meaningful abstraction and over-partition the shapes. 
We attribute this to a reduced batch size during the training of their model.
Unfortunately, no pre-trained model weights are released.

\begin{table}[]
    \centering
    \newcommand{\graycell}{\cellcolor[HTML]{EFEFEF}}
\newcommand{\redcell}{\cellcolor[HTML]{FAD1D0}}

\begin{tabular}{cc|ccccc}
\toprule
&        & \textbf{plane}         & \textbf{chair}              & \textbf{table}              & \textbf{human}    \\
\midrule

\multirow{4}{*}{\rotatebox{90}{\textbf{Num↓}}}         
& HCA                   & 7.14                   & \textbf{6.38}               & \textbf{4.44}               & -                 \\
& CAS                   & 10.64                  & 9.77                        & 7.64                        & 7.50              \\
& $\text{DPF}_{PPM}$    & 15.93                  & 11.42                       & 9.45                        & 16.00             \\
& Ours \graycell        & \textbf{6.03} \graycell & 8.37 \graycell             & 5.67 \graycell              & \textbf{6.02} \graycell    \\

\midrule

\multirow{4}{*}{\rotatebox{90}{\textbf{CD↓}}}            
& HCA                   & 0.040                 & 0.054                        & 0.058                       & -                 \\
& CAS                   & 0.029                 & \textbf{0.036}               & 0.044                       & \textbf{0.029}    \\
& $\text{DPF}_{PPM}$    & 0.035                 & 0.056                        & 0.056                       & 0.040             \\
& Ours \graycell        & \textbf{0.026} \graycell & \textbf{0.036} \graycell  & \textbf{0.035} \graycell    & 0.032 \graycell   \\

\midrule

\multirow{4}{*}{\rotatebox{90}{\textbf{IoU\%↑}}}            
& HCA                   & 36.2                  & 39.3                        & 29.7                        & -                 \\
& CAS                   & 38.1                  & 49.5                        & 37.1                        & 32.0              \\
& $\text{DPF}_{PPM}$    & 55.4                  & 44.3                        & 42.3                        & 44.4              \\
& Ours \graycell        & \textbf{56.0} \graycell  & \textbf{54.9} \graycell  & \textbf{ 45.1} \graycell    & \textbf{58.8} \graycell    \\
\bottomrule
\end{tabular}

    \caption{Quantitative evaluation of shape reconstruction on the test set. We report Chamfer distance (CD), Intersection of Union (IoU) and mean number of primitives (Num) . Best highlighted in bold.}
    \label{tab:reconstruction}
\end{table}


\subsection{Shape Co-Segmentation}
\label{sec:shape_co-segmentation}

To demonstrate the structural consistency of the generated abstractions with semantic annotations we evaluate our method on the co-segmentation task defined in \cite{BSP_Chen}. 
We compute the mean per-label Accuracy (mAcc) and Intersection over Union (mIoU) as the evaluation criterion, using 4 parts for plane (body, tail, wing, engine), 4 parts for chair (back, seat, leg, arm) and 2 parts for table (top, support).
To obtain a semantic mapping from class labels to primitive indices, we use all annotated shapes from the training set and assign each primitive to a label via a voting scheme.
Next, we propagate the primitive class labels to point clouds of the test set via nearest neighbor and evaluate the segmentation quality.
Results are reported in \Cref{tab:co_seg}.

Our method performs best on the table, second best on the plane and third on the chair category.
\cite{CAS_Yang} model their abstraction through a segmentation module and use a higher number of cuboids, allowing them to segment the shape with finer details.
The same holds true for \cite{DPF-Net_Shuai}.
This is most obvious in comparison to \cite{HCA_Sun} who use the least number of primitives resulting in poor performance on this benchmark.
Even thought we use a lower number of primitives than \cite{CAS_Yang} our scores are similar.
The reported metrics demonstrate that our model is able to recover geometric parts aligned with semantics of the shapes.

\begin{table}[]
    \centering
    \newcommand{\graycell}{\cellcolor[HTML]{EFEFEF}}
\newcommand{\redcell}{\cellcolor[HTML]{FAD1D0}}

\begin{tabular}{c|ccc|ccc}
\toprule
\multirow{2}{*}{} & \multicolumn{3}{c}{\textbf{mAcc\%↑}}         & \multicolumn{3}{c}{\textbf{mIoU\%↑}}    \\
                                 & \textbf{plane}  & \textbf{chair} & \textbf{table}     & \textbf{plane}  & \textbf{chair} & \textbf{table}\\
\midrule

HCA               & 61.0  & 63.3 & 52.2  & 41.4  & 48.4 & 36.3  \\
CAS               & 84.5  & \textbf{91.3} & 92.6  & \textbf{74.9}  & \textbf{84.8} & 85.9  \\
$\text{DPF}_{PPM}$& 78.9  & 83.4    & 86.6  & 62.9   & 74.7    & 76.5     \\

Ours \graycell    & \textbf{85.5} \graycell & 82.0 \graycell & \textbf{93.6} \graycell & 66.3  \graycell & 73.7 \graycell & \textbf{87.1} \graycell \\
\bottomrule
\end{tabular}

    \caption{Quantitative evaluation of co-segmentation performance on the test set. We report mean per-label Accuracy (mAcc) and Intersection over Union (mIoU). Best highlighted in bold.}
    \label{tab:co_seg}
\end{table}


\subsection{Shape Clustering and Retrieval}
\label{sec:shape_clustering_and_retrieval}

\begin{figure}
    \centering
    \resizebox{0.475\linewidth}{!}{\begin{tabular}{ccccc}

    \begin{overpic}[]{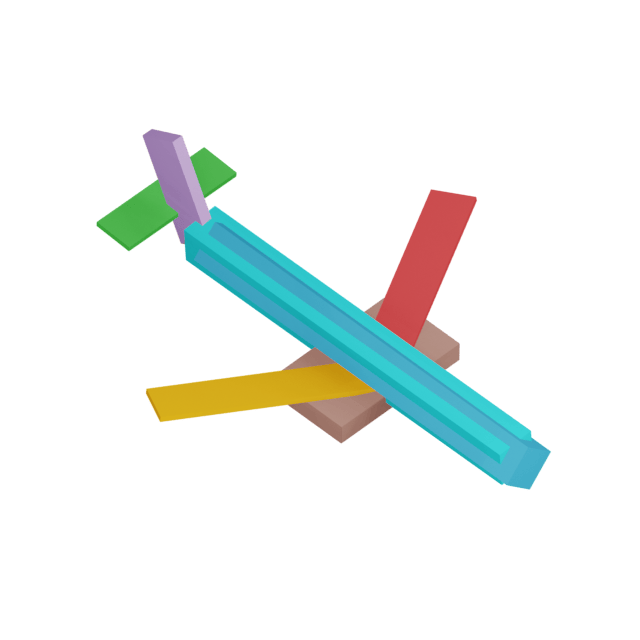}
        \put(-25,-5){\includegraphics[scale=0.8]{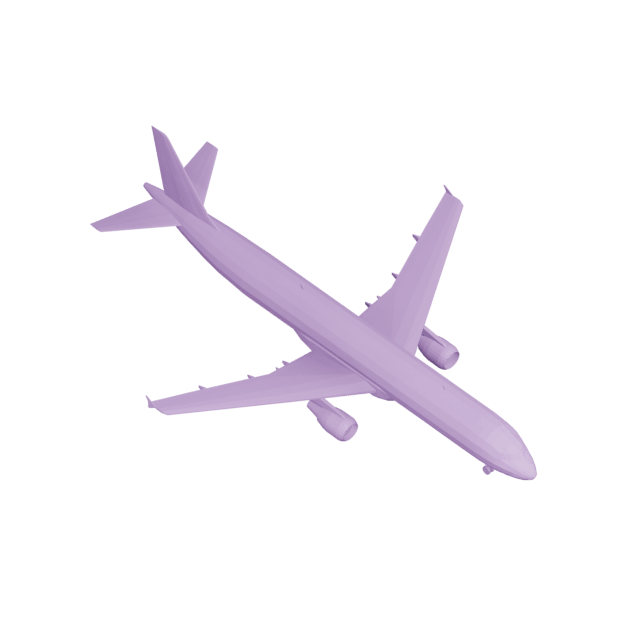}}
    \end{overpic} &  
    \begin{overpic}[]{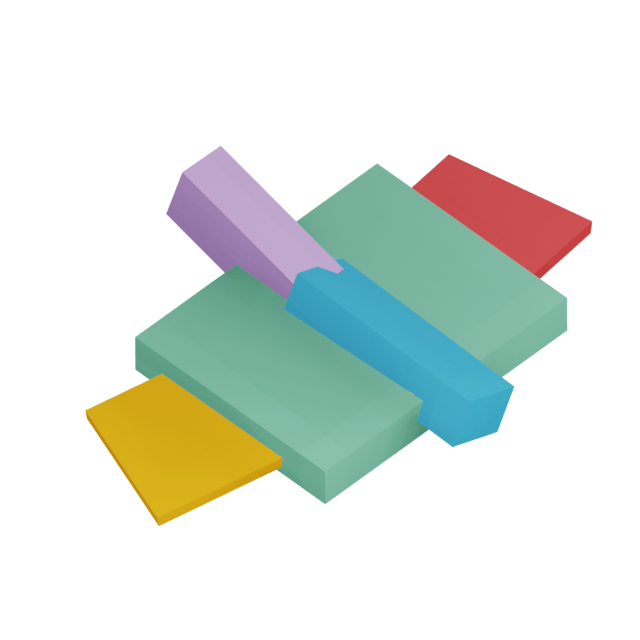}
        \put(-25, -5){\includegraphics[scale=0.8]{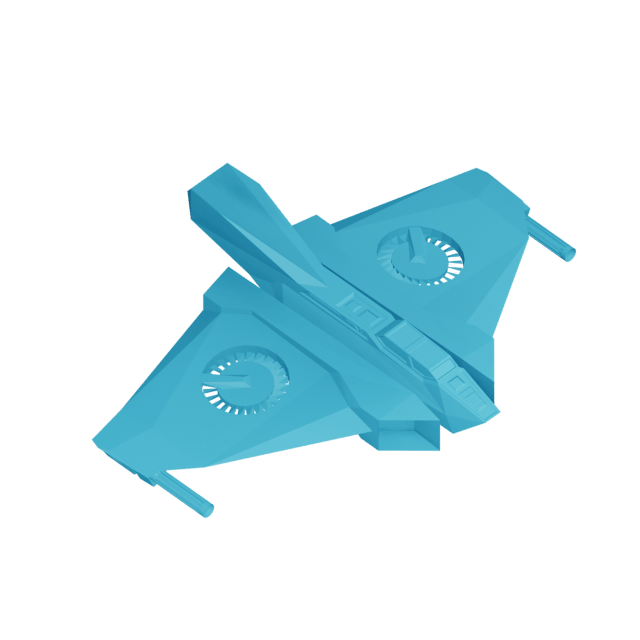}}
    \end{overpic} &  
    \begin{overpic}[]{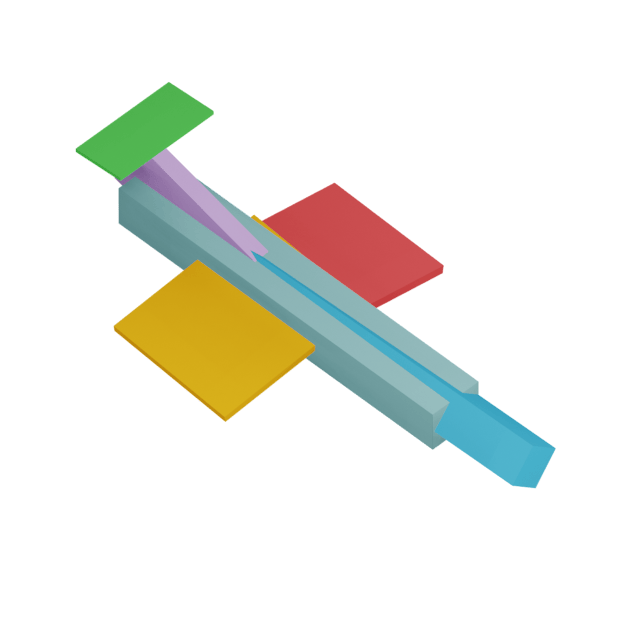}
        \put(-25,-5){\includegraphics[scale=0.8]{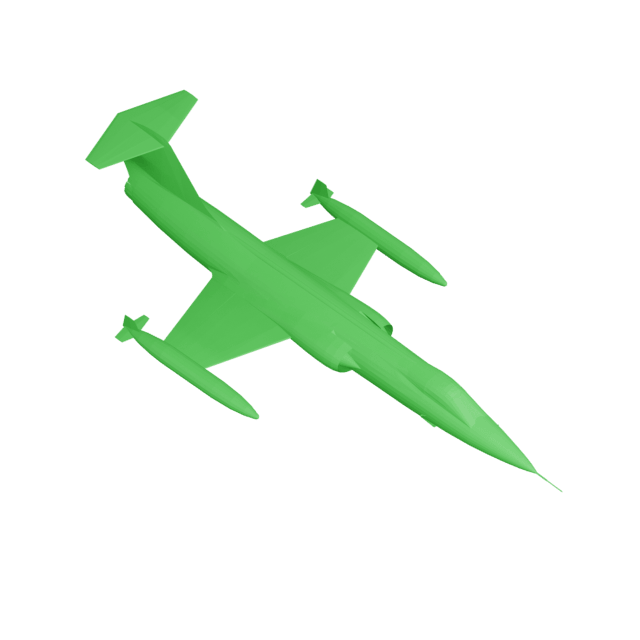}}
    \end{overpic} &  
    \begin{overpic}[]{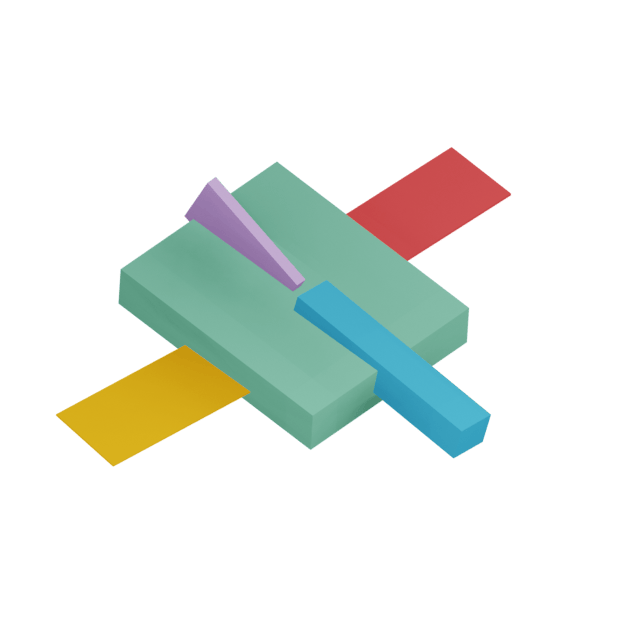}
        \put(-25,-5){\includegraphics[scale=0.8]{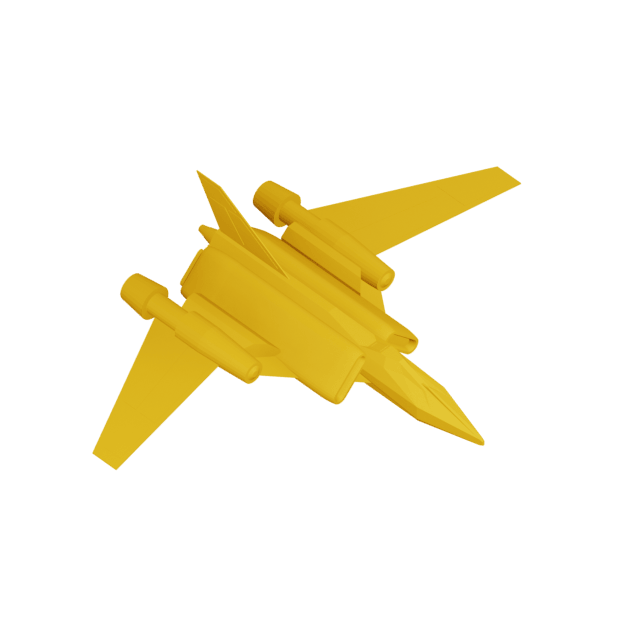}}
    \end{overpic} & 
    \begin{overpic}[]{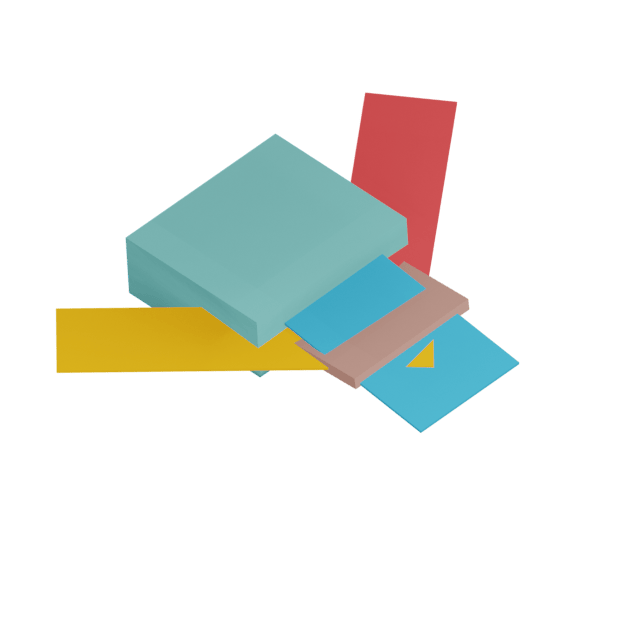}
        \put(-25, -5){\includegraphics[scale=0.8]{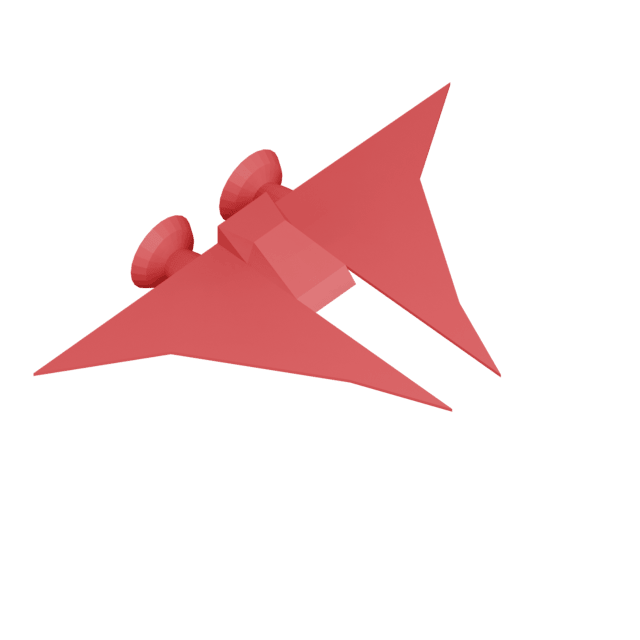}}
    \end{overpic} \\

    \begin{overpic}[]{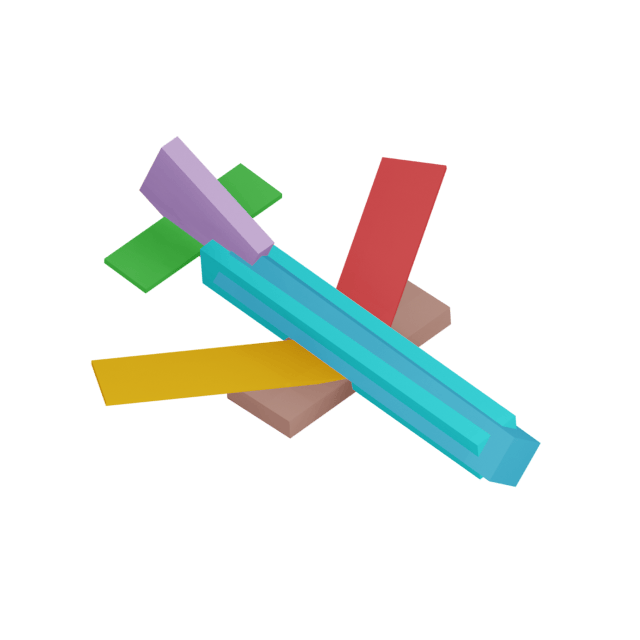}
        \put(-25,-5){\includegraphics[scale=0.8]{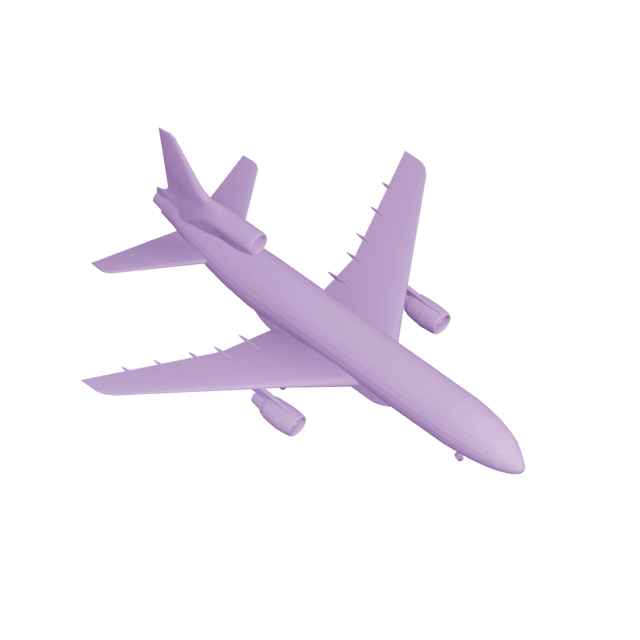}}
    \end{overpic} &  
    \begin{overpic}[]{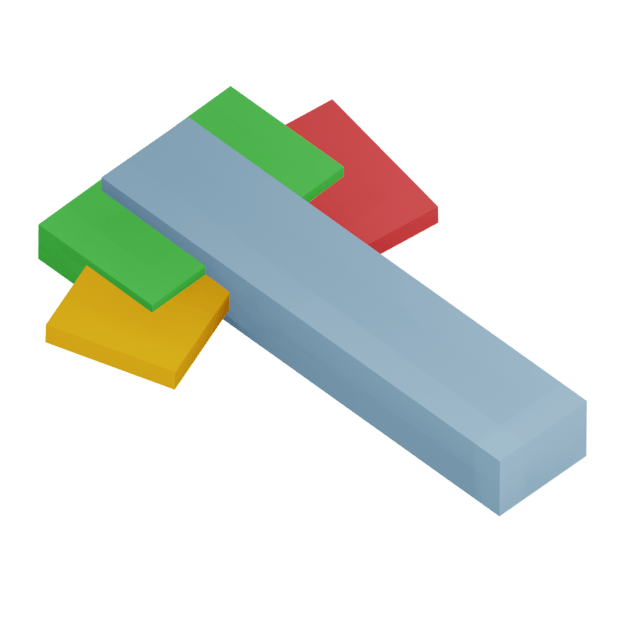}
        \put(-25,-5){\includegraphics[scale=0.8]{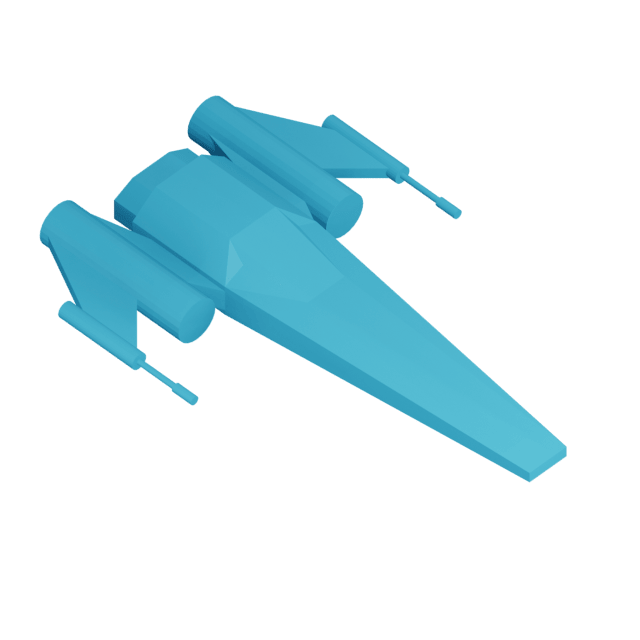}}
    \end{overpic}&  
    \begin{overpic}[]{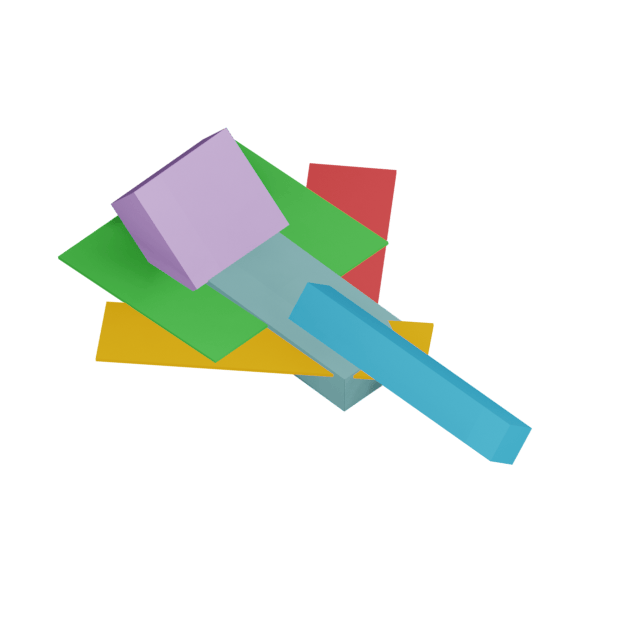}
        \put(-25,-5){\includegraphics[scale=0.8]{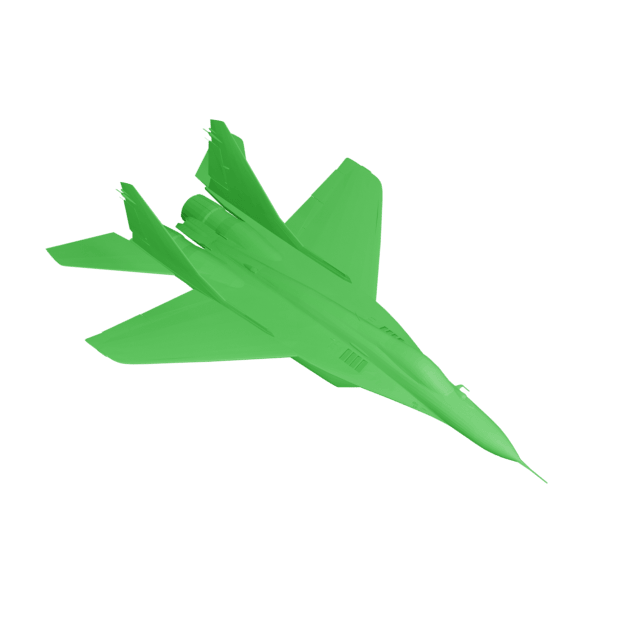}}
    \end{overpic}&  
    \begin{overpic}[]{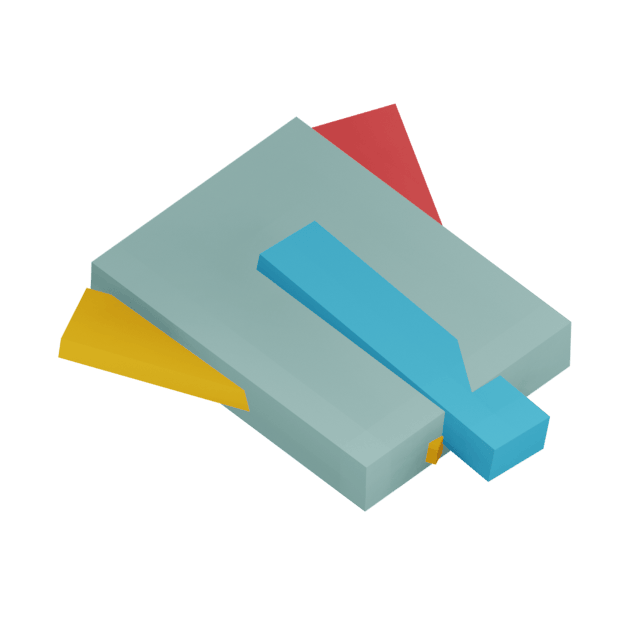}
        \put(-25,-5){\includegraphics[scale=0.8]{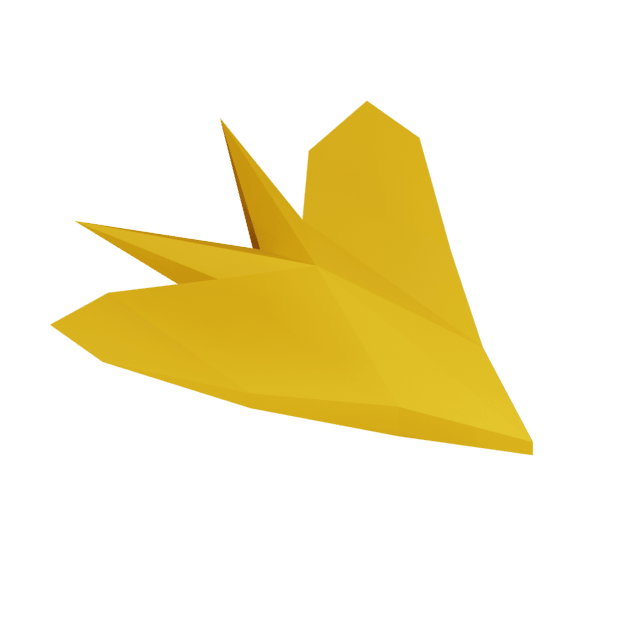}}
    \end{overpic}& 
    \begin{overpic}[]{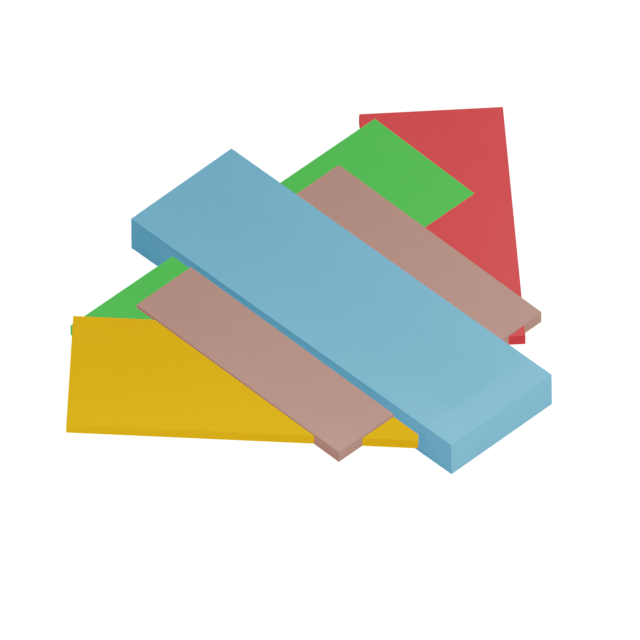}
        \put(-25,-5){\includegraphics[scale=0.8]{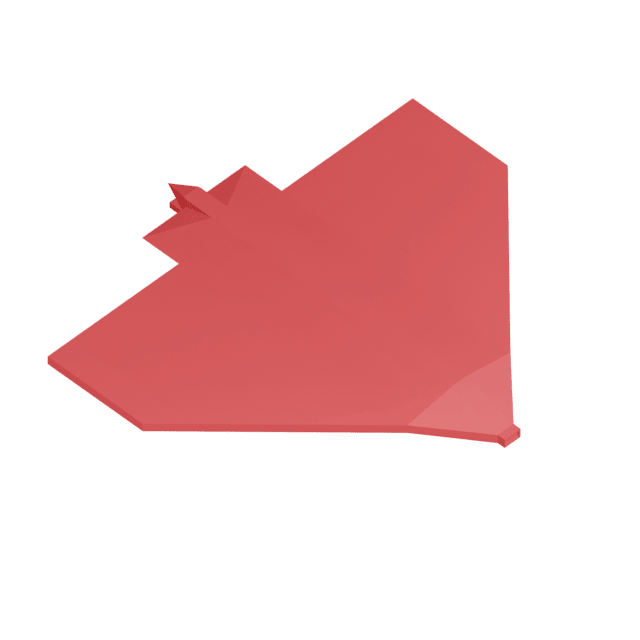}}
    \end{overpic} \\
    
    \begin{overpic}[]{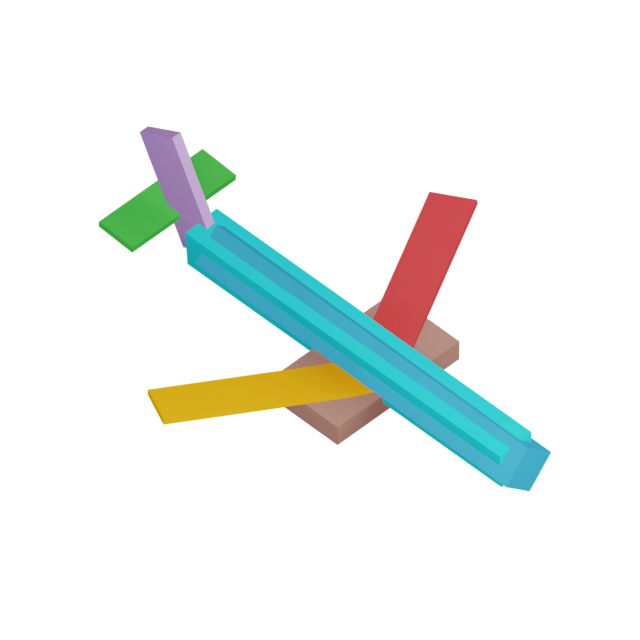}
        \put(-25,-5){\includegraphics[scale=0.8]{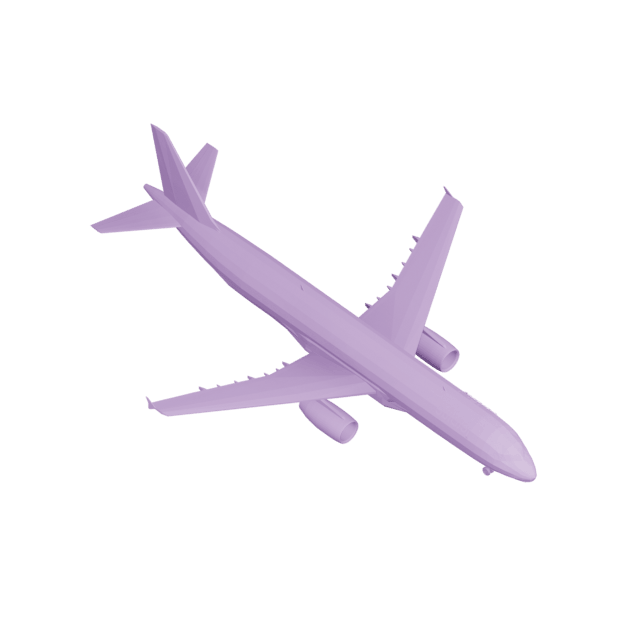}}
    \end{overpic} &  
    \begin{overpic}[]{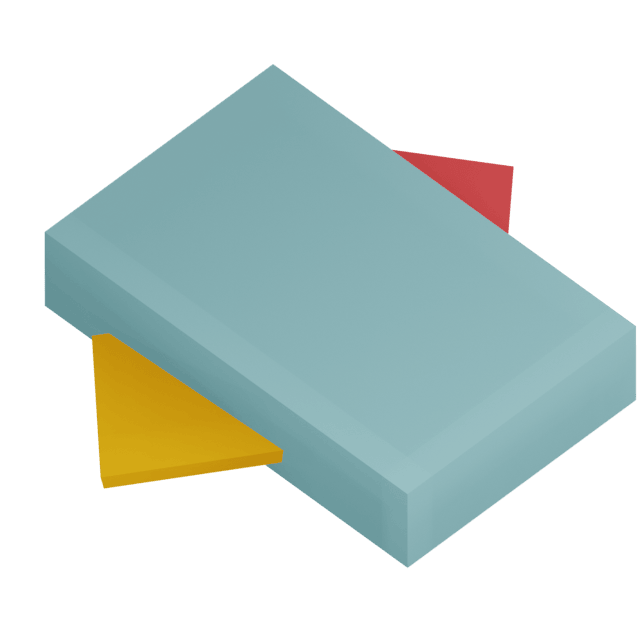}
        \put(-25,-5){\includegraphics[scale=0.8]{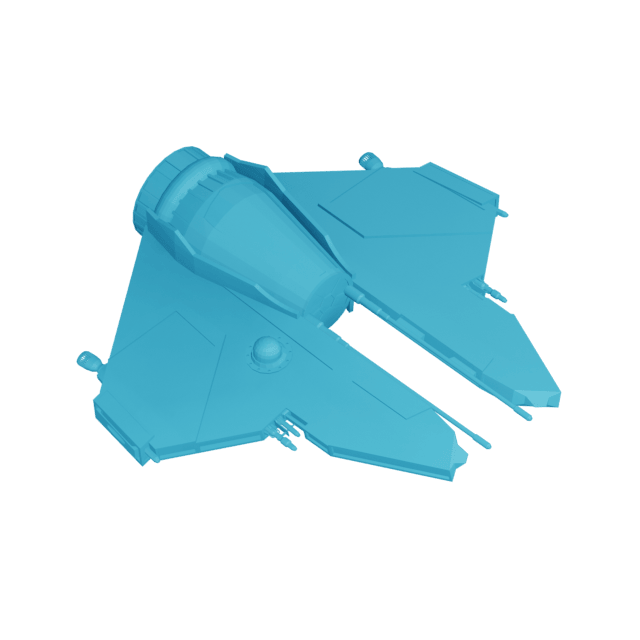}}
    \end{overpic} & 
    \begin{overpic}[]{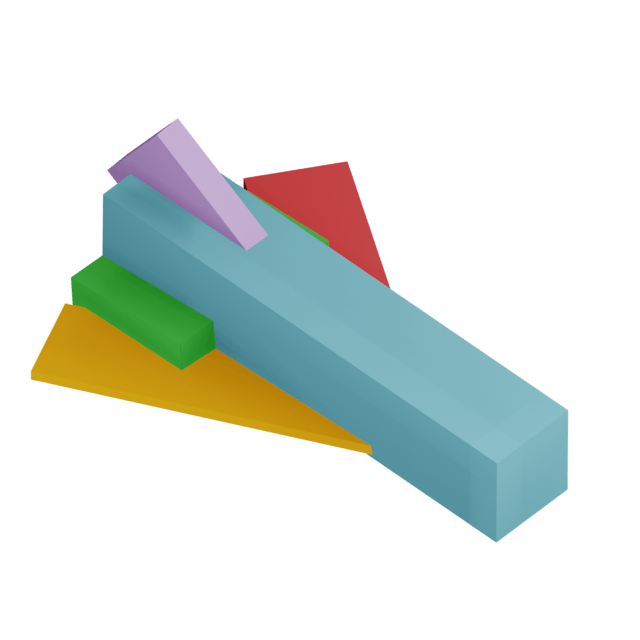}
        \put(-25,-5){\includegraphics[scale=0.8]{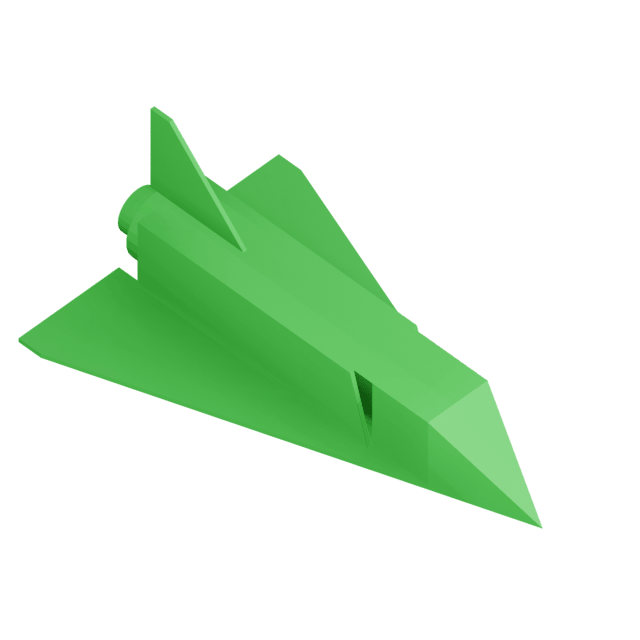}}
    \end{overpic} &  
    \begin{overpic}[]{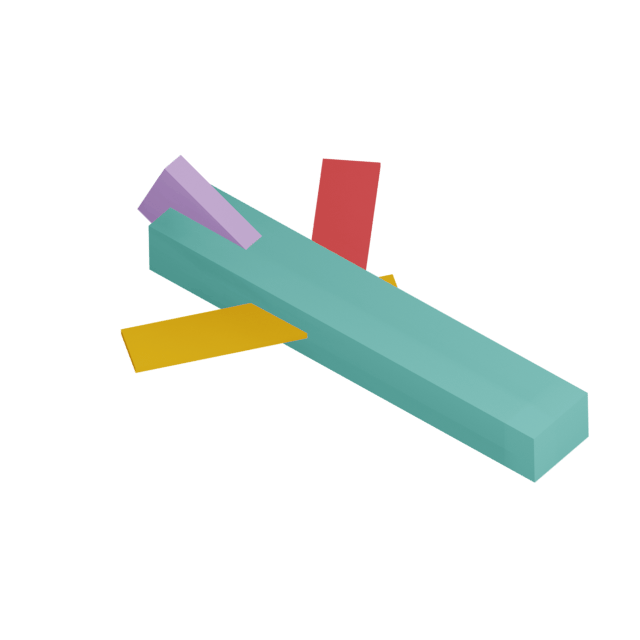}
        \put(-25,-5){\includegraphics[scale=0.8]{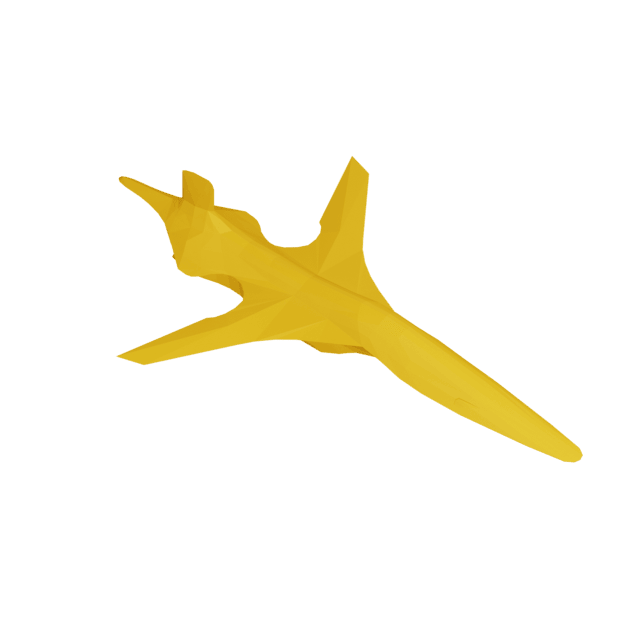}}
    \end{overpic} & 
    \begin{overpic}[]{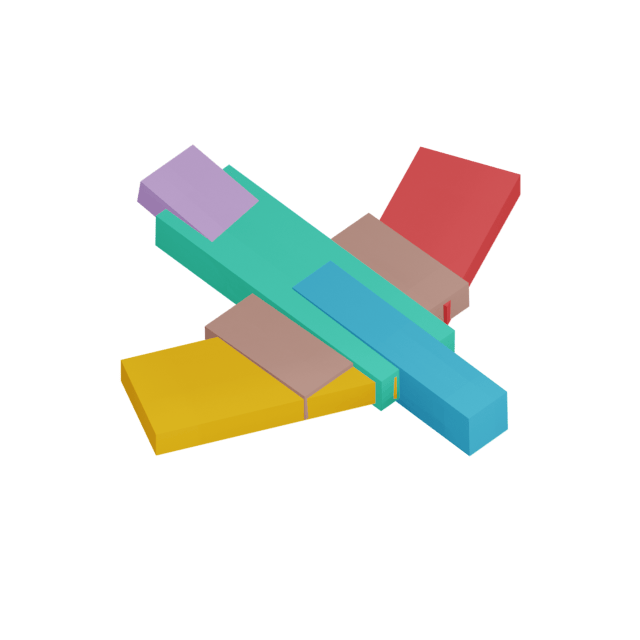}
        \put(-25,-5){\includegraphics[scale=0.8]{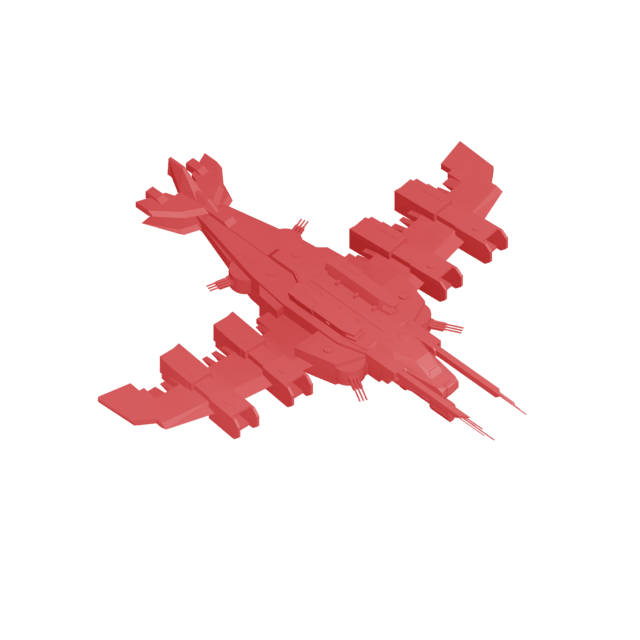}}
    \end{overpic} \\

    \begin{overpic}[]{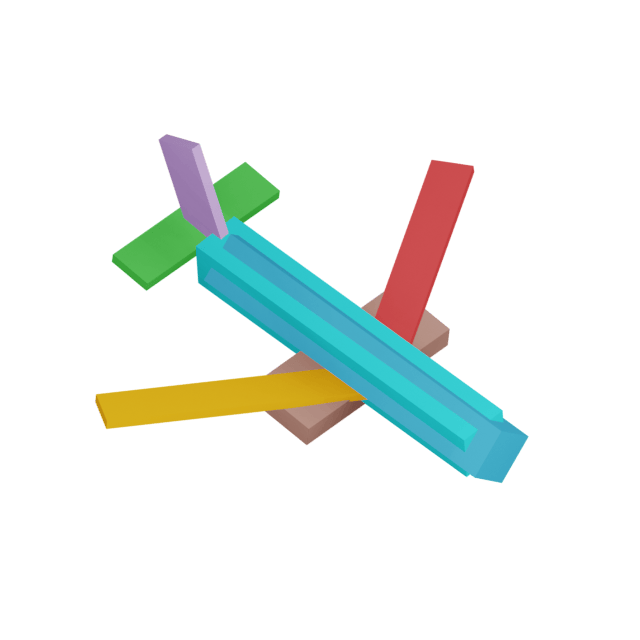}
        \put(-25,-5){\includegraphics[scale=0.8]{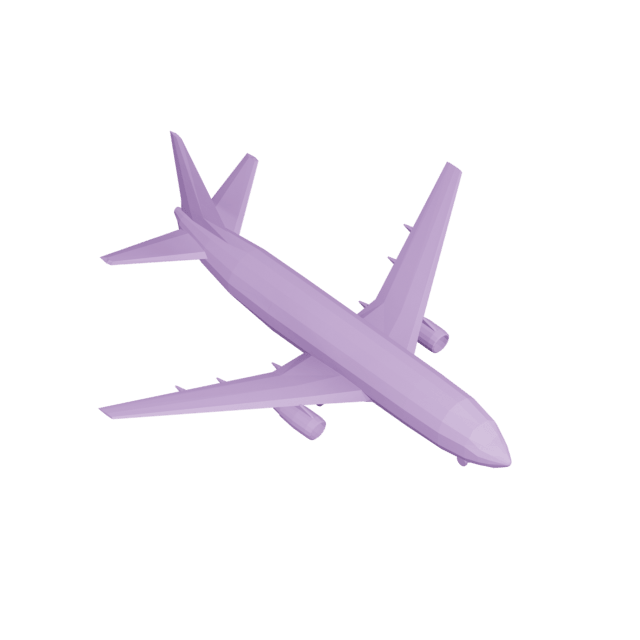}}
    \end{overpic} &  
    \begin{overpic}[]{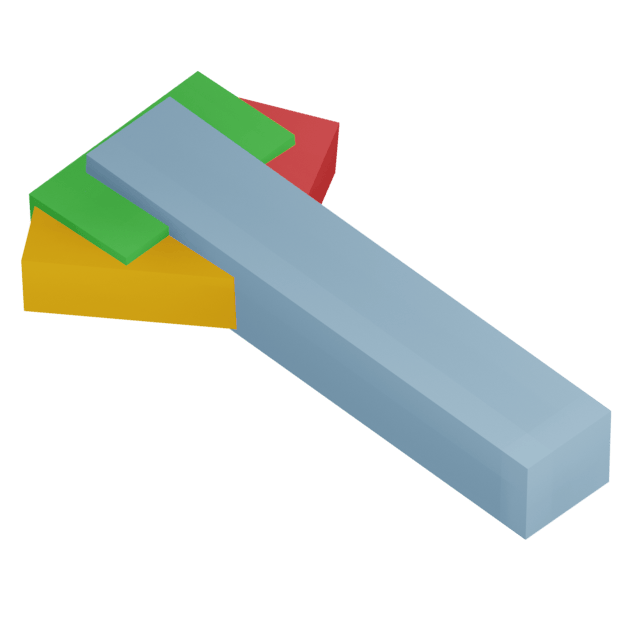}
        \put(-25,-5){\includegraphics[scale=0.8]{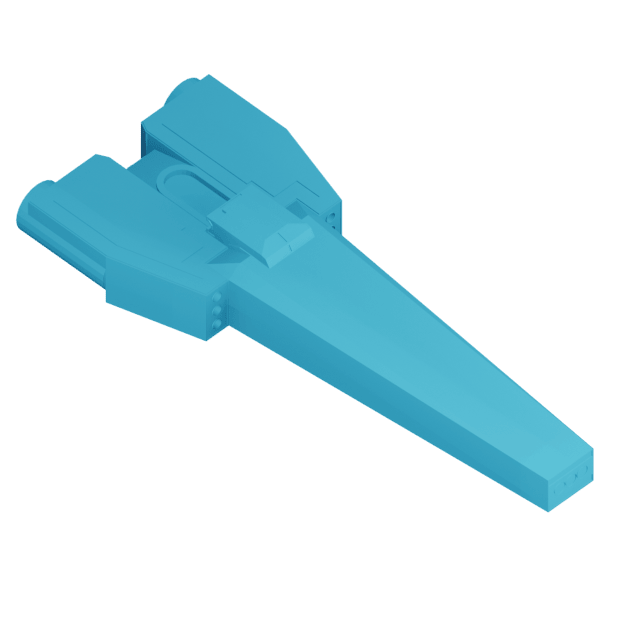}}
    \end{overpic} &  
    \begin{overpic}[]{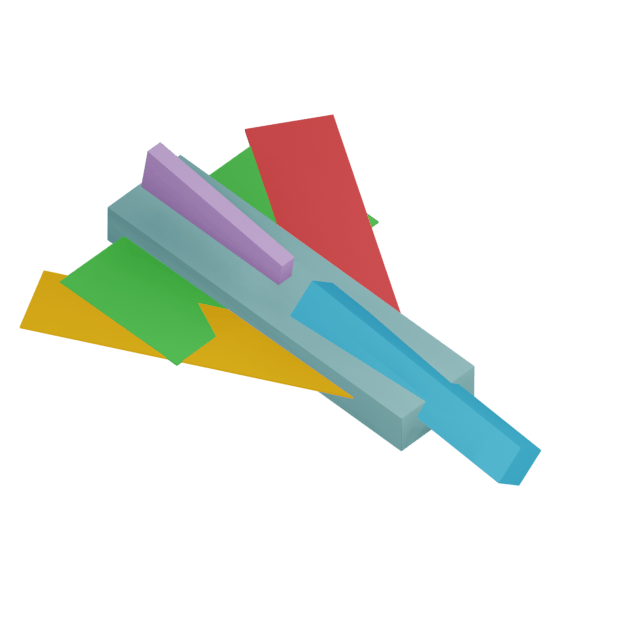}
        \put(-25,-5){\includegraphics[scale=0.8]{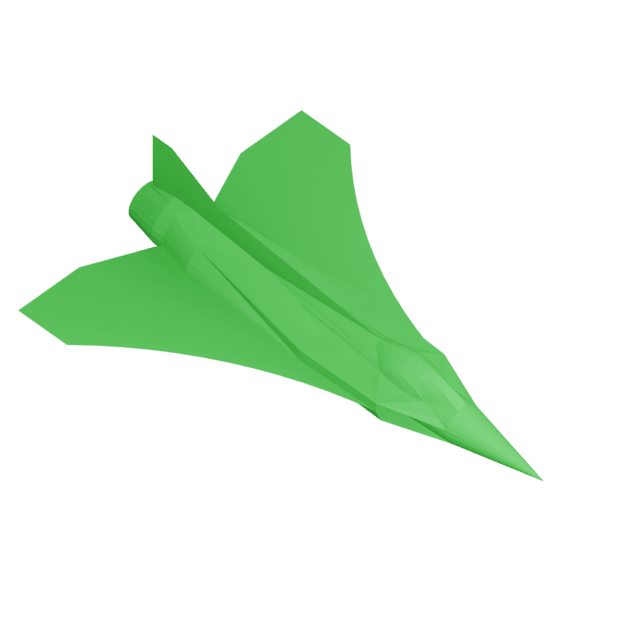}}
    \end{overpic} &  
    \begin{overpic}[]{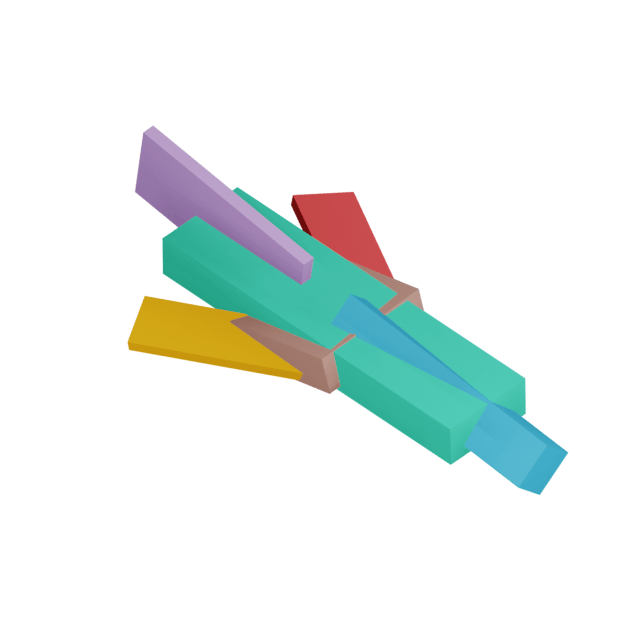}
        \put(-25,-5){\includegraphics[scale=0.8]{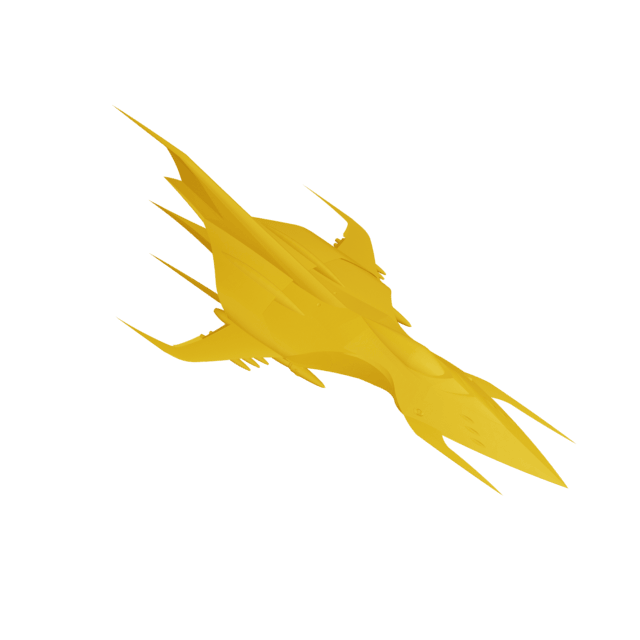}}
    \end{overpic} & 
    \begin{overpic}[]{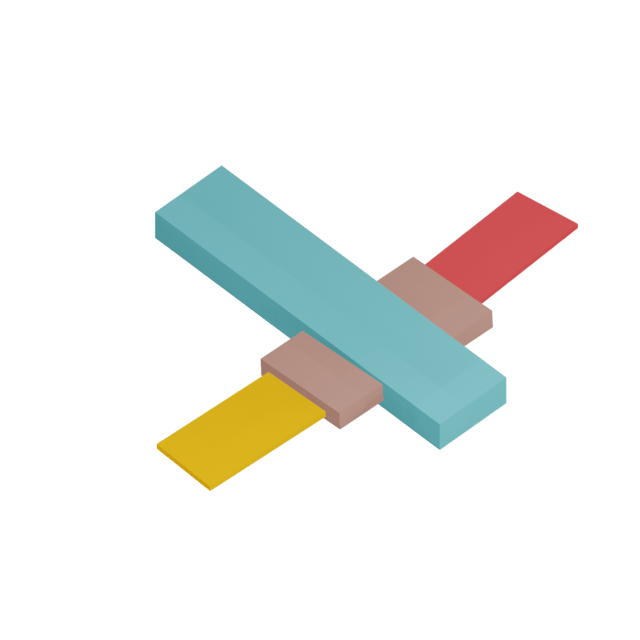}
        \put(-25,-5){\includegraphics[scale=0.8]{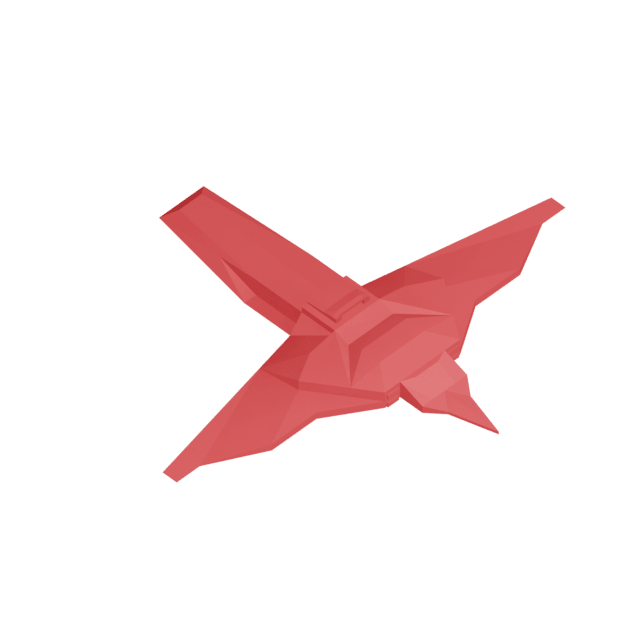}}
    \end{overpic} \\

    \begin{overpic}[]{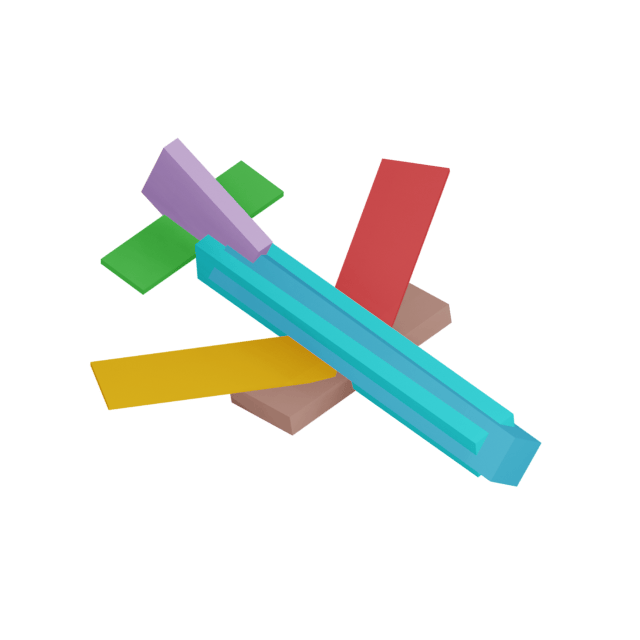}
        \put(-25,-5){\includegraphics[scale=0.8]{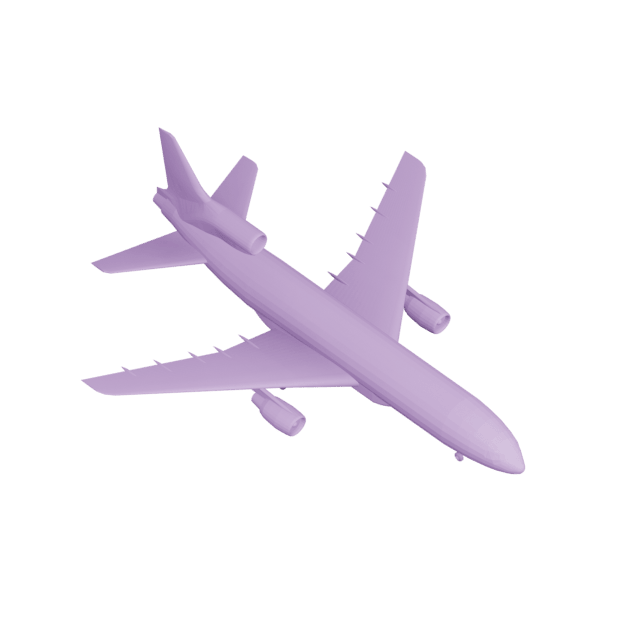}}
    \end{overpic} &      
    \begin{overpic}[]{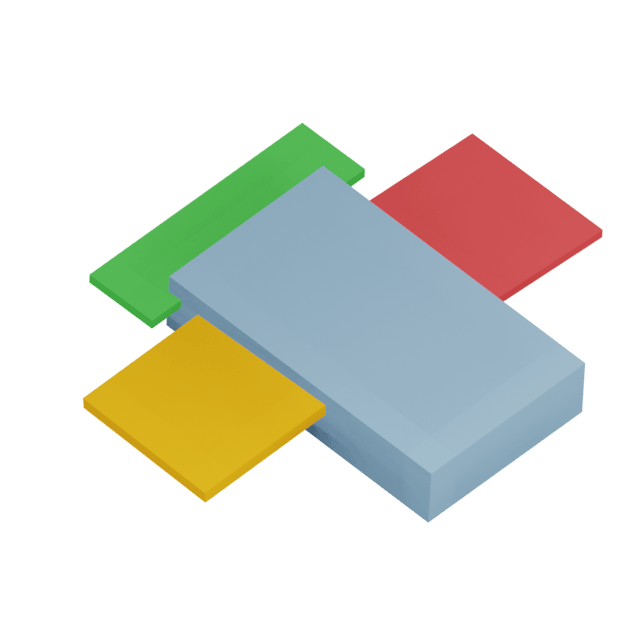}
        \put(-25,-5){\includegraphics[scale=0.8]{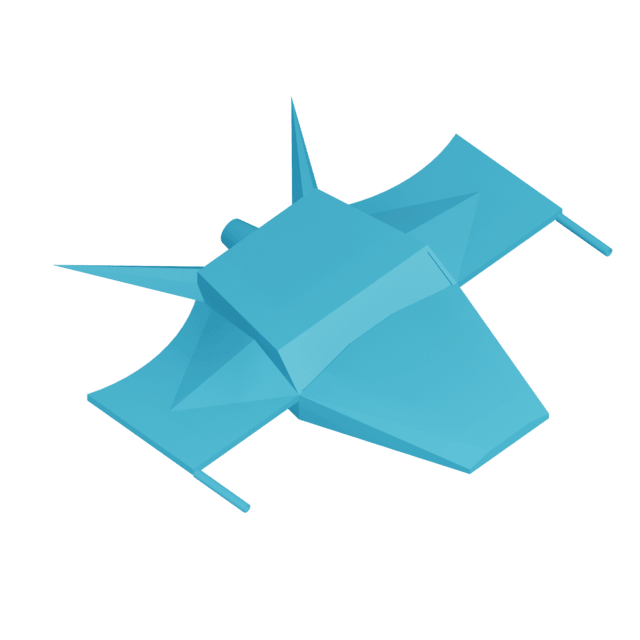}}
    \end{overpic} &  
    \begin{overpic}[]{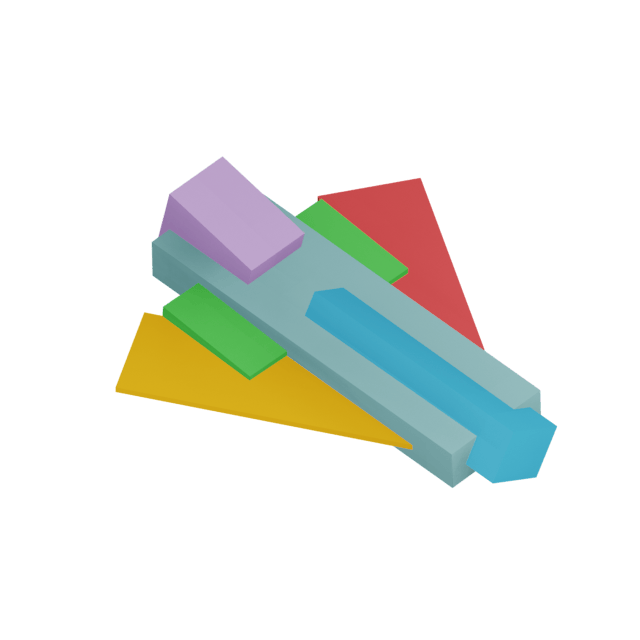}
        \put(-25,-5){\includegraphics[scale=0.8]{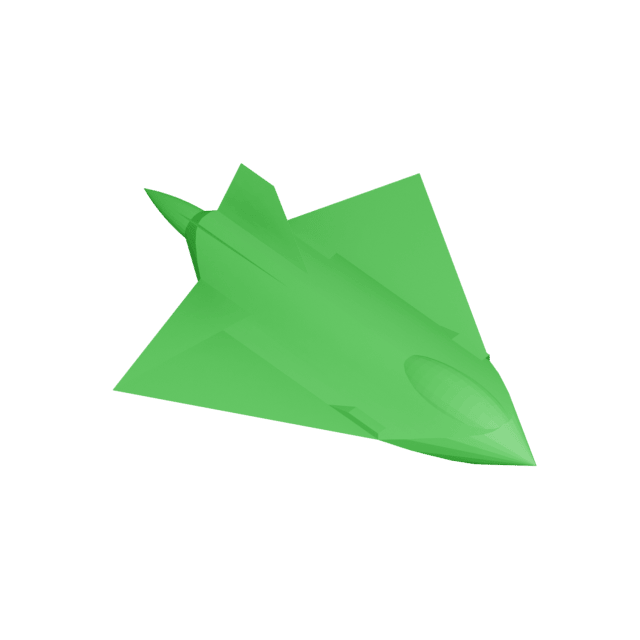}}
    \end{overpic} &  
    \begin{overpic}[]{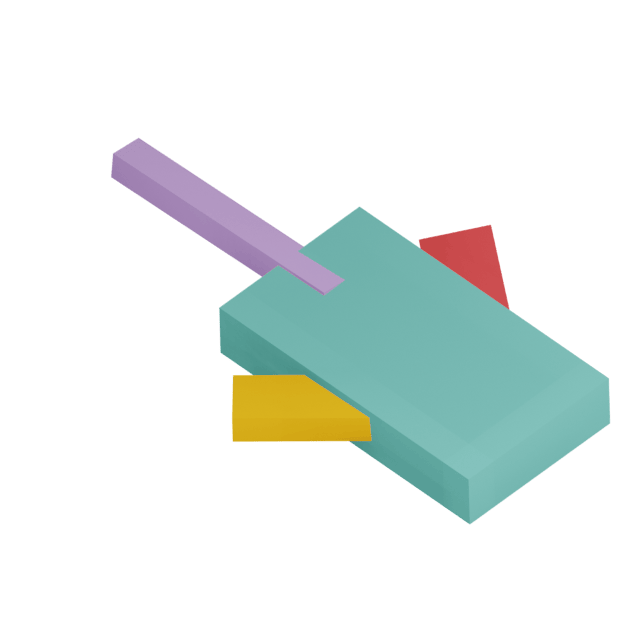}
        \put(-25,-5){\includegraphics[scale=0.8]{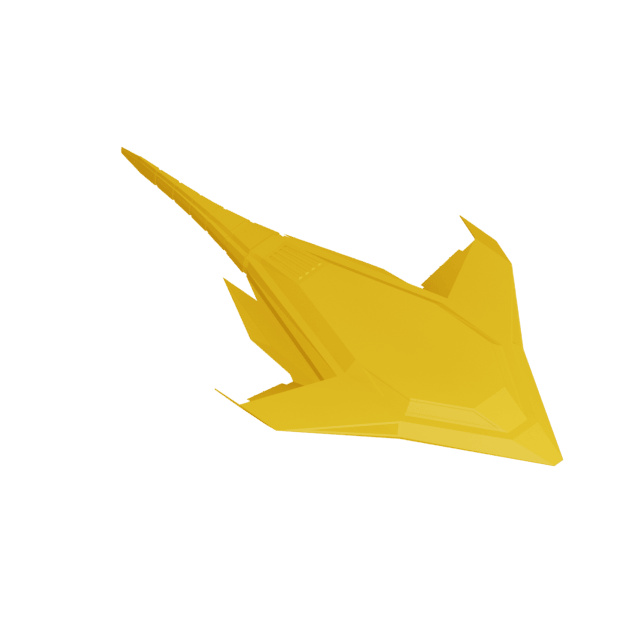}}
    \end{overpic} & 
    \begin{overpic}[]{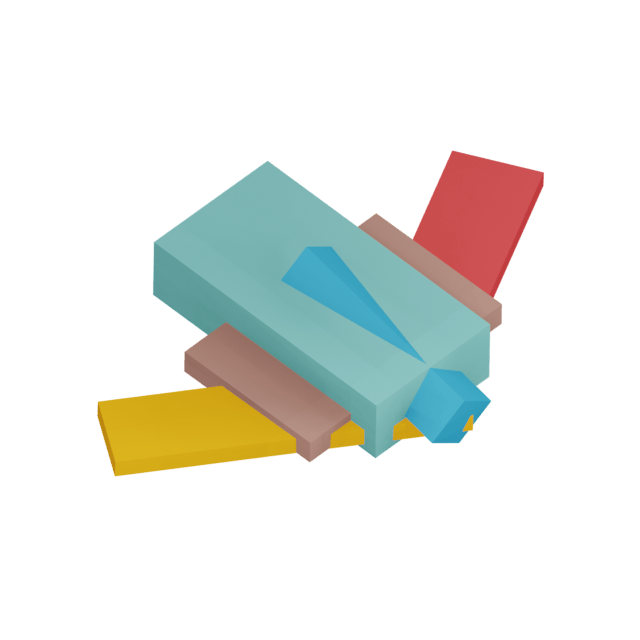}
        \put(-25,-5){\includegraphics[scale=0.8]{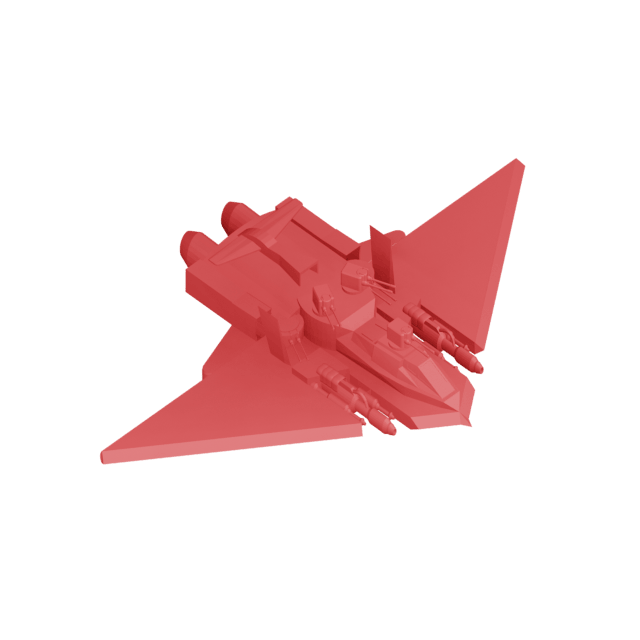}}
    \end{overpic} \\
\end{tabular}
}
    \hfill
    \resizebox{0.475\linewidth}{!}{\begin{tabular}{ccccc}
    \begin{overpic}[]{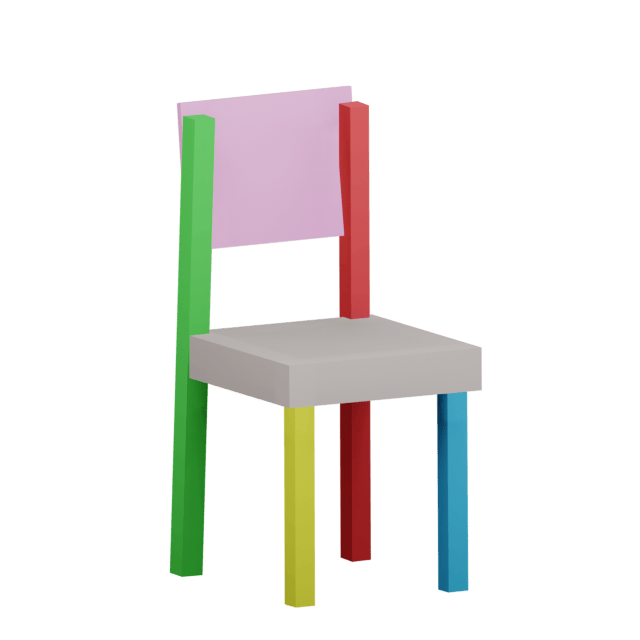}
        \put(-25,-5){\includegraphics[scale=0.9]{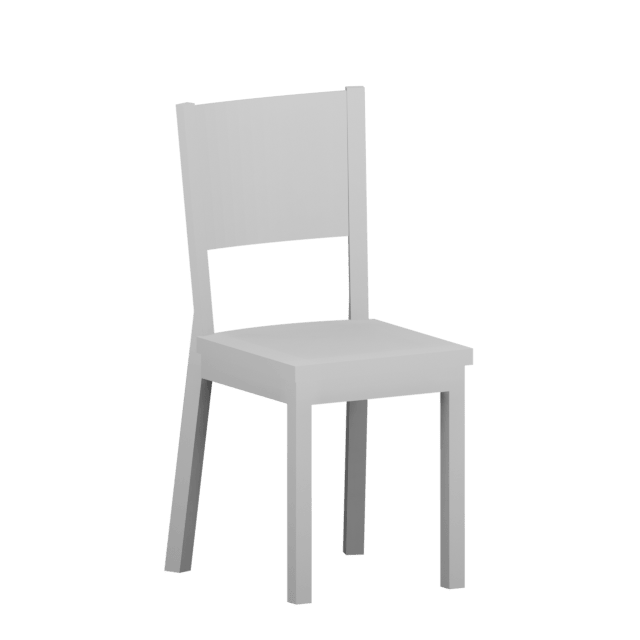}}
    \end{overpic} &  
    \begin{overpic}[]{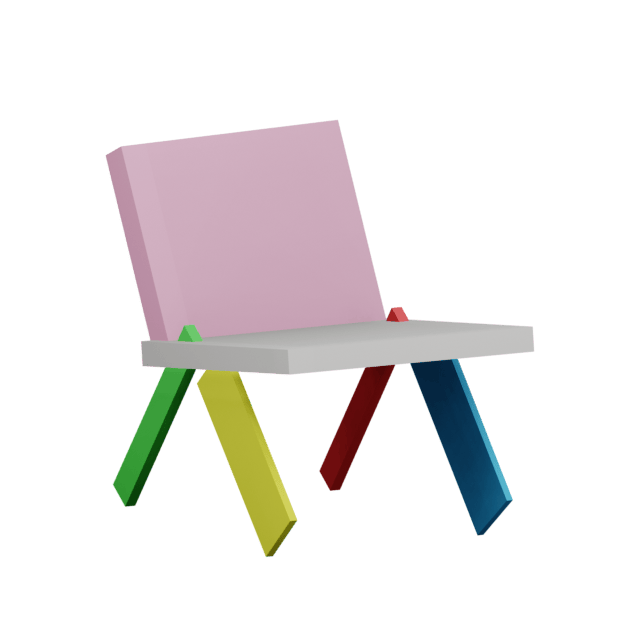}
        \put(-25, -5){\includegraphics[scale=0.9]{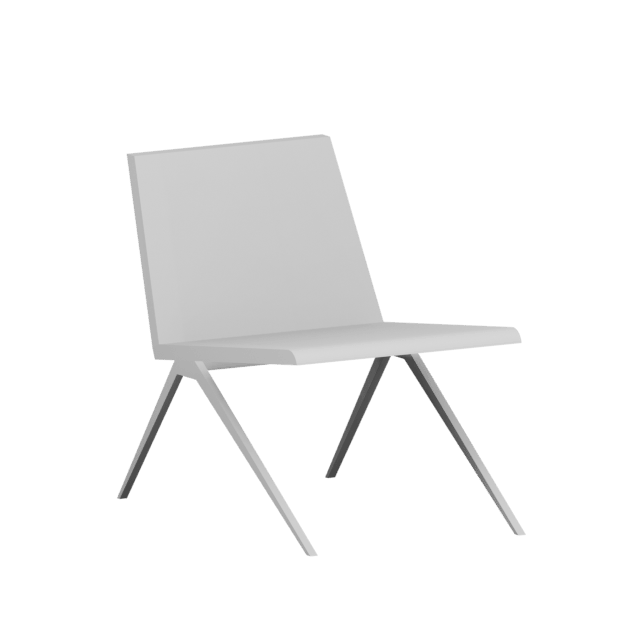}}
    \end{overpic} &  
    \begin{overpic}[]{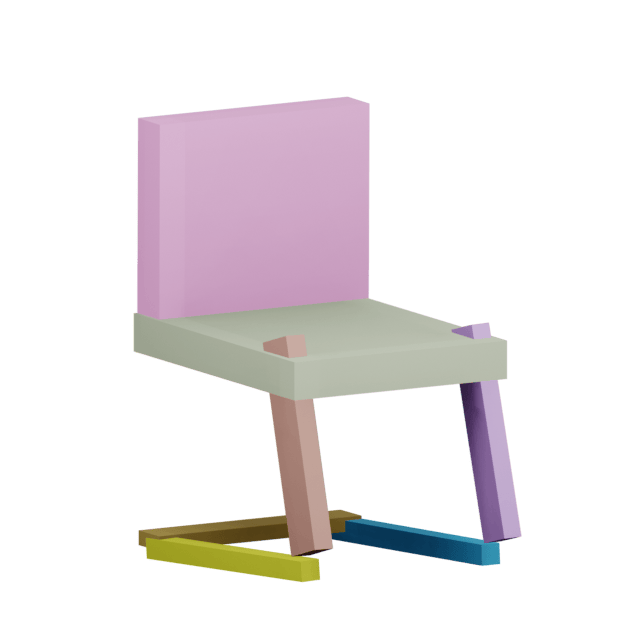}
        \put(-25,-5){\includegraphics[scale=0.9]{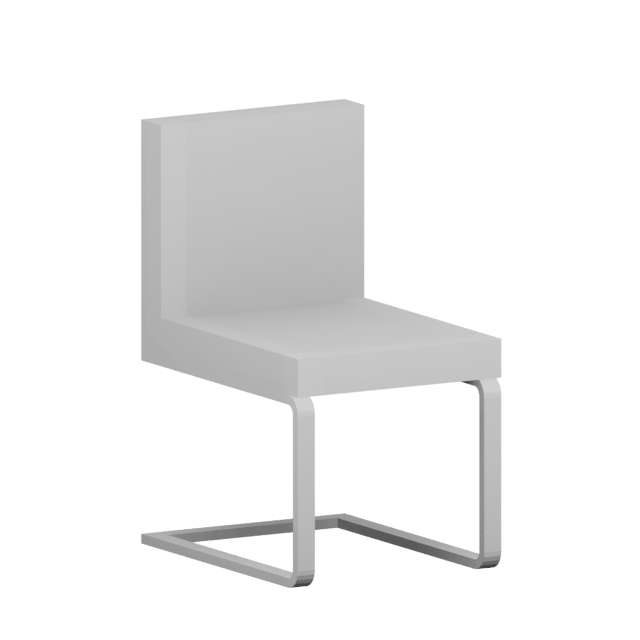}}
    \end{overpic} &  
    \begin{overpic}[]{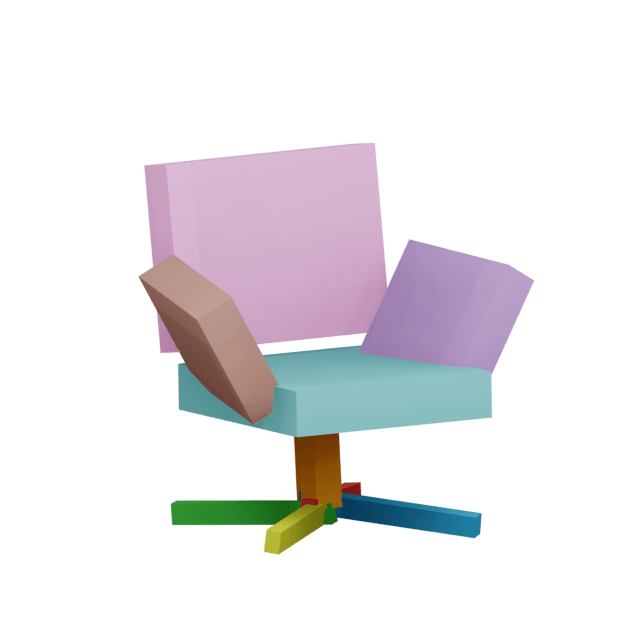}
        \put(-25,-5){\includegraphics[scale=0.9]{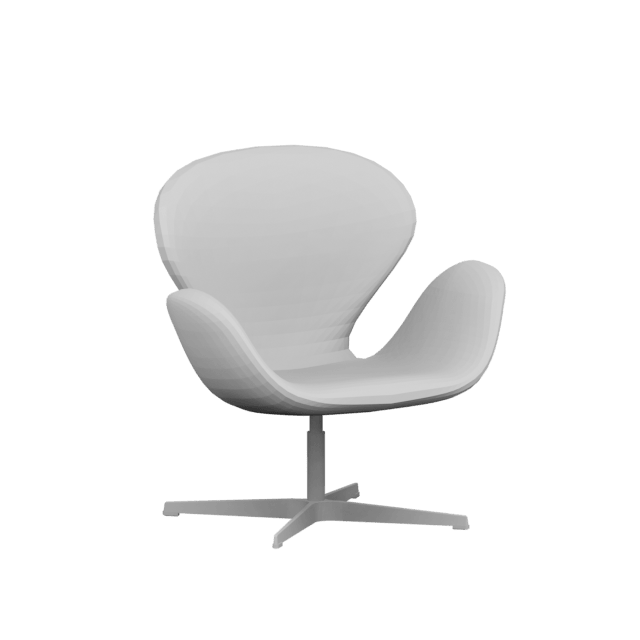}}
    \end{overpic} & 
    \begin{overpic}[]{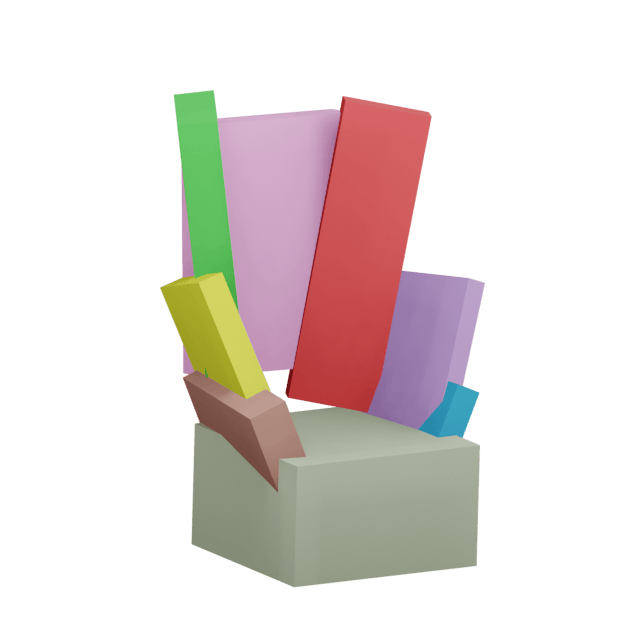}
        \put(-25, -5){\includegraphics[scale=0.9]{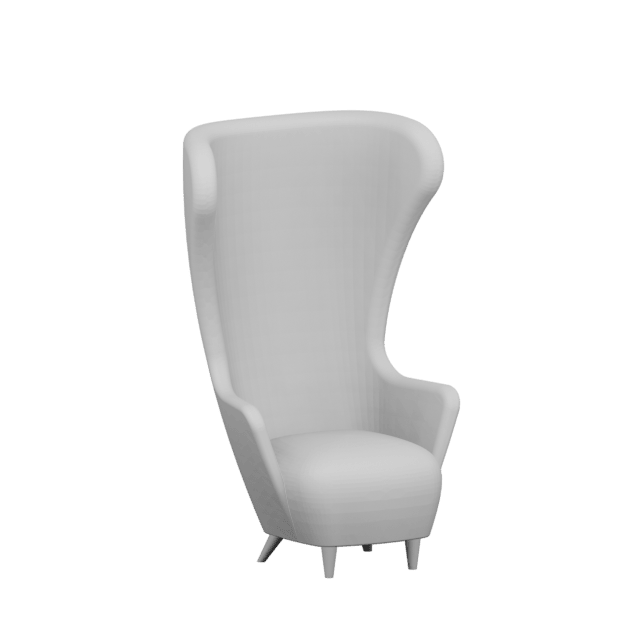}}
    \end{overpic} \\
    \midrule
     
    \begin{overpic}[]{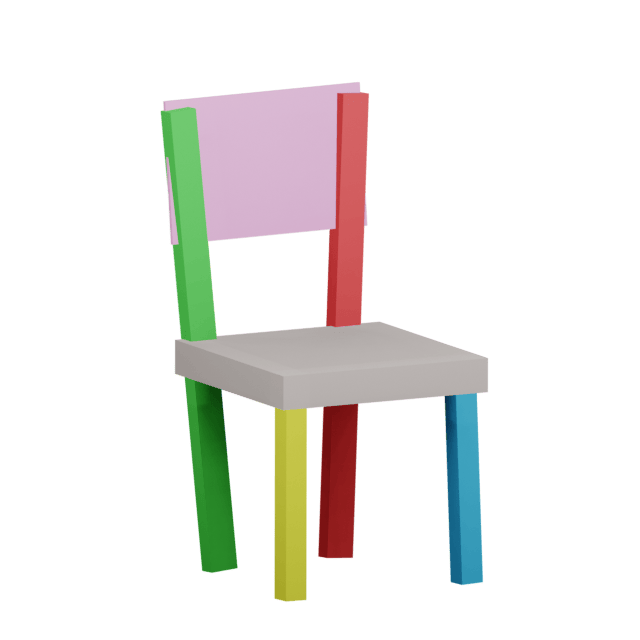}
        \put(-25,-5){\includegraphics[scale=0.9]{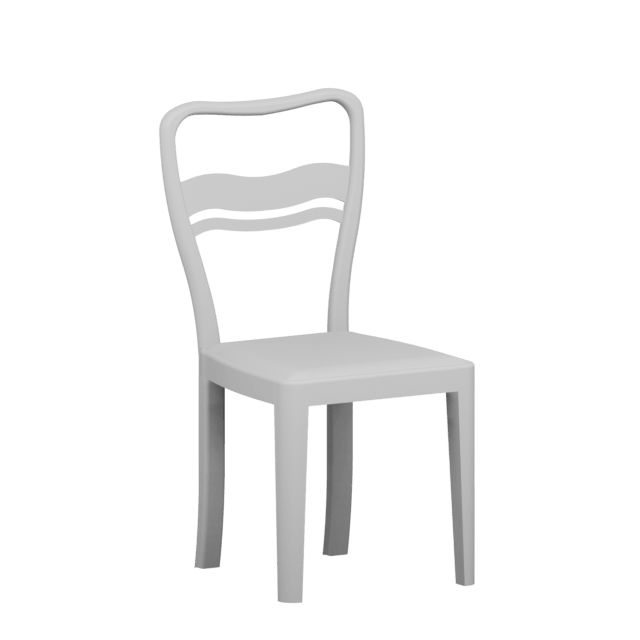}}
    \end{overpic} &  
    \begin{overpic}[]{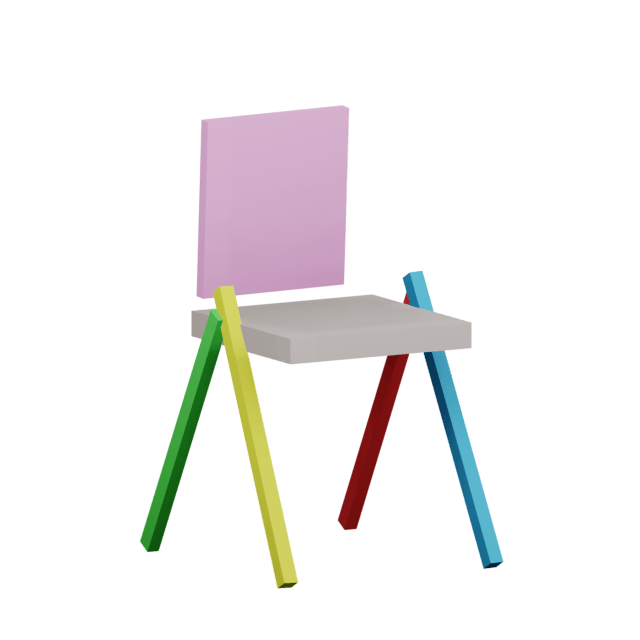}
        \put(-25,-5){\includegraphics[scale=0.9]{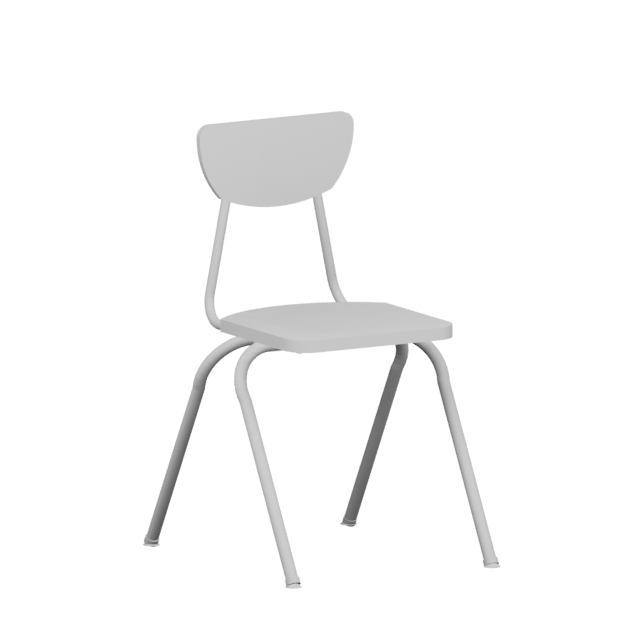}}
    \end{overpic}&  
    \begin{overpic}[]{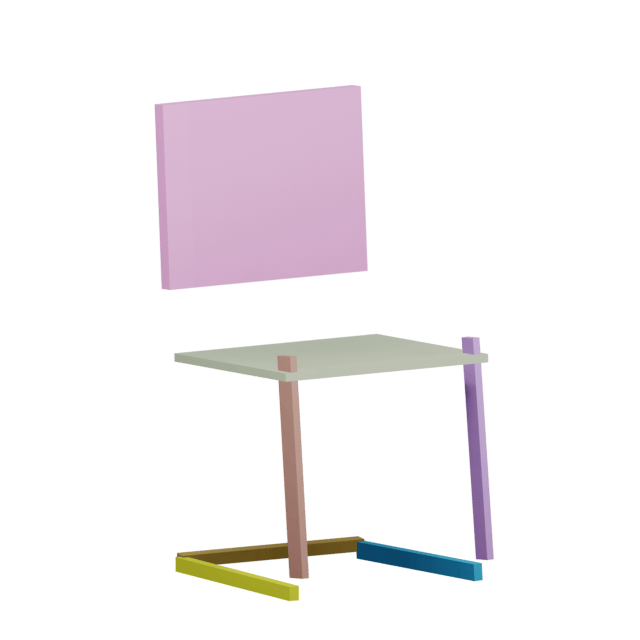}
        \put(-25,-5){\includegraphics[scale=0.9]{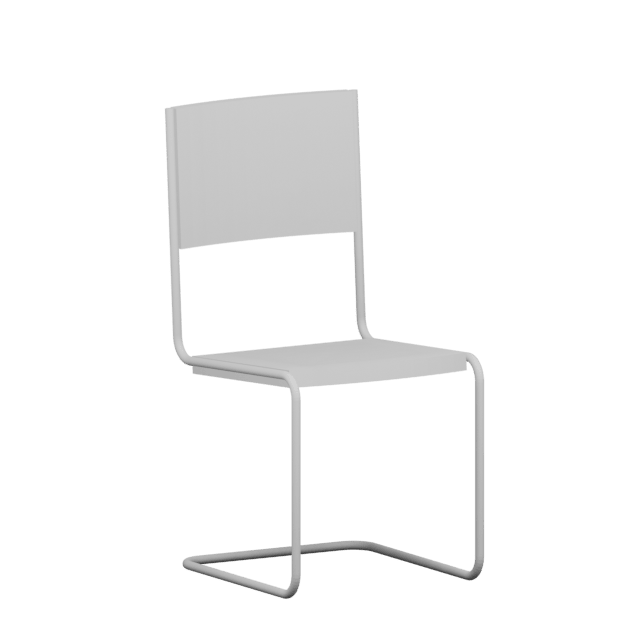}}
    \end{overpic}&  
    \begin{overpic}[]{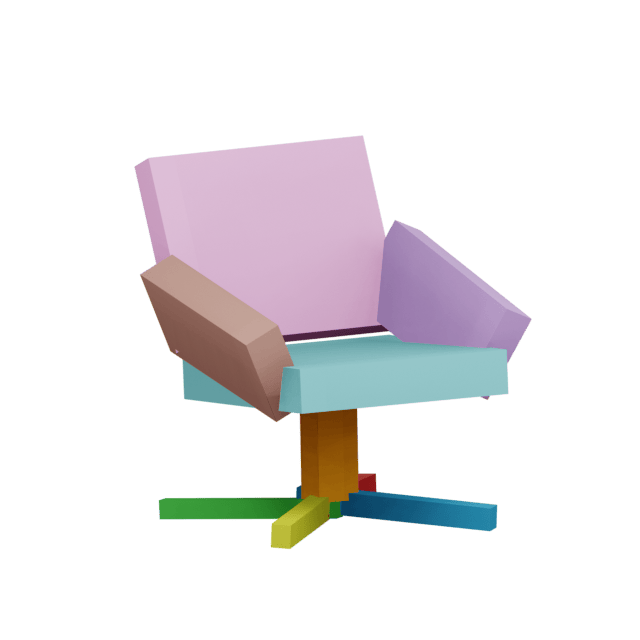}
        \put(-25,-5){\includegraphics[scale=0.9]{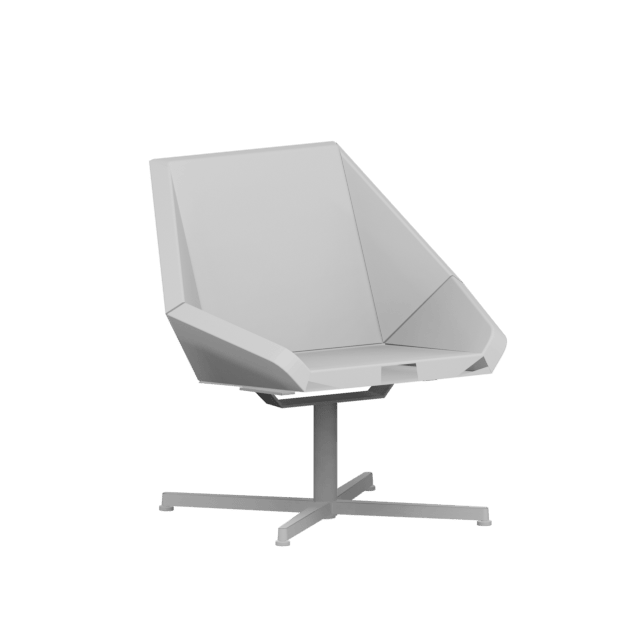}}
    \end{overpic}& 
    \begin{overpic}[]{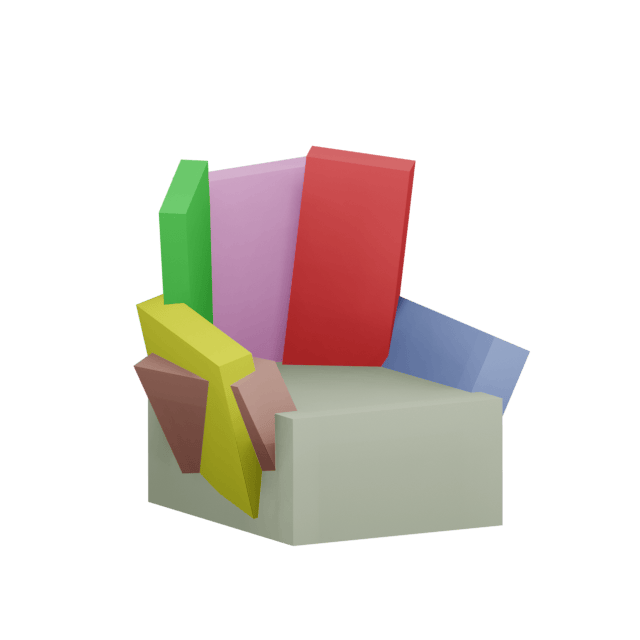}
        \put(-25,-5){\includegraphics[scale=0.9]{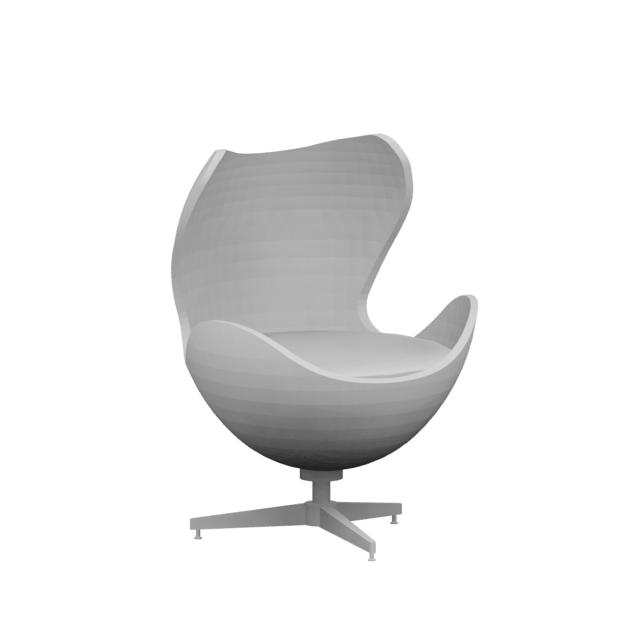}}
    \end{overpic} \\

    \begin{overpic}[]{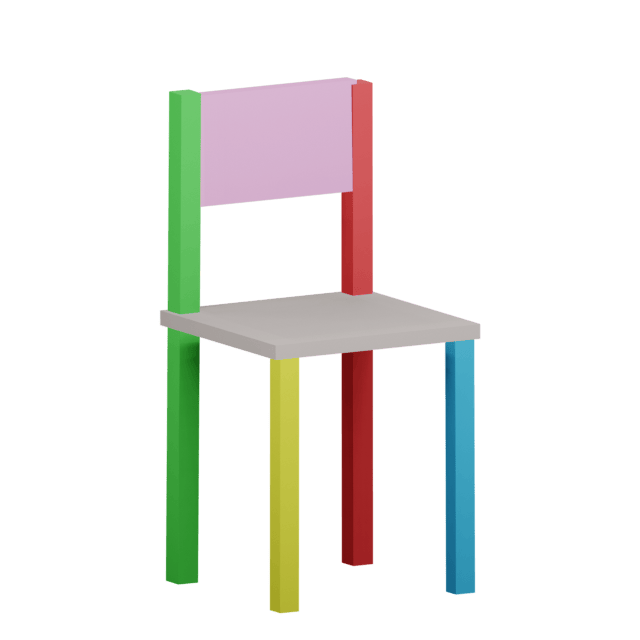}
        \put(-25,-5){\includegraphics[scale=0.9]{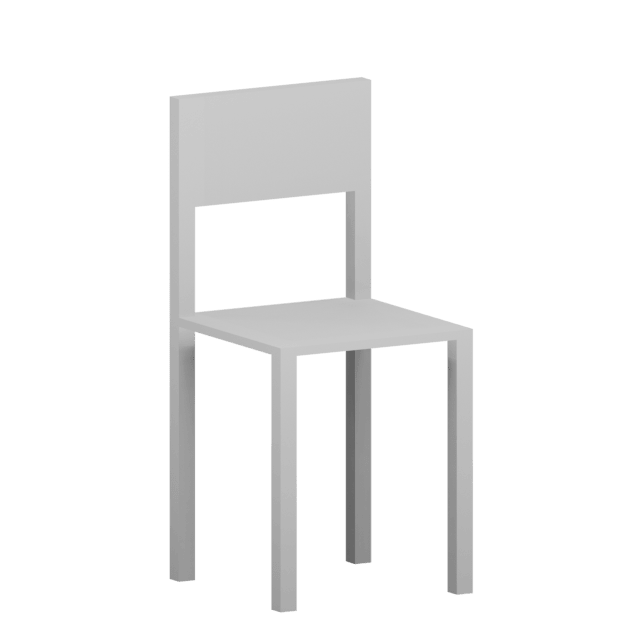}}
    \end{overpic} &  
    \begin{overpic}[]{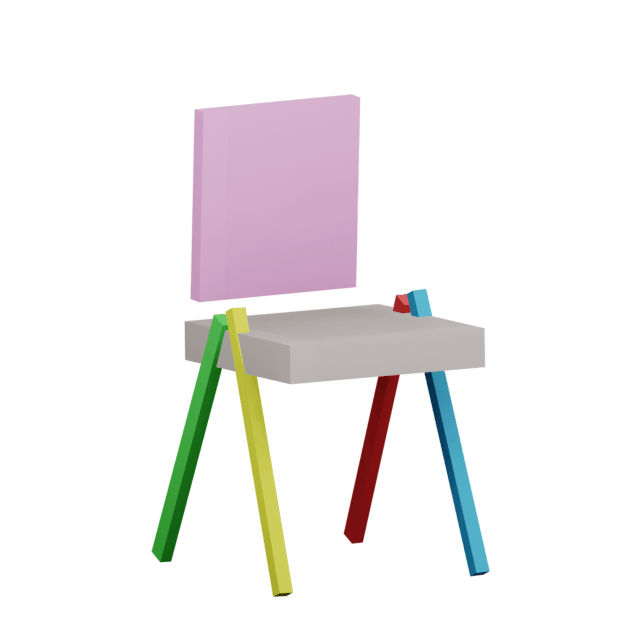}
        \put(-25,-5){\includegraphics[scale=0.9]{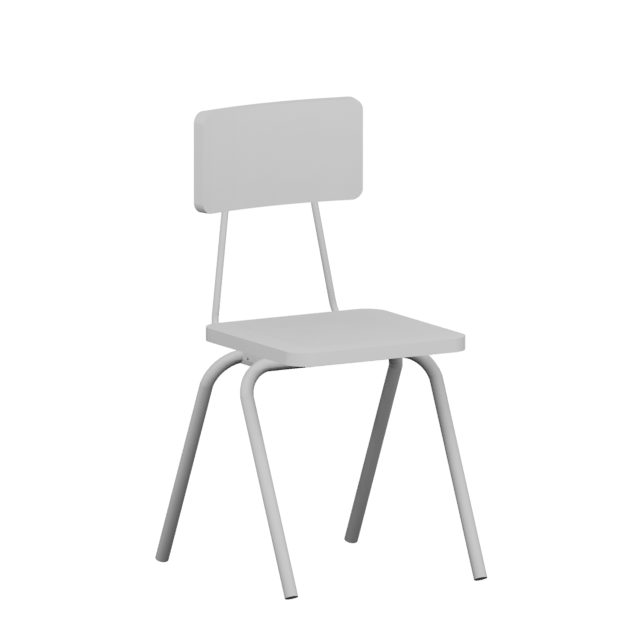}}
    \end{overpic} & 
    \begin{overpic}[]{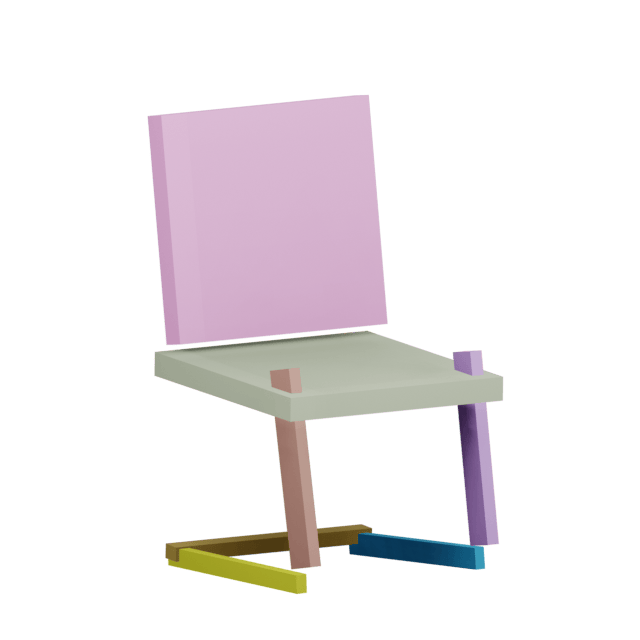}
        \put(-25,-5){\includegraphics[scale=0.9]{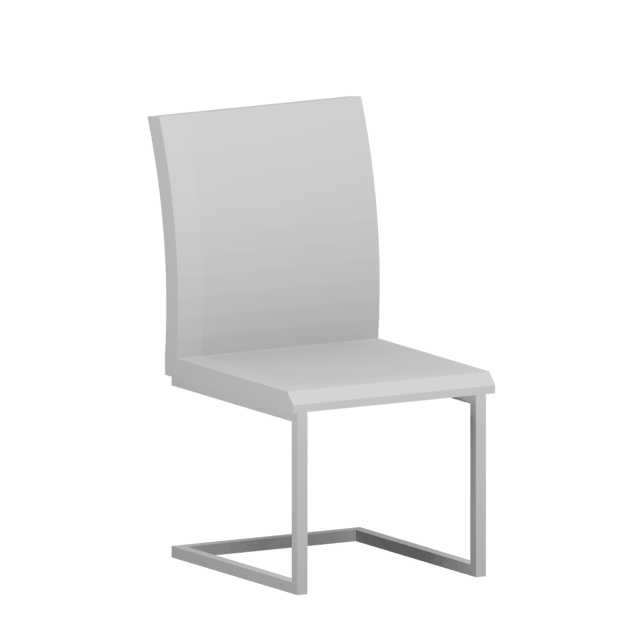}}
    \end{overpic} &  
    \begin{overpic}[]{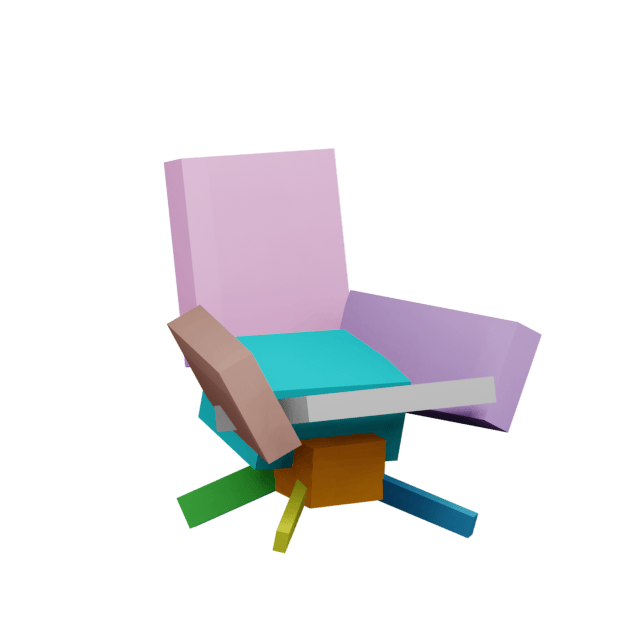}
        \put(-25,-5){\includegraphics[scale=0.9]{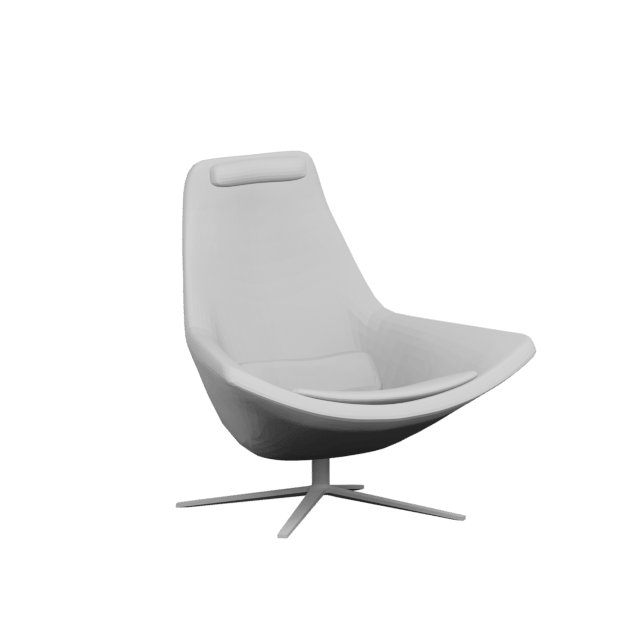}}
    \end{overpic} & 
    \begin{overpic}[]{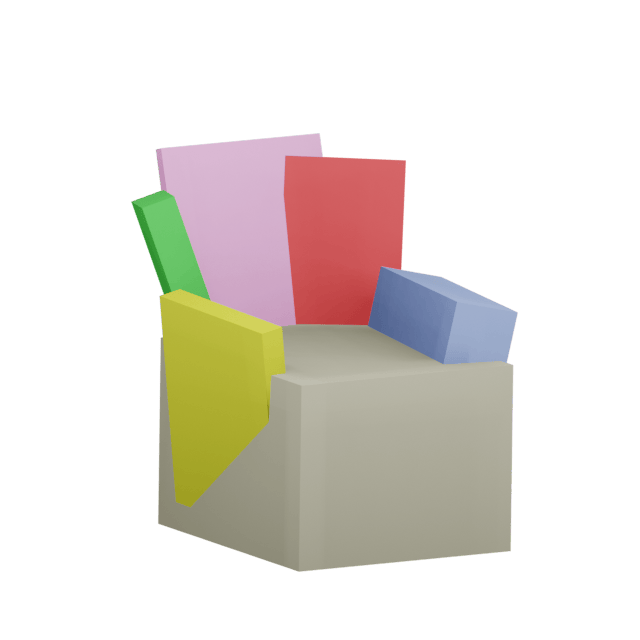}
        \put(-25,-5){\includegraphics[scale=0.9]{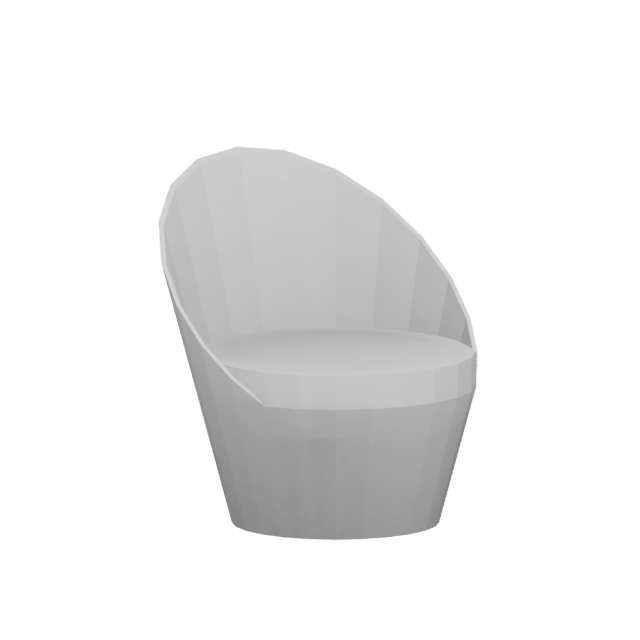}}
    \end{overpic} \\
    
    \begin{overpic}[]{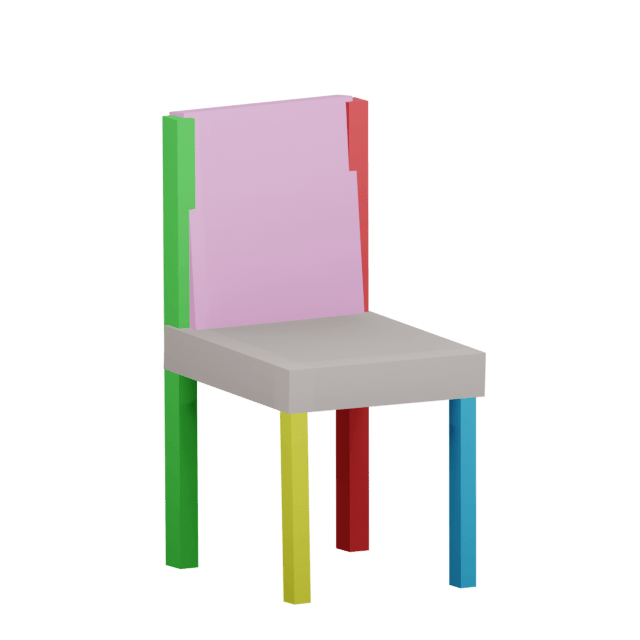}
        \put(-25,-5){\includegraphics[scale=0.9]{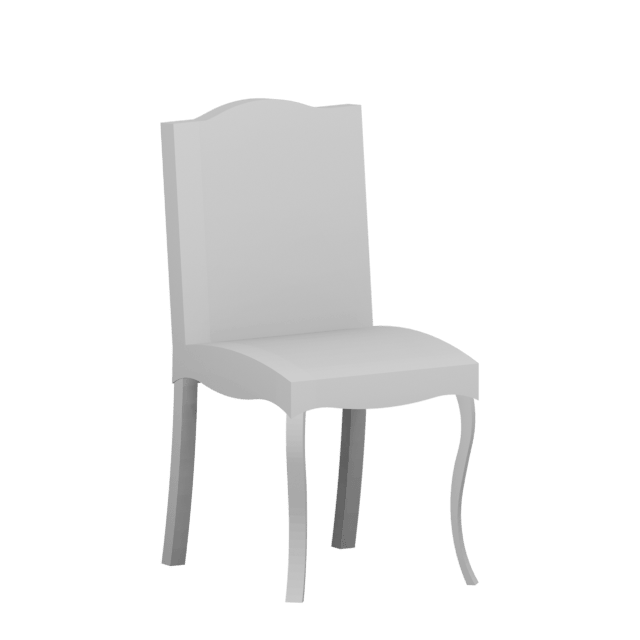}}
    \end{overpic} &  
    \begin{overpic}[]{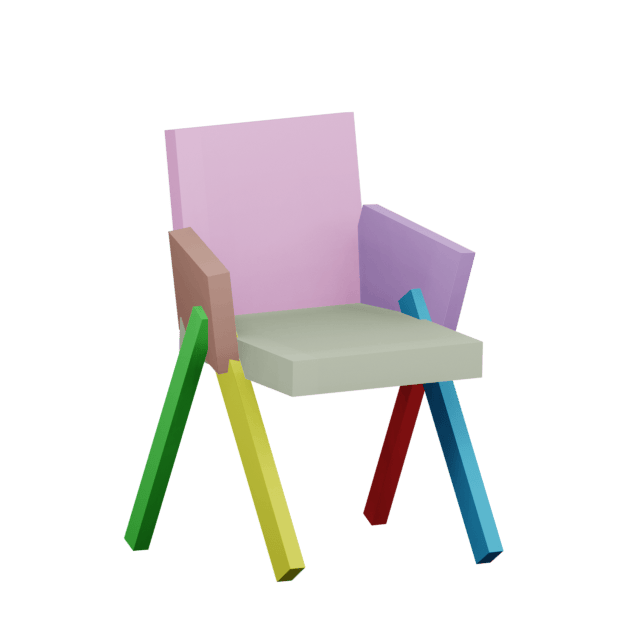}
        \put(-25,-5){\includegraphics[scale=0.9]{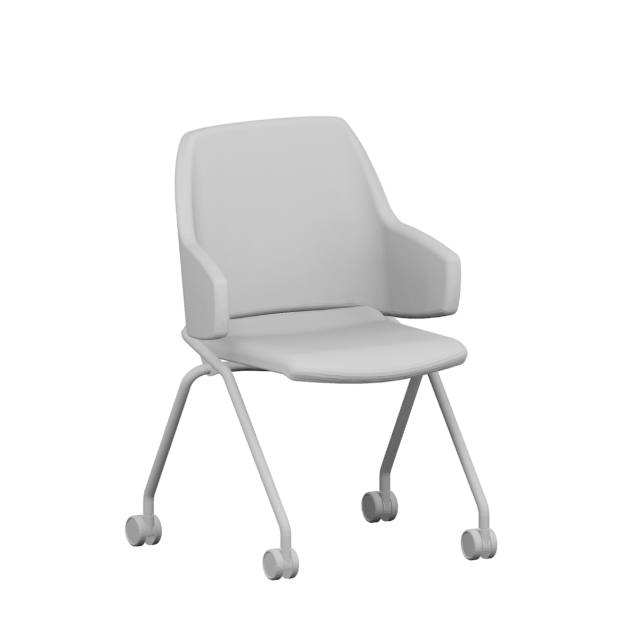}}
    \end{overpic} &  
    \begin{overpic}[]{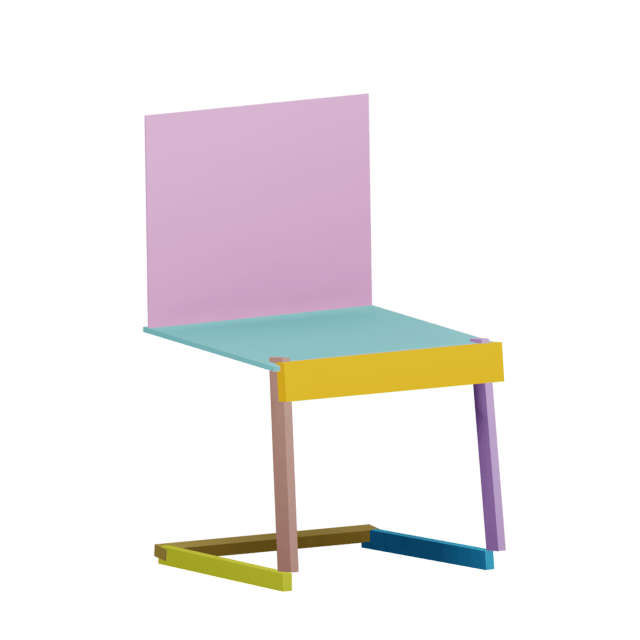}
        \put(-25,-5){\includegraphics[scale=0.9]{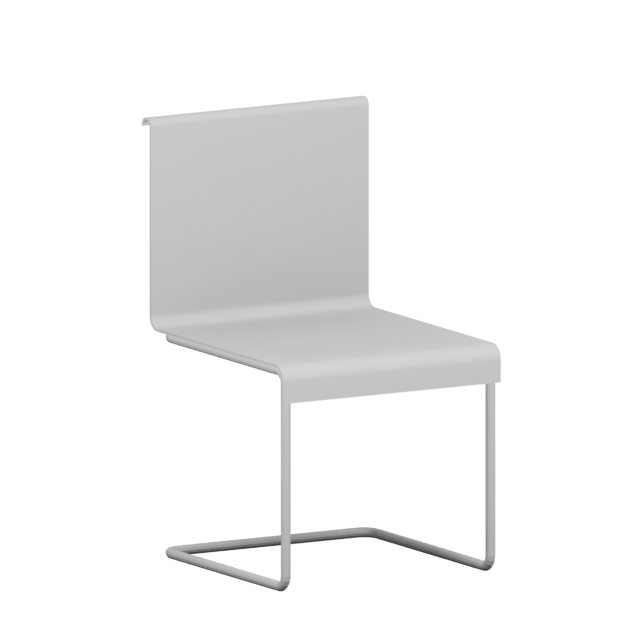}}
    \end{overpic} &  
    \begin{overpic}[]{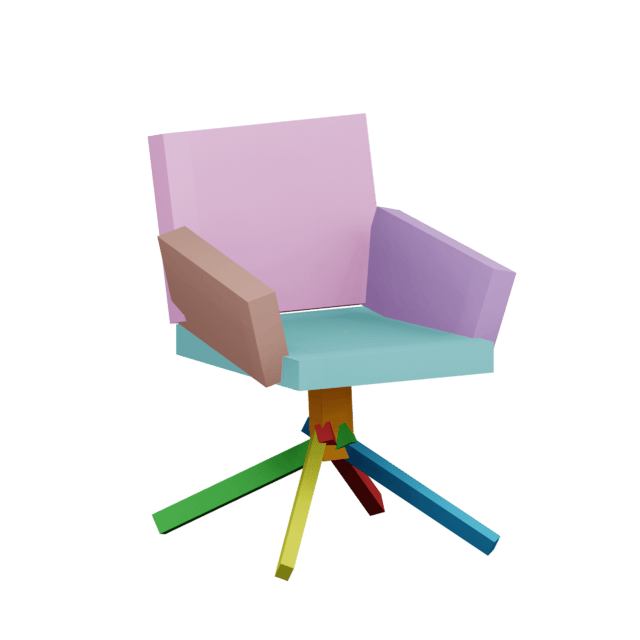}
        \put(-25,-5){\includegraphics[scale=0.9]{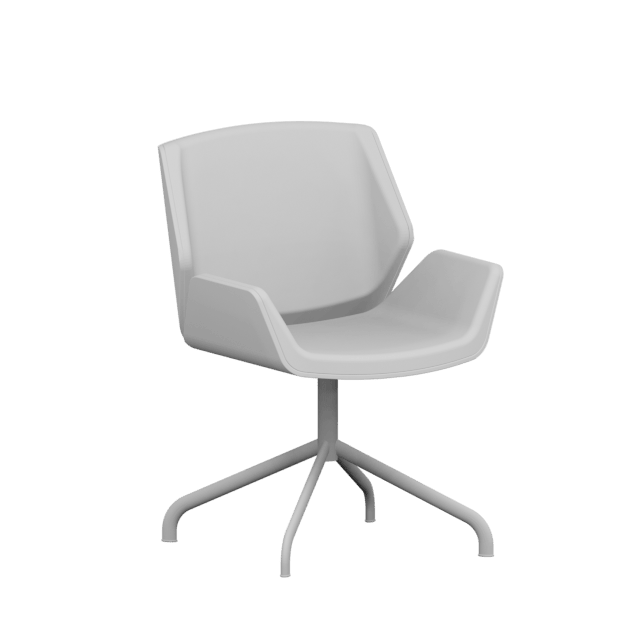}}
    \end{overpic} & 
    \begin{overpic}[]{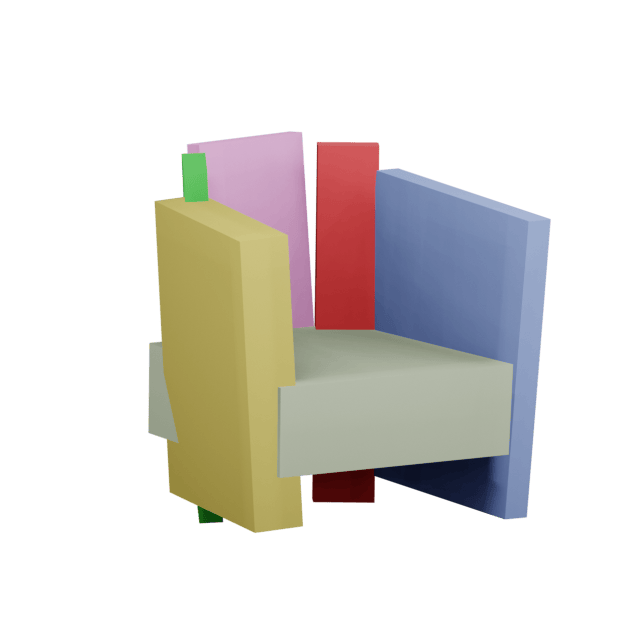}
        \put(-25,-5){\includegraphics[scale=0.9]{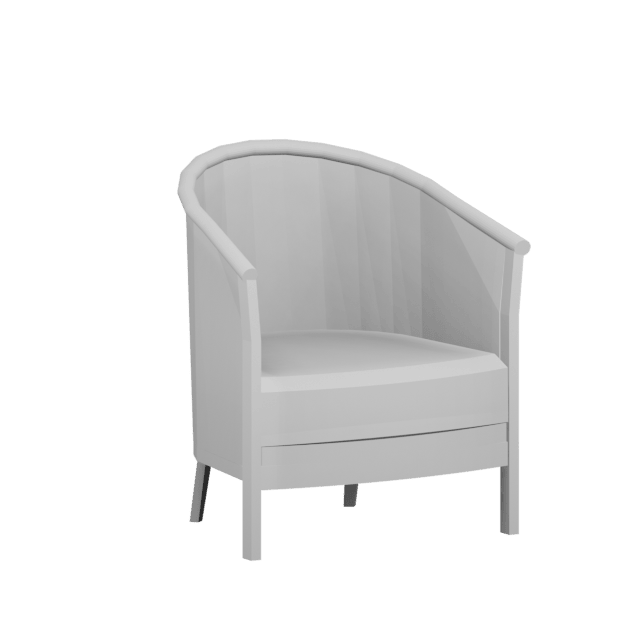}}
    \end{overpic} \\
    
    \begin{overpic}[]{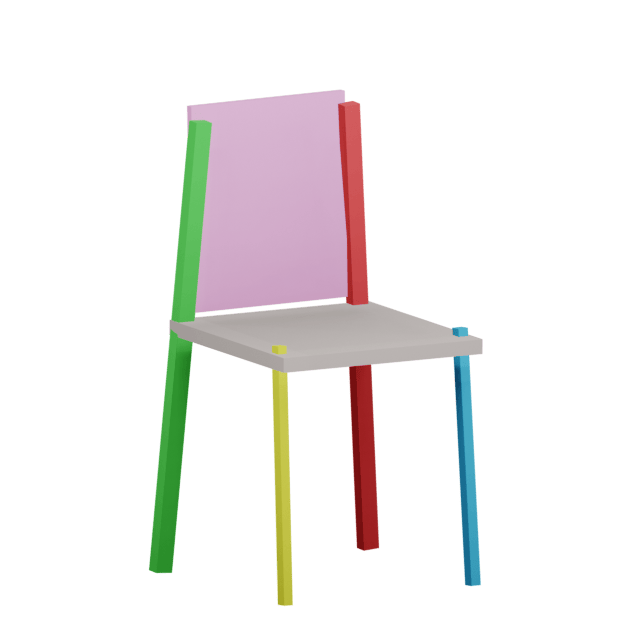}
        \put(-25,-5){\includegraphics[scale=0.9]{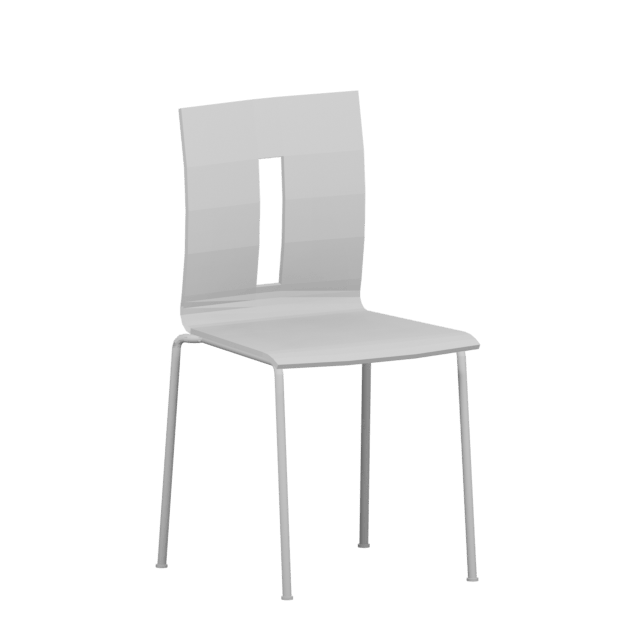}}
    \end{overpic} &      
    \begin{overpic}[]{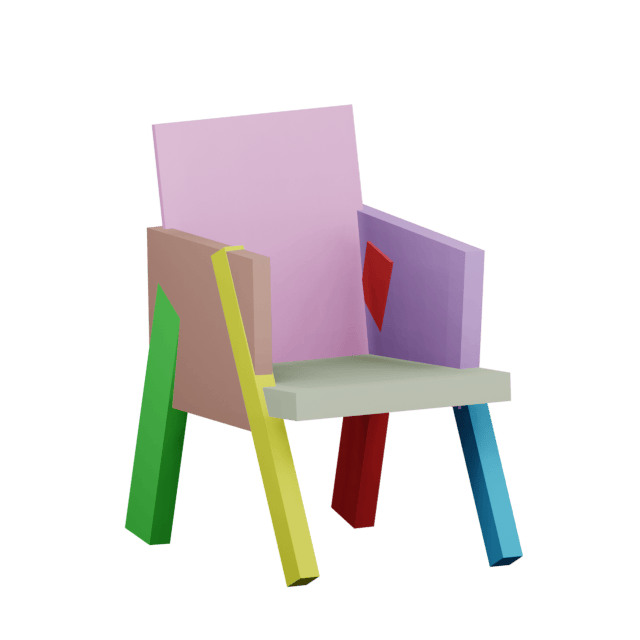}
        \put(-25,-5){\includegraphics[scale=0.9]{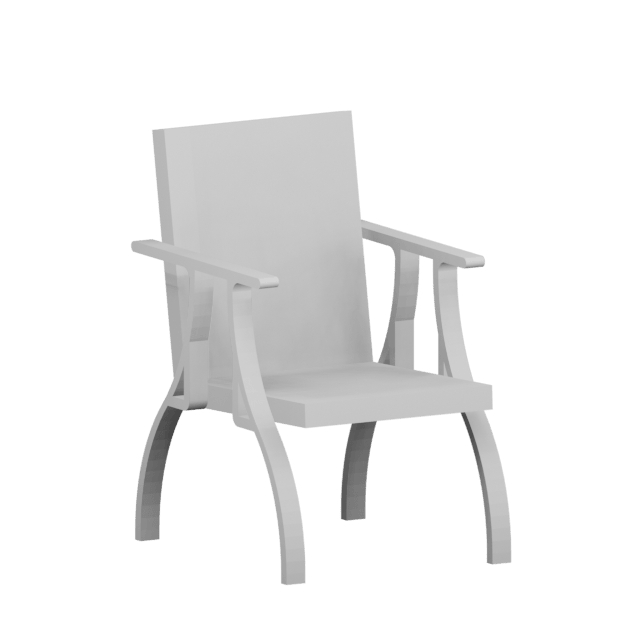}}
    \end{overpic} &  
    \begin{overpic}[]{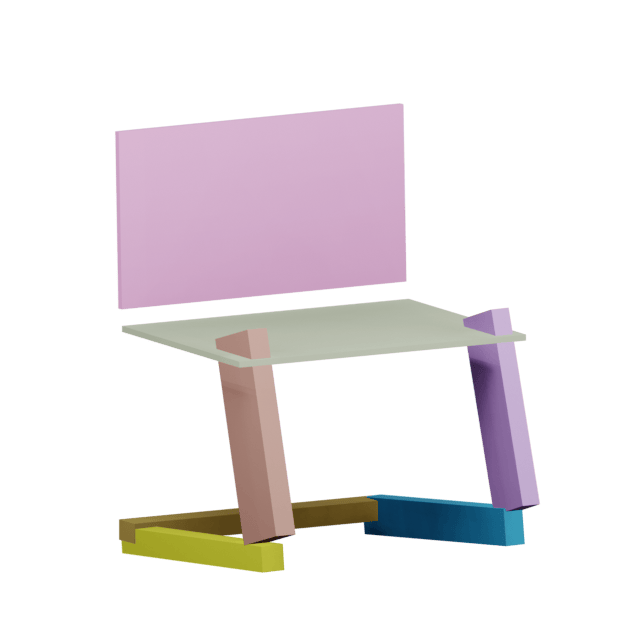}
        \put(-25,-5){\includegraphics[scale=0.9]{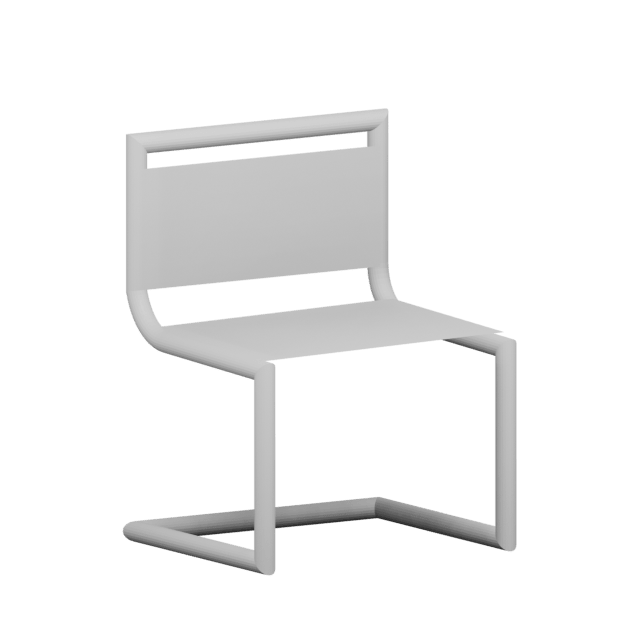}}
    \end{overpic} &  
    \begin{overpic}[]{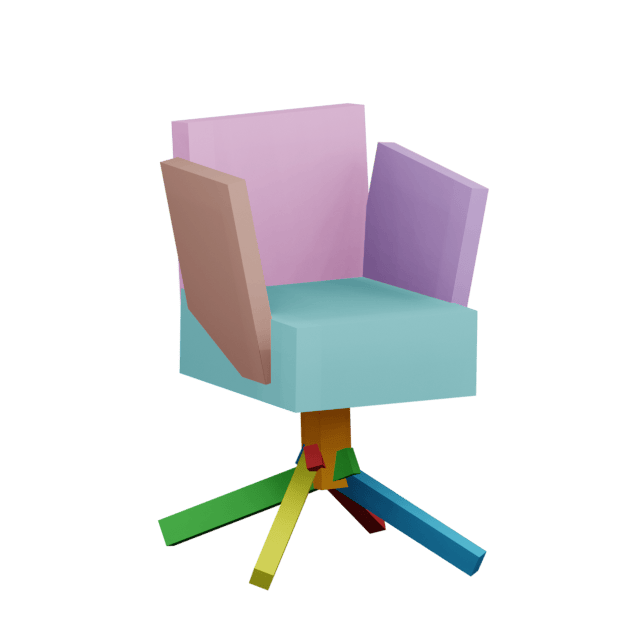}
        \put(-25,-5){\includegraphics[scale=0.9]{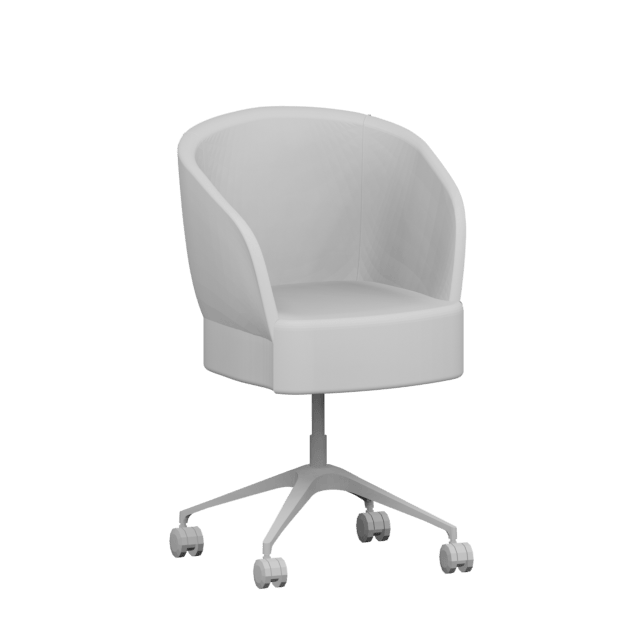}}
    \end{overpic} & 
    \begin{overpic}[]{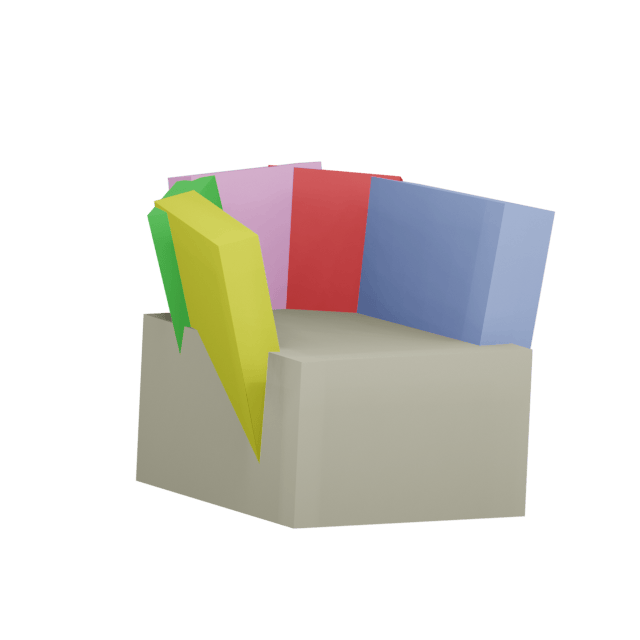}
        \put(-25,-5){\includegraphics[scale=0.9]{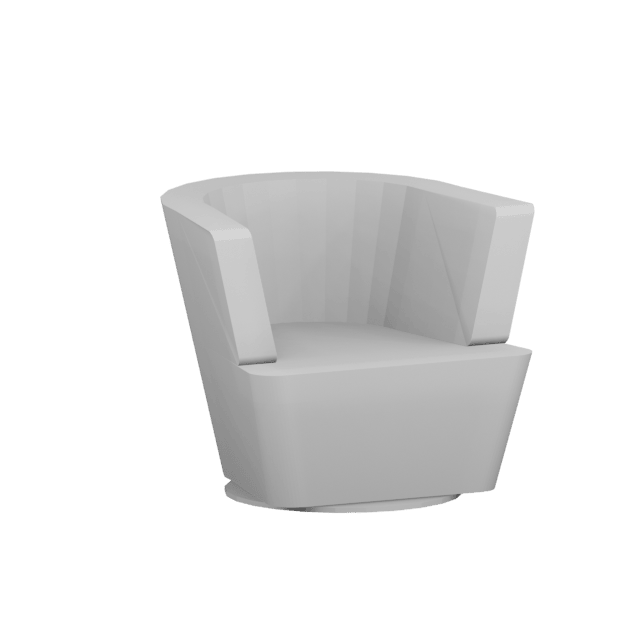}}
    \end{overpic} \\
\end{tabular}
}
    \caption{Left: k-means clustering ($k = 5$) on top of our cuboid parameters $[r,t,s, \gamma]$. Each column represents a cluster. Right: Retrieval of the nearest neighbors from a collection of shapes. Top row shows the query cuboids.}
    \label{fig:clustering_and_retrieval}
\end{figure}

We apply the output of our network to the downstream task of shape clustering and retrieval. 
First, we concatenate the cuboid parameters [$r,t,s$] into a single vector and multiply them with their corresponding existence probability $\gamma$. 
Next, we perform k-means clustering ($k = 5$) and compute the pairwise Euclidean distances of these vectors.
The results are presented in \Cref{fig:clustering_and_retrieval}.
Based solely on the cuboid abstraction we can discover distinct plane subcategories like passenger airplanes (purple), spacecrafts (blue) or jet fighters (green).
The retrieval results show that our abstract representation can be used to search for structurally similar shapes in collections of 3D models.

\subsection{Partial Symmetry Detection}
\label{sec:partial_symmetry_detection}

\begin{figure}
    \centering
    \resizebox{3.33in}{!}{\begin{tabular}{cccc}

    \begin{overpic}[]{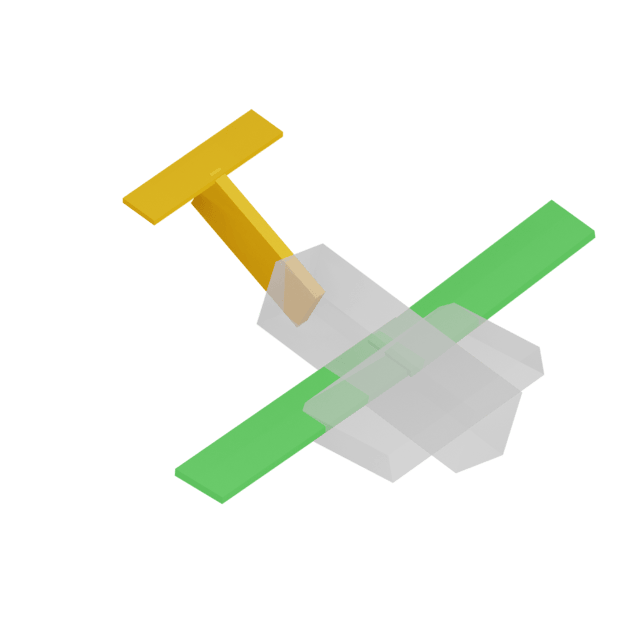}
        \put(-20,-20){\includegraphics[scale=0.8]{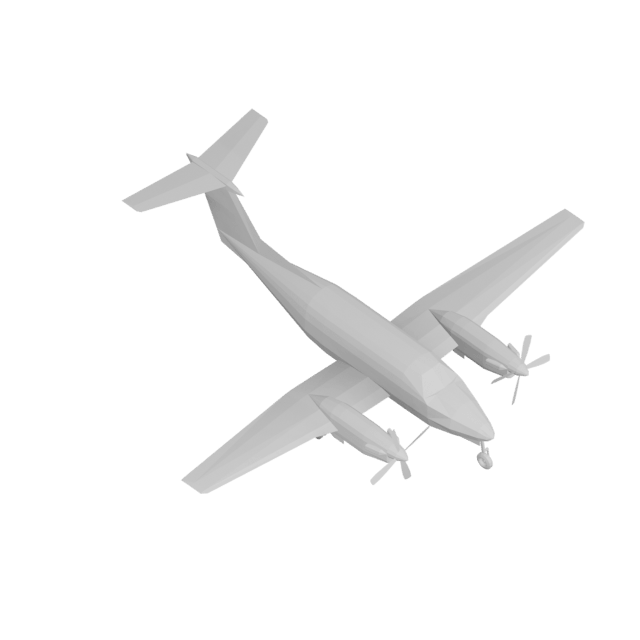}}
    \end{overpic} &
    \begin{overpic}[]{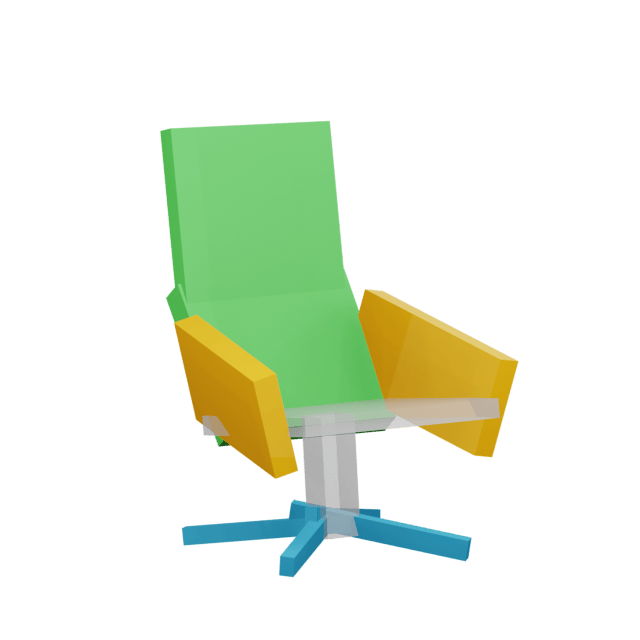}
        \put(-20,-20){\includegraphics[scale=0.7]{Figures/teaser/GroundTruth/chair/383ab6330284af461fc4ae93e00c18e5.png}}
    \end{overpic} &  
    \begin{overpic}[]{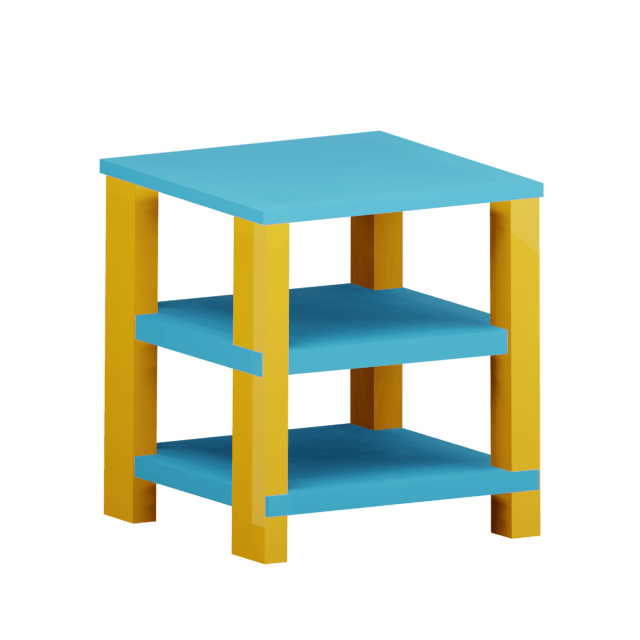}
        \put(-20,-20){\includegraphics[scale=0.7]{Figures/teaser/GroundTruth/table/667a88cc3ca1cef8f37af16b2893f1d4.png}}
    \end{overpic} & 
    \begin{overpic}[]{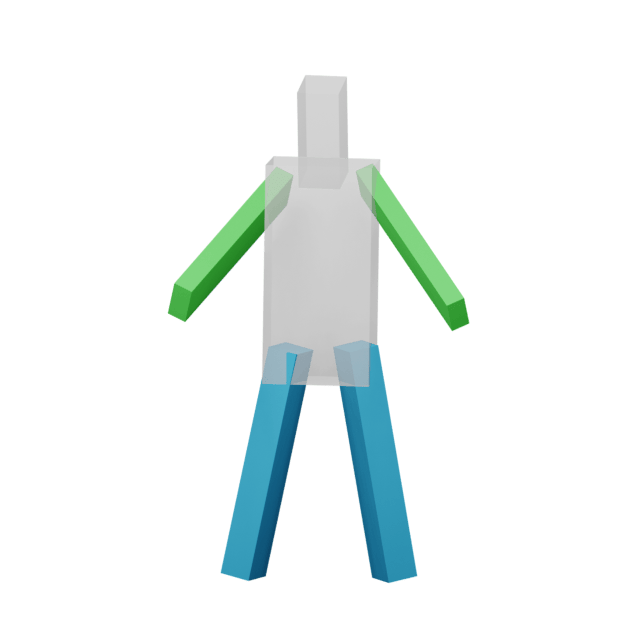}
        \put(-20,-20){\includegraphics[scale=0.7]{Figures/teaser/GroundTruth/human/32.png}}
    \end{overpic} \\

    \begin{overpic}[]{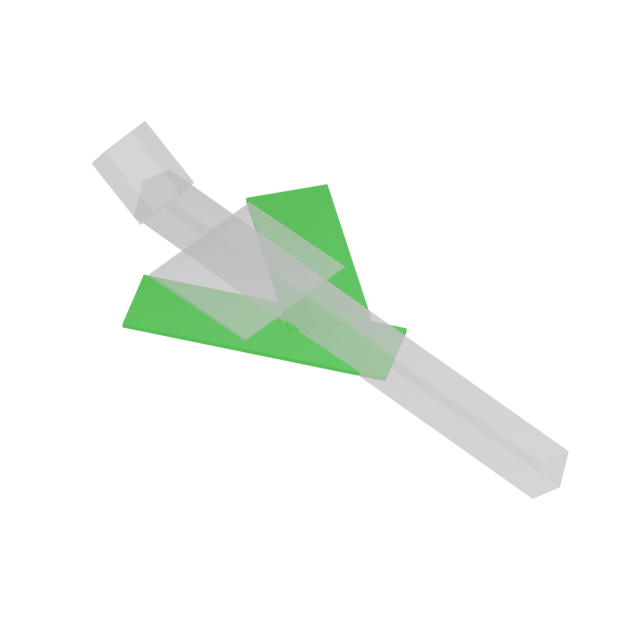}
        \put(-20,-20){\includegraphics[scale=0.8]{Figures/teaser/GroundTruth/airplane/32e7224d196e5866bd564bd76cf3cbec.png}}
    \end{overpic} &
    \begin{overpic}[]{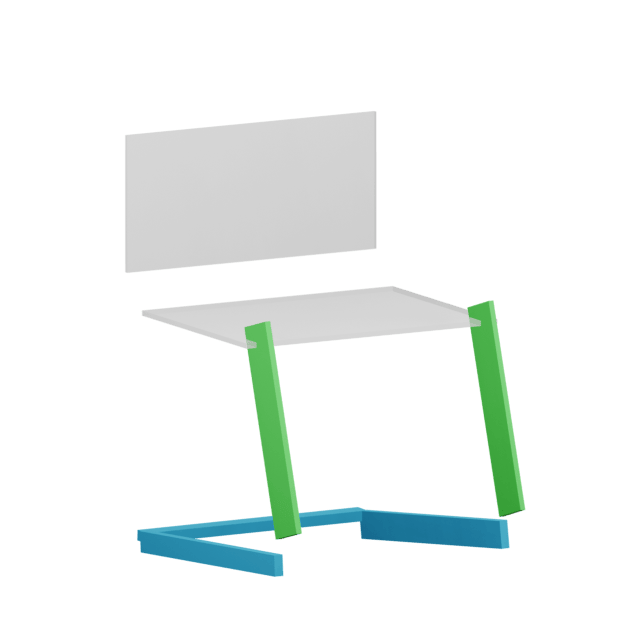}
        \put(-20,-20){\includegraphics[scale=0.7]{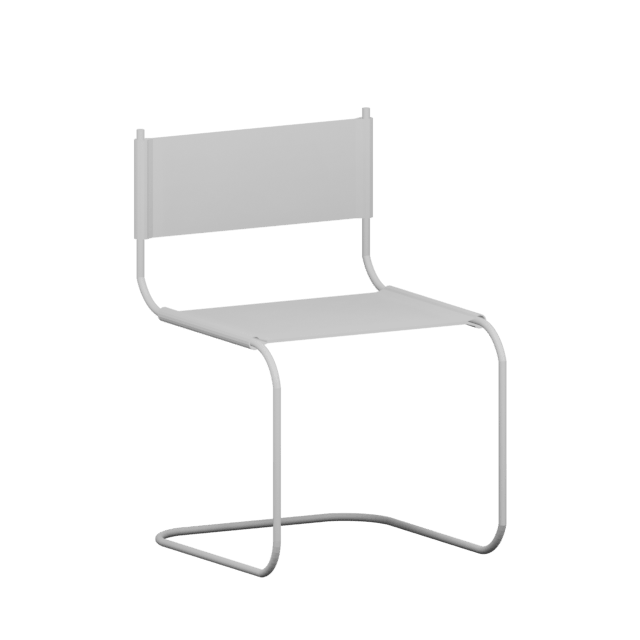}}
    \end{overpic} & 
    \begin{overpic}[]{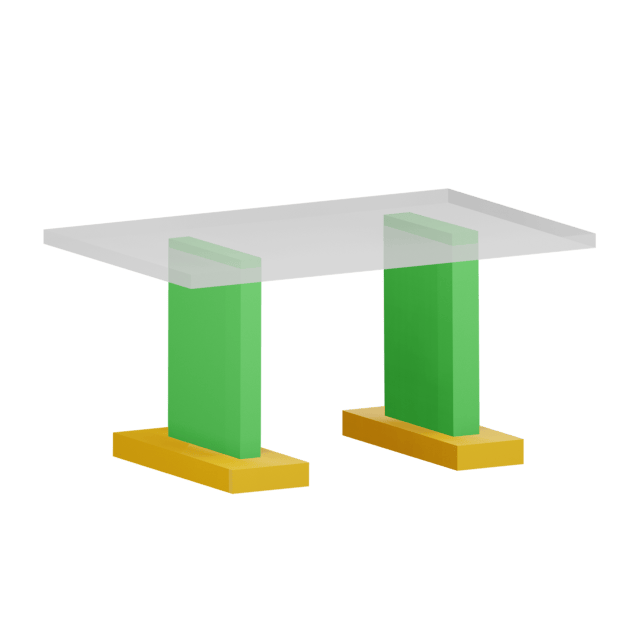}
        \put(-20,-20){\includegraphics[scale=0.7]{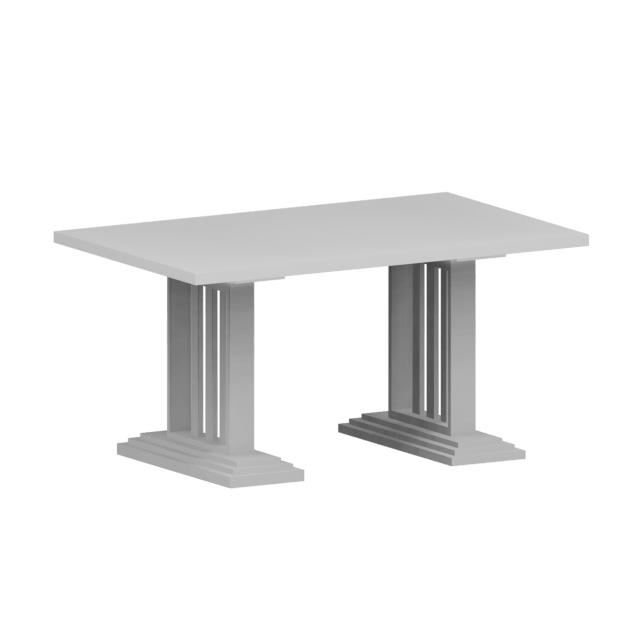}}
    \end{overpic} & 
    \begin{overpic}[]{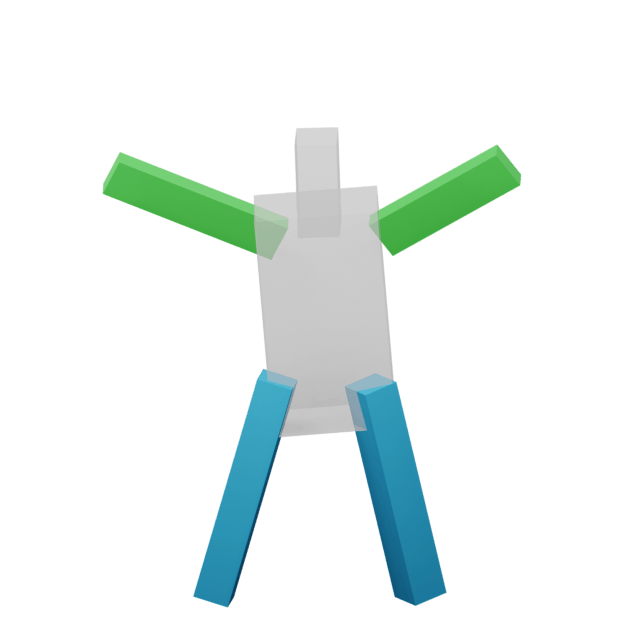}
        \put(-20,-20){\includegraphics[scale=0.7]{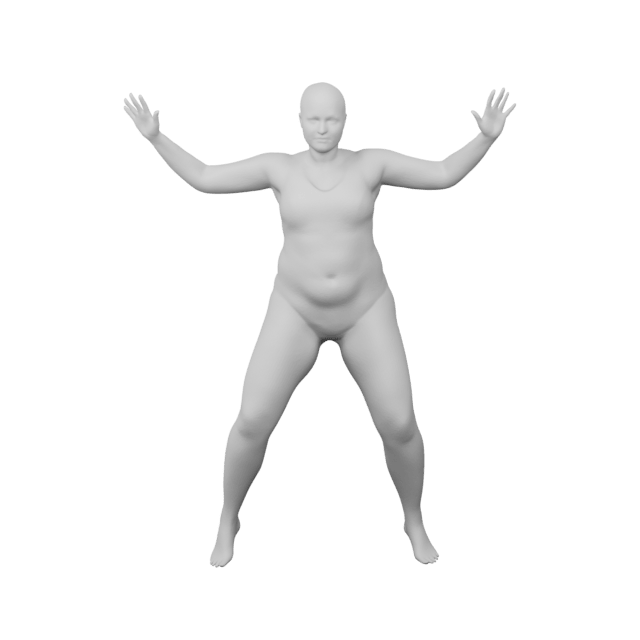}}
    \end{overpic} \\
\end{tabular}
}
    \caption{Partial symmetry detection: Using our cuboid abstraction to decompose shapes into its geometric parts enables us to perform partial symmetry detection. Colors indicate groups of symmetric parts.}
    \label{fig:symmetry_detection}
\end{figure}

Furthermore, we apply the predicted cuboid abstraction to the task of partial symmetry detection. 
Given a geometrically motivated abstraction of 3D shapes we are able to decompose them into their parts.
We do so by sampling a surface point cloud $X_S$ of each shape and extract points $X_m \subseteq X_S$ enclosed by each cuboid $C_m$.
To ensure that each cuboid encloses its corresponding surface points we add a small epsilon $\epsilon_{size}$ to thier size.
Next, we register pairs of point clouds $X_i, X_j$ on top of each other using the iterative closest points algorithm \cite{ICP_Besl} to compute a symmetry matrix $M^{sym}$.
We mark $M^{sym}_{i,j}$ as symmetric if the Chamfer distance $\Delta_{i,j}$ after the registration is below a merge threshold $\Delta_{i,j} < \theta_{sym}$.
Finally, we extract symmetry groups by computing connected components of the symmetry matrix $M^{sym}$.
We use $\theta_{sym} = 5e\text{-}4$, except for the table class where we set $\theta_{sym} = 1e\text{-}3$.
Results are visualized in \Cref{fig:symmetry_detection} with colors indicating groups of symmetric parts.
We are able to extract meaningful partial symmetries based on our cuboid shape abstraction.

\subsection{Ablation Study and Analysis}
\label{sec:ablation_study_and_analysis}

We perform an ablation study our architecture (\Cref{tab:ablation_architecture}) without primitive merging.
Due to limited computational resources we perform these studies only representatively on the plane class. 
We train each configuration five times and report the best run in terms of Chamfer distance (CD).

First, we evaluate the performance using different abstraction loss weights $\lambda_{abs}$. 
We observe that setting a larger $\lambda_{abs}$ leads to degradation of reconstruction quality as the network puts too much emphasis on the abstraction and not enough on the reconstruction of the shape at the beginning of the training leading to suboptimal starting configuration.
In contrast, setting a smaller $\lambda_{abs}$ leads to an improved reconstruction quality at the cost of compactness. 
The network focuses mainly on minimizing the reconstruction error using more primitives.

Second, we use only surface $\mathcal{L}_{surf}$ or only volume loss $\mathcal{L}_{vol}$ as reconstruction loss $\mathcal{L}_{rec}$.
Leaving out $\mathcal{L}_{vol}$ leads to degenerate primitives, which approximate the surface shape with thin lines.
In most cases the network does not recover and diverges.
Contrary, leaving out $\mathcal{L}_{surf}$ does not lead to degenerated or failed training, but the cuboids reconstruct the surface less accurately.
These findings confirm that a combination of surface reconstruction and volume preservation leads to the best results.

Third, we evaluate different cuboid target functions
\begin{equation}
    \Gamma(\phi) =
    \begin{cases}
      \kappa_{max} - (\kappa_{max} - \kappa_{min}) \cdot \phi, & \text{if}\ \text{linear}\\
      \kappa_{max} \cdot (\phi + 1)^{-4}, & \text{if}\ \text{exp}\\
      \kappa_{min} + (\kappa_{max} - \kappa_{min}) \cdot \frac{1}{2}(\cos(\pi \phi) + 1), & \text{if}\ \text{cosine}\\
      \kappa_{min} + (\kappa_{max} - \kappa_{min}) \cdot (\cos({\frac{\pi}{2} + \frac{\pi}{2} \phi}) + 1). & \text{if}\ \text{half-cos}\\
    \end{cases}
    \label{eq:target_fuction}
\end{equation}
We observe that the linear and the exponential function do not adhere to the desired number of primitives at the end of the training.
In the case of the linear function, the network has very limited time to adjust its weights in the final epochs, as the target function changes during the whole training at the same rate.
In the case of the exponential function the network is forced to discard a large number of primitives already in the first stage of training before establishing a stable initial state.
Thus, the network cannot discover shape details first and minimizes the reconstruction loss using more primitives.
The cosine and half-cosine functions behave very similar.
At the end of training the network produces the target number of primitives $\kappa_{min} = 7$ resulting in the best performance in terms of CD and IoU.

Fourth, we perform an ablation of our architecture and the cuboid latents $l_c$.
In the first case we directly concatenate $l_c$ with the local features.
In the second case we do not use $l_c$ at all and define our cuboid features $f_c$ as all output tokens of the ViT.
In both cases we consequently use only one large ViT with twice the number of layers.
We observe that the training without using $l_c$ explicitly diverges.
In the case of [$p_c, f_l$] 80\% of runs did not converge, otherwise the results were similar.
We argue, that our architecture design helps the network to learn a global description of the shape first.
Using learnarble cuboid latents $l_c$ allows the network to learn a template of the whole collection which it modifies based on the input conditions.
This enables us to learn a structurally consistent and geometrically meaningful abstraction across the whole dataset.


\begin{table}[]
    \centering
    \newcommand{\graycell}{\cellcolor[HTML]{EFEFEF}}
\newcommand{\redcell}{\cellcolor[HTML]{FAD1D0}}

\centering
\resizebox{.48\textwidth}{!}{

\begin{tabular}{c|ccccc}
\toprule
\multirow{2}{*}{}   & \textbf{Abstr.}   & \multicolumn{2}{c}{\textbf{Reconstr.}}    & \multicolumn{2}{c}{\textbf{Co-Segment.}}  \\
                    & \textbf{Num↓}    & \textbf{CD↓}  & \textbf{IoU\%↑}           & \textbf{mAcc\%↑}  & \textbf{mIoU\%↑}      \\
\midrule
$\lambda_{abs} = 1e\text{-}2$    & 7.00   & 0.026 & 58.6  & 88.1  & 67.2  \\
$\lambda_{abs} = 1e\text{-}4$    & 10.00  & 0.023 & 62.6  & 85.7  & 66.6  \\
\midrule
$\mathcal{L}_{rec} = \mathcal{L}_{surf}$    & 17.3  & 0.058 & 0.0   & 80.3  & 64.6  \\
$\mathcal{L}_{rec} = \mathcal{L}_{vol}$     & 7.00  & 0.032 & 51.3  & 72.3  & 53.6  \\
\midrule
$\Gamma(\phi) = $ linear    & 8.64  & 0.029 & 52.1  & 81.9  & 61.8  \\
$\Gamma(\phi) = $ exp       & 7.98  & 0.026 & 58.9  & 73.1  & 57.6  \\
$\Gamma(\phi) = $ cosine    & 7.00  & 0.025 & 60.3  & 81.3  & 66.9  \\
\midrule
w/o $l_c$         & 37.32  & 0.028 & 28.8  & 35.3  & 20.1  \\
$[p_c, f_l]$      & 7.00  & 0.026 & 57.5  & 85.2  & 64.9  \\
\midrule
baseline \graycell         & 7.00 \graycell & 0.024 \graycell & 61.1 \graycell & 81.8 \graycell & 66.9 \graycell \\
\bottomrule
\end{tabular}

} 
    \caption{Ablation study using different configuration of abtraction loss weight $\lambda_{abs}$, reconstruction loss $\mathcal{L}_{rec}$, cuboid target function $\Gamma(\phi)$, and architecture. The baseline model uses $\lambda_{abs} = 1e\text{-}3$, $\mathcal{L}_{rec} = \mathcal{L}_{surf} + \mathcal{L}_{vol}$, $\Gamma(\phi) =$ half-cos and $[l_c, f_g]$. All metrics evaluated on the plane class.}
    \label{tab:ablation_architecture}
\end{table}

Last, we analyze the influence of the merge threshold $\theta_{merge}$ on the cuboid reconstruction quality.
We report metrics in \Cref{tab:ablation_merge} evaluated on the plane and chair class.
Using a smaller value for $\theta_{merge}$ leads to merging of more cuboids.
Using less primitives results naturally in a less accurate reconstruction.
Especially, when we merge only highly overlapping cuboids (large $\theta_{merge}$) the reconstruction quality degrades only marginally.
This can be attributed to structurally redundant cuboids.
Overlapping cuboids can more easily satisfy the reconstruction loss $\mathcal{L}_{rec}$ when approximating non-square geometry, but they are structurally redundant from an abstraction point of view and should be merged.

\begin{table}[]
    \centering
    \newcommand{\graycell}{\cellcolor[HTML]{EFEFEF}}
\newcommand{\redcell}{\cellcolor[HTML]{FAD1D0}}

\begin{tabular}{c|ccc|ccc}
\toprule
\multirow{2}{*}{\textbf{$\theta_{merge}$}} 
& \multicolumn{3}{c}{\textbf{plane}}               
& \multicolumn{3}{c}{\textbf{chair}}  \\
                                        
                                         & \textbf{Num↓}    & \textbf{CD↓}  & \textbf{IoU\%↑}  & \textbf{Num↓}    & \textbf{CD↓}  & \textbf{IoU\%↑}       \\
\midrule
1.0  & 5.33 & 0.031 & 48.9    & 7.03  & 0.040 & 52.3  \\
1.1  & 5.75 & 0.028 & 53.5    & 7.49  & 0.038 & 53.7  \\
1.2  & 6.03 & 0.026 & 56.0    & 7.79  & 0.038 & 54.5  \\
1.3  & 6.22 & 0.026 & 58.1    & 8.07  & 0.037 & 55.1  \\
1.4  & 6.35 & 0.025 & 58.6    & 8.37  & 0.036 & 54.9  \\    
1.5  & 6.54 & 0.025 & 58.8    & 8.78  & 0.036 & 55.5  \\
\midrule
w/o  & 7.00 & 0.024 & 61.1    & 10.00  & 0.035 & 58.8  \\
\bottomrule
\end{tabular}
    \caption{Comparison of reconstruction quality using different merge threshold $\theta_{merge}$ values. All metrics evaluated on the plane and chair class.}
    \label{tab:ablation_merge}
\end{table}

\subsection{Limitations and Future Work}
\label{sec:failure_cases_and_future_work}

Similar to previous work we assume our dataset to be axis-aligned. 
This allows the network to learn the position and shape of common parts, but does not ensure that the network is aware of the relationships between them.
Introducing rotational-invariance to the prediction of primitives would enable the network to be more approachable in real-world scenarios.

Although hybrid-approaches combining learning-based and classical shape analysis techniques are a valid choice, formulating primitive merging in terms of a loss function would enable to predict the abstraction in a single inference step without additional post-processing.
Using a binary encoding to track ancestors of merged cuboids introduces novel cuboid indices. 
In the worst case this can lead to $2^n$ unique indices, where $n = \kappa_{min}$.
Re-indexing these indices by clustering geometrically similar primitives based on the cuboid parameters $[r, t, s, \gamma]$ could improve the performance on the co-segmentation benchmark.
We leave this to future work.

Another line of work could extend our method from cuboids to super-quadrics by predicting two additional parameters or deformable templates by extending the decoder network further.
It would be highly interesting to see an extension of this work towards 3D shape generative models conditioned on automatically predicted cuboid primitives instead of manually annotated ones.


\section{CONCLUSION}
\label{sec:conclusion}

In this paper we propose a novel fine-to-coarse unsupervised method for cuboid shape abstraction. 
Our loss formulation allows to approximate a collection of shapes with an extensive number of primitives first and continually reduce their count to form an abstraction second.
We show that preserving distinct geometric details is easier than learning them using a limited number of primitives from the beginning.
We compare with previous work on common structural shape reconstruction and co-segmentation benchmarks to demonstrate state-of-the-art performance.
Furthermore, we apply our abstraction to downstream task, like shape clustering, retrieval or partial symmetry detection.

\begin{acks}
We would like to thank Julius Nehring-Wirxel, Alexandra Heuschling, Zain Selman and Lin Gao for helpful discussions during the project.
\end{acks}

\bibliographystyle{ACM-Reference-Format}
\bibliography{references}


\appendix

\begin{figure*}[hpt]
    \centering
    \begin{tabular}{c}
        \rotatebox[origin=c]{90}{\hspace{0.6cm} GT \hspace{0.6cm}} \\
        \rotatebox[origin=c]{90}{\hspace{0.6cm} HCA \hspace{0.6cm}} \\
        \rotatebox[origin=c]{90}{\hspace{0.6cm} CAS \hspace{0.6cm}} \\
        \rotatebox[origin=c]{90}{\hspace{0.2cm} $\text{DPF}_{PPM}$ \hspace{0.2cm}} \\
        \rotatebox[origin=c]{90}{\hspace{0.6cm} Ours \hspace{0.6cm}} \\
    \end{tabular}
    \resizebox{0.96\textwidth}{!}{\begin{tabular}{ccccccccc}
    \includegraphics[]{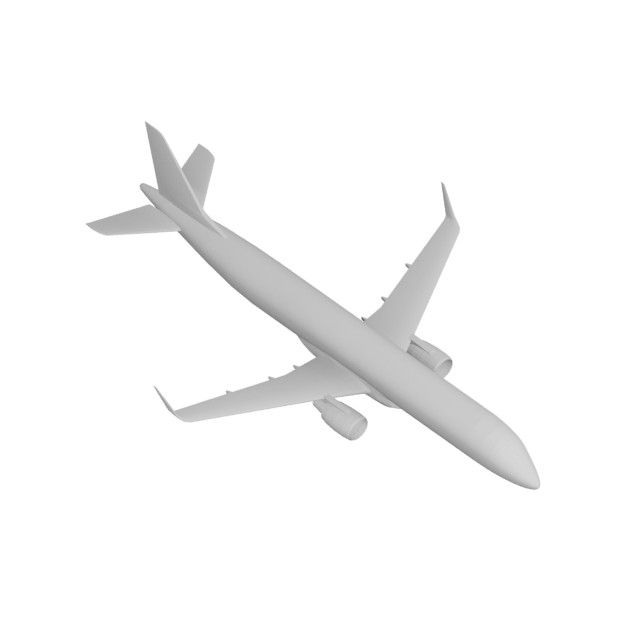} &
    \includegraphics[]{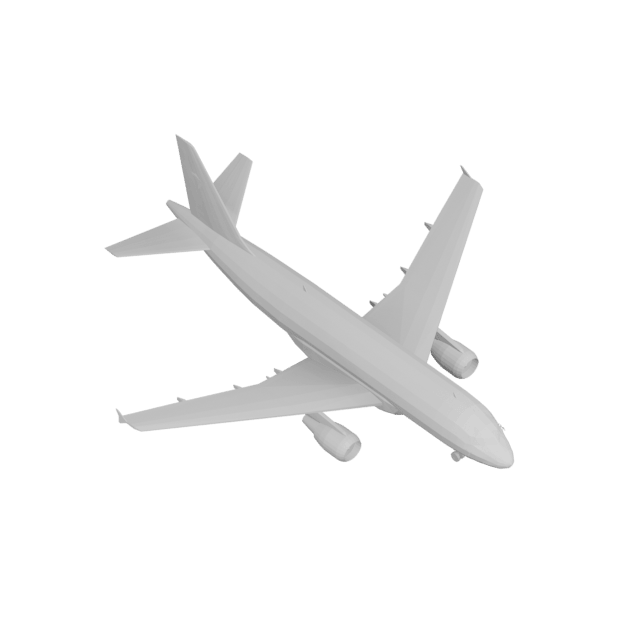} & 
    \includegraphics[]{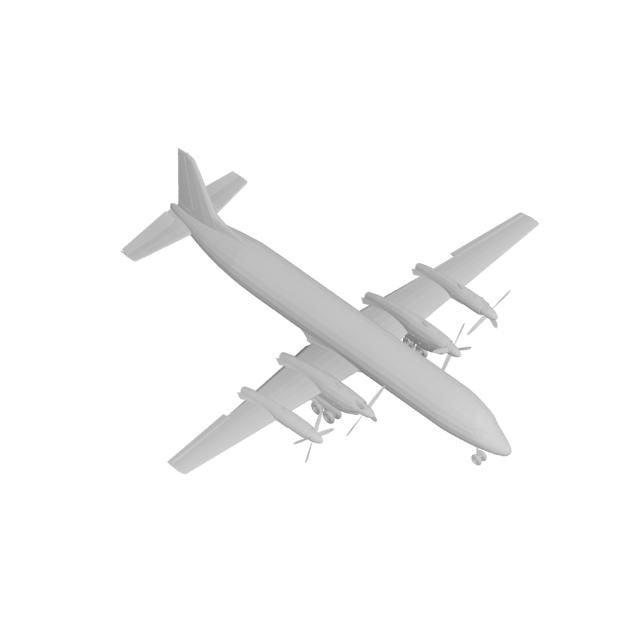} &
    
    \includegraphics[]{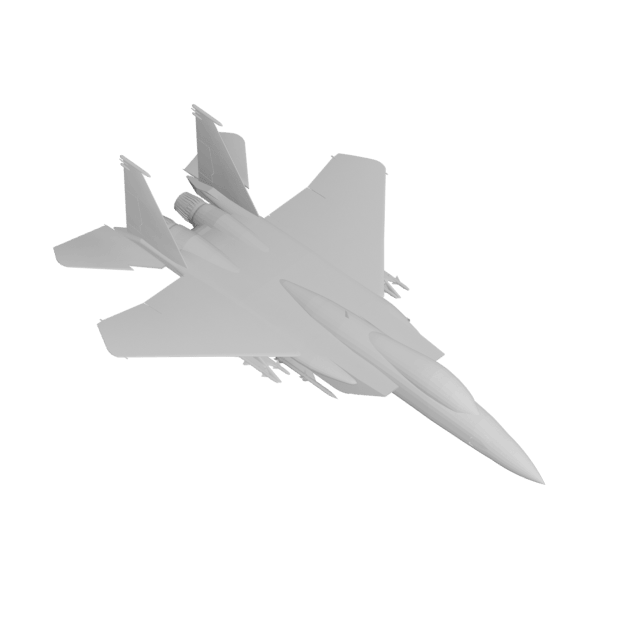} & 
    \includegraphics[]{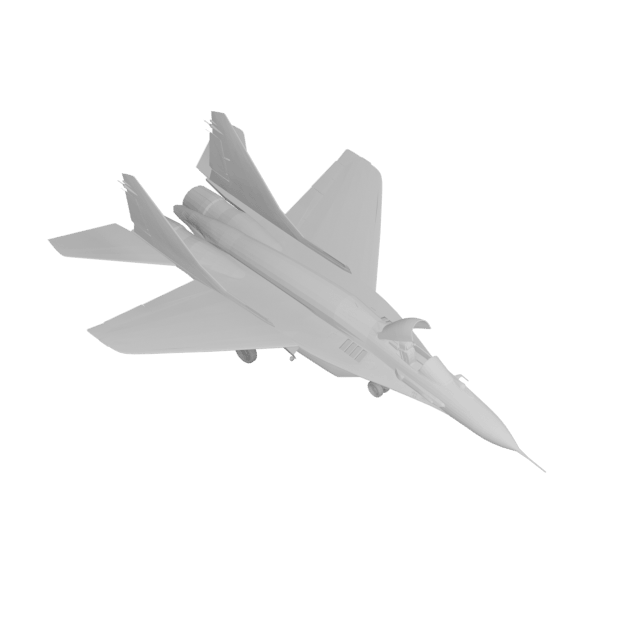} & 
    \includegraphics[]{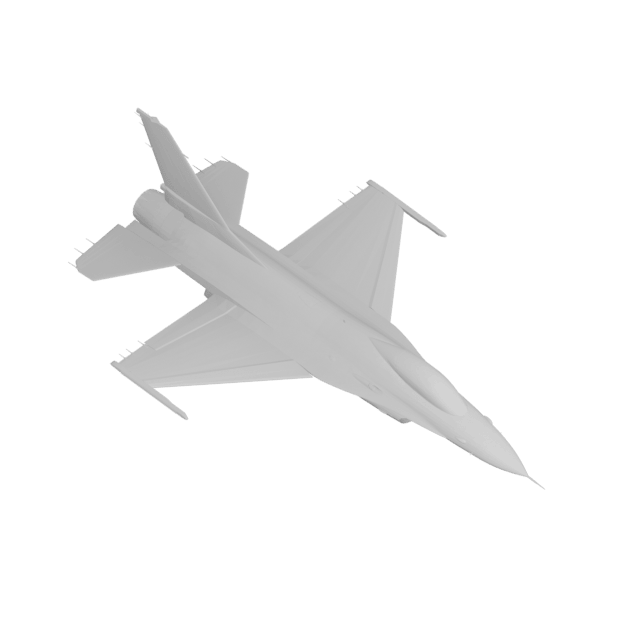} &
    
    \includegraphics[]{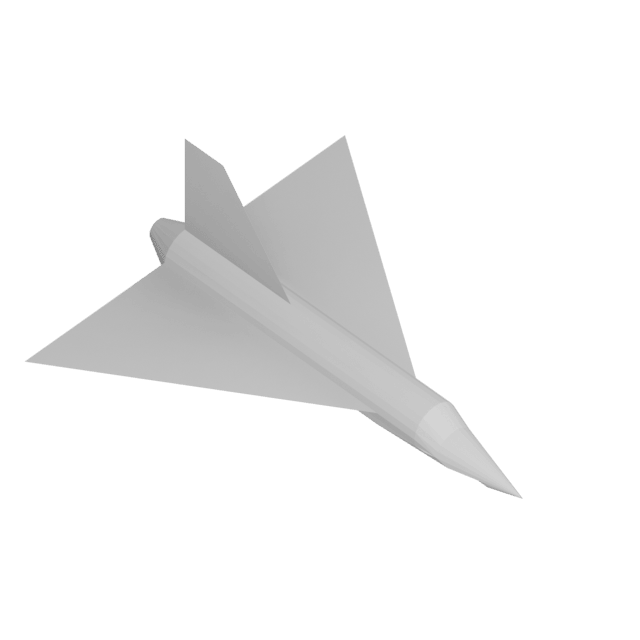} & 
    \includegraphics[]{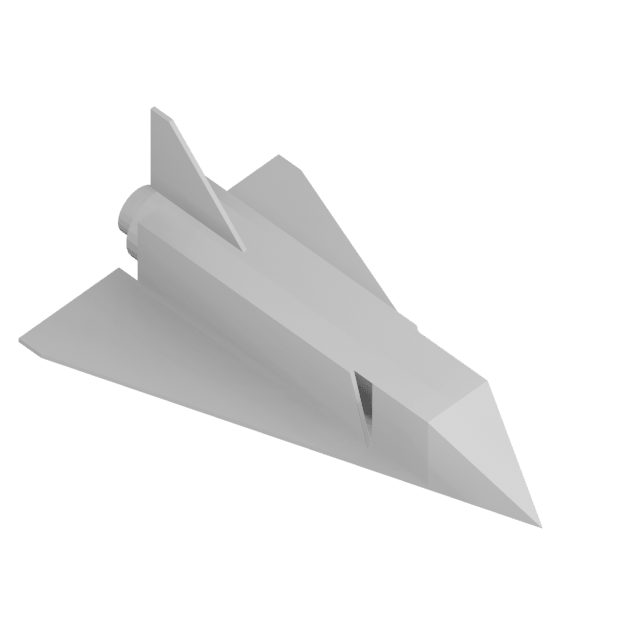} & 
    \includegraphics[]{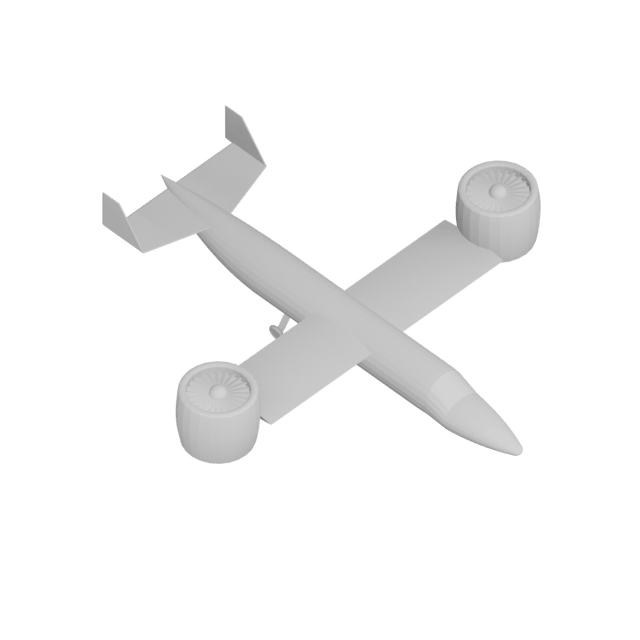} \\

    \includegraphics[]{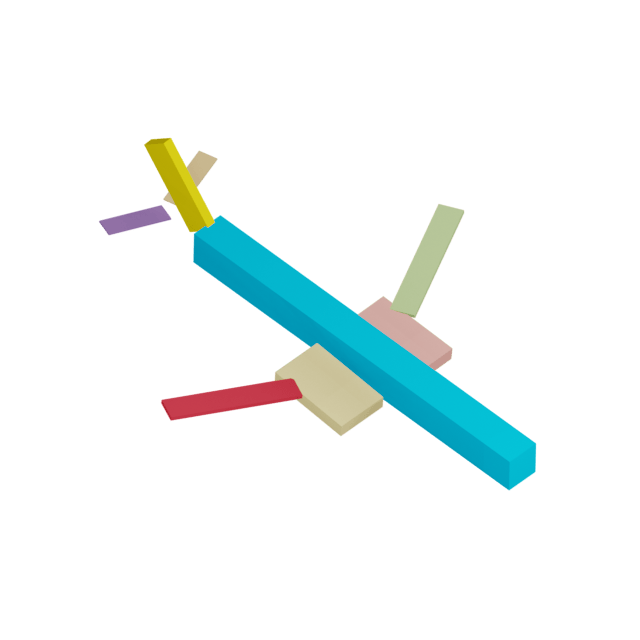} &
    \includegraphics[]{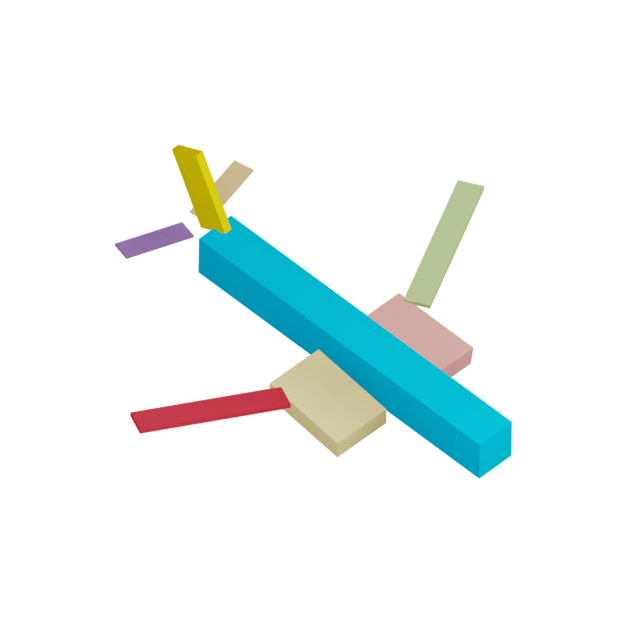} & 
    \includegraphics[]{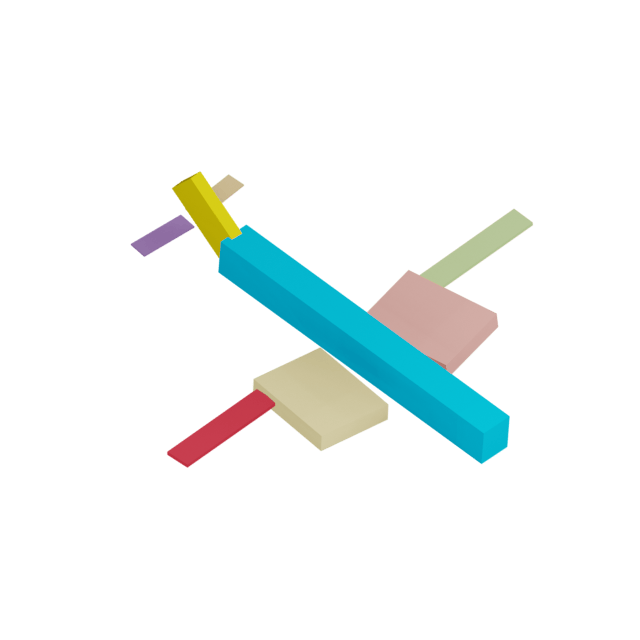} &
    
    \includegraphics[]{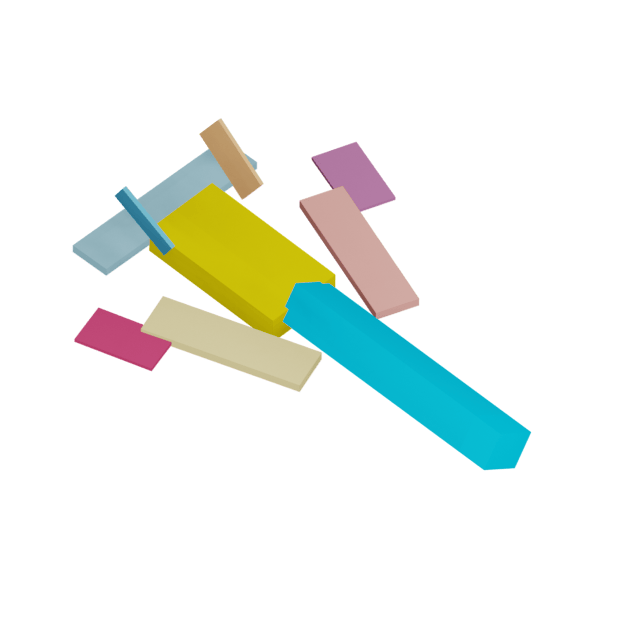} & 
    \includegraphics[]{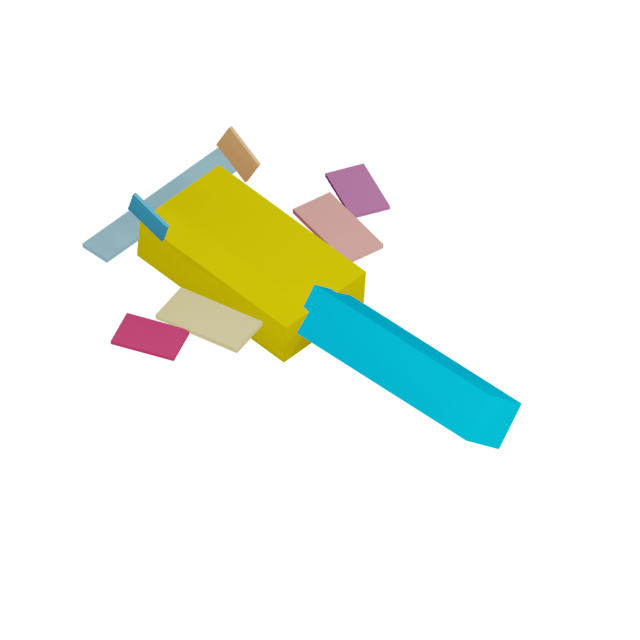} & 
    \includegraphics[]{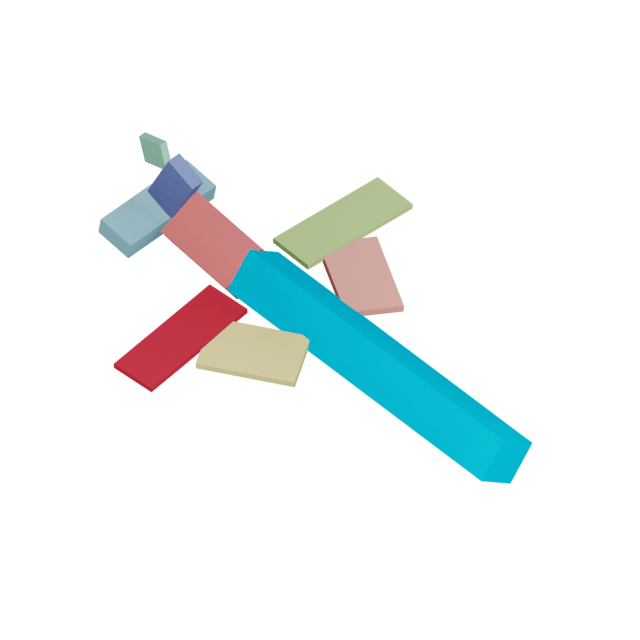} &
    
    \includegraphics[]{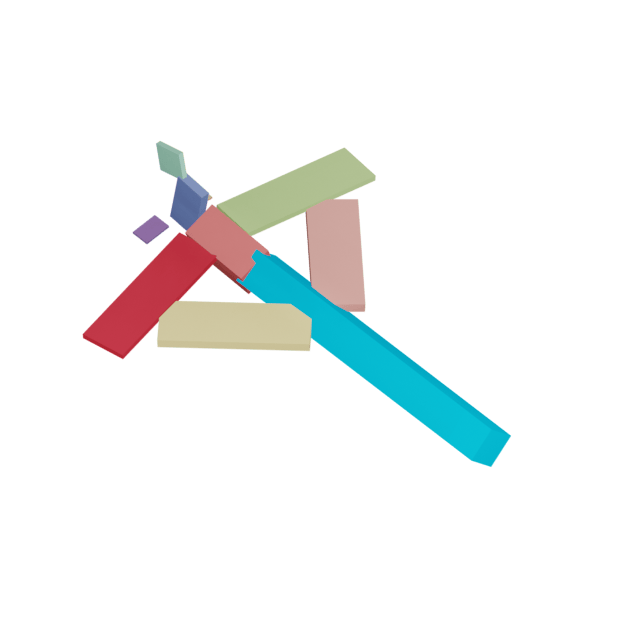} & 
    \includegraphics[]{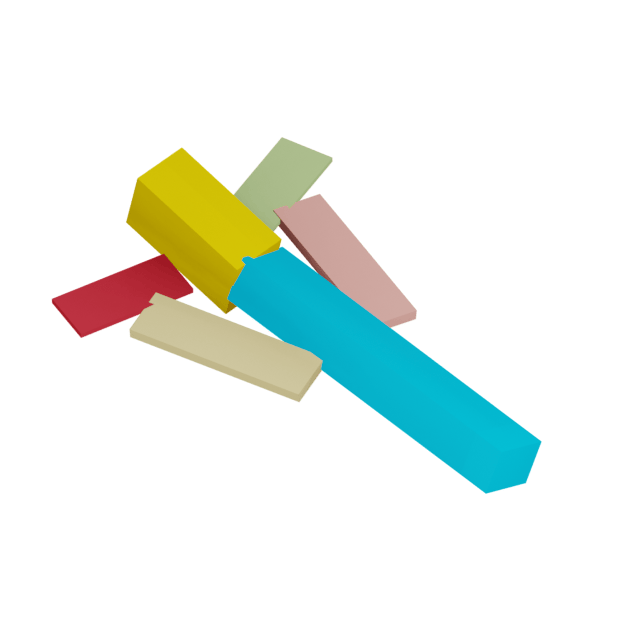} & 
    \includegraphics[]{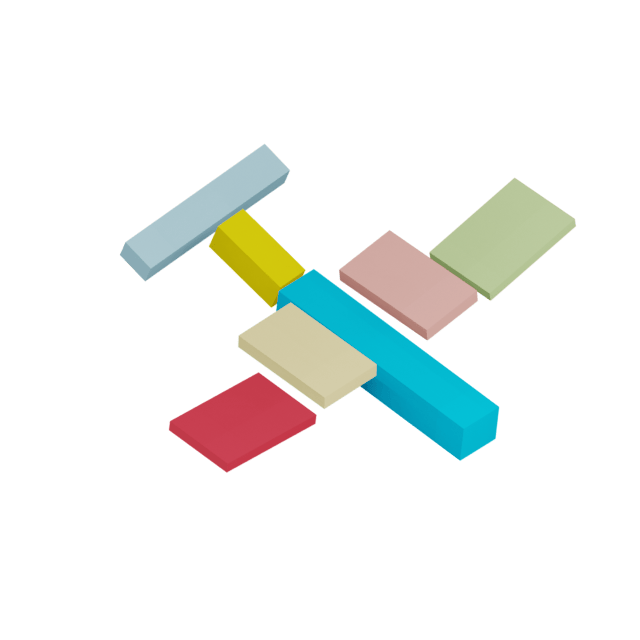} \\

    \includegraphics[]{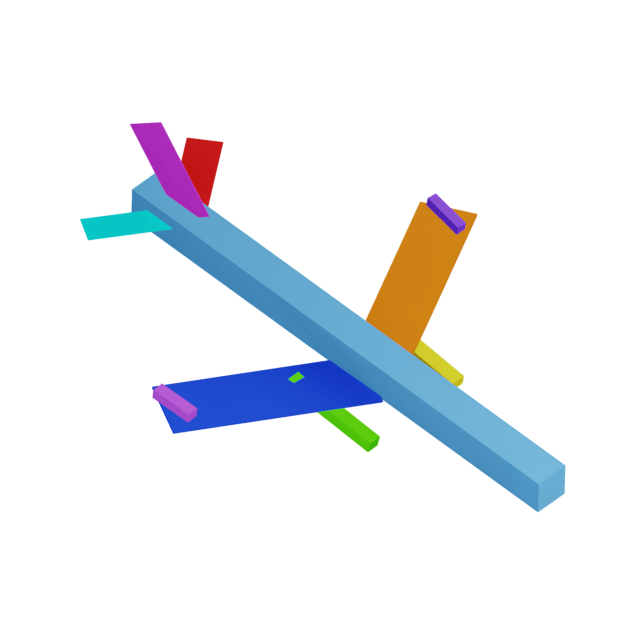} &
    \includegraphics[]{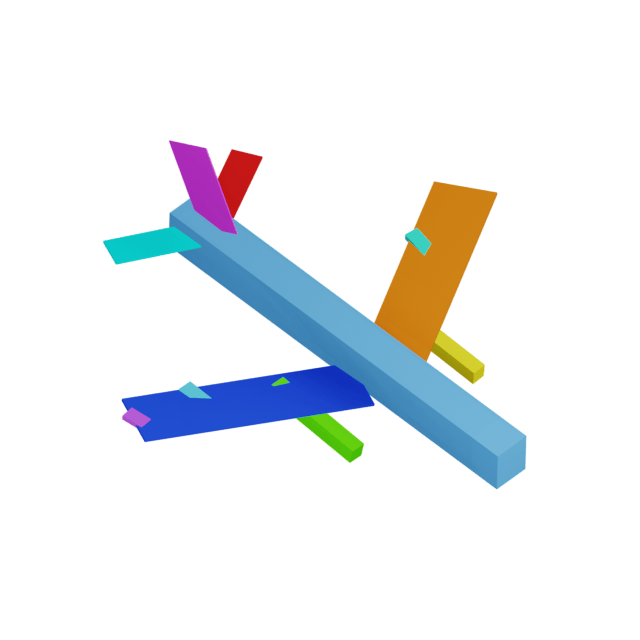} & 
    \includegraphics[]{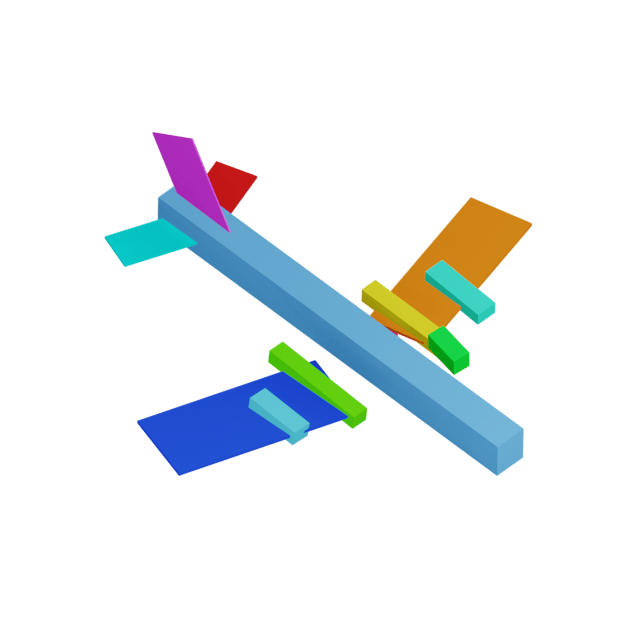} &
    
    \includegraphics[]{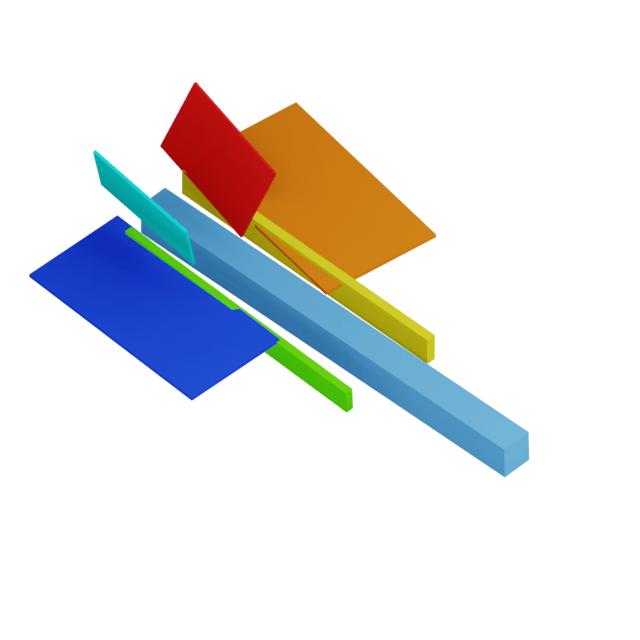} & 
    \includegraphics[]{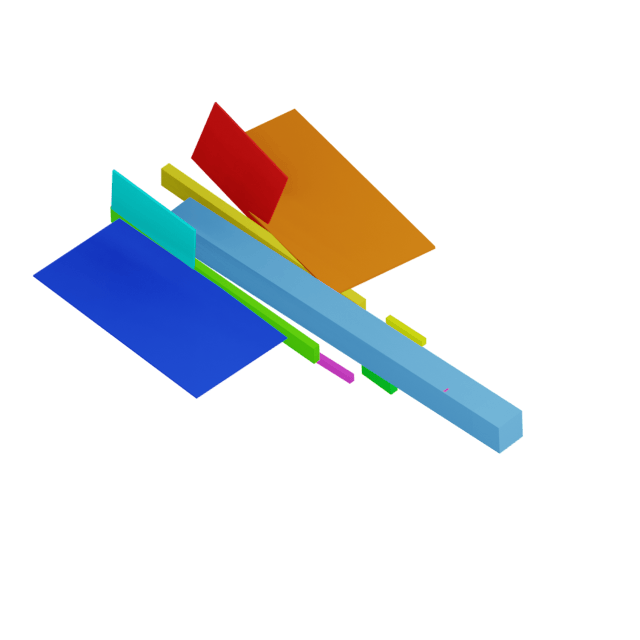} & 
    \includegraphics[]{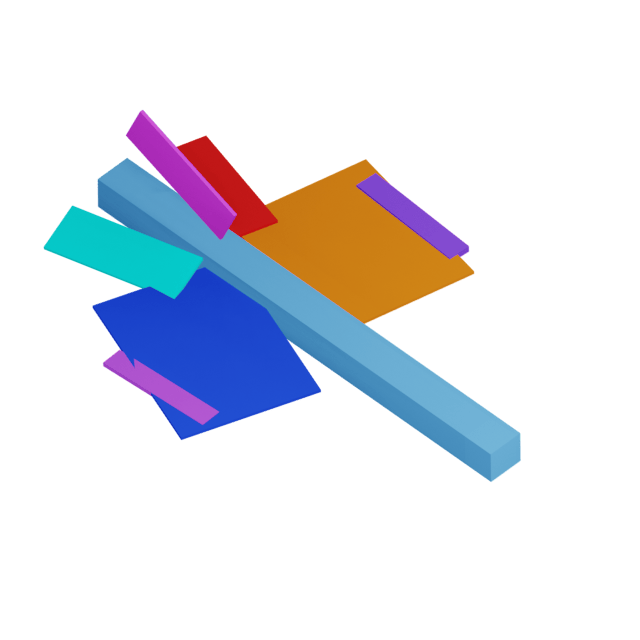} &
    
     \includegraphics[]{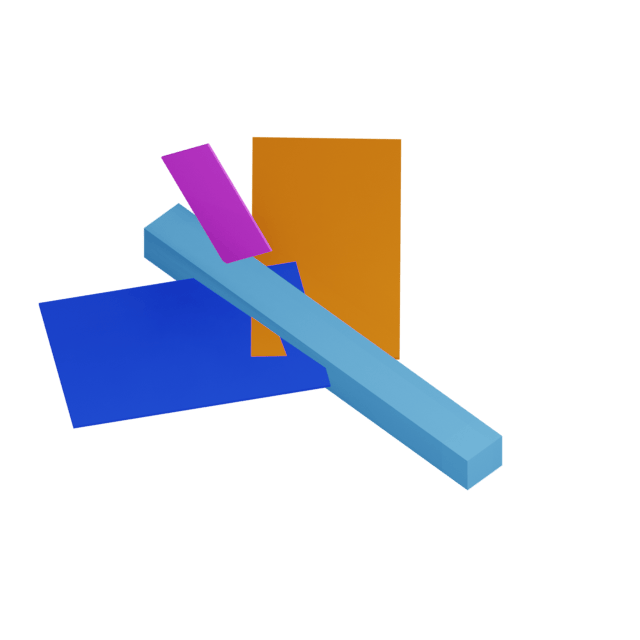} & 
     \includegraphics[]{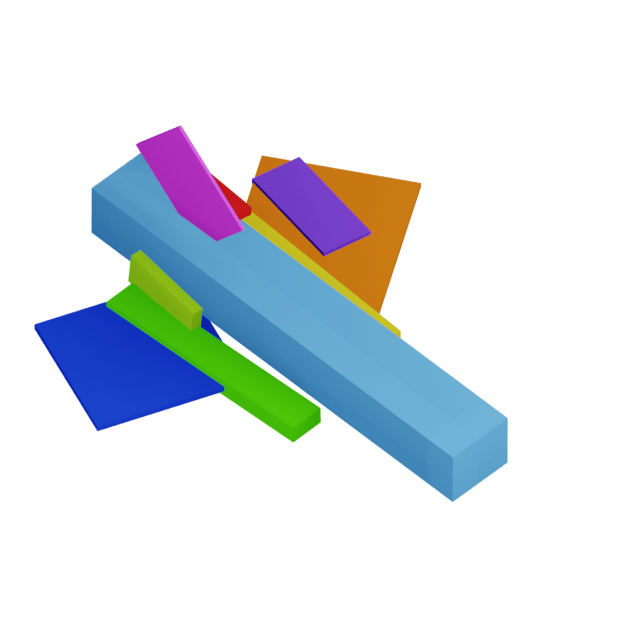} & 
     \includegraphics[]{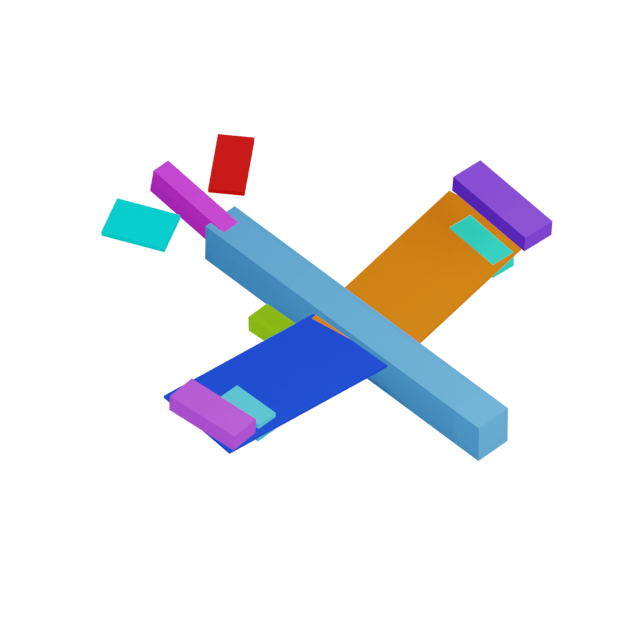} \\

    \includegraphics[]{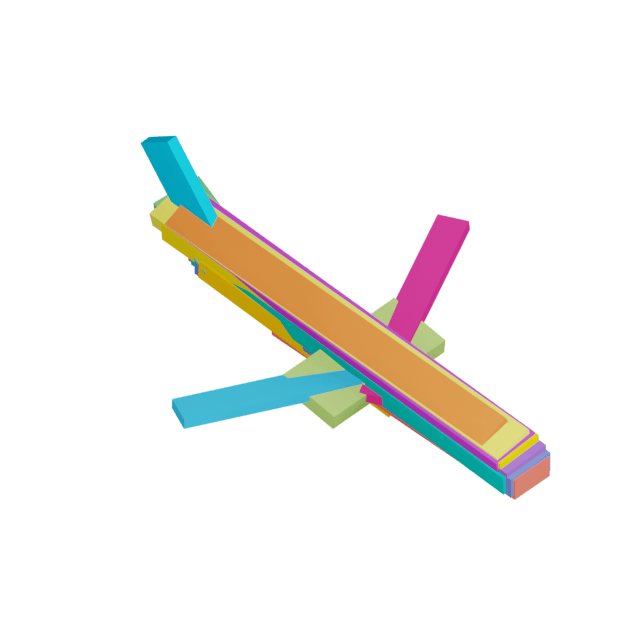} &
    \includegraphics[]{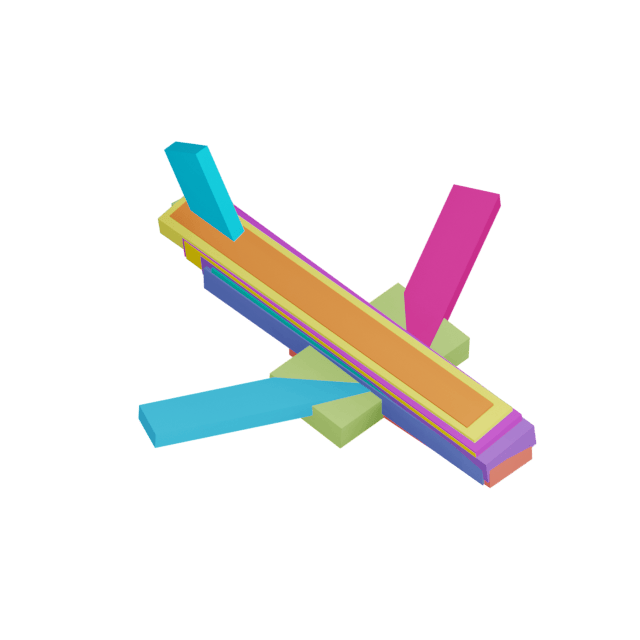} & 
    \includegraphics[]{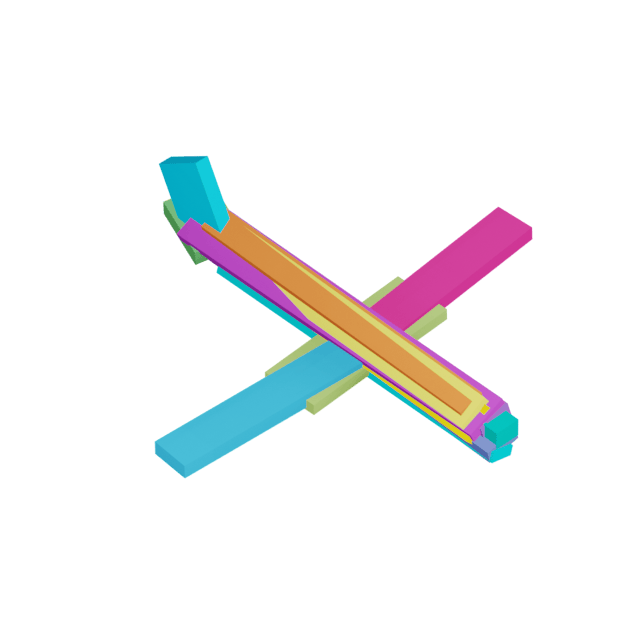} &
    
    \includegraphics[]{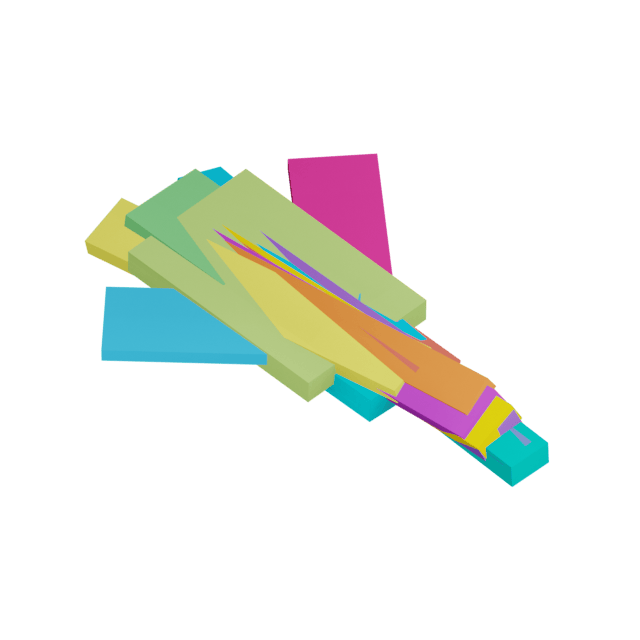} & 
    \includegraphics[]{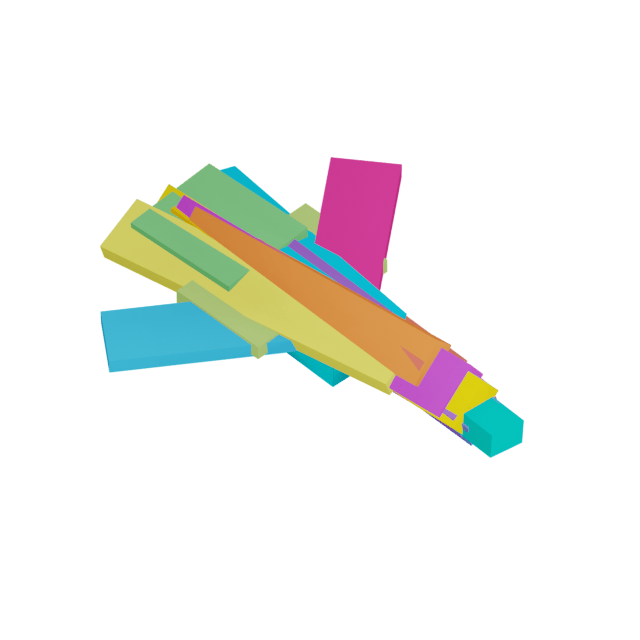} & 
    \includegraphics[]{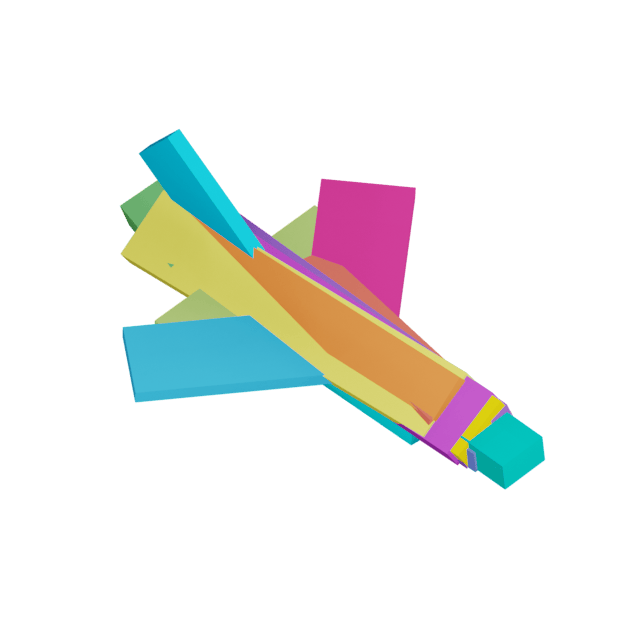} &
    
    \includegraphics[]{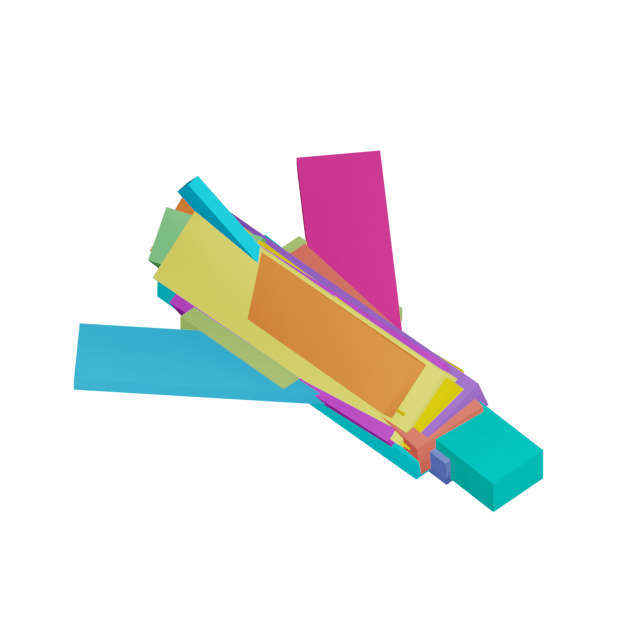} & 
    \includegraphics[]{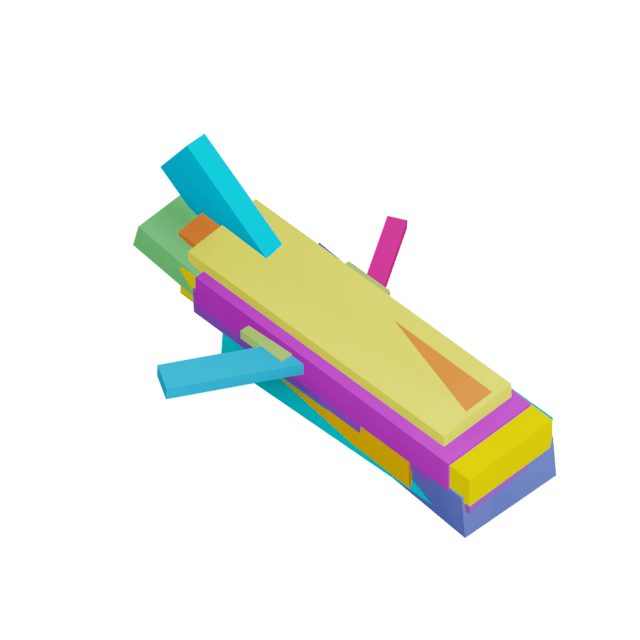} & 
    \includegraphics[]{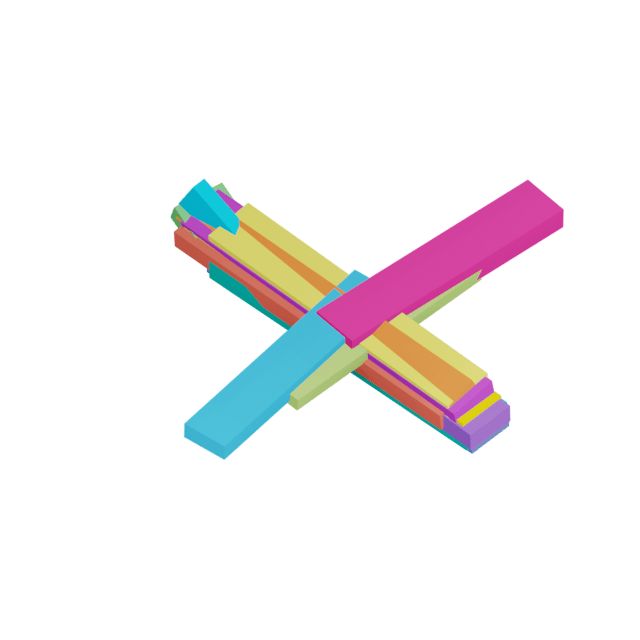} \\

    \includegraphics[]{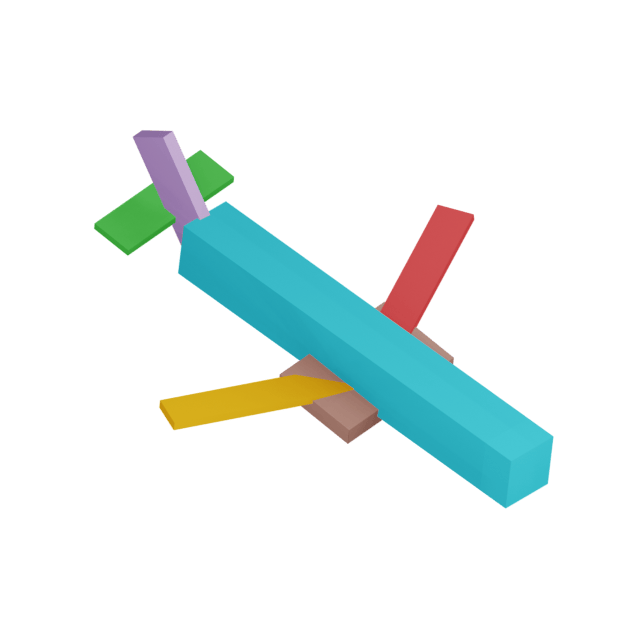} &
    \includegraphics[]{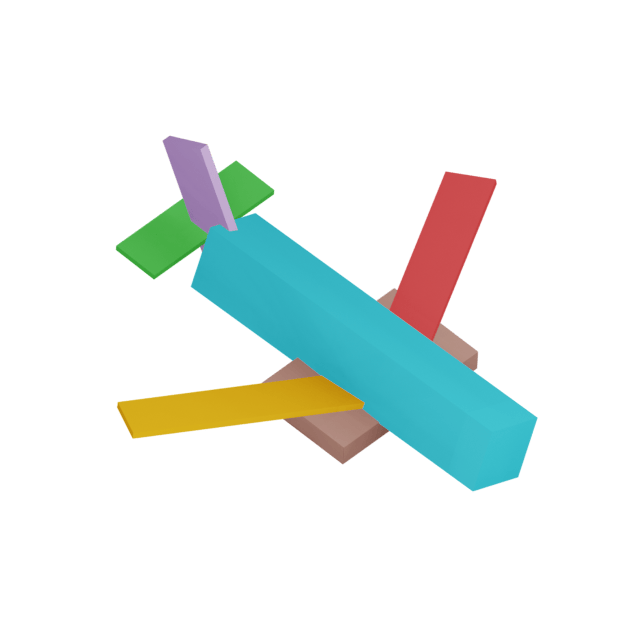} & 
    \includegraphics[]{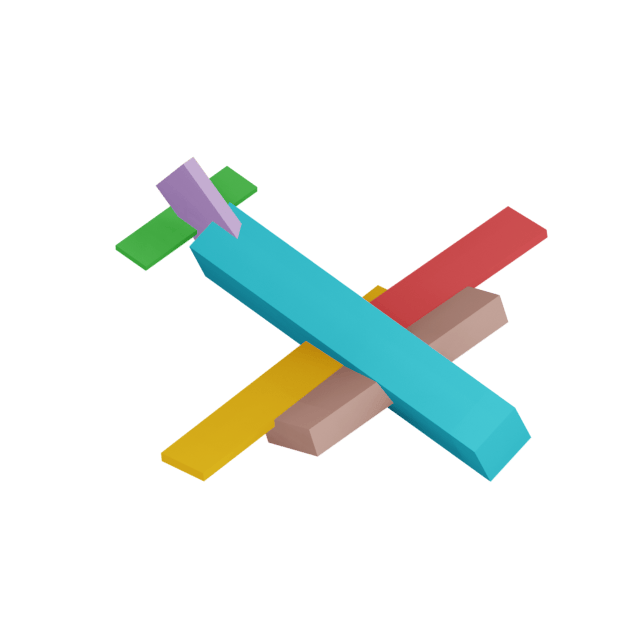} &
    
    \includegraphics[]{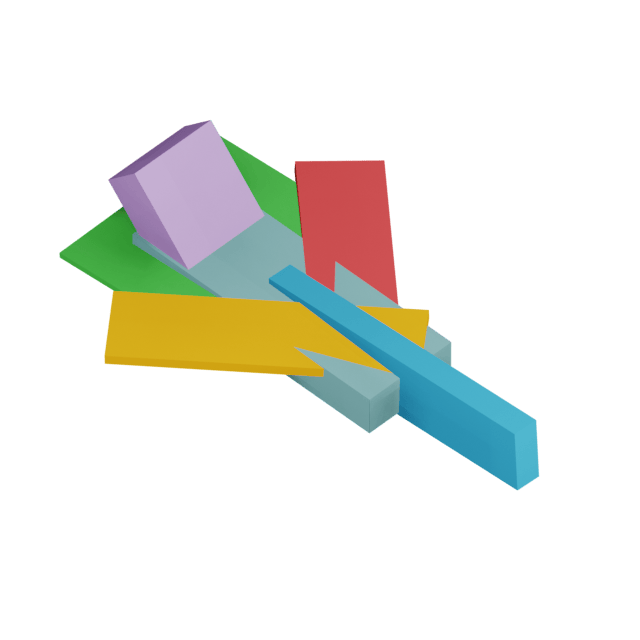} & 
    \includegraphics[]{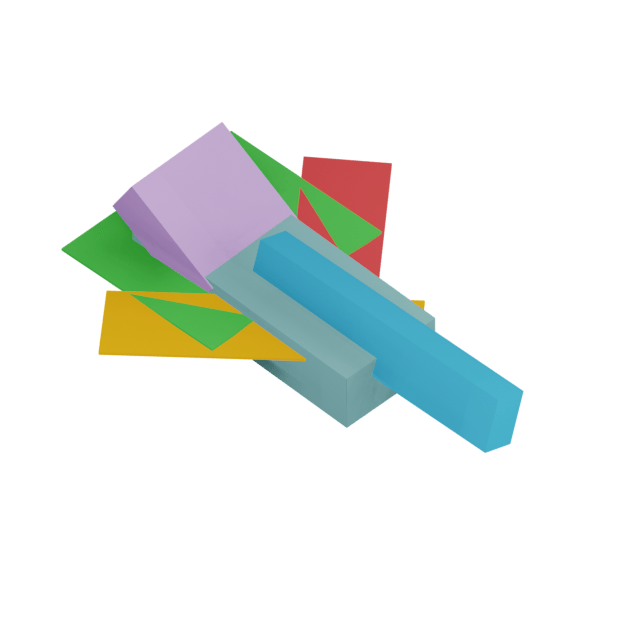} & 
    \includegraphics[]{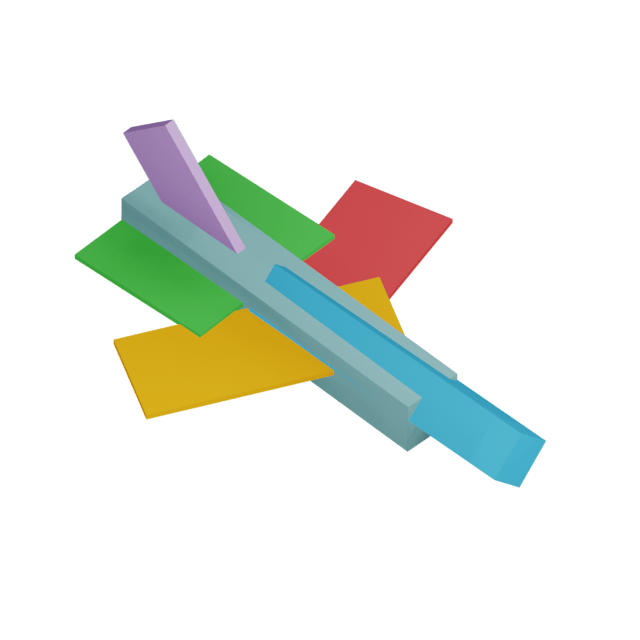} &
    
    \includegraphics[]{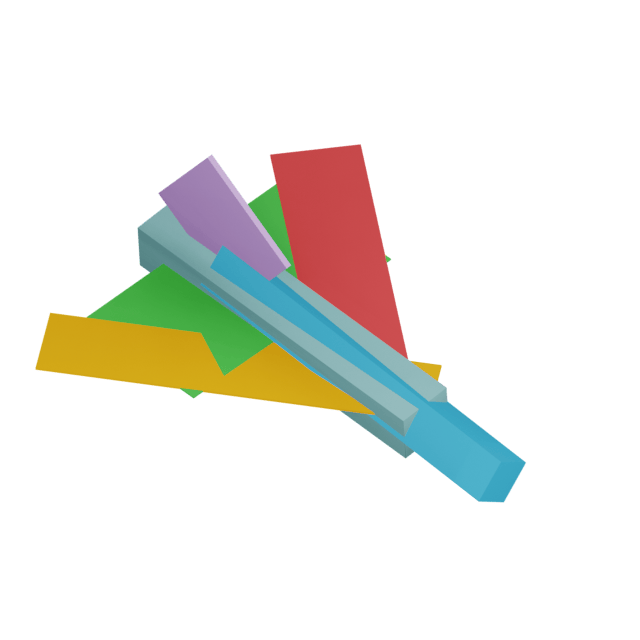} & 
    \includegraphics[]{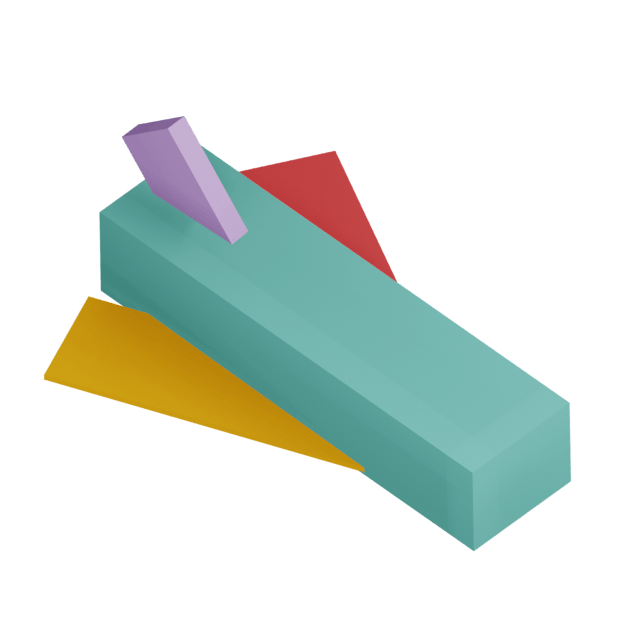} & 
    \includegraphics[]{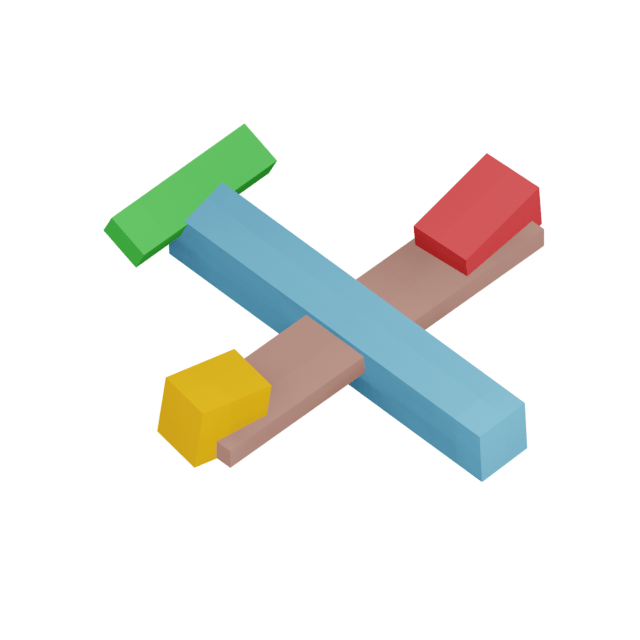} \\
 
\end{tabular}

    }
    
    \rule{0.9\textwidth}{0.4pt}\\
    
    \begin{tabular}{c}
        \rotatebox[origin=c]{90}{\hspace{0.6cm} GT \hspace{0.6cm}} \\
        \rotatebox[origin=c]{90}{\hspace{0.6cm} HCA \hspace{0.6cm}} \\
        \rotatebox[origin=c]{90}{\hspace{0.6cm} CAS \hspace{0.6cm}} \\
        \rotatebox[origin=c]{90}{\hspace{0.2cm} $\text{DPF}_{PPM}$ \hspace{0.2cm}} \\
        \rotatebox[origin=c]{90}{\hspace{0.6cm} Ours \hspace{0.6cm}} \\
    \end{tabular}
    \resizebox{0.96\textwidth}{!}{\begin{tabular}{ccccccccc}
    \includegraphics[]{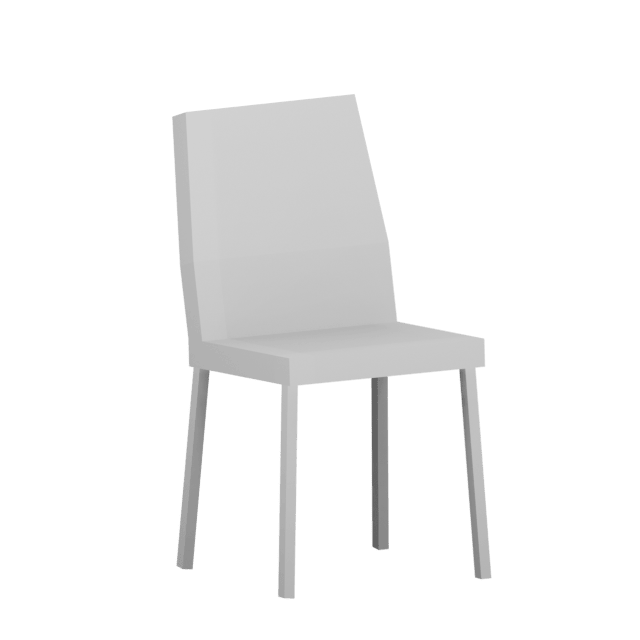} &
    \includegraphics[]{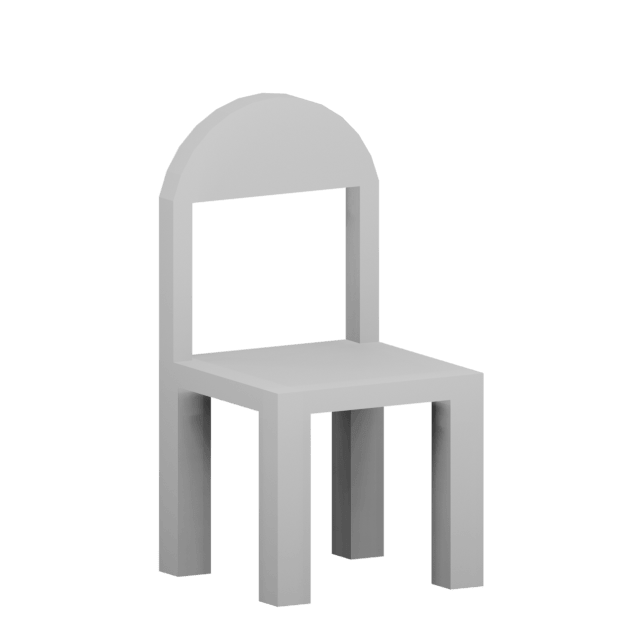} & 
    \includegraphics[]{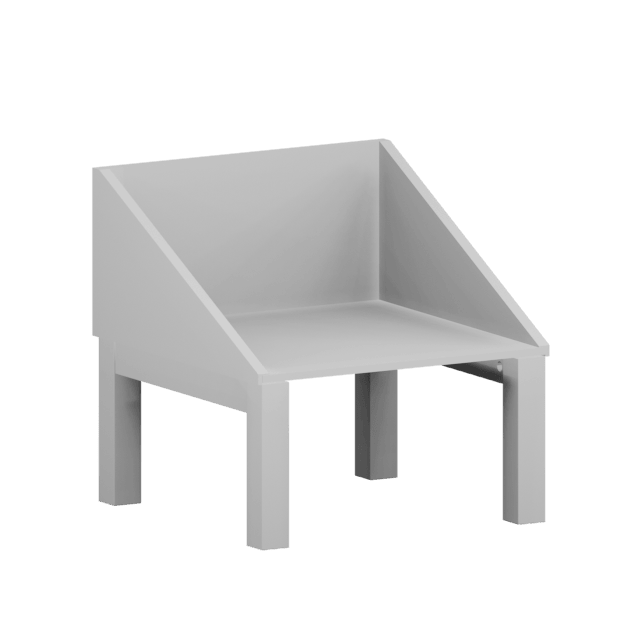} &

    \includegraphics[]{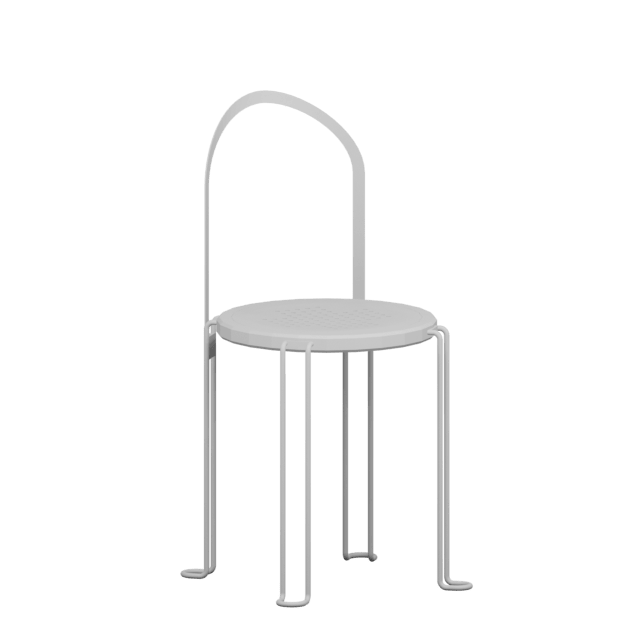} & 
    \includegraphics[]{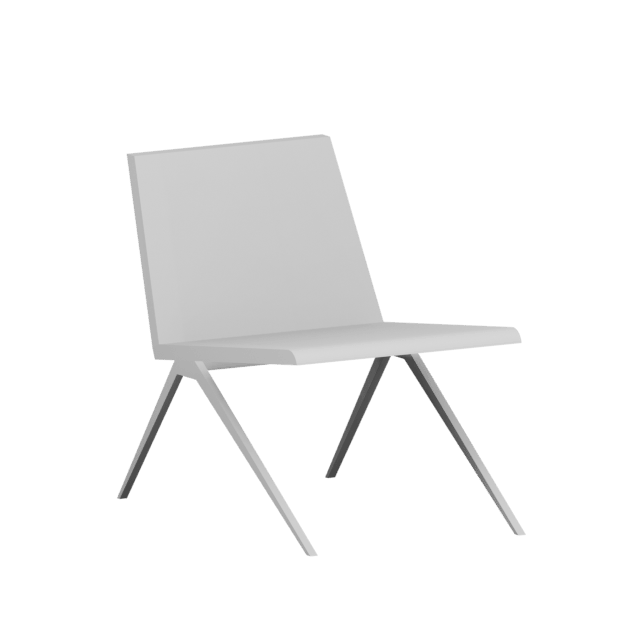} & 
    \includegraphics[]{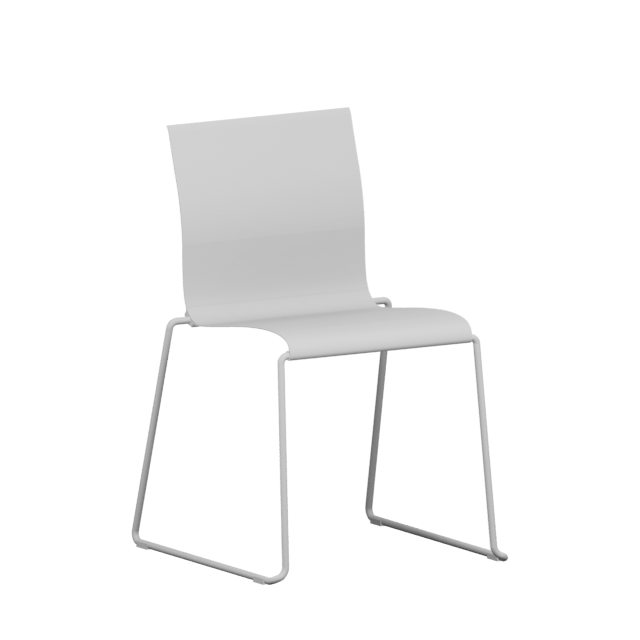} &
    
    \includegraphics[]{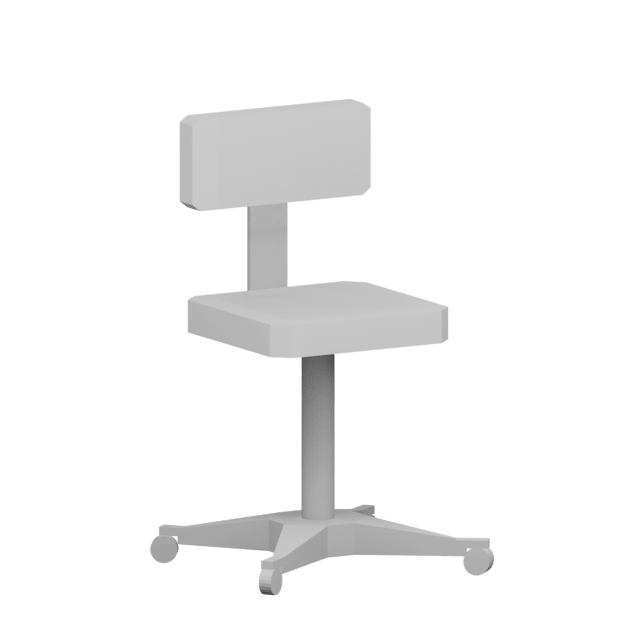} & 
    \includegraphics[]{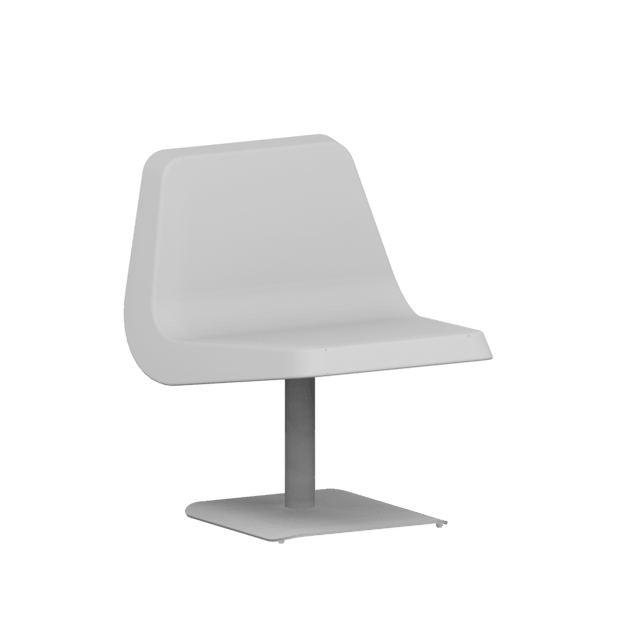} & 
    \includegraphics[]{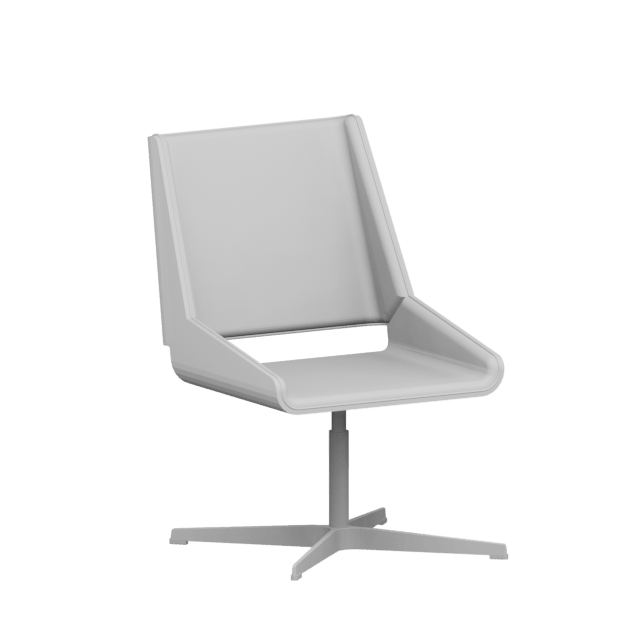} \\

    \includegraphics[]{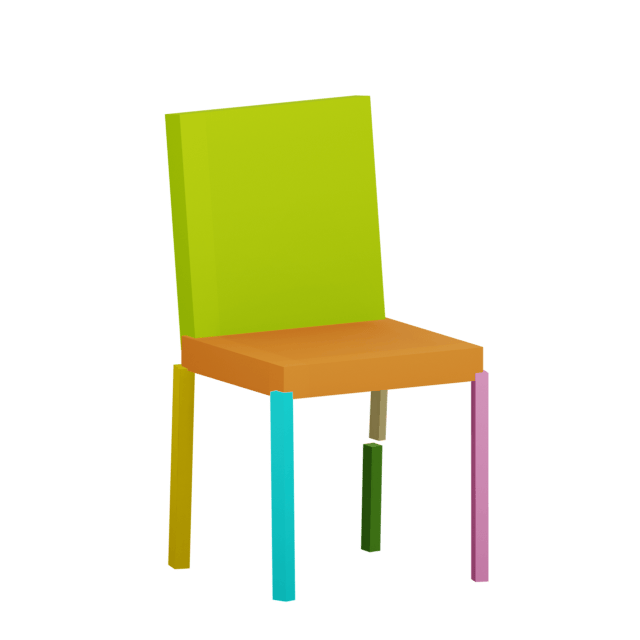} &
    \includegraphics[]{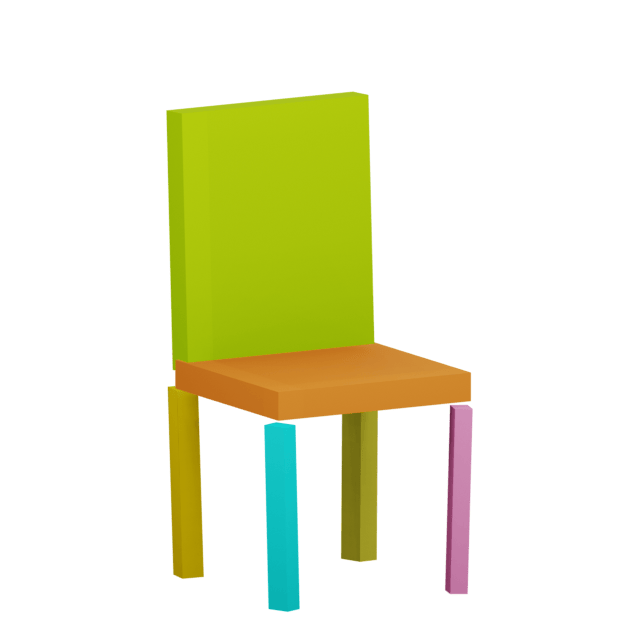} & 
    \includegraphics[]{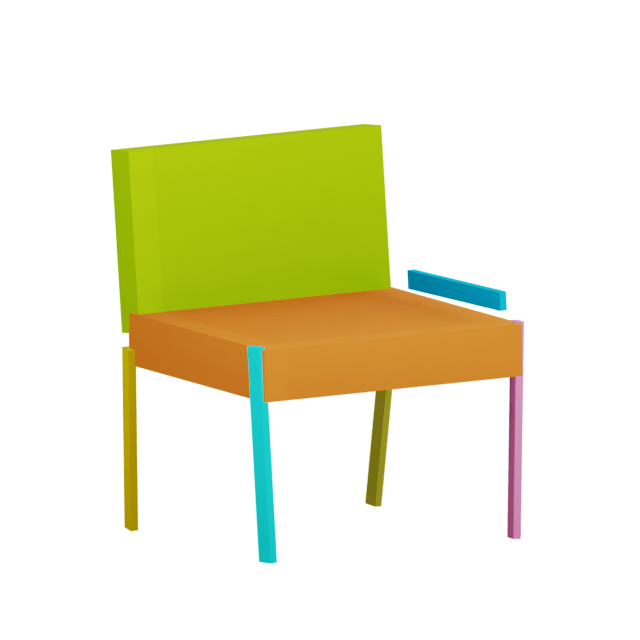} &

    \includegraphics[]{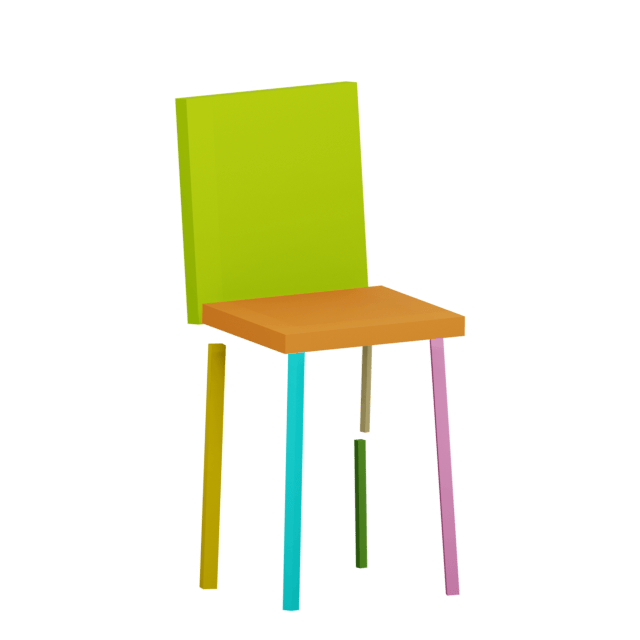} & 
    \includegraphics[]{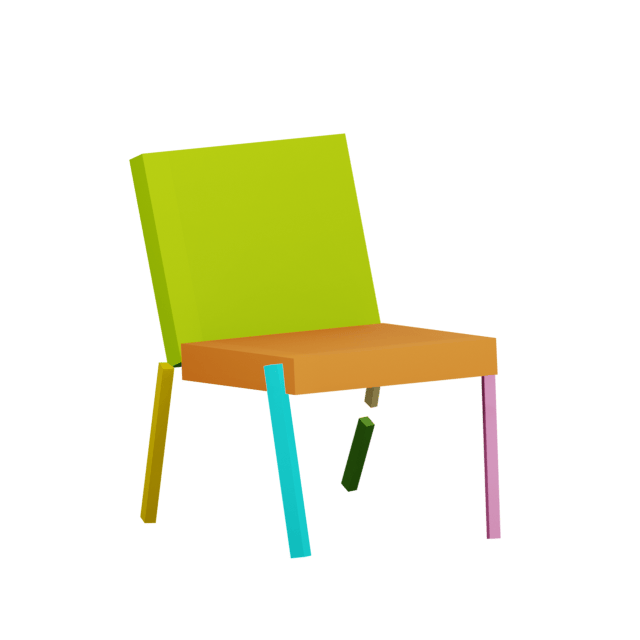} & 
    \includegraphics[]{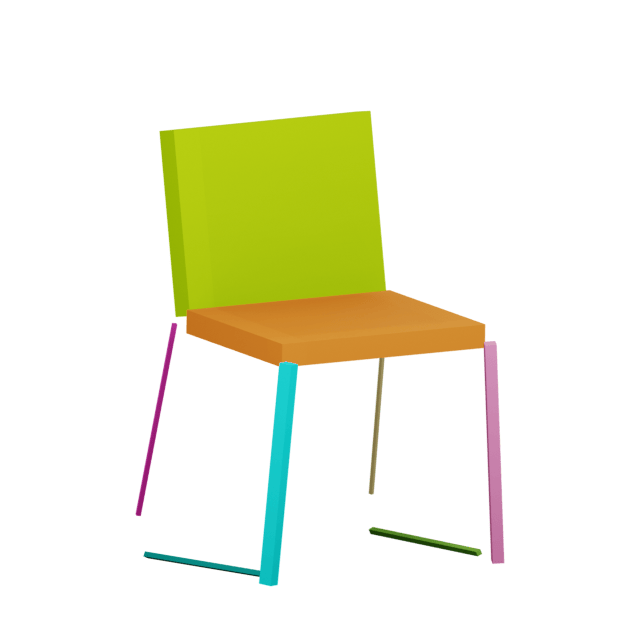} &
    
    \includegraphics[]{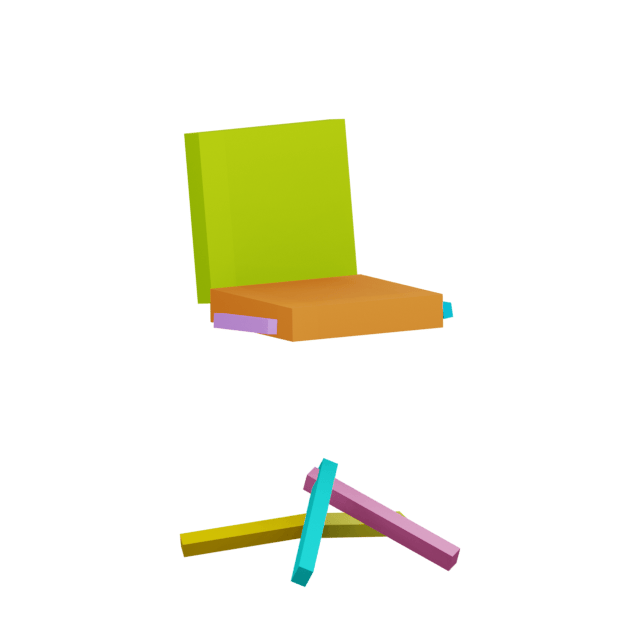} & 
    \includegraphics[]{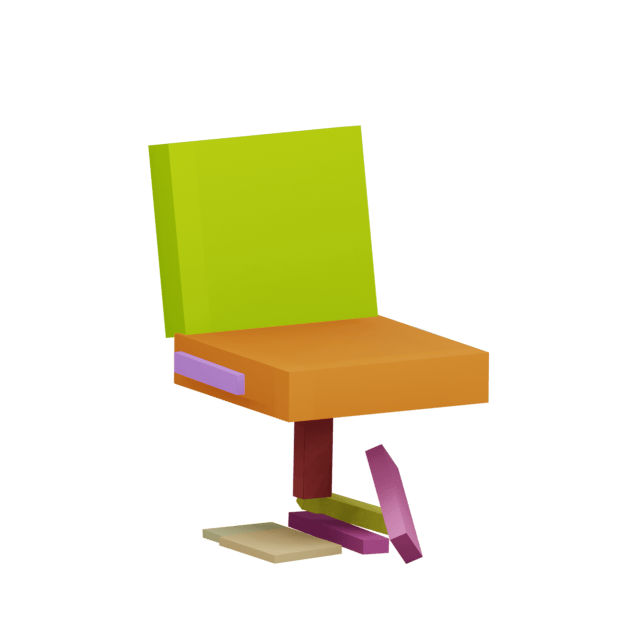} & 
    \includegraphics[]{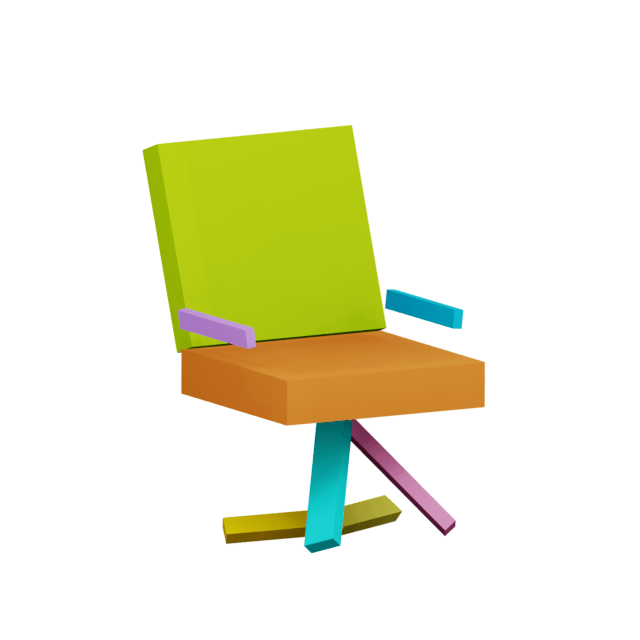} \\

    \includegraphics[]{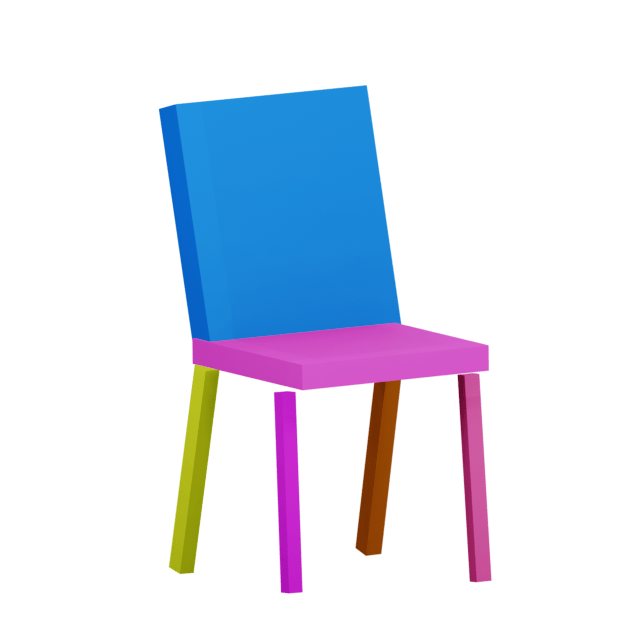} &
    \includegraphics[]{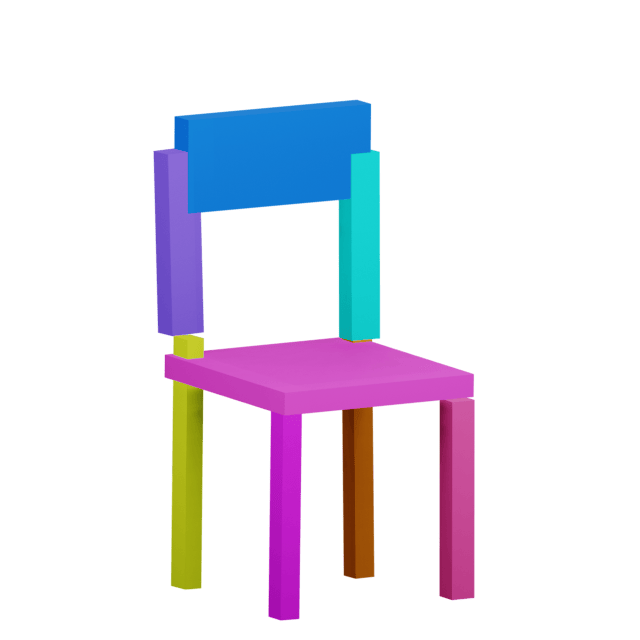} & 
    \includegraphics[]{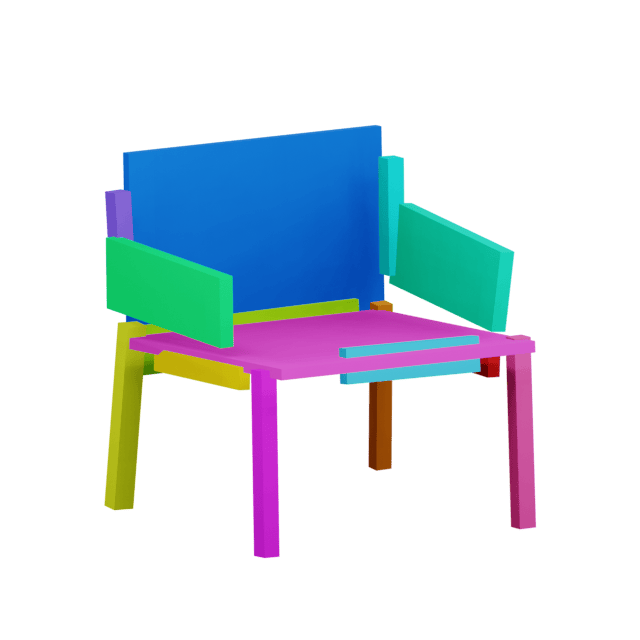} &
    
    \includegraphics[]{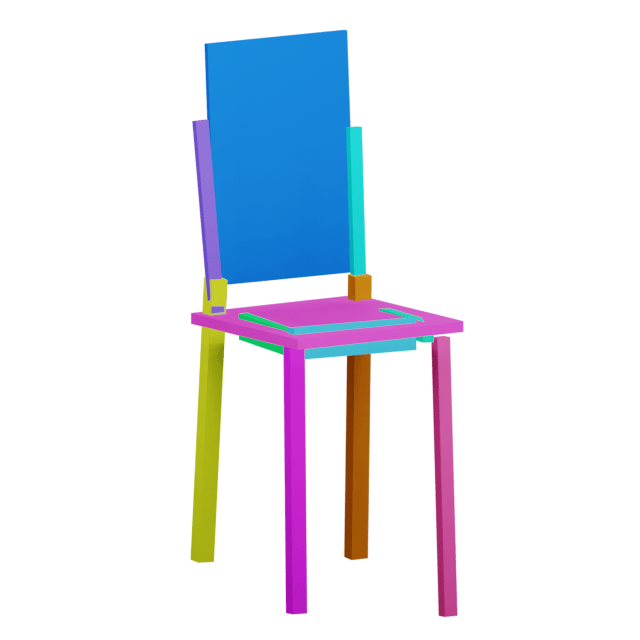} & 
    \includegraphics[]{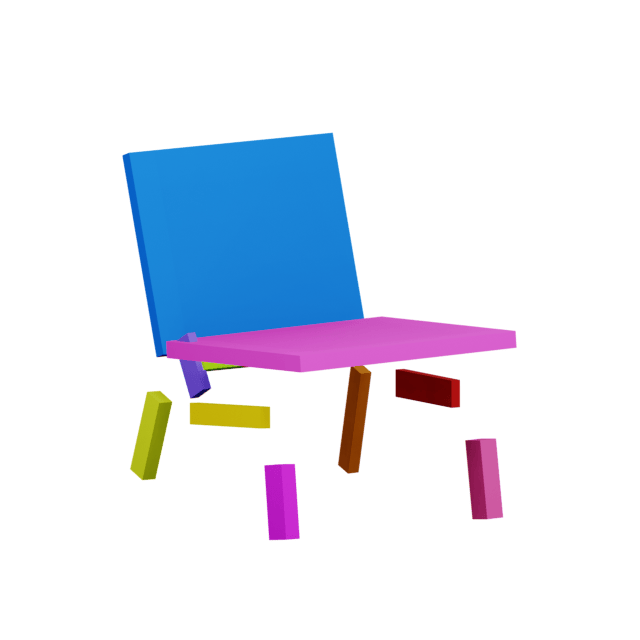} & 
    \includegraphics[]{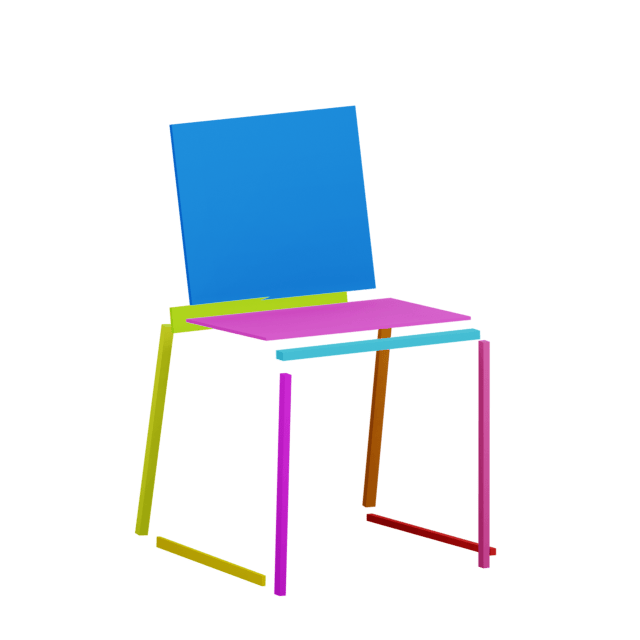} &

    \includegraphics[]{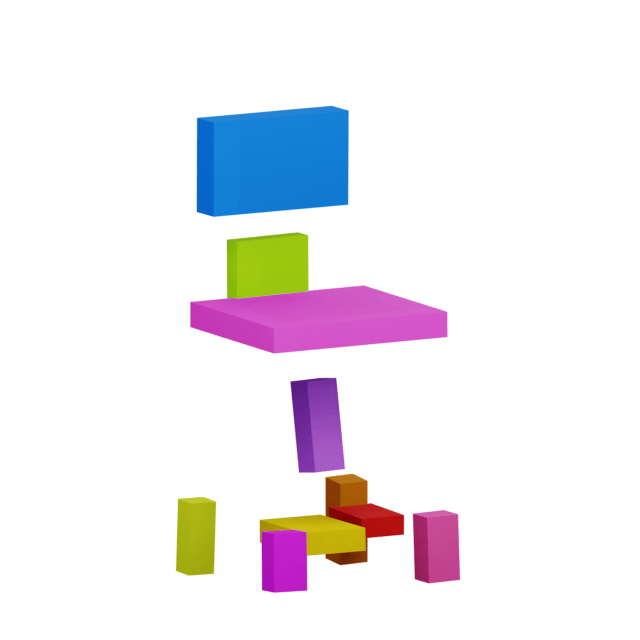} & 
    \includegraphics[]{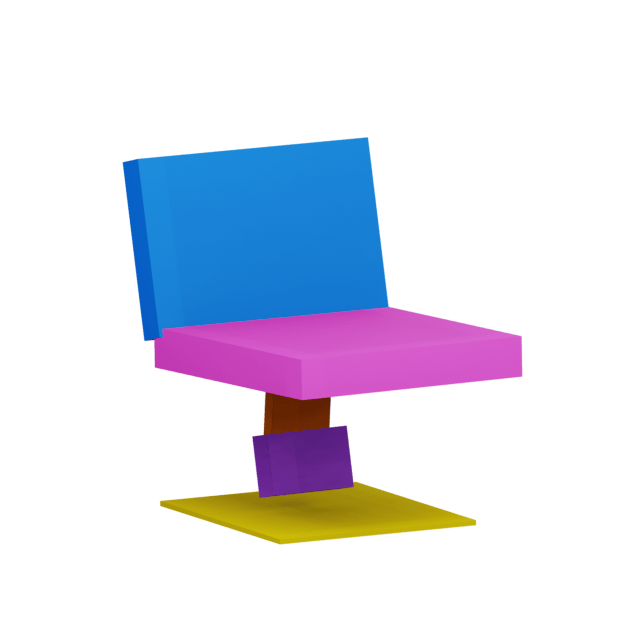} & 
    \includegraphics[]{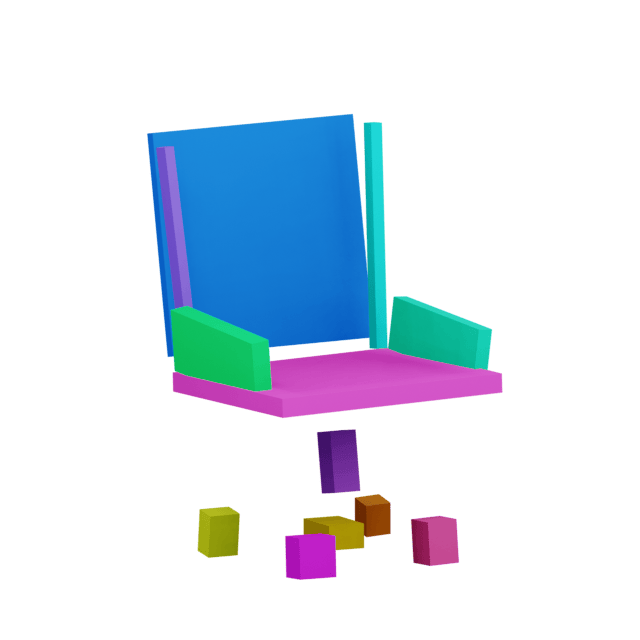} \\

    \includegraphics[]{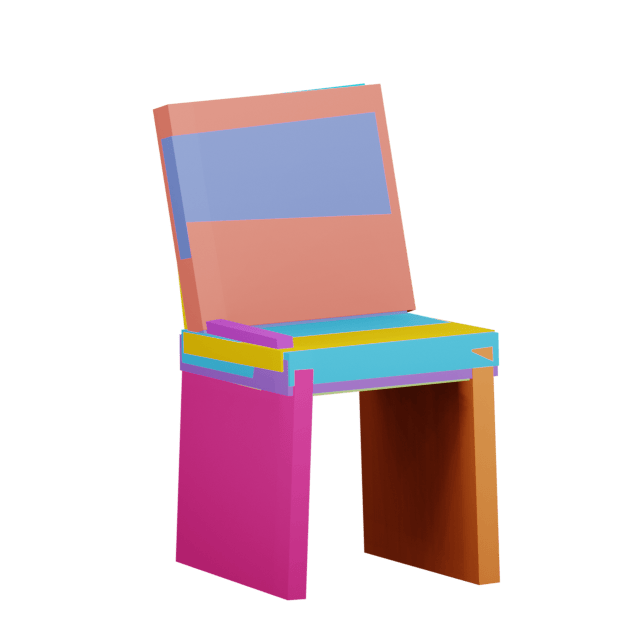} &
    \includegraphics[]{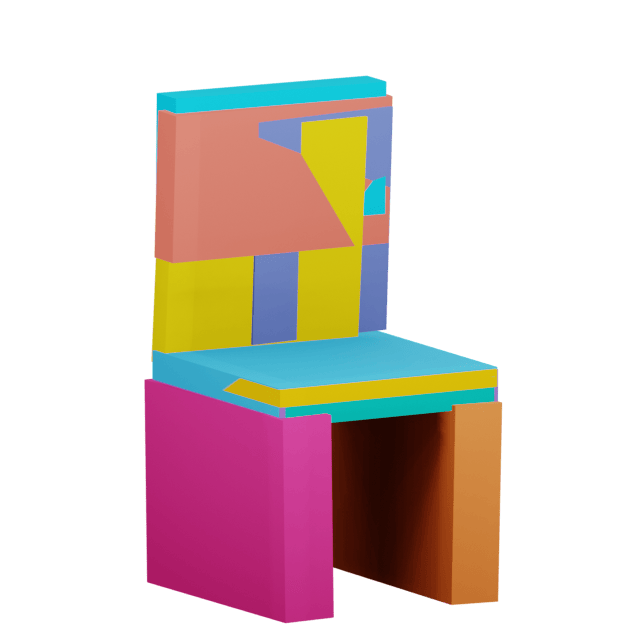} & 
    \includegraphics[]{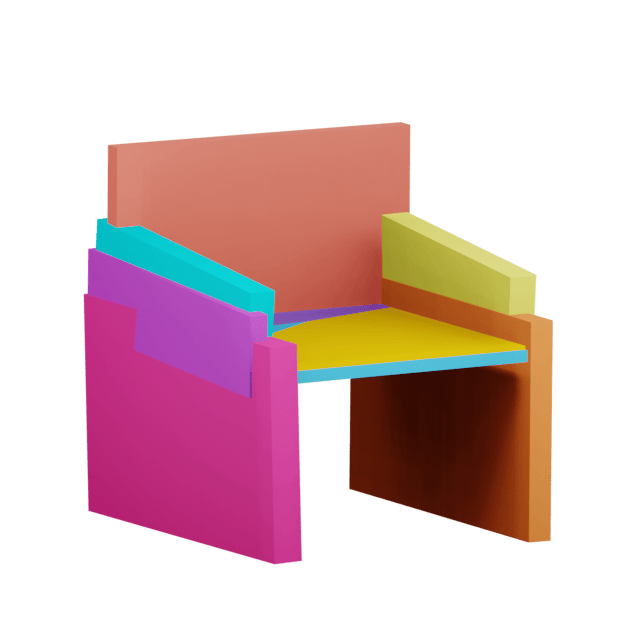} &    
    
    \includegraphics[]{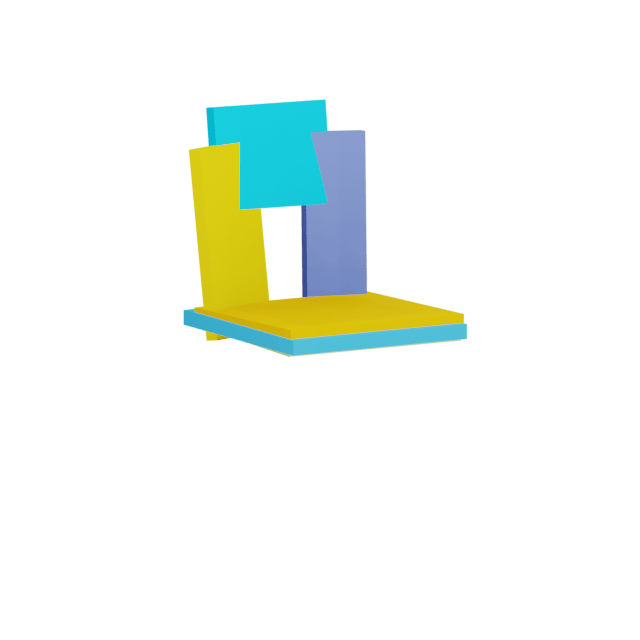} & 
    \includegraphics[]{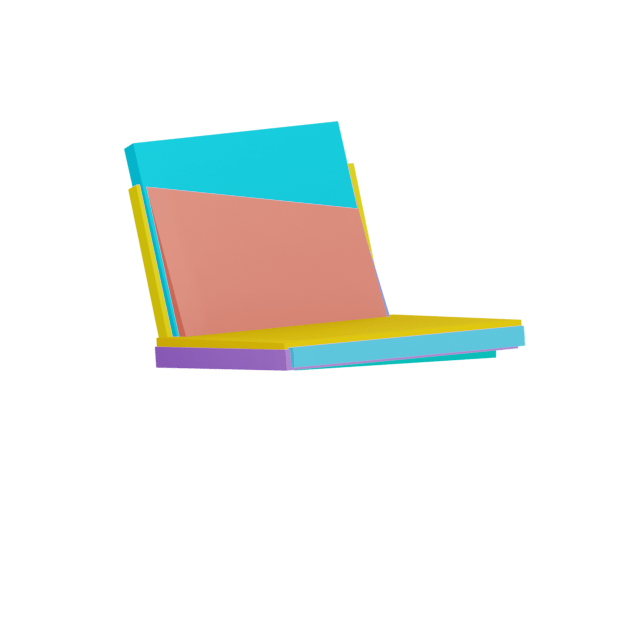} & 
    \includegraphics[]{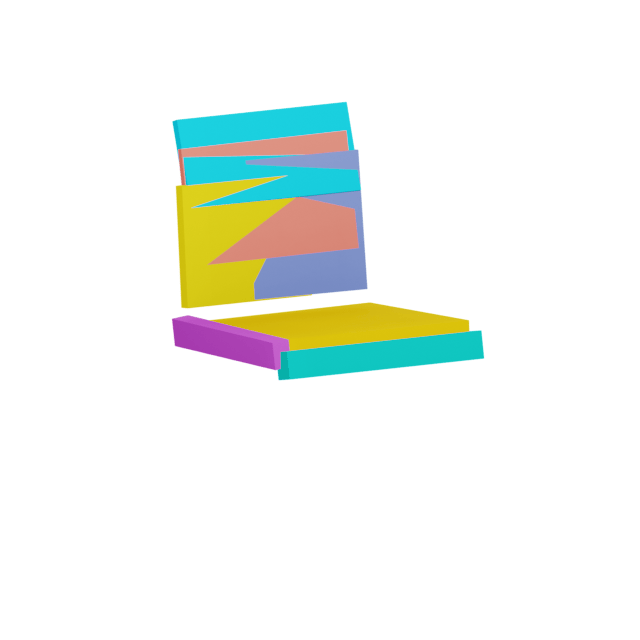} &

    \includegraphics[]{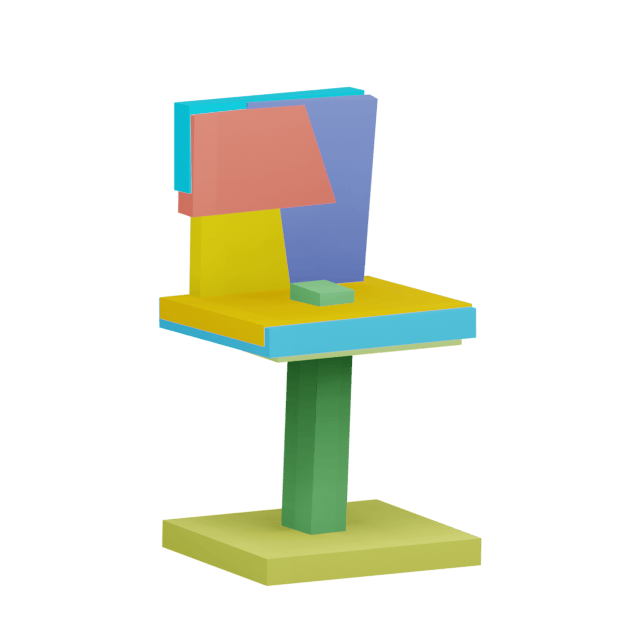} & 
    \includegraphics[]{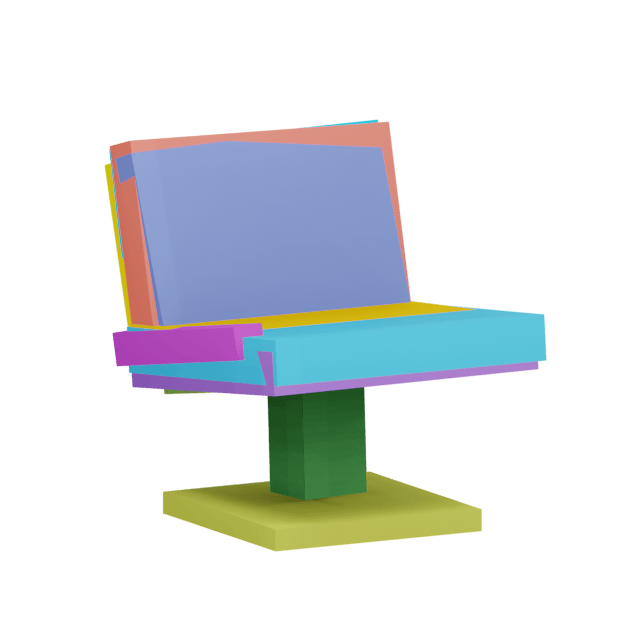} & 
    \includegraphics[]{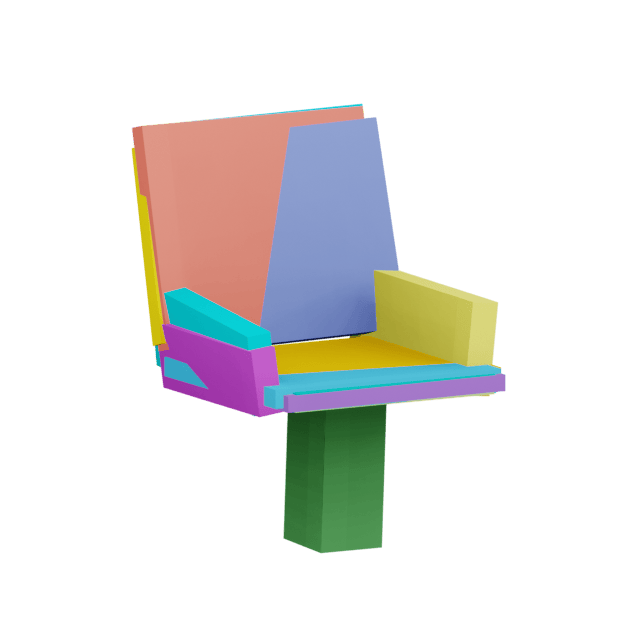} \\

    \includegraphics[]{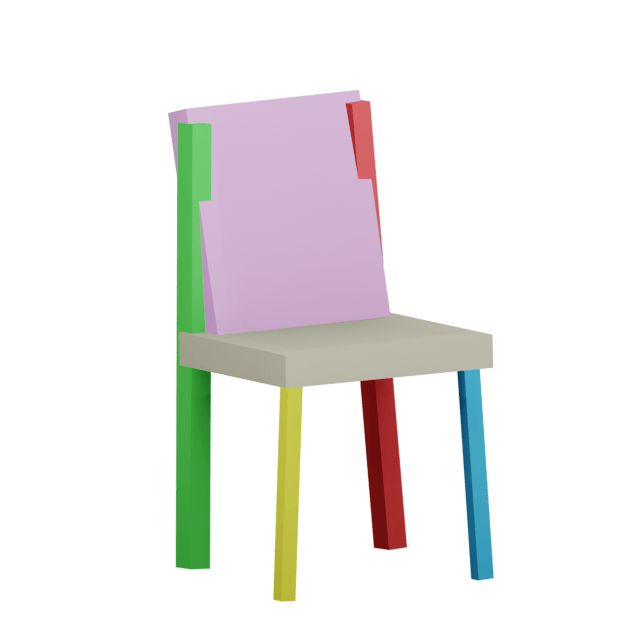} &
    \includegraphics[]{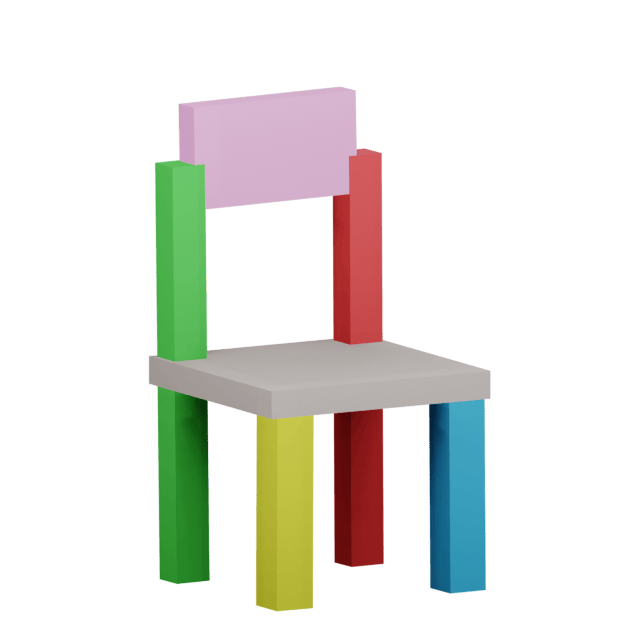} & 
    \includegraphics[]{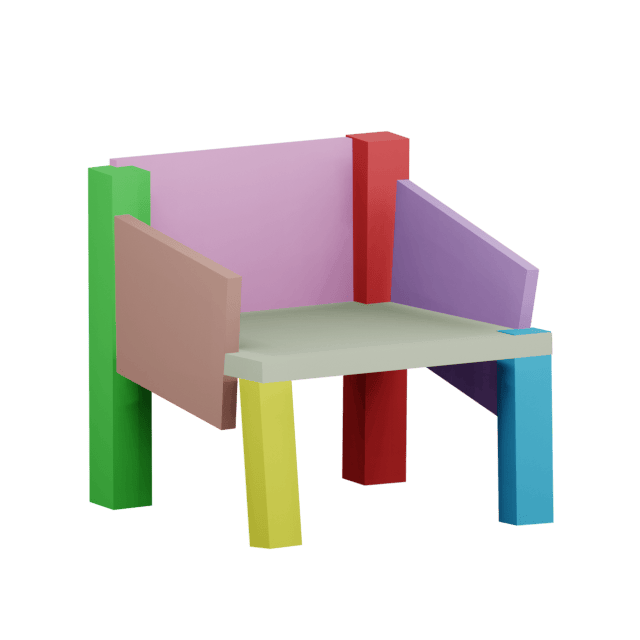} &    
    
    \includegraphics[]{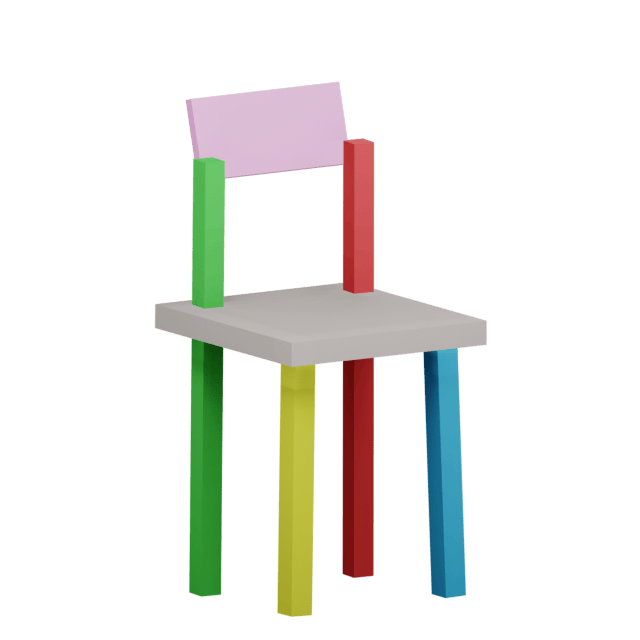} & 
    \includegraphics[]{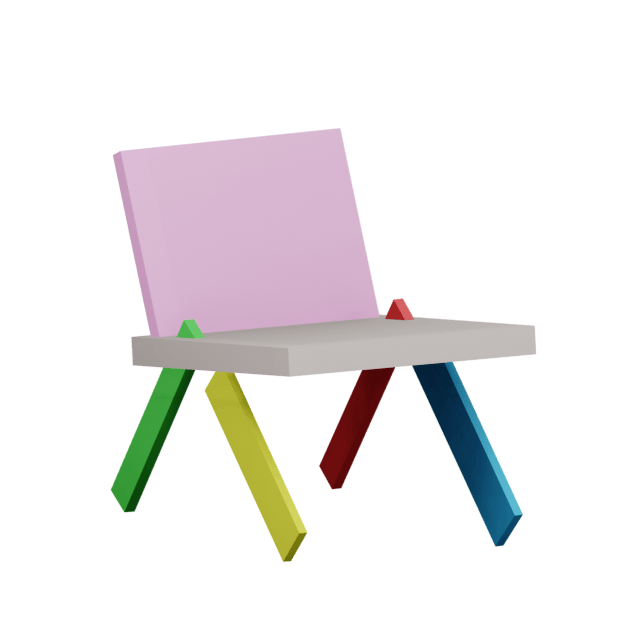} & 
    \includegraphics[]{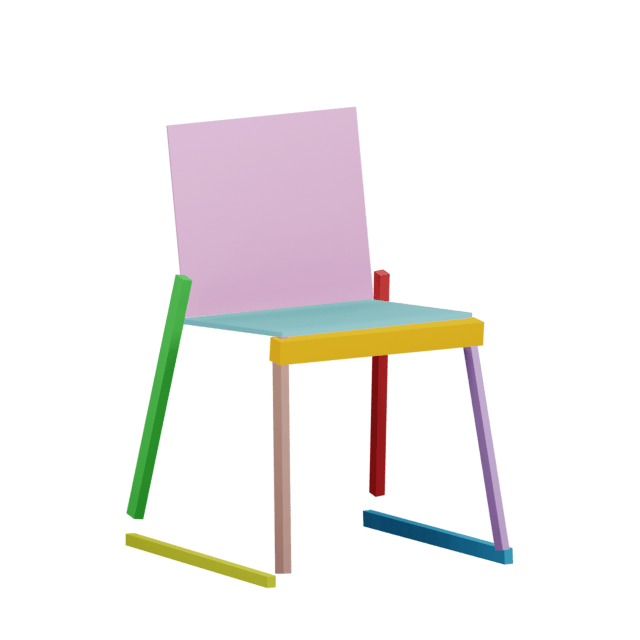} &

    \includegraphics[]{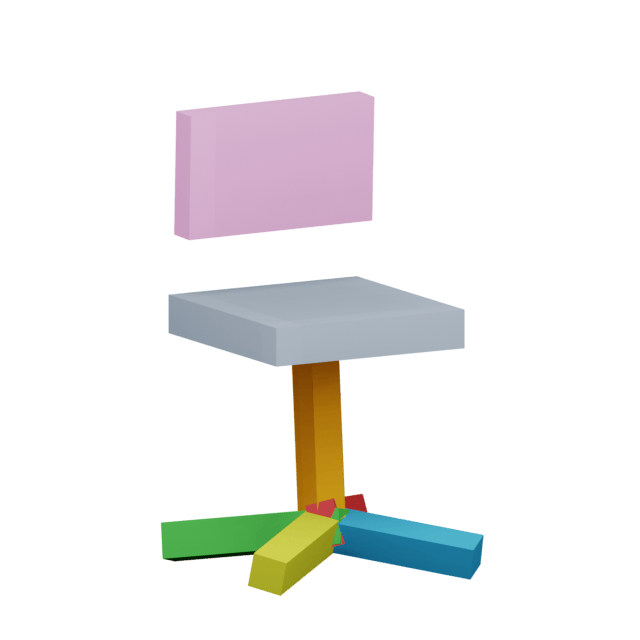} & 
    \includegraphics[]{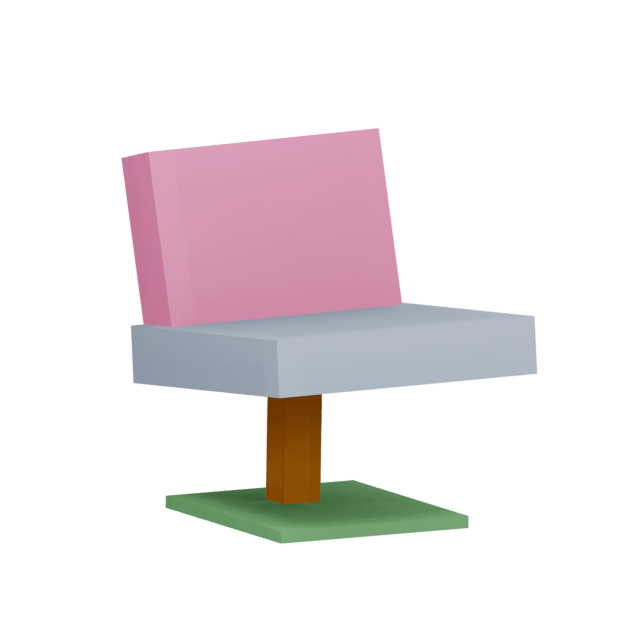} & 
    \includegraphics[]{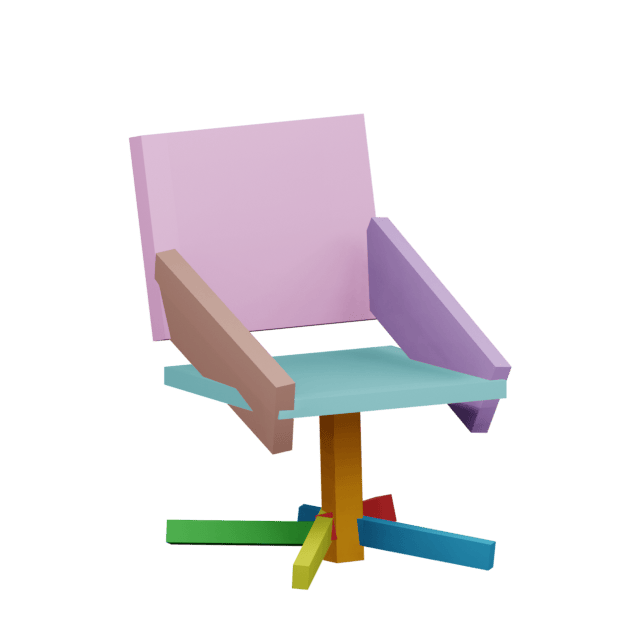} \\
    
\end{tabular}}
    \caption{Qualitative samples of the plane and chair class. We visualize the ground truth (GT) shape and the cuboid abstraction of \cite{HCA_Sun} (HCA), \cite{CAS_Yang} (CAS), \cite{DPF-Net_Shuai} ($\text{DPF}_{PPM}$) and Ours. Note, our abstraction preserves the geometry of the ground truth shape more closely using less cuboid primitives. Colors indicate distinc parts per category.}
    \label{fig:qualitative_airplane_chair}
\end{figure*}

\begin{figure*}[hpt]
    \centering
    \begin{tabular}{c}
        \rotatebox[origin=c]{90}{\hspace{0.6cm} GT \hspace{0.6cm}} \\
        \rotatebox[origin=c]{90}{\hspace{0.6cm} HCA \hspace{0.6cm}} \\
        \rotatebox[origin=c]{90}{\hspace{0.6cm} CAS \hspace{0.6cm}} \\
        \rotatebox[origin=c]{90}{\hspace{0.2cm} $\text{DPF}_{PPM}$ \hspace{0.2cm}} \\
        \rotatebox[origin=c]{90}{\hspace{0.6cm} Ours \hspace{0.6cm}} \\
    \end{tabular}
    \resizebox{0.96\textwidth}{!}{\begin{tabular}{ccccccccc}
    \includegraphics[]{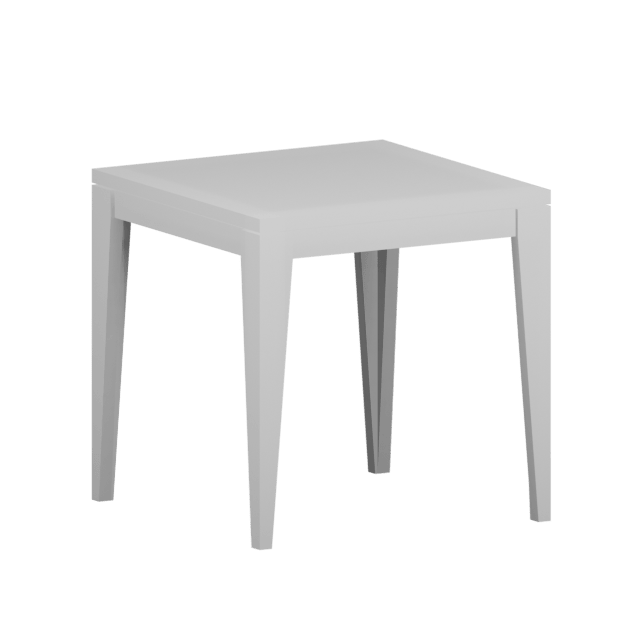} &
    \includegraphics[]{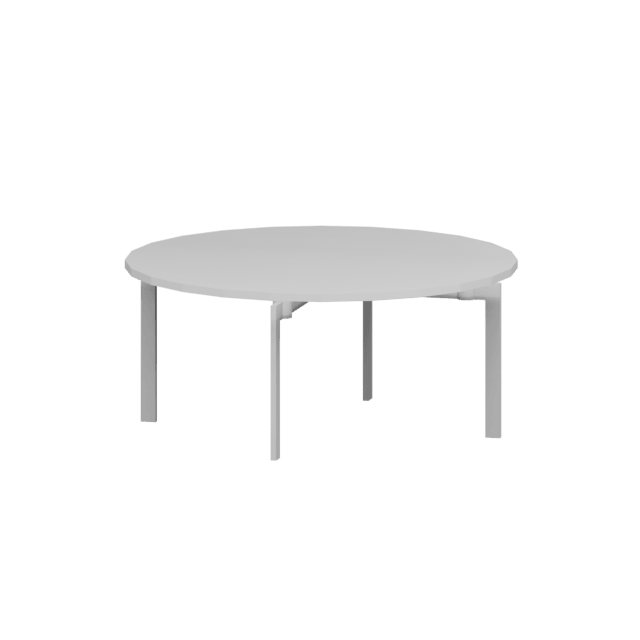} & 
    \includegraphics[]{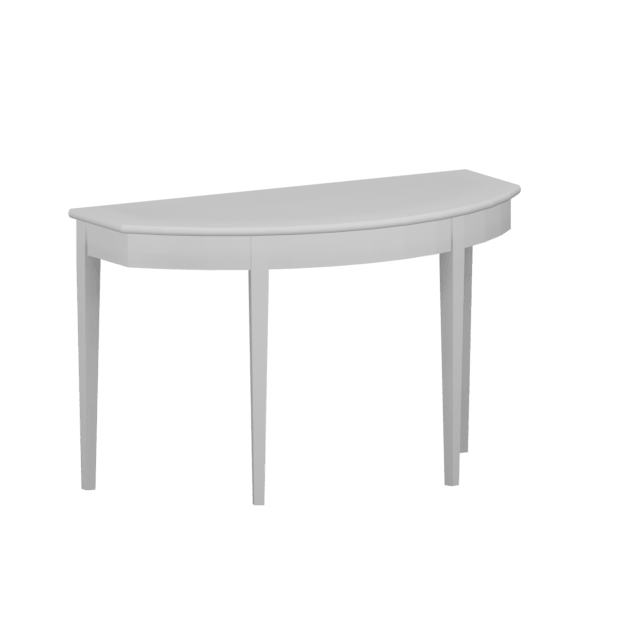} &
    
    \includegraphics[]{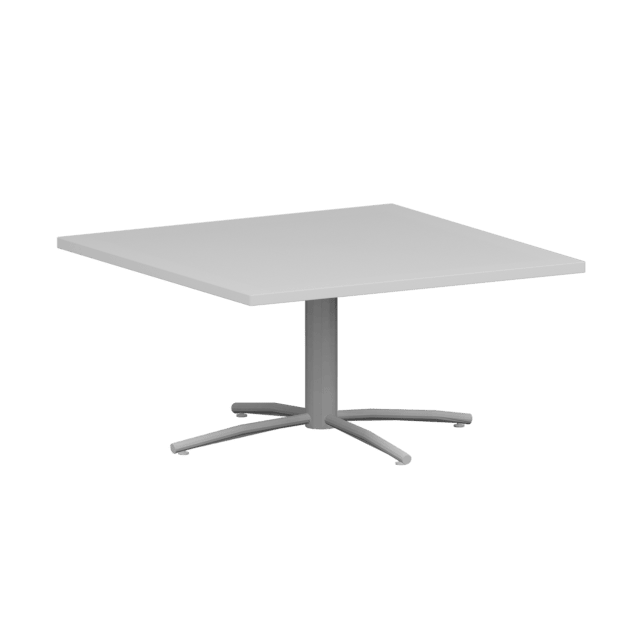} & 
    \includegraphics[]{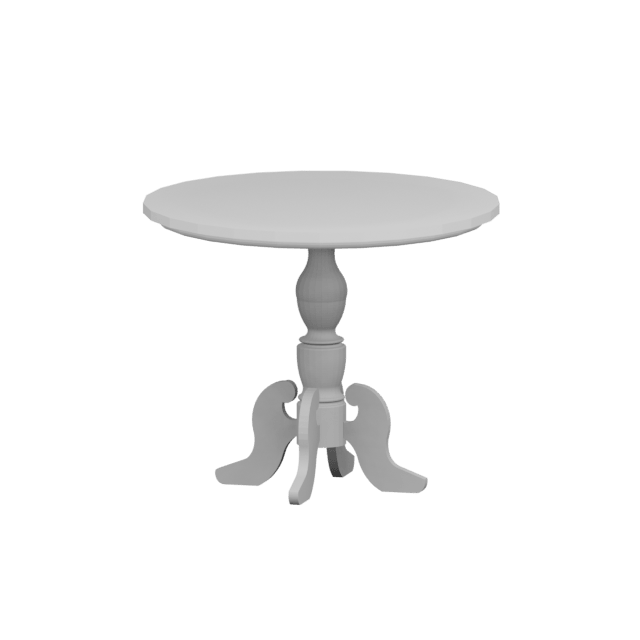} & 
    \includegraphics[]{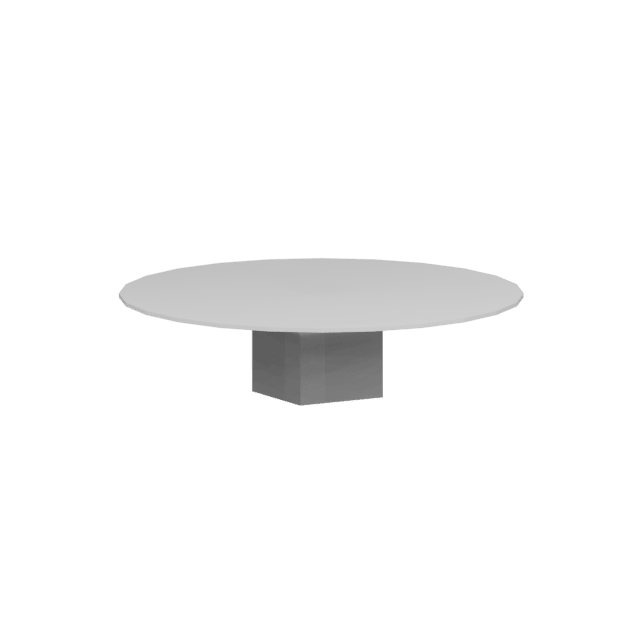} &
    
    \includegraphics[]{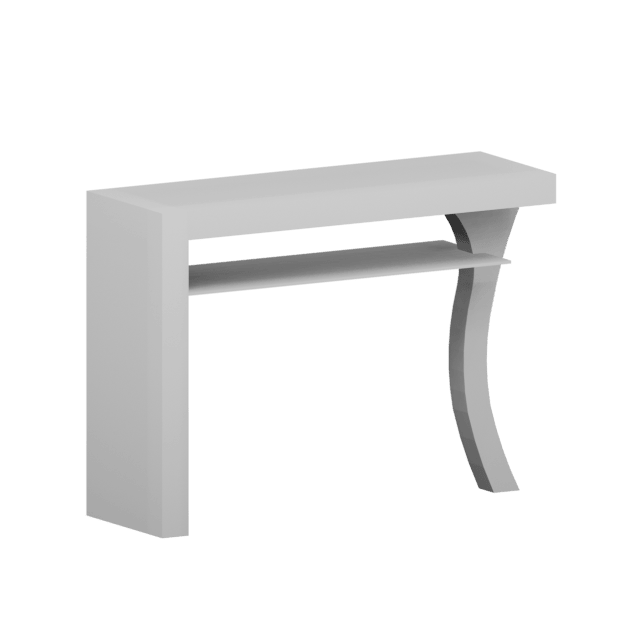} & 
    \includegraphics[]{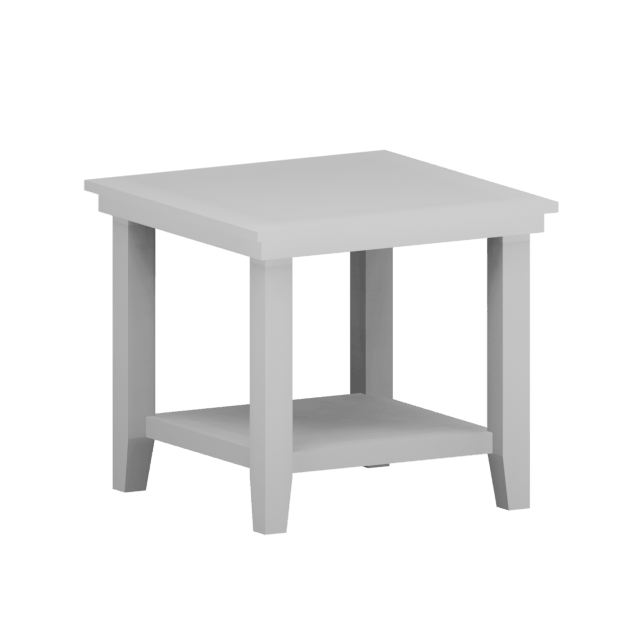} & 
    \includegraphics[]{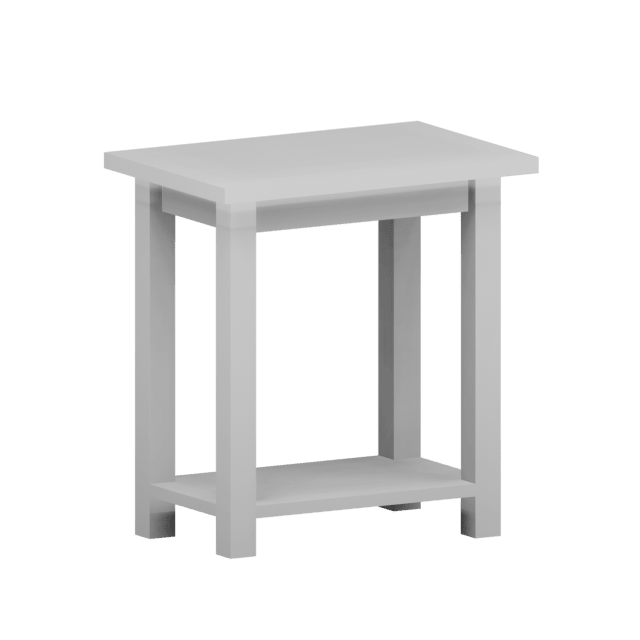} \\

    \includegraphics[]{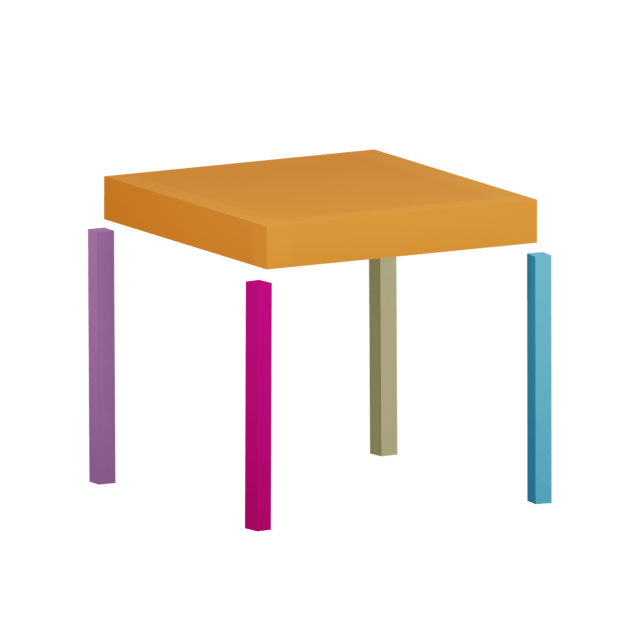} &
    \includegraphics[]{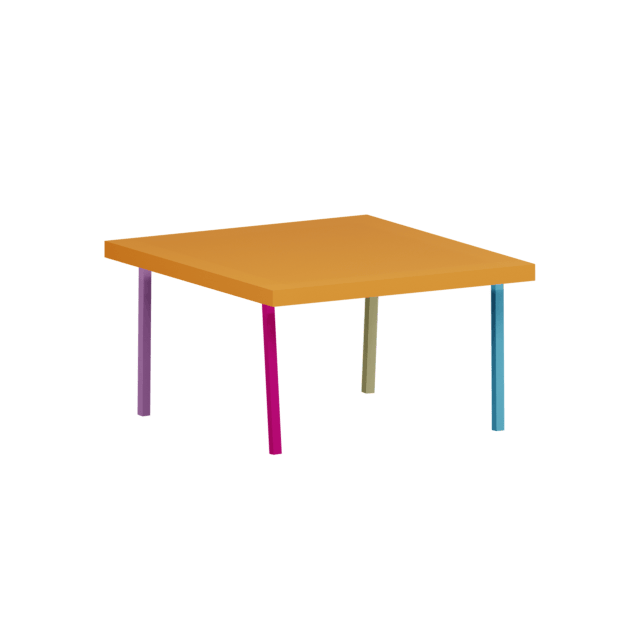} & 
    \includegraphics[]{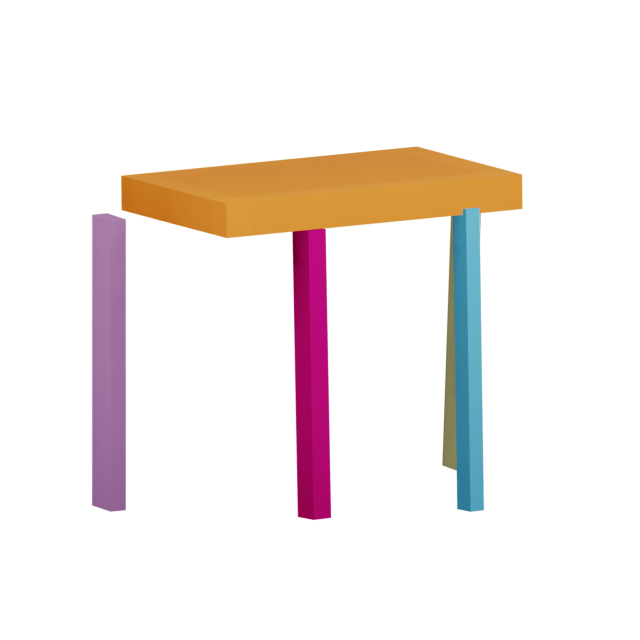} &
    
    \includegraphics[]{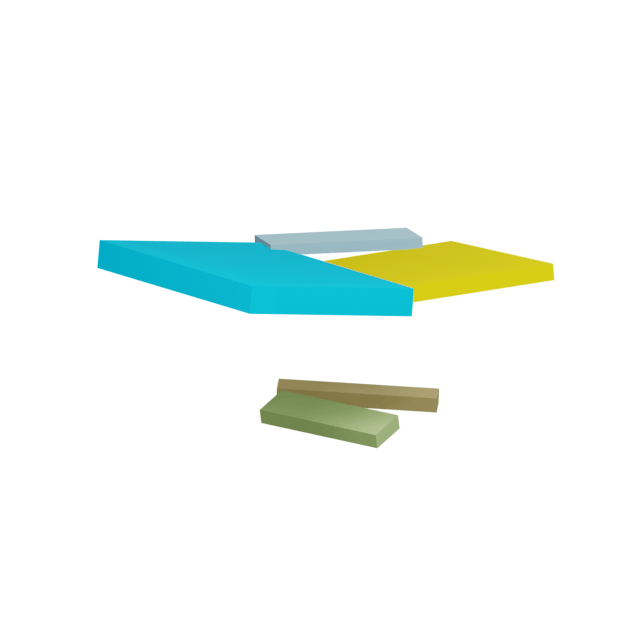} & 
    \includegraphics[]{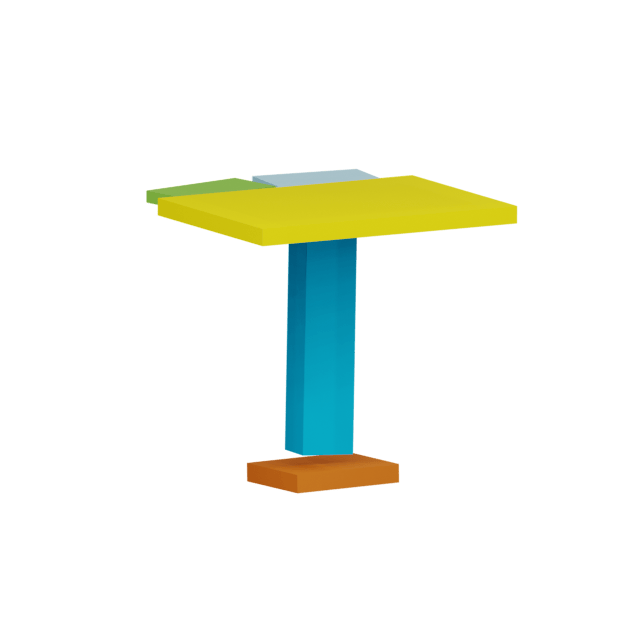} & 
    \includegraphics[]{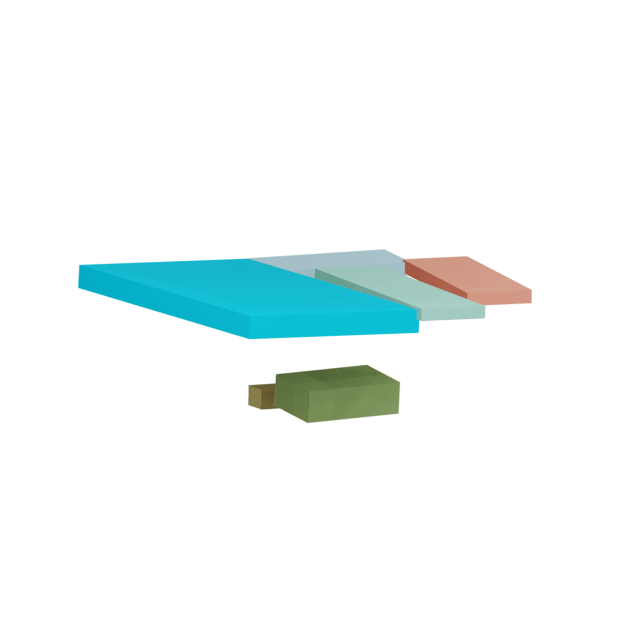} &
    
    \includegraphics[]{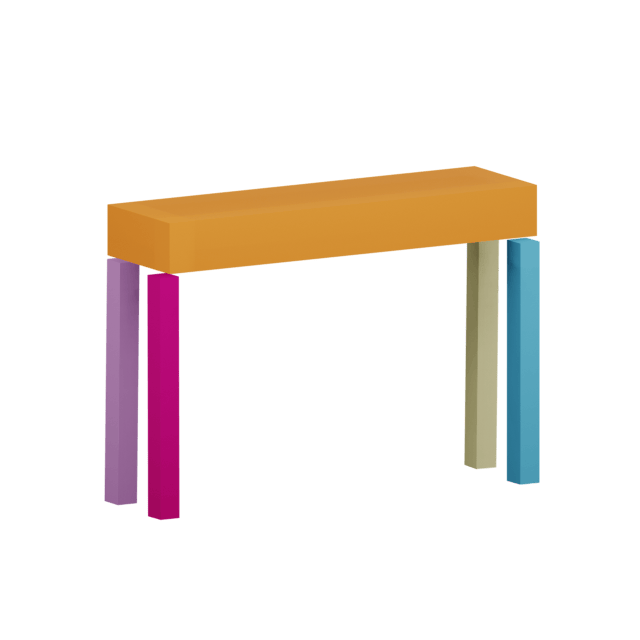} & 
    \includegraphics[]{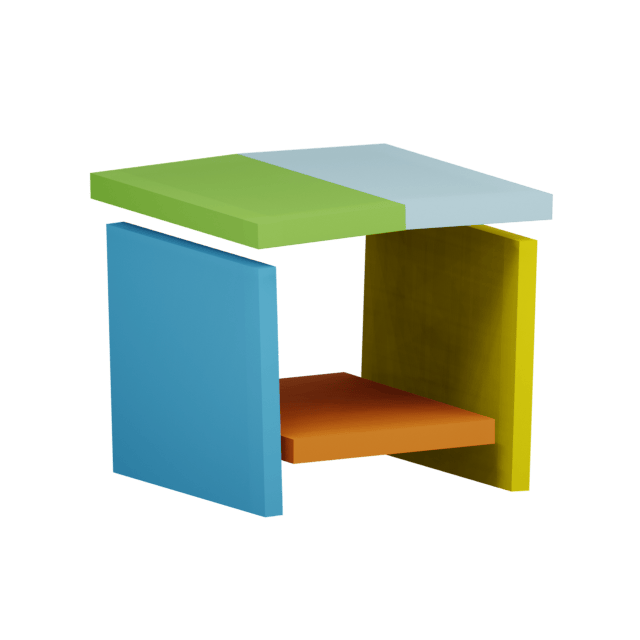} & 
    \includegraphics[]{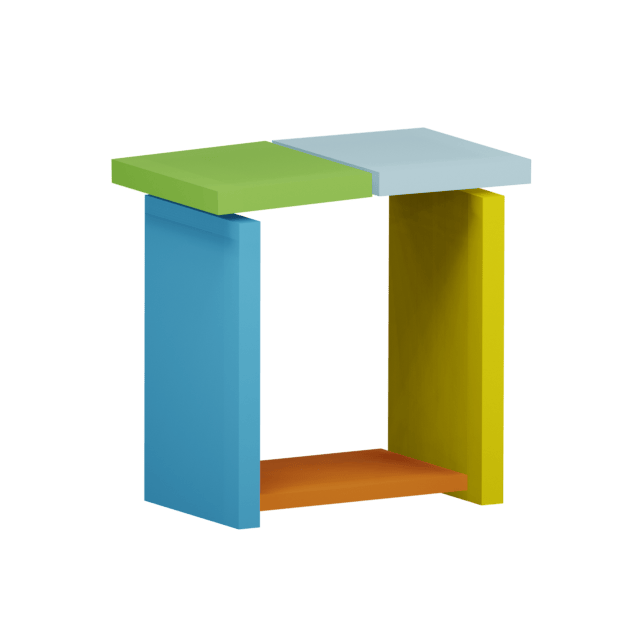} \\

    \includegraphics[]{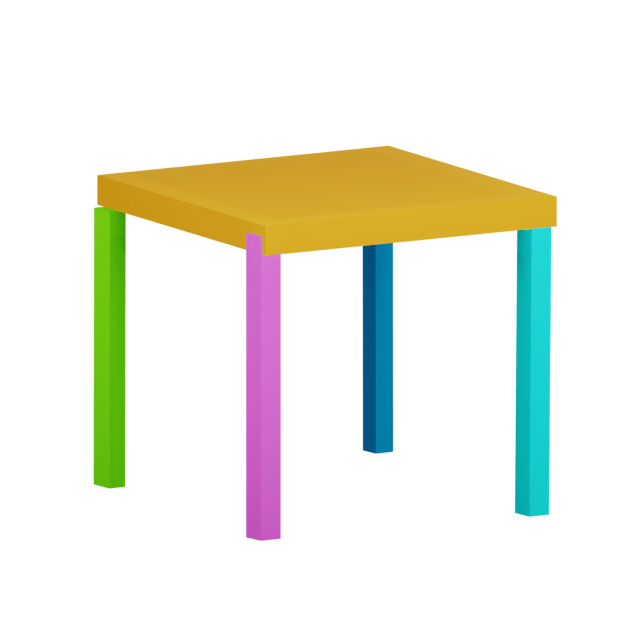} &
    \includegraphics[]{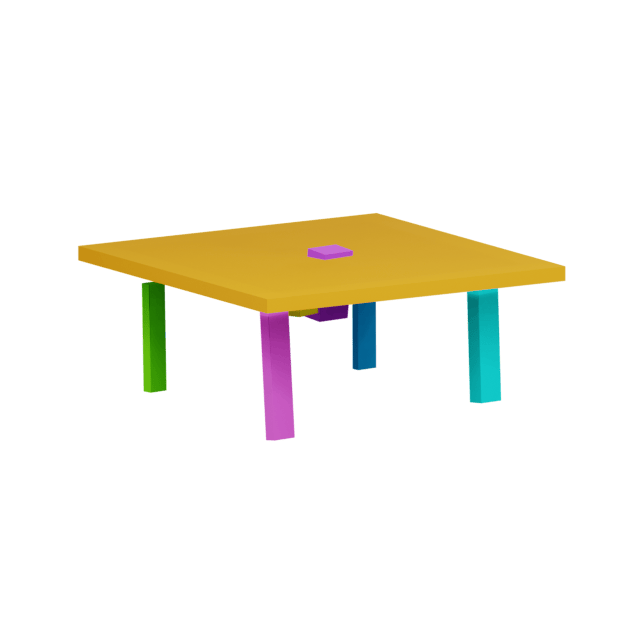} & 
    \includegraphics[]{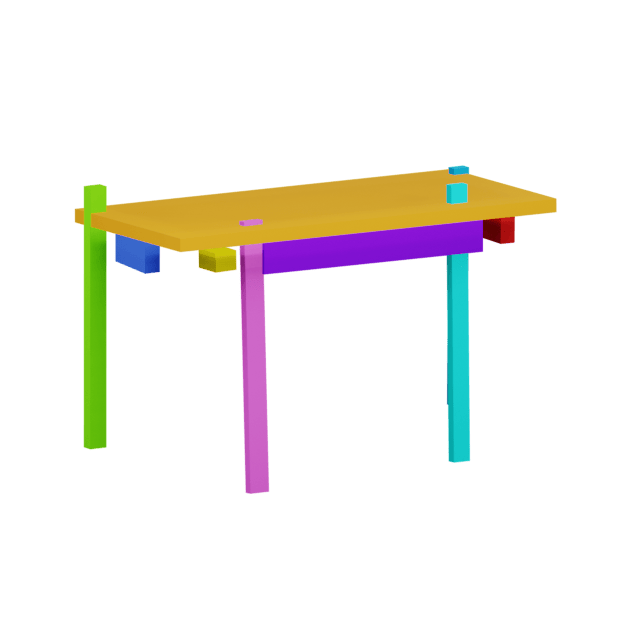} &
    
    \includegraphics[]{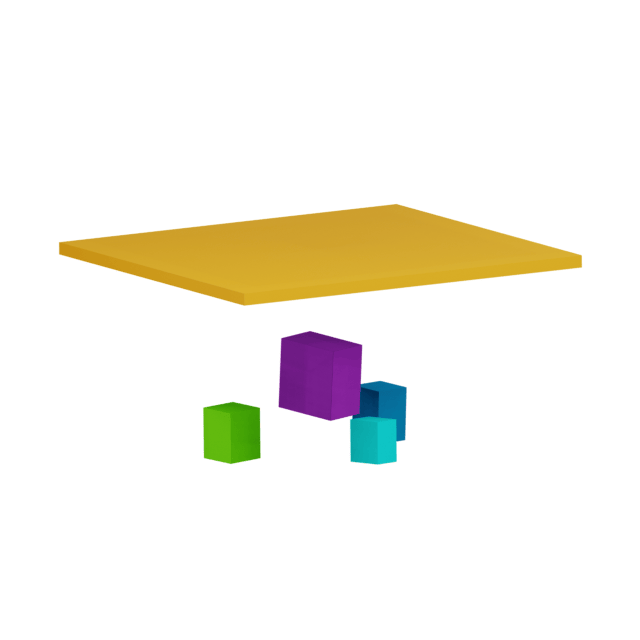} & 
    \includegraphics[]{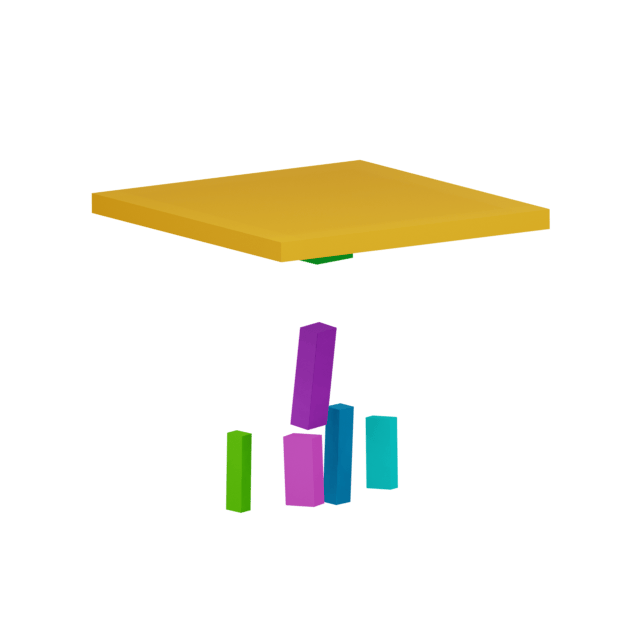} & 
    \includegraphics[]{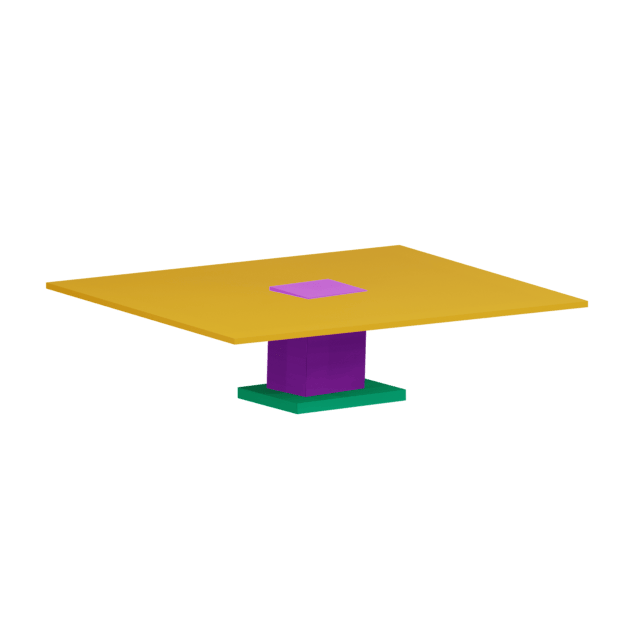} &
    
    \includegraphics[]{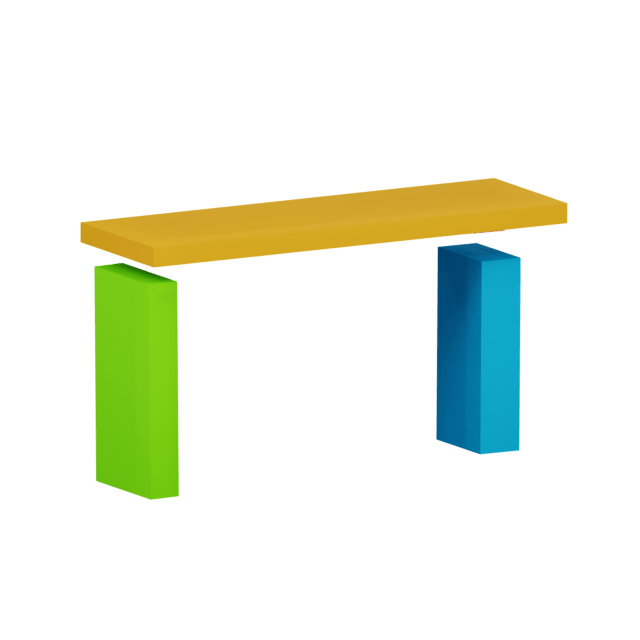} & 
    \includegraphics[]{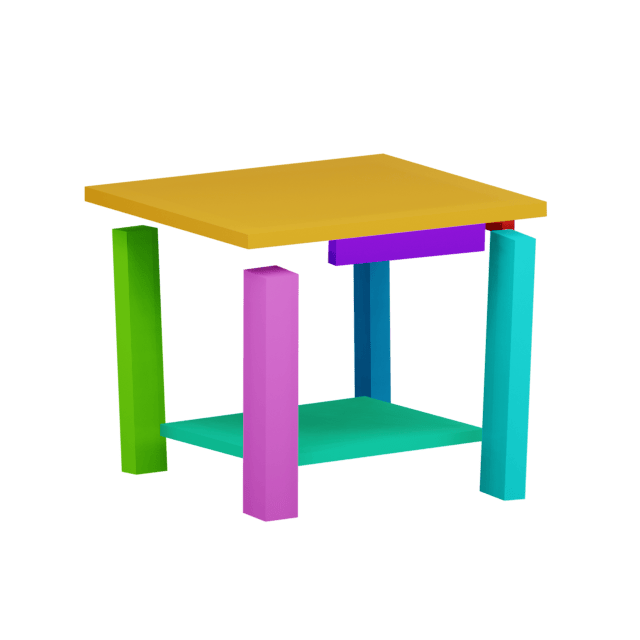} & 
    \includegraphics[]{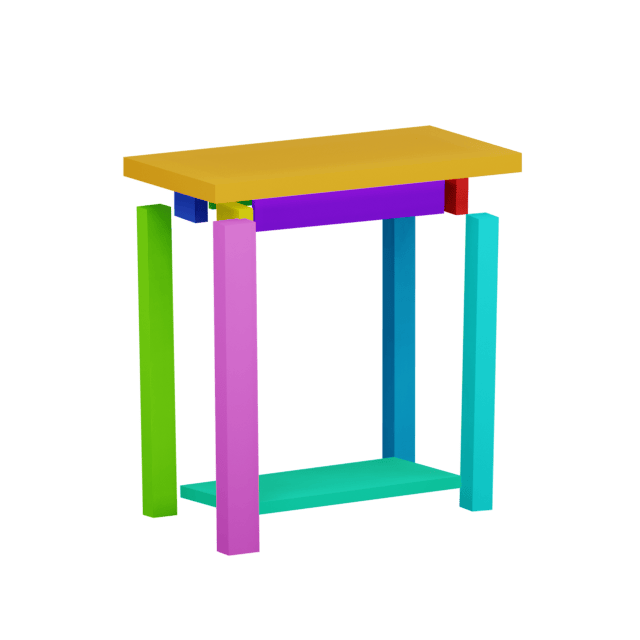} \\

    \includegraphics[]{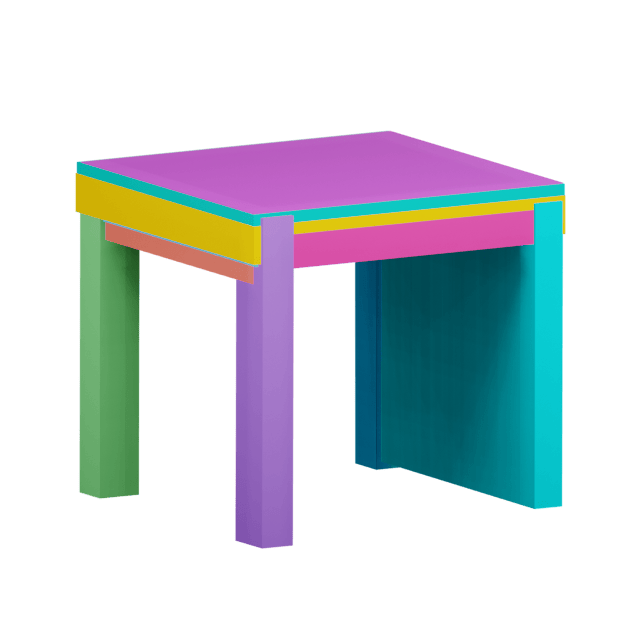} &
    \includegraphics[]{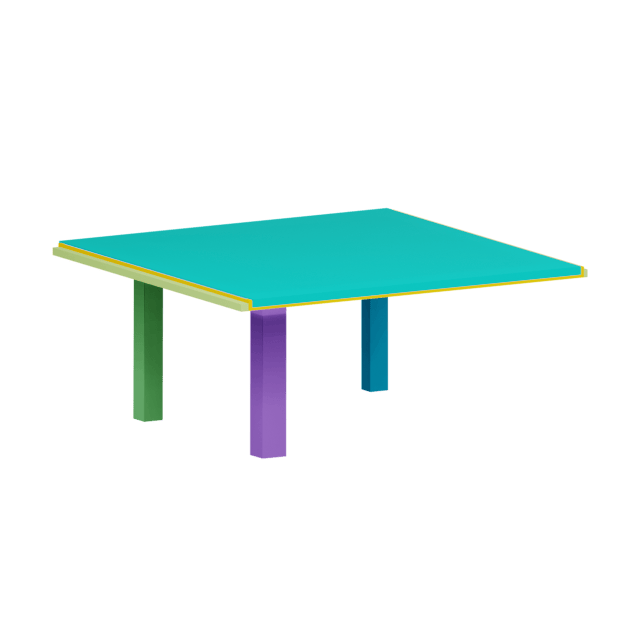} & 
    \includegraphics[]{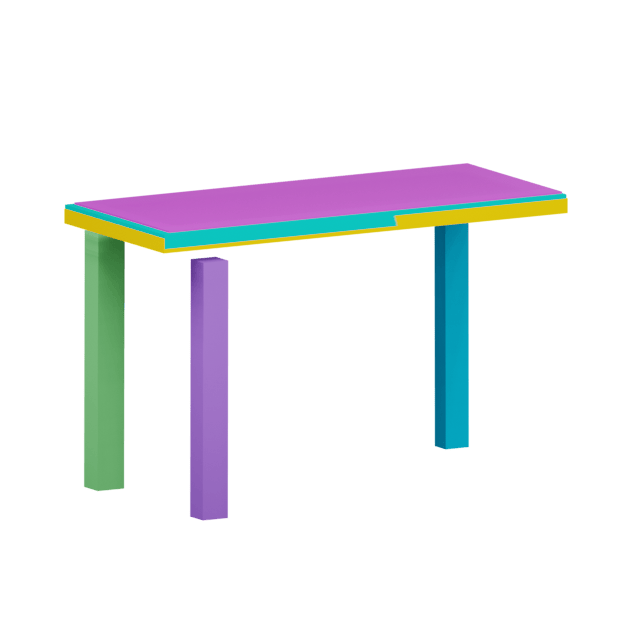} &
    
    \includegraphics[]{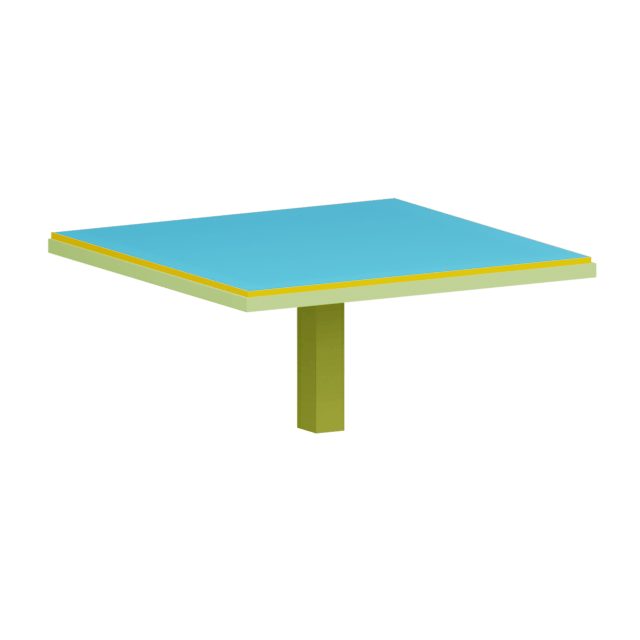} & 
    \includegraphics[]{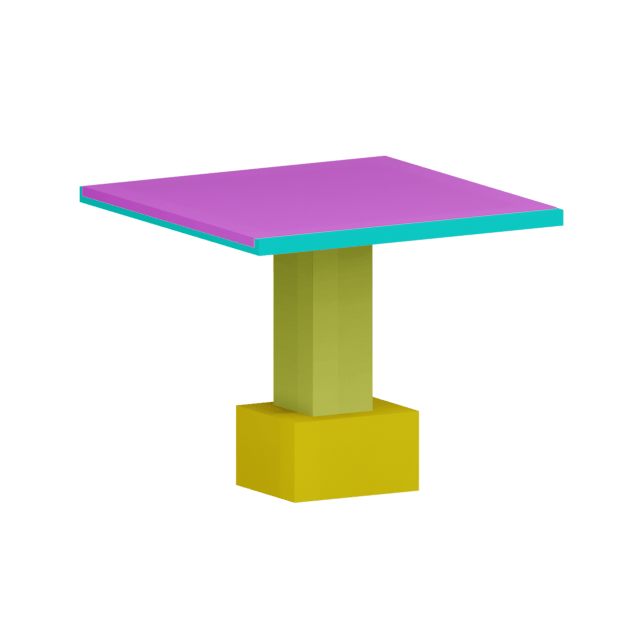} & 
    \includegraphics[]{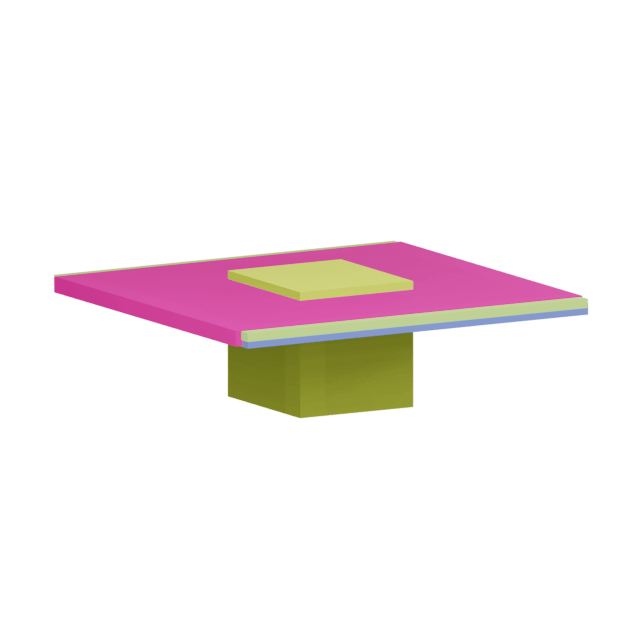} &
    
    \includegraphics[]{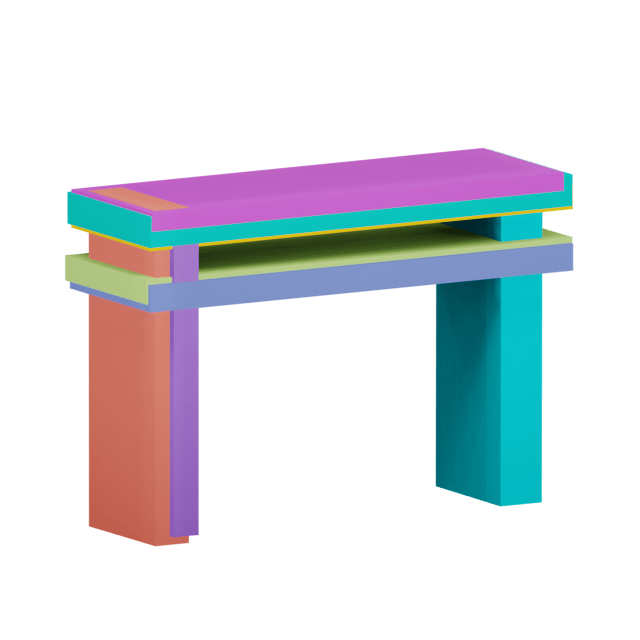} & 
    \includegraphics[]{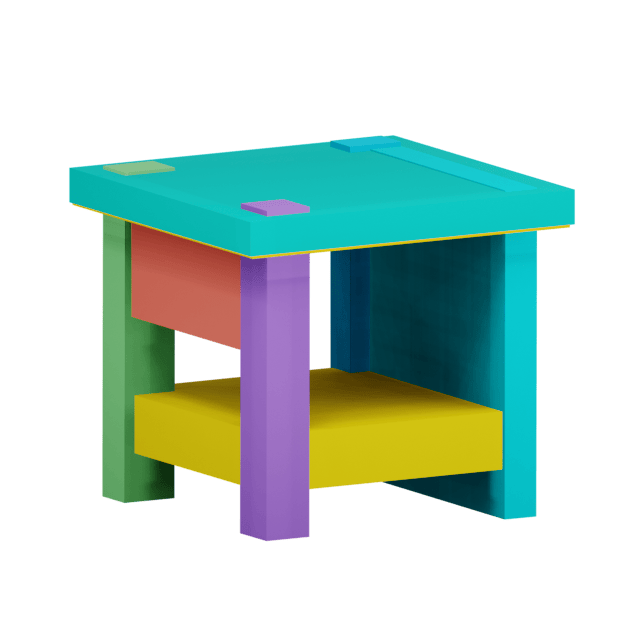} & 
    \includegraphics[]{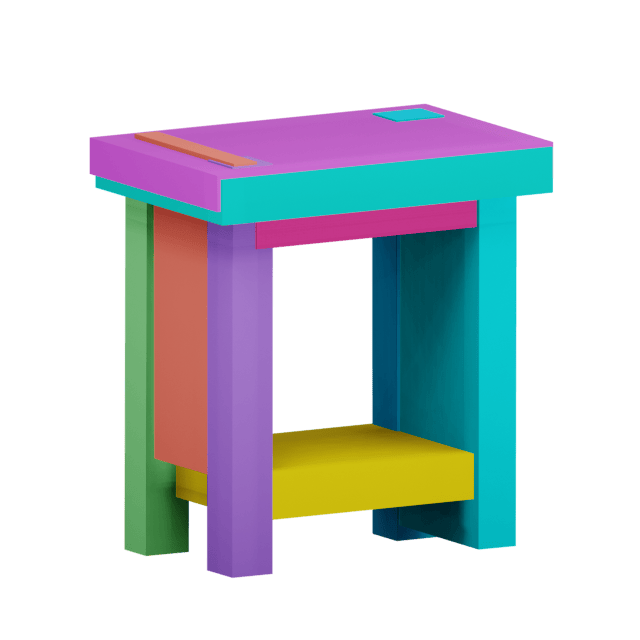} \\

    \includegraphics[]{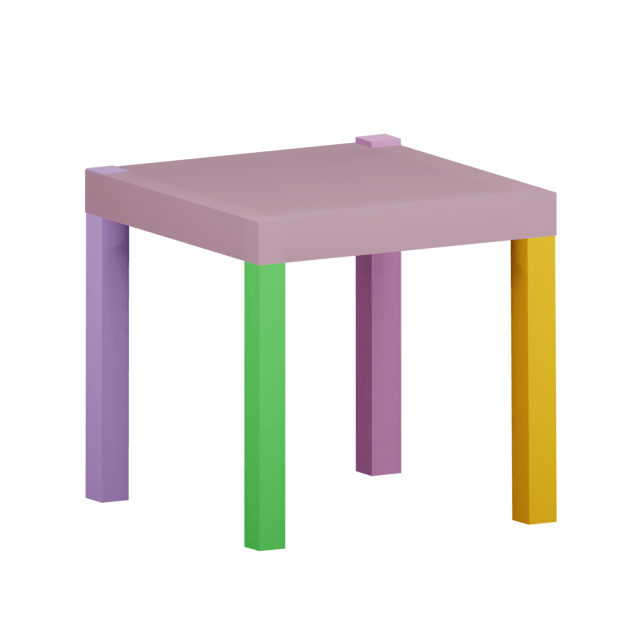} &
    \includegraphics[]{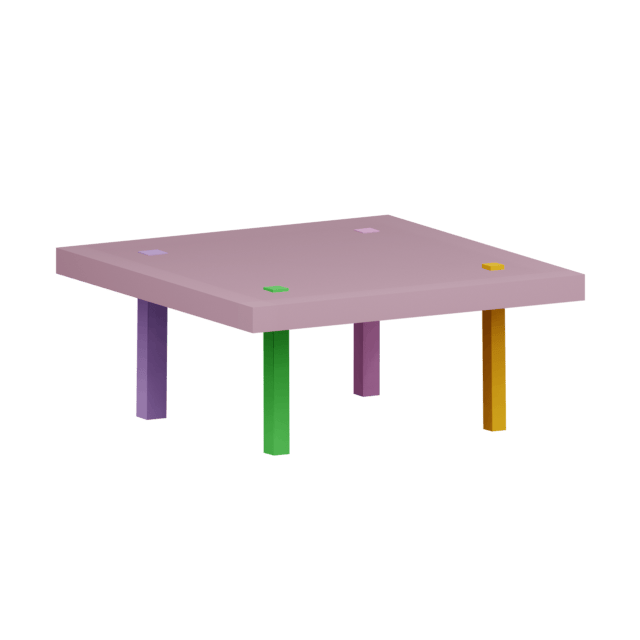} & 
    \includegraphics[]{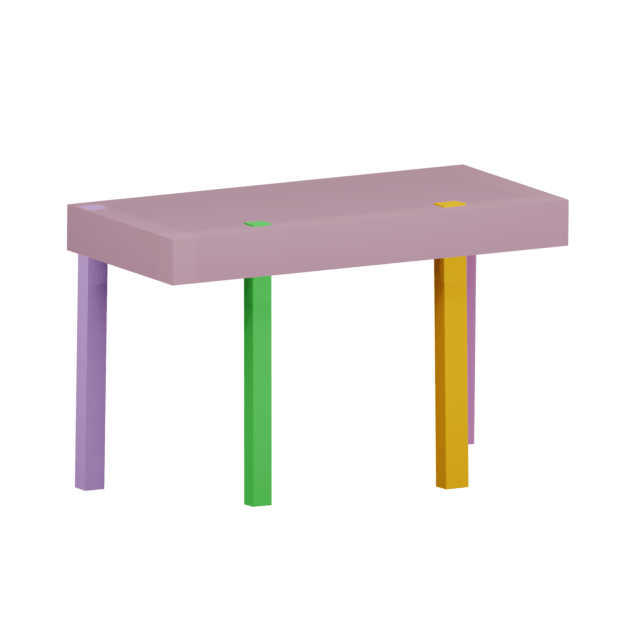} &
    
    \includegraphics[]{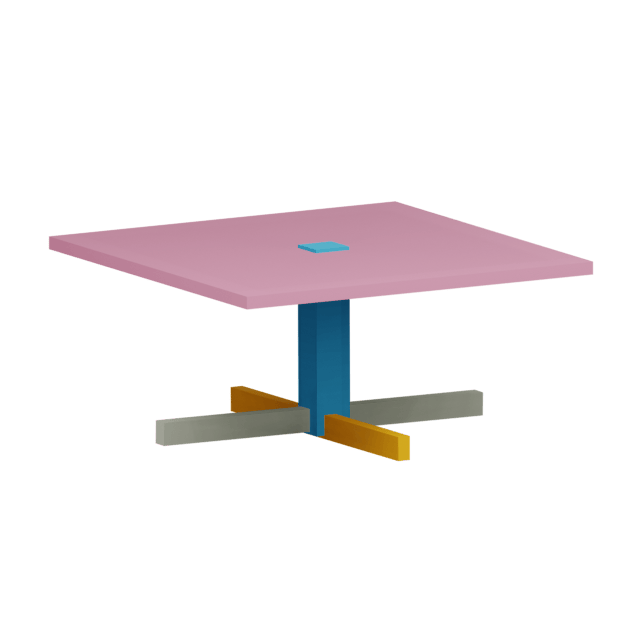} & 
    \includegraphics[]{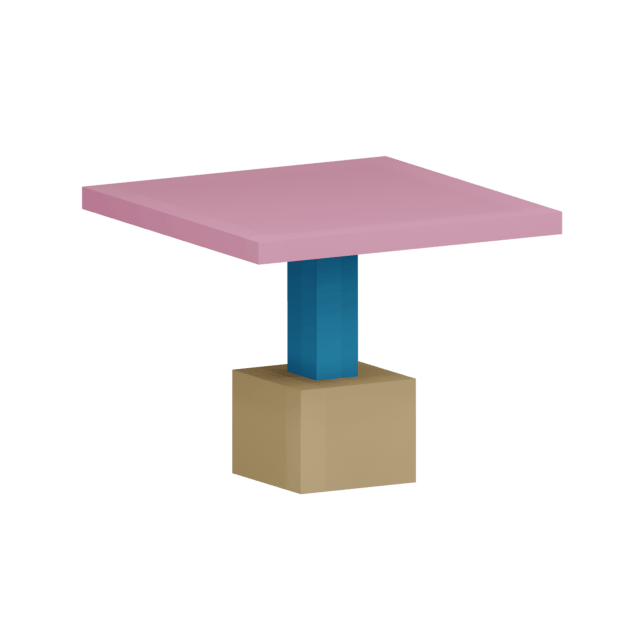} & 
    \includegraphics[]{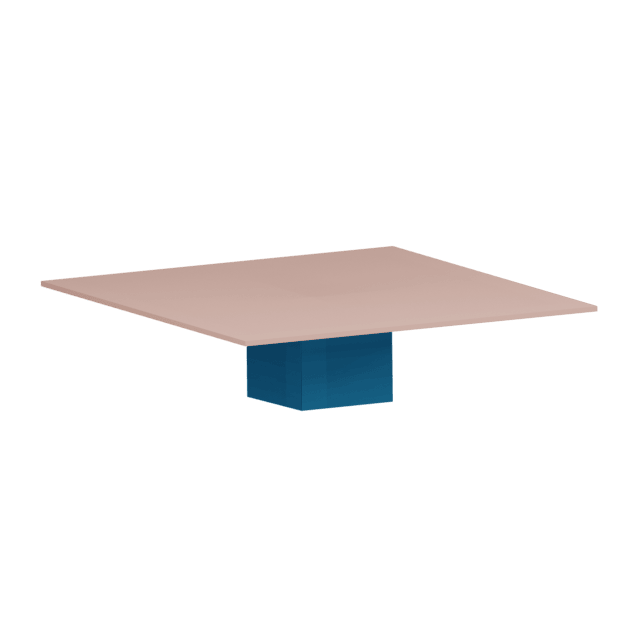} &
    
    \includegraphics[]{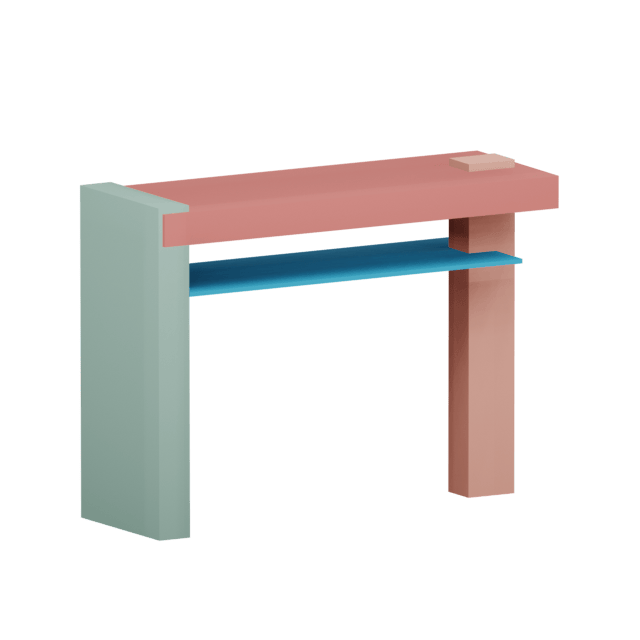} & 
    \includegraphics[]{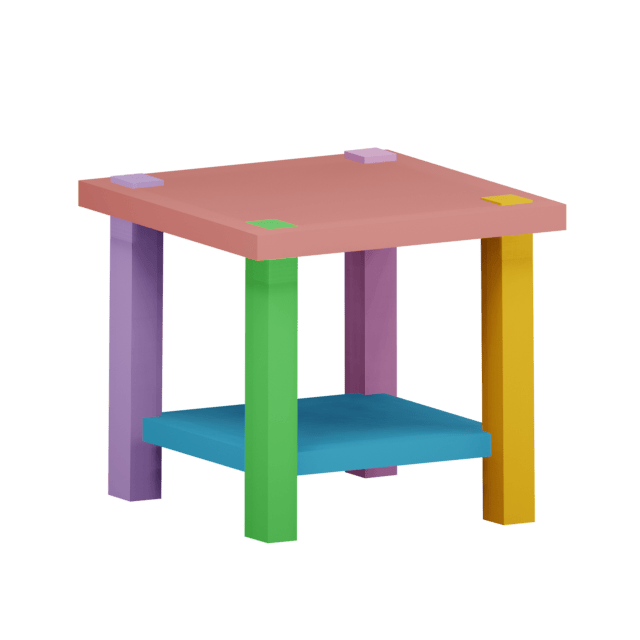} & 
    \includegraphics[]{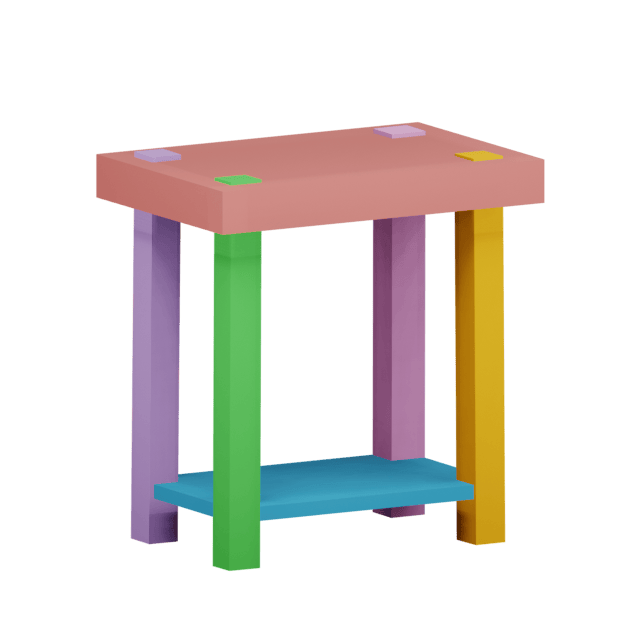} \\
    
\end{tabular}}
    
    \rule{0.9\textwidth}{0.4pt} \\
    
    \begin{tabular}{c}
        \rotatebox[origin=c]{90}{\hspace{0.6cm} GT \hspace{0.6cm}} \\
        \rotatebox[origin=c]{90}{\hspace{0.6cm} CAS \hspace{0.6cm}} \\
        \rotatebox[origin=c]{90}{\hspace{0.2cm} $\text{DPF}_{PPM}$ \hspace{0.2cm}} \\
        \rotatebox[origin=c]{90}{\hspace{0.6cm} Ours \hspace{0.6cm}} \\
    \end{tabular}
    \resizebox{0.96\textwidth}{!}{\begin{tabular}{cccccccccc}
    \includegraphics[]{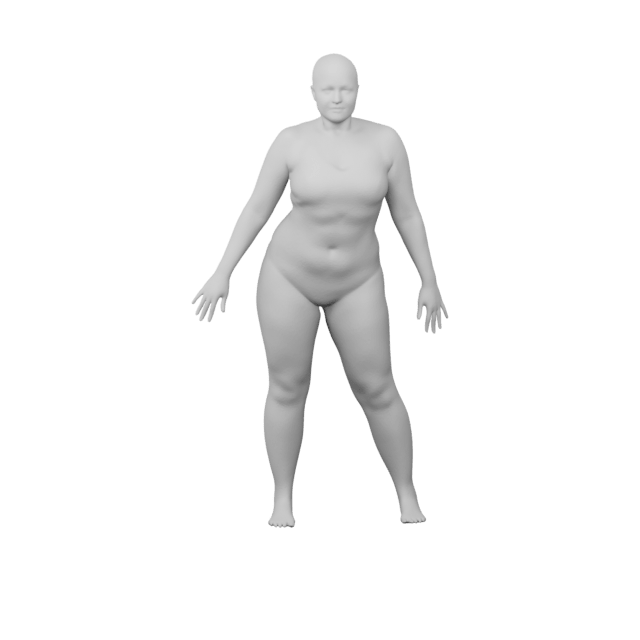} &
    \includegraphics[]{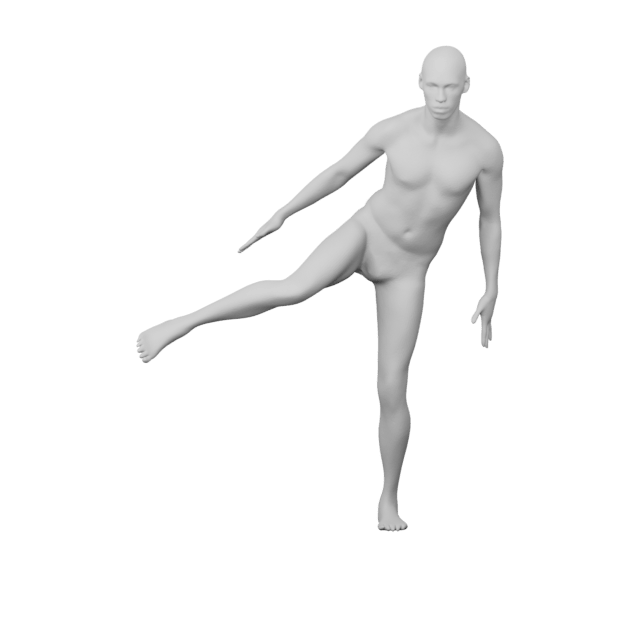} & 
    \includegraphics[]{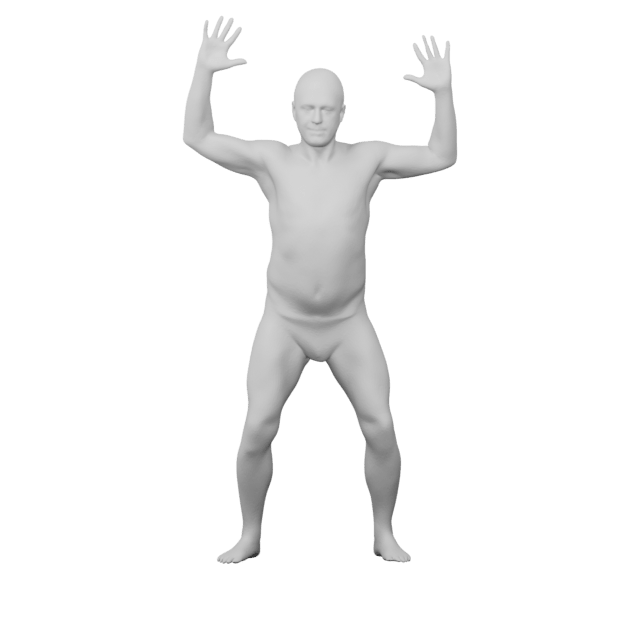} &
    
    \includegraphics[]{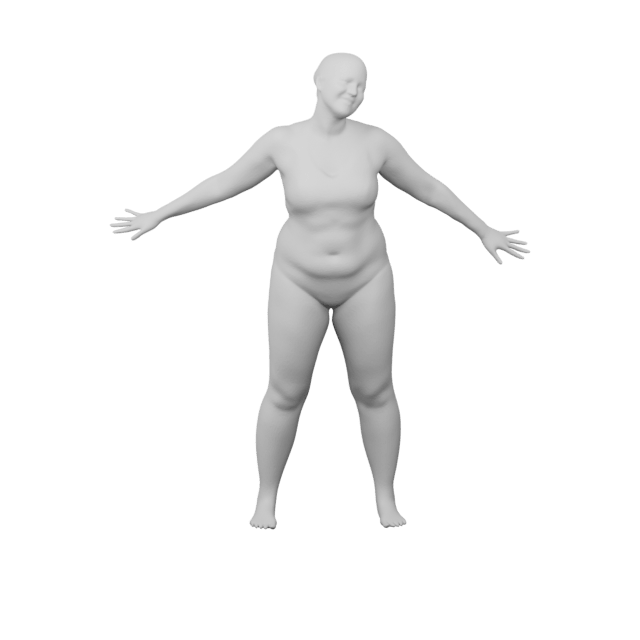} & 
    \includegraphics[]{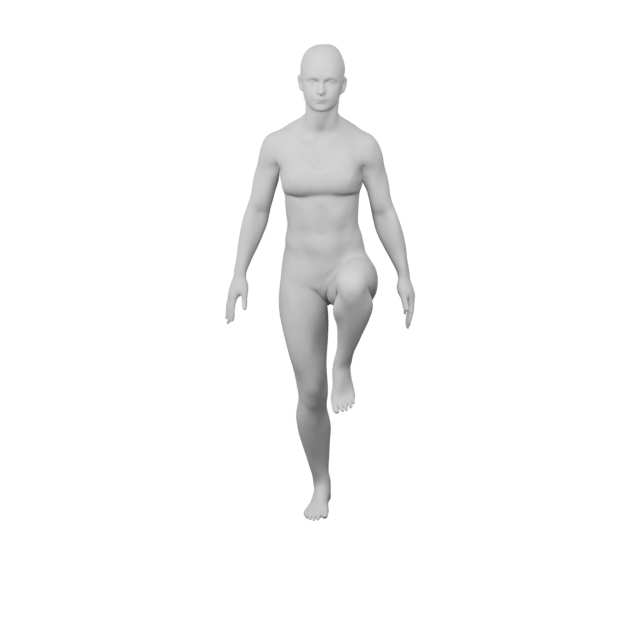} & 
    \includegraphics[]{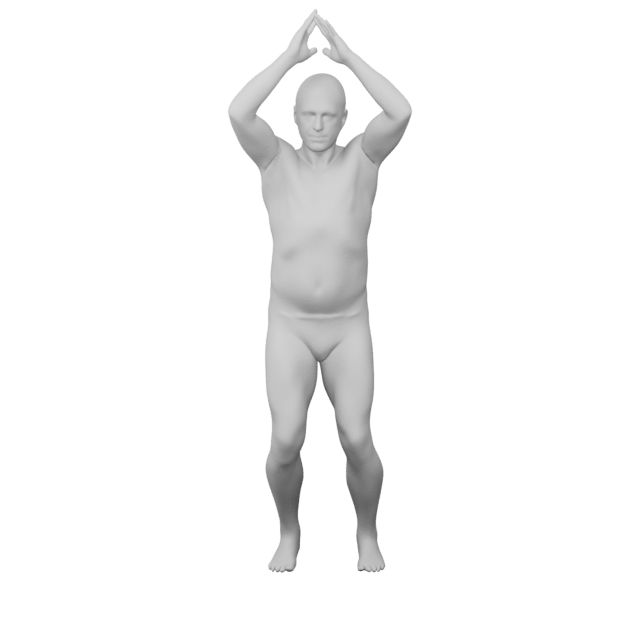} &
    
    \includegraphics[]{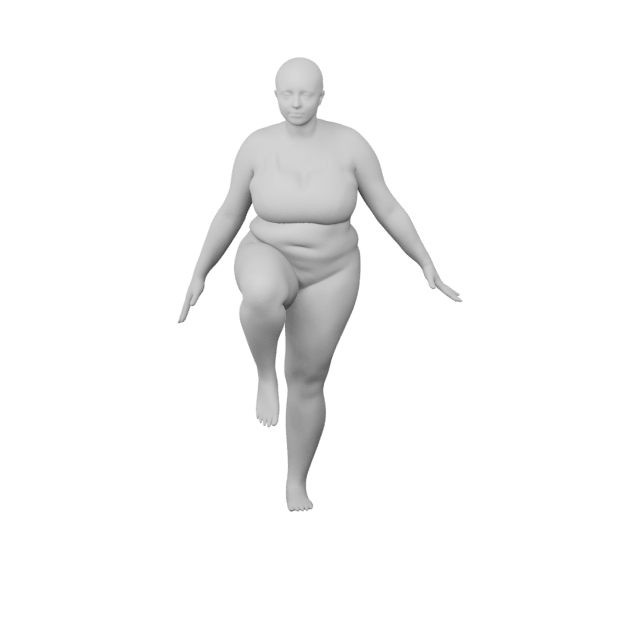} & 
    \includegraphics[]{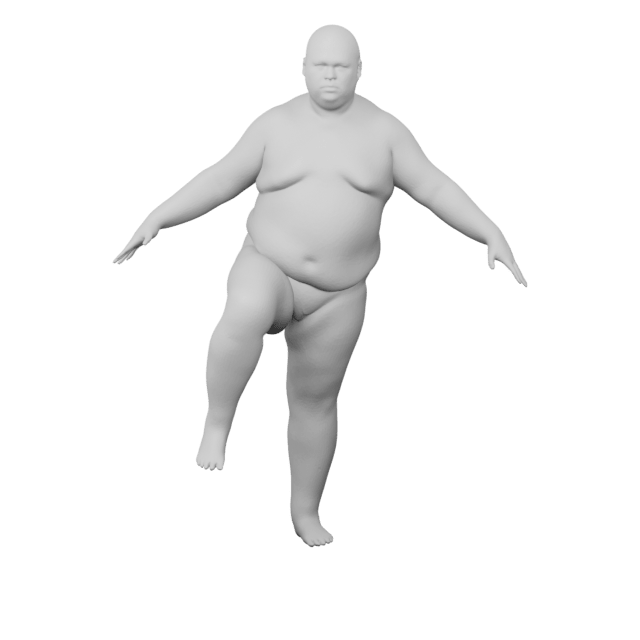} & 
    \includegraphics[]{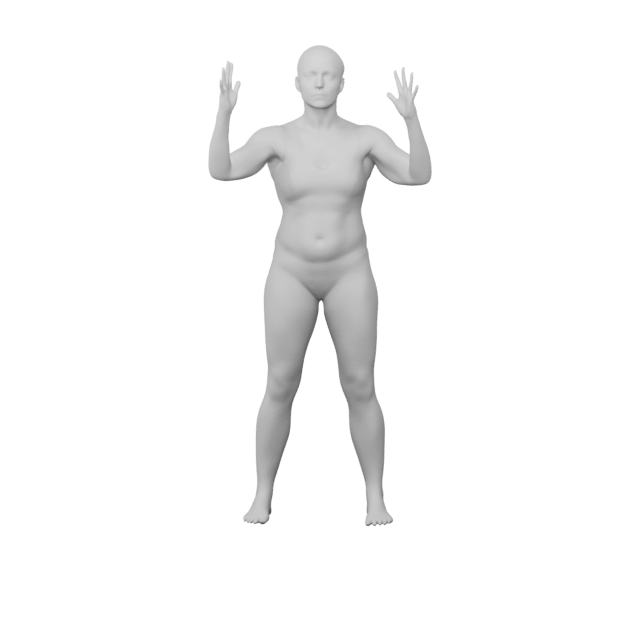} \\

    
    

    \includegraphics[]{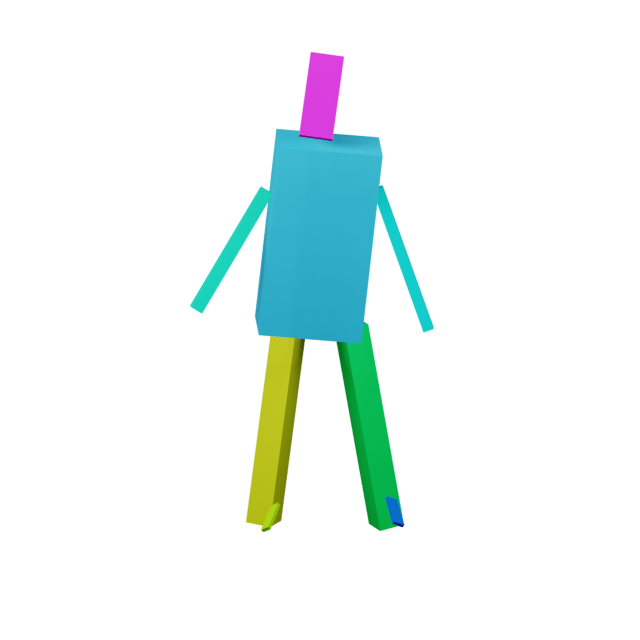} &
    \includegraphics[]{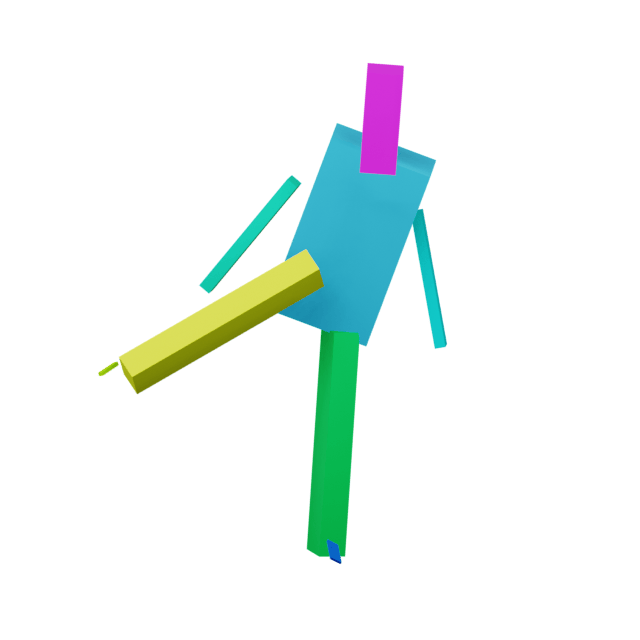} & 
    \includegraphics[]{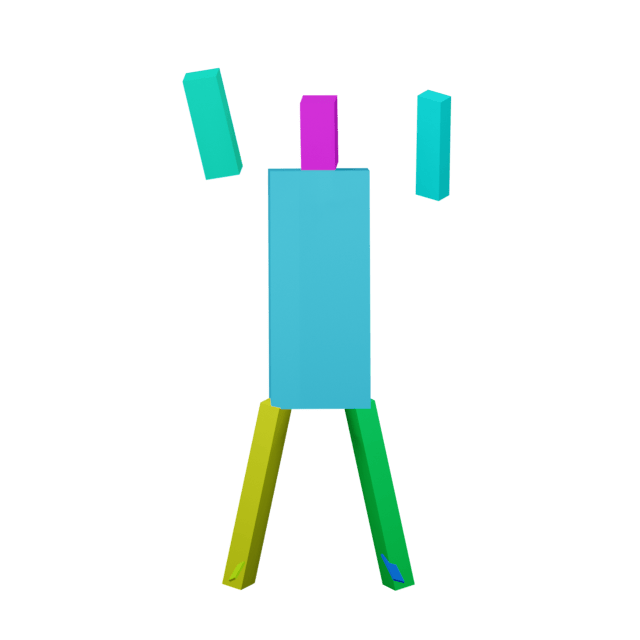} &
    
    \includegraphics[]{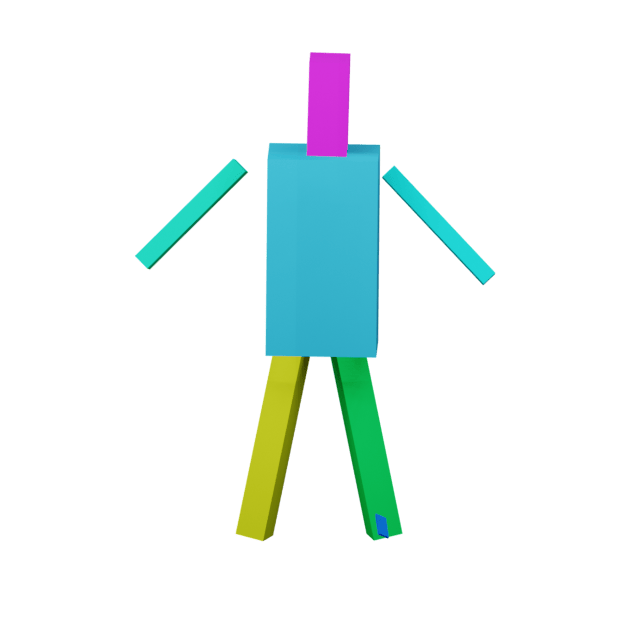} & 
    \includegraphics[]{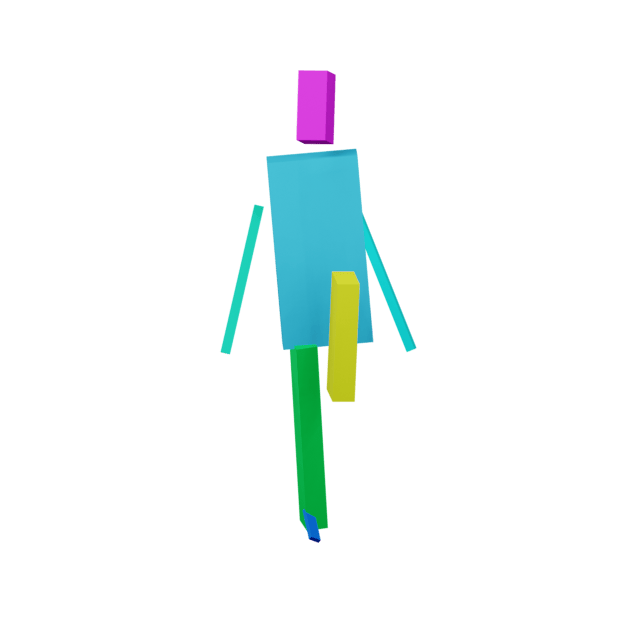} & 
    \includegraphics[]{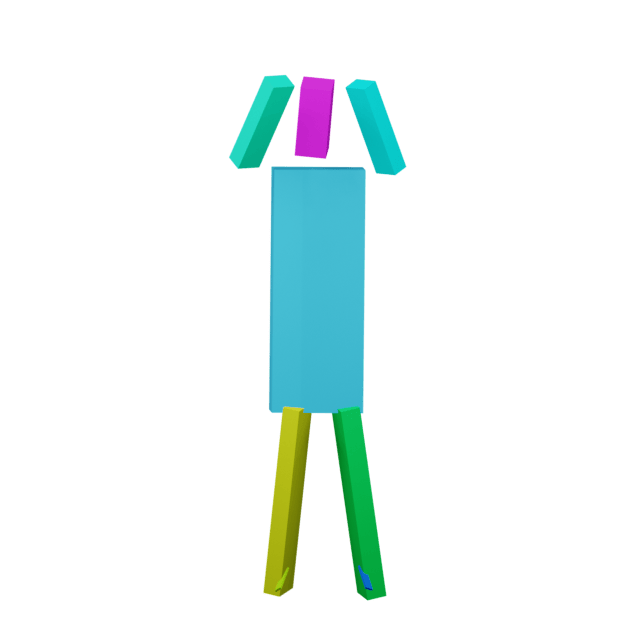} &
    
    \includegraphics[]{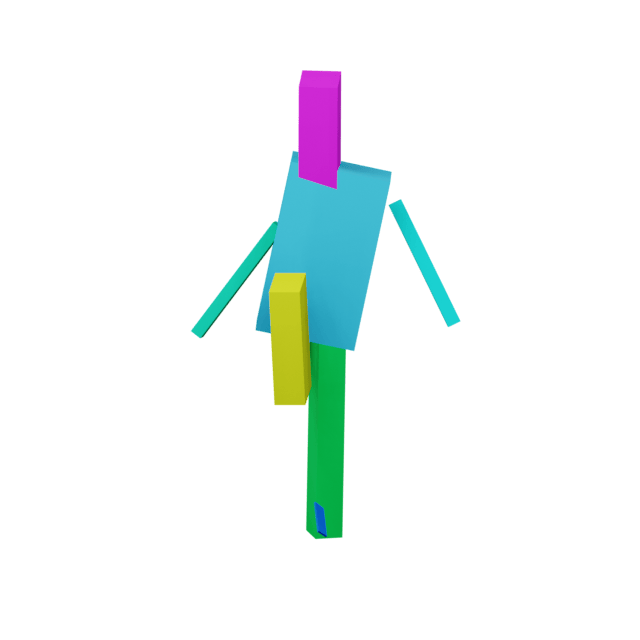} & 
    \includegraphics[]{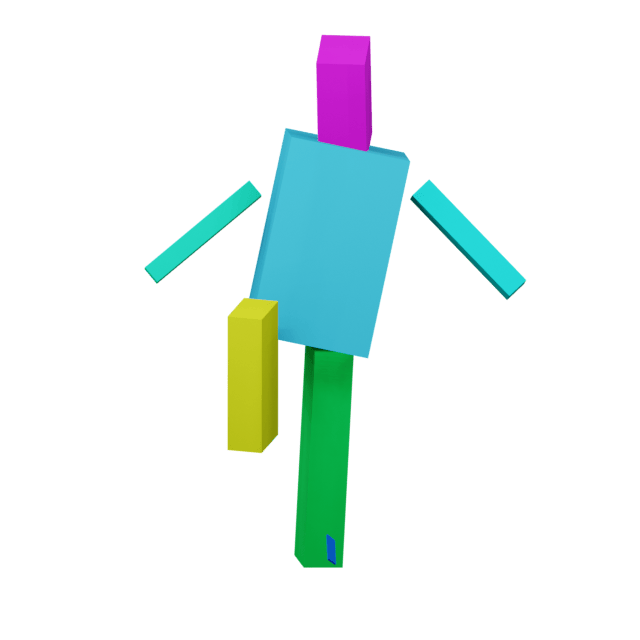} & 
    \includegraphics[]{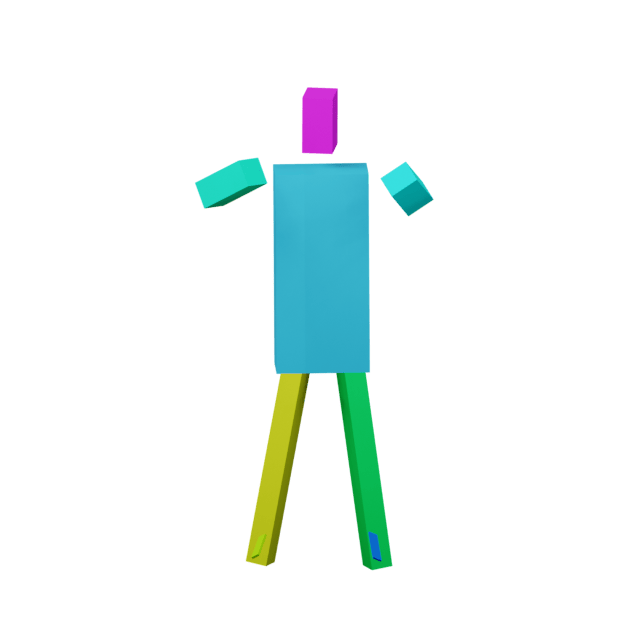} \\

    \includegraphics[]{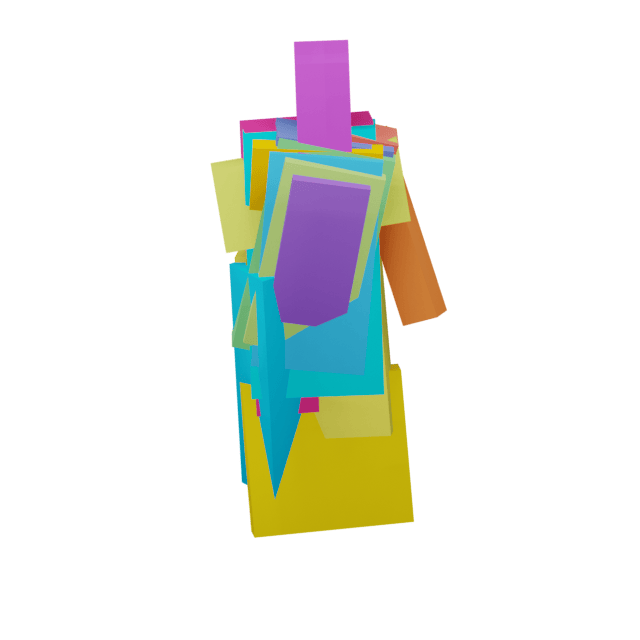} &
    \includegraphics[]{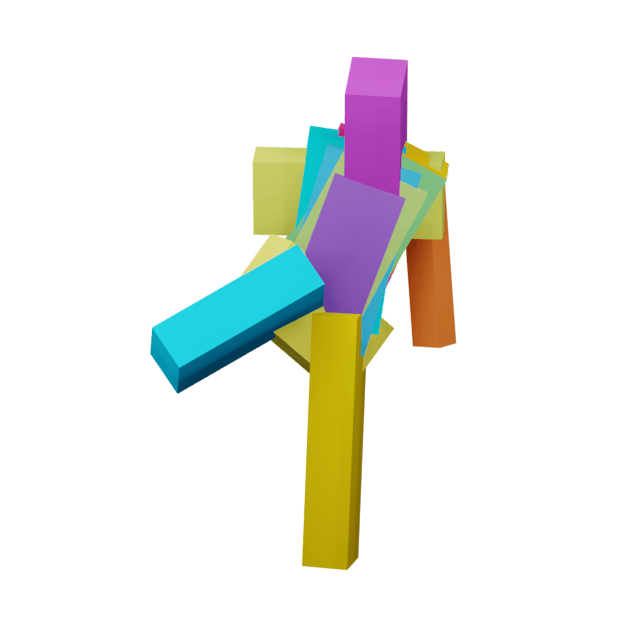} & 
    \includegraphics[]{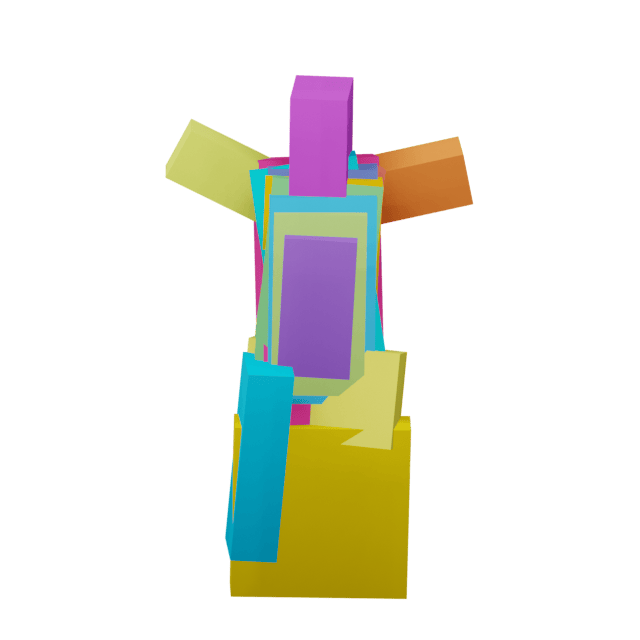} &
    
    \includegraphics[]{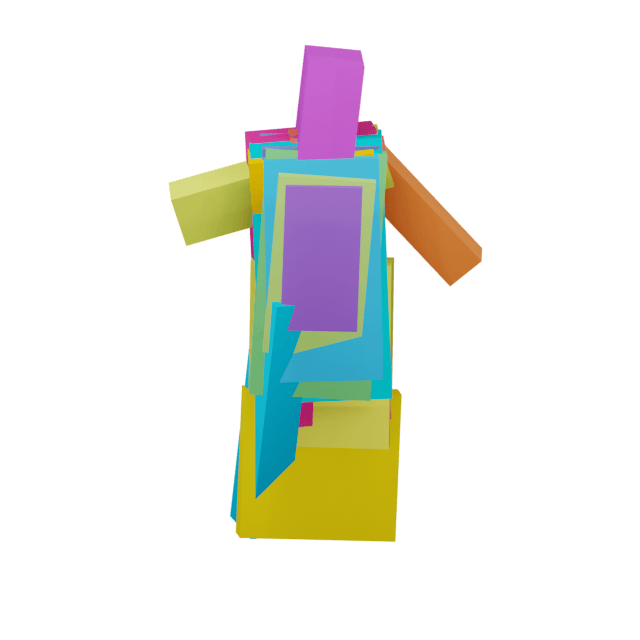} & 
    \includegraphics[]{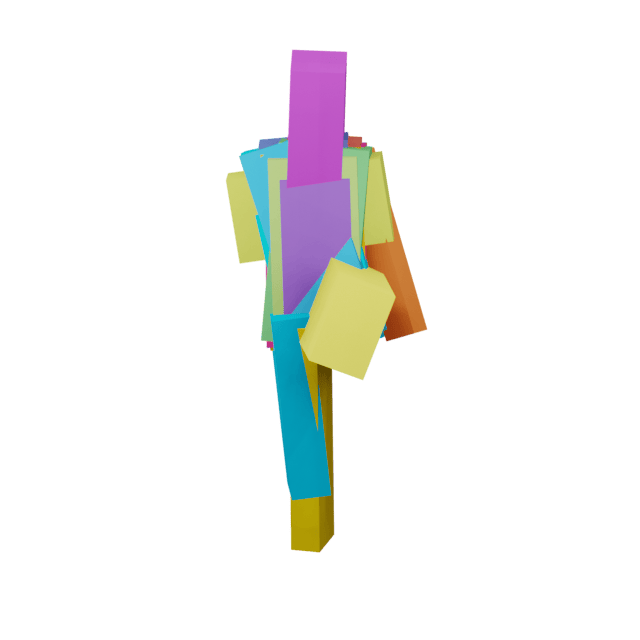} & 
    \includegraphics[]{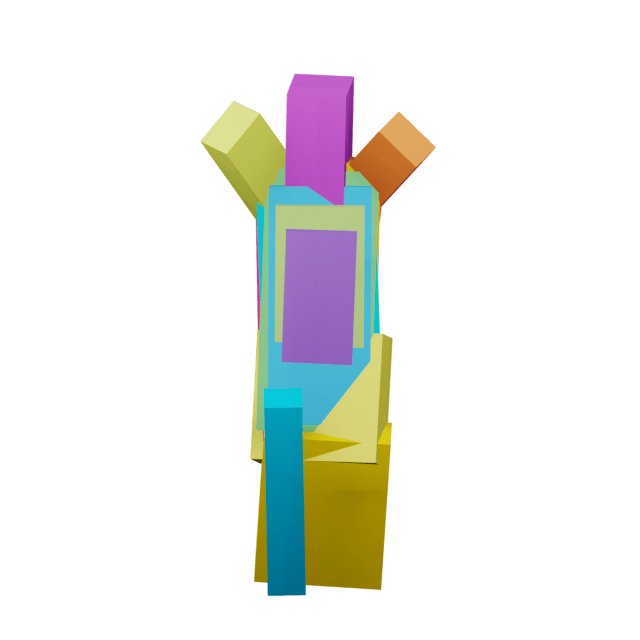} &
    
    \includegraphics[]{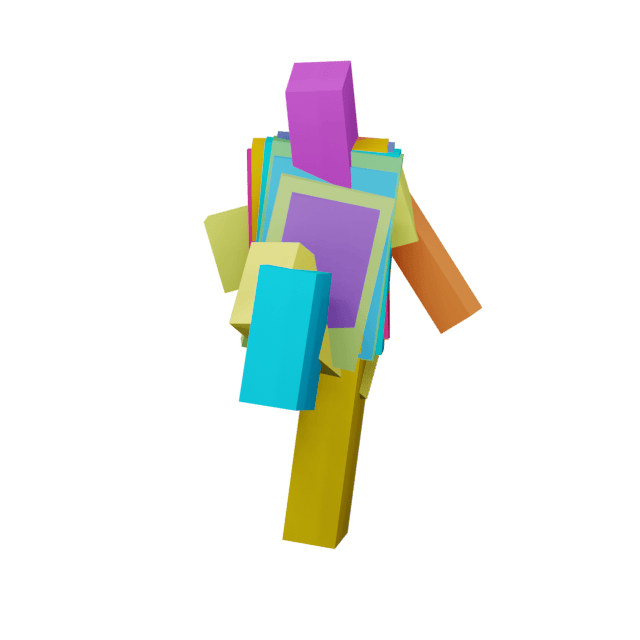} & 
    \includegraphics[]{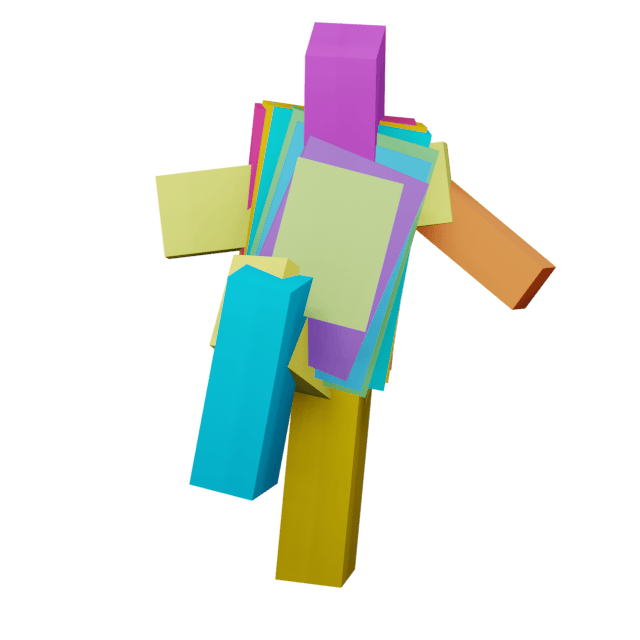} & 
    \includegraphics[]{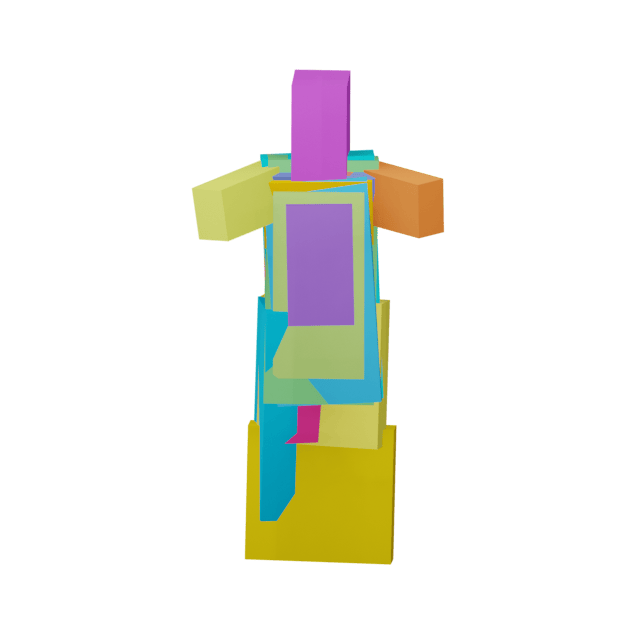} \\

    \includegraphics[]{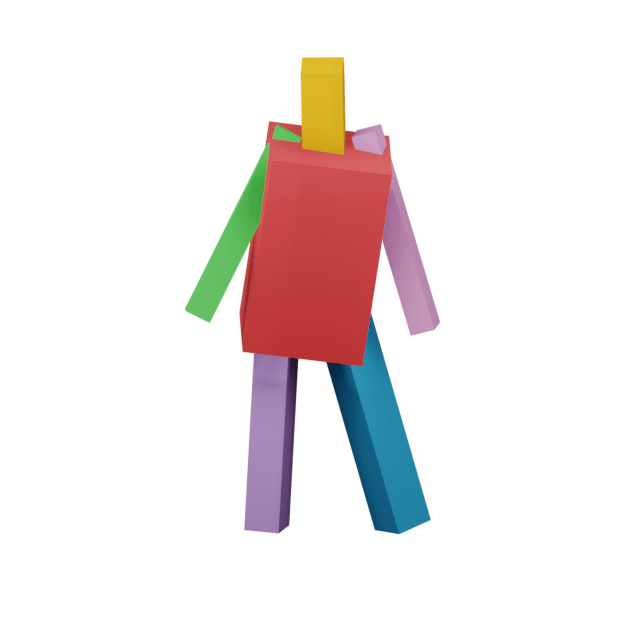} &
    \includegraphics[]{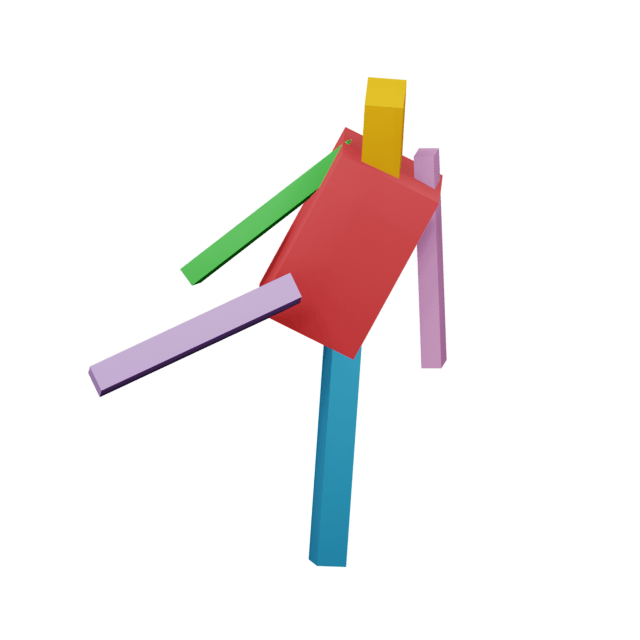} & 
    \includegraphics[]{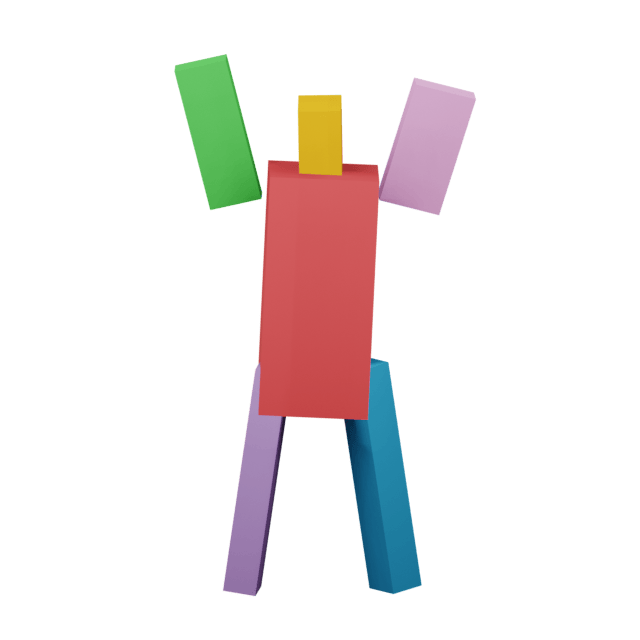} &
    
    \includegraphics[]{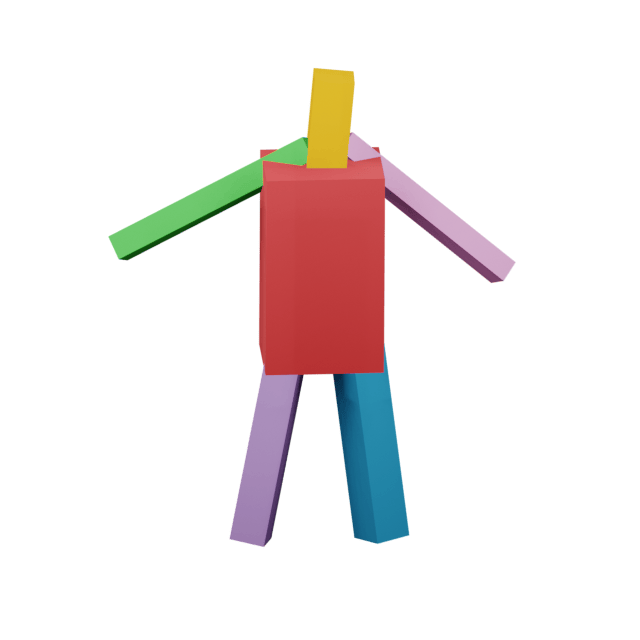} & 
    \includegraphics[]{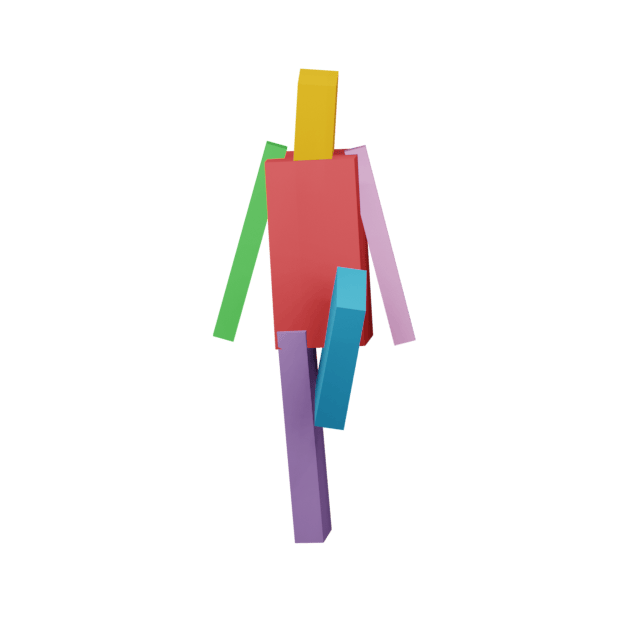} & 
    \includegraphics[]{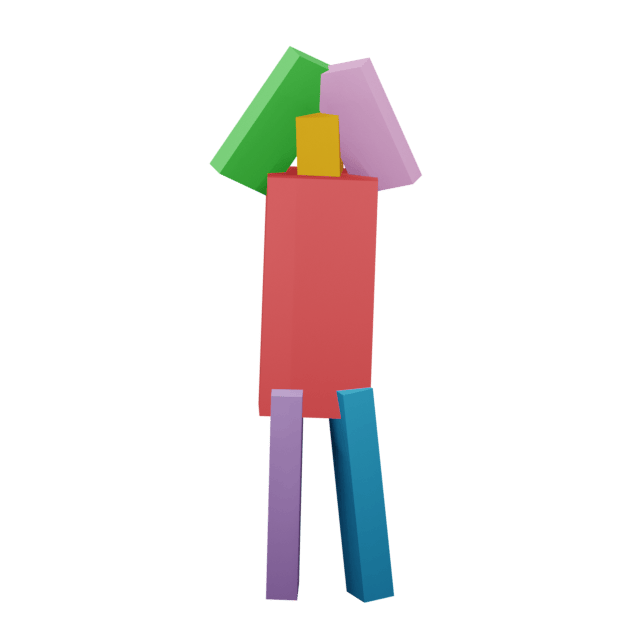} &
    
    \includegraphics[]{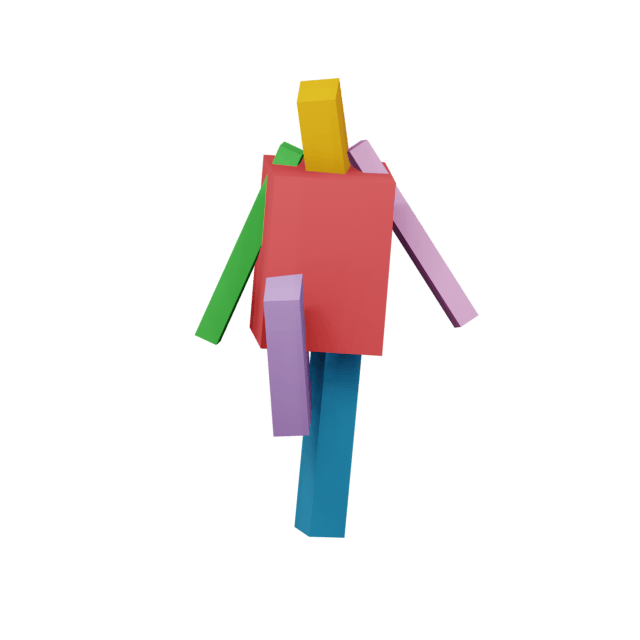} & 
    \includegraphics[]{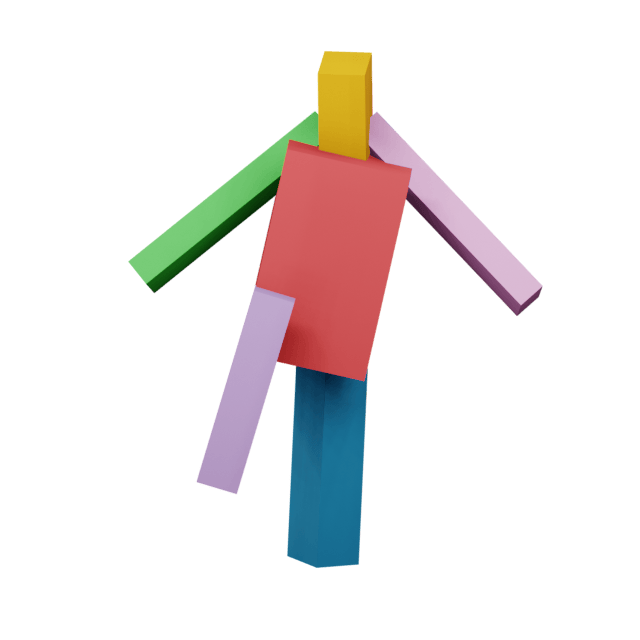} & 
    \includegraphics[]{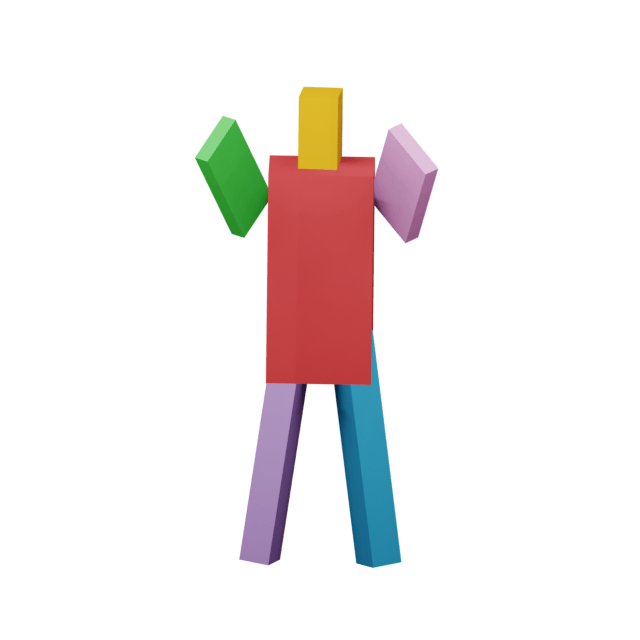} \\
    
\end{tabular}}
    \caption{Qualitative samples of the table and human class. We visualize the ground truth (GT) shape and the cuboid abstraction of \cite{HCA_Sun} (HCA), \cite{CAS_Yang} (CAS), \cite{DPF-Net_Shuai} ($\text{DPF}_{PPM}$) and Ours. Note, our abstraction preserves the geometry of the ground truth shape more closely using less cuboid primitives. Colors indicate distinc parts per category.}
    \label{fig:qualitative_table_human}
\end{figure*}

\end{document}